\documentclass[10pt,twocolumn,letterpaper]{article}

\usepackage{wacv}              %

\newif\ifsigg
\siggfalse

\newif\ifarxiv
\arxivtrue

\newenvironment{acks}{\section*{Acknowledgments}}{}

\usepackage{multirow}
\usepackage{pgfplots}
\usepackage{amssymb}
\usepgfplotslibrary{groupplots}
\usepackage{pgfplotstable}
\usepackage{soul}
\usepackage{subcaption}
\pgfplotsset{compat=newest}
\usepackage{siunitx}
\usepackage{arydshln}
\usepackage{multicol}
\usepackage{appendix}

\definecolor{wacvblue}{rgb}{0.21,0.49,0.74}
\usepackage[pagebackref,breaklinks,colorlinks,allcolors=wacvblue]{hyperref}

\def\wacvPaperID{1018} %
\def\confName{WACV}
\def\confYear{2027}

\title{Latent-Identity Tuning in Text-to-Image Personalization Models}

\author{
        Daniel Garibi$^1$ \hspace{4mm}
        Ronen Kamenetsky$^1$ \hspace{4mm}
        Hadar Averbuch-Elor$^2$ \hspace{4mm}
        Daniel Cohen-Or$^1$ \hspace{4mm}
        Or Patashnik$^1$
        \\[4pt]
        $^1$Tel Aviv University \hspace{10mm} $^2$Cornell University
        \\
        \small\url{https://garibida.github.io/IdentityTuning/}
         \\
}

\begin{document}
\ifsigg
    \begin{teaserfigure}
        \centering
        \includegraphics[width=1.0\textwidth]{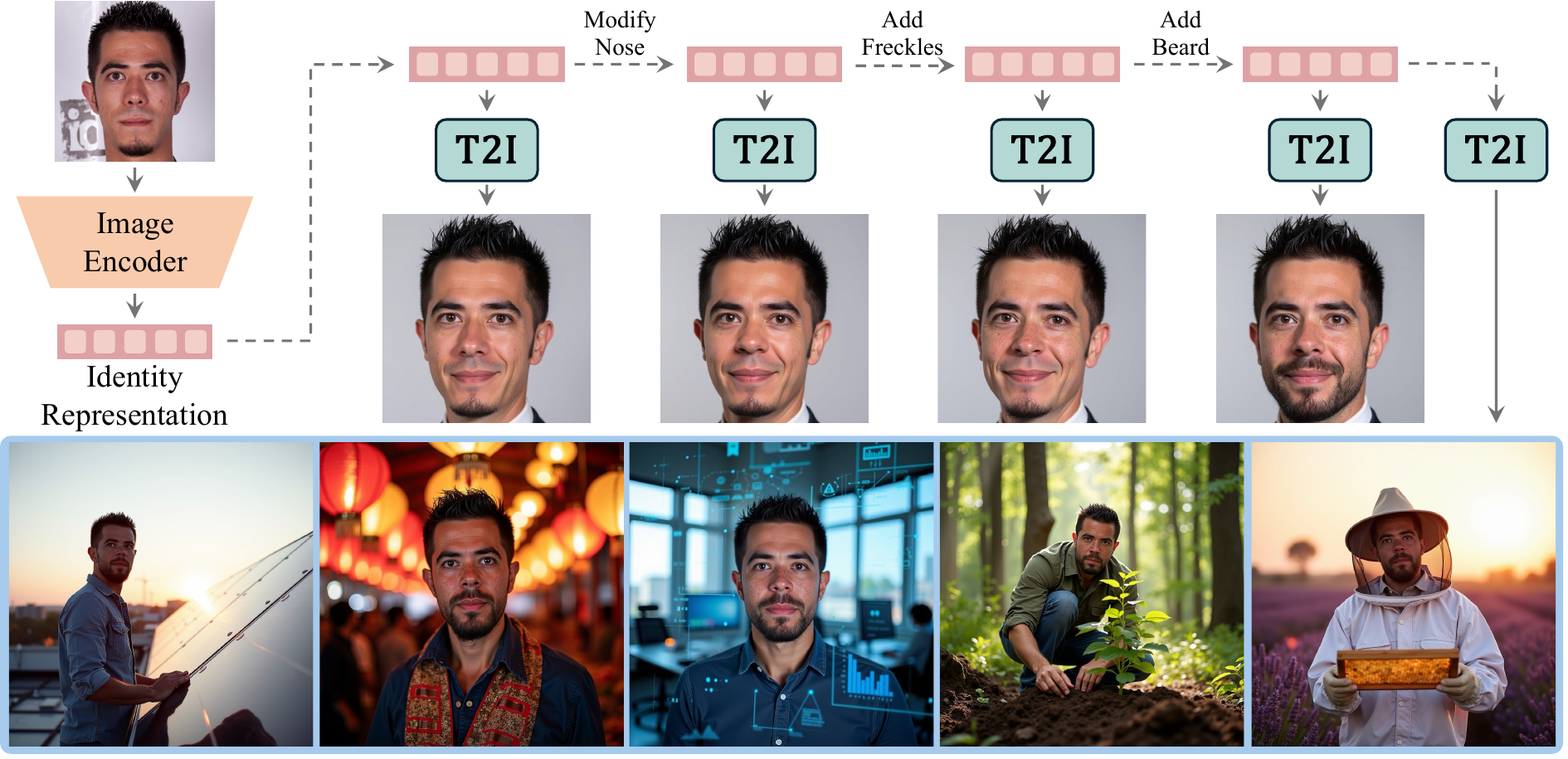}
        \vspace{-20pt}
        \caption{
            We present methods for directly tuning the identity tokens of a personalization encoder, enabling fine-grained control of facial attributes, for example modifying the nose, adding freckles, or a beard (top).
            The edited identity can then be used across diverse prompts to generate the same tuned subject consistently in new scenes (bottom).
        }
        \label{fig:teaser}
    \end{teaserfigure}
\else
\twocolumn[{%
    \renewcommand\twocolumn[1][]{#1}%
    \maketitle
    \begin{center}
        \vspace{-27pt}
        \includegraphics[width=1.0\textwidth]{images/teaser2.pdf}
        \vspace{-20pt}
        \captionof{figure}{
        We present methods for directly tuning the identity tokens of a personalization encoder, enabling fine-grained control of facial attributes, for example modifying the nose, adding freckles, or a beard (top).
            The edited identity can then be used across diverse prompts to generate the same tuned subject consistently in new scenes (bottom).
        }
    \vspace{-5pt}
    \label{fig:teaser}
    \end{center}
}] 
\fi

\begin{abstract}
    \vspace{-25pt}

Generating and editing a person's face demands high precision, as even minor modifications can significantly alter a subject’s perceived identity. 
Current personalization and editing methods built on general-purpose text-to-image models, however, often lack the precision required for fine-grained facial edits.
We present a method for fine-grained identity tuning in text-to-image personalization models.
Unlike standard image editing, which operates on a given image, identity tuning modifies the latent representation of a specific identity, enabling the generation of diverse images that consistently depict the same edited identity.
To enable fine-grained latent identity tuning, we explore the latent space of a pre-trained, frozen encoder for text-to-image personalization.
Our approach requires no additional training. Instead, it leverages the existing architecture of a frozen encoder to uncover latent semantic directions.
This space consists of a set of latent tokens that play distinct roles in capturing different aspects of an identity
and often correspond to specific spatial or semantic facial regions.
We show that meaningful directions can be identified within this space and within subspaces defined by selected tokens, enabling localized, fine-grained, and semantically coherent edits.
We validate our approach through qualitative and quantitative experiments that demonstrate diverse localized facial edits while preserving cross-image identity consistency.

\end{abstract}

\vspace{-15pt}
\section{Introduction}
\vspace{-5pt}
\label{sec:intro}

Advances in text-to-image generation~\cite{ho2020denoising, song2020score, rombach2022highresolution, lipman2023flowmatchinggenerativemodeling} have enabled highly personalized synthesis, where models can generate new images of a person from a single or a few reference images. Such personalization methods allow the depiction of an individual in various contexts that vary in background, style, and pose~\cite{gal2022imageworthwordpersonalizing, ruiz2024hyperdreambooth, ye2023ipadaptertextcompatibleimage, labs2025flux1kontextflowmatching}. However, while existing methods faithfully reproduce identity, they offer little control over modifying it, limiting users’ ability to refine, customize, or creatively reinterpret how a given identity is depicted. 
For instance, a person may wish to appear with a beard, freckles, or a modified nose, requiring this edited appearance to remain consistent across all subsequent generations 
(\cref{fig:teaser}).

We define identity tuning as the task of modifying the latent representation that encodes a person’s identity in a pre-trained text-to-image personalization model.
In contrast to image editing, which manipulates individual images, 
identity tuning operates on the latent identity representation itself. This ensures that the modified attributes remain consistent across all generated samples, establishing a persistent, altered identity that can be reliably deployed across varied prompts and settings~\cite{dravid2024interpretingweightspacecustomized, rishubh2024precisecontrol}.
As this task involves human faces, it requires exceptional precision, as even minute differences can alter the perception of the identity.
Achieving this level of precision requires enabling localized and continuous control over how the identity is portrayed or intended to appear.

Existing personalization methods typically learn an embedding or a set of latent tokens that capture a person’s identity from one or more reference images~\cite{qian2025omniidholisticidentityrepresentation, ye2023ipadaptertextcompatibleimage, guo2024pulid}. While these representations are primarily used to reproduce identity during generation, their internal structure has remained largely unexplored. In this work, we analyze the identity latent space of a personalization encoder, and show that 
it contains meaningful editing directions that can be explicitly identified.
These directions enable fine-grained and semantically coherent tuning of a person’s identity across generated samples.

Specifically, the identity latent space consists of a set of latent tokens that play distinct roles in representing different generative aspects of facial identity, with some corresponding to specific spatial or semantic regions of the face~\cite{li2023blip2bootstrappinglanguageimagepretraining, qian2025omniidholisticidentityrepresentation}, as illustrated in 
\cref{fig:attn_tokens}.
To enable control within this space, we investigate how manipulating individual identity tokens or all tokens jointly affects the encoded identity, and introduce a method to identify the tokens most relevant to a given localized facial attribute.
We then apply several supervised and unsupervised techniques to the selected tokens to discover semantic directions in the latent space.
These directions enable targeted and continuous fine-grained edits, ranging from localized feature modifications to smooth interpolations between identities.

We demonstrate the effectiveness of our framework across a range of identity tuning scenarios and different personalization models.
In particular, we show that our approach enables fine-grained and semantically meaningful edits that are difficult to achieve through alternative text-guided methods, such as controlling the size or openness of the eyes, adding a specific beard, or transferring eyebrows from one identity to another.
Moreover, by producing consistent identity representations across diverse images, our framework allows users to adjust an identity once and obtain results that align with their preferred appearance.

\ifsigg
\begin{figure}[t]
\else
\begin{figure}[t]
\fi
    \centering
    \footnotesize
    
    \setlength{\tabcolsep}{0pt}
    \begin{tabular}{@{}c@{\hspace{6pt}}c@{}c@{}c@{}}
    
        \includegraphics[width=0.24\linewidth]{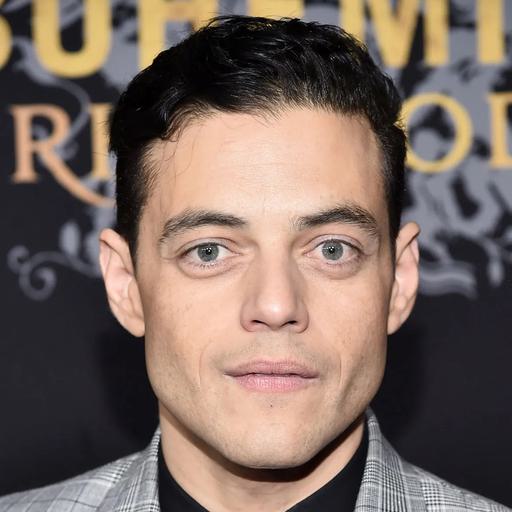} &
        \includegraphics[width=0.24\linewidth]{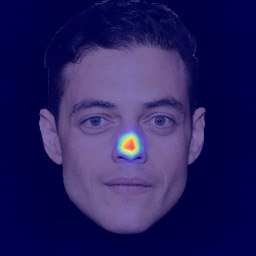} &
        \includegraphics[width=0.24\linewidth]{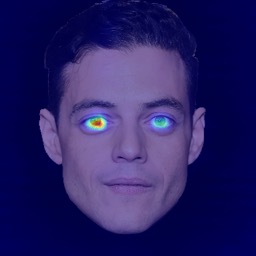} &
        \includegraphics[width=0.24\linewidth]{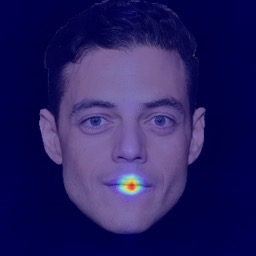} \\
        
        \includegraphics[width=0.24\linewidth]{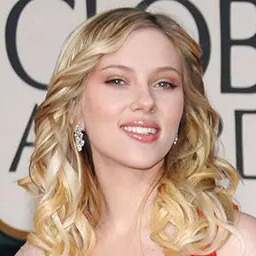} &
        \includegraphics[width=0.24\linewidth]{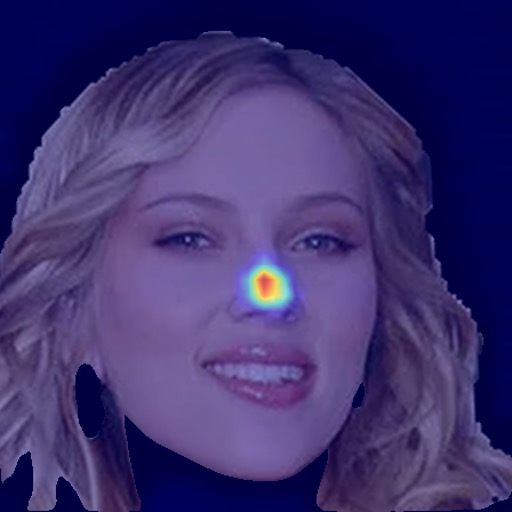} &
        \includegraphics[width=0.24\linewidth]{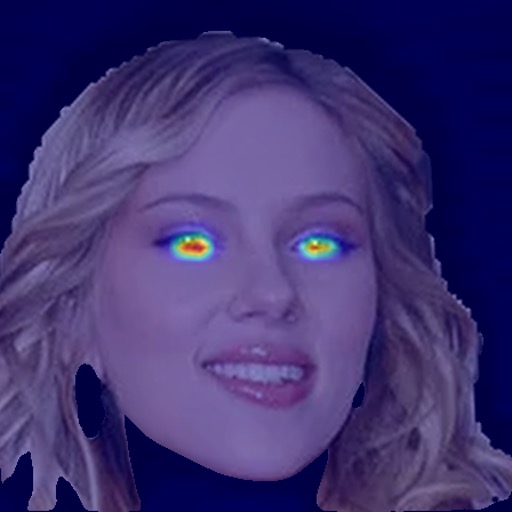} &
        \includegraphics[width=0.24\linewidth]{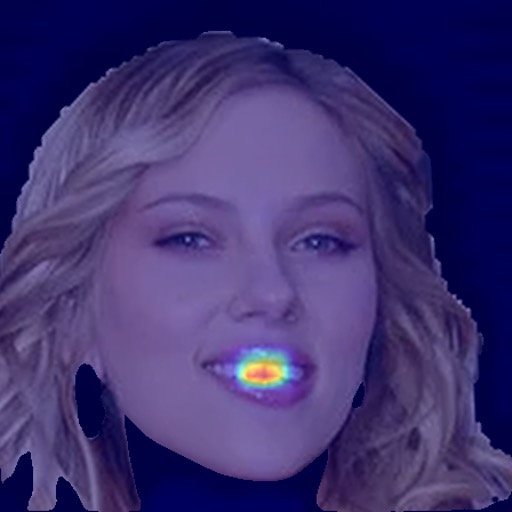} \\

        Input Image & \multicolumn{3}{c}{Q-Former attention maps} \\

    \end{tabular}
    \vspace{-8pt}
    \caption{
        Token behavior in the face encoder. Q-Former attention maps for three learned query tokens reveal localized semantics: one focuses on the eyes, one on the nose, and one on the lips. The same token keeps the same role across both identities, showing consistent token-to-region mapping.
    }
    \vspace{-15pt}
    \label{fig:attn_tokens}
\ifsigg
\end{figure}
\else
\end{figure}
\fi

\section{Related Work}
\label{sec:related_work}
\vspace{-5pt}

\paragraph{Text-to-Image Personalization}
Given an image or a small set of images depicting a single concept, text-to-image personalization enables generating new images of that concept, guided by a text prompt that determines style, composition, and interactions with other subjects. Early works achieved this through per-subject optimization
~\cite{gal2022imageworthwordpersonalizing, ruiz2023dreamboothfinetuningtexttoimage, kumari2023multiconceptcustomizationtexttoimagediffusion, tewel2024keylockedrankeditingtexttoimage, alaluf2023neuralspacetimerepresentationtexttoimage, voynov2023pextendedtextualconditioning, garibi2025tokenverseversatilemulticonceptpersonalization, arar2024palp, Avrahami_2023, po2023orthogonal}.
Later works trained encoders that directly map the input images to concept embeddings, which are then used to condition the denoising process~\cite{gal2023encoderbaseddomaintuningfast, ye2023ipadaptertextcompatibleimage, arar2023domainagnostic, ruiz2024hyperdreambooth, tan2025ominicontrol, kumari2025generating, cai2025diffusion, labs2025flux1kontextflowmatching, li2023blip, subjectdiffusion}.

Personalization of human faces has been a central focus~\cite{wang2024moa, wang2024instantid, guo2024pulid, li2024photomaker, patashnik2025nestedattentionsemanticawareattention, gal2024lcm, qian2025omniidholisticidentityrepresentation}, as it enables generating personalized depictions of individuals but also introduces unique challenges. Faithfully representing identity requires learning fine-grained details to which humans are highly sensitive, and users often find that personalized generations do not fully align with how they perceive themselves or wish to appear.
We introduce a method for localized, continuous tuning of the identity representation, enabling fine-grained refinement or modification of a specific identity.

\vspace{-13pt}
\paragraph{Image Editing}
Recent advances in generative models have substantially improved image editing capabilities. Early works showed that the latent spaces of GANs~\cite{goodfellow2014generative, karras2019style} are organized in a semantically meaningful and disentangled manner, enabling controlled manipulation of diverse semantic attributes across images~\cite{patashnik2021styleclip, shen2020interpreting, abdal2020styleflow, abdal2019image2stylegan, richardson2021encoding, harkonen2020ganspace, ling2021editgan}. While many GAN-based editing methods explored human faces, their latent representations correspond to individual images rather than persistent identities, making them more suitable for single-image edits such as expressions or poses.
Text-to-image diffusion models further broadened editing capabilities, enabling manipulation across diverse domains and direct control through text prompts~\cite{hertz2022prompttoprompt, mokady2022nulltext, epstein2023selfguidance, alaluf2023crossimage, hubermanspiegelglas2023edit, garibi2024renoiserealimageinversion, deutch2024turboedittextbasedimageediting, parmar2023zeroshot, tumanyan2023plug, brooks2023instructpix2pix, labs2025flux1kontextflowmatching, wu2025qwen, kulikov2025flowedit, dalva2024fluxspace, ge2023expressive, patashnik2023localizing, meng2022sdedit}.
Although these methods support intuitive natural-language interfaces and powerful manipulations, achieving fine-grained, localized, and continuous control remains challenging and often requires specialized approaches~\cite{gandikota2023conceptslidersloraadaptors, gandikota2025sliderspace, kamenetsky2025saedit, parihar2025kontinuous, golan2026paretosliderdiffusionmodelsposttraining, zarei2025slideredit, wolf2026continuouscontroleditingmodels, dahary2026repulsion, sella2026looseropecontentawareattentionmanipulation}. In particular, controlling identity-related 
attributes through textual interfaces remains a challenge.

\vspace{-13pt}
\paragraph{Identity Tuning}
Compared to image editing, identity editing has received limited attention. Weights2Weights~\cite{dravid2024interpretingweightspacecustomized} explores the weight space of LoRA-based personalized models, showing that it can be used to invert images, sample new identities, and edit existing ones. However, it requires per-image optimization to obtain the identity representation. In contrast, our method builds on a personalization encoder and operates without any optimization.
PreciseControl~\cite{rishubh2024precisecontrol} uses a GAN encoder~\cite{tov2021designing} for personalization alongside a text-to-image model, performing per-image optimization to improve expressiveness and identity preservation.
By aligning the GAN latent space with the personalization space, it enables GAN-based edits within text-to-image personalization.
However, this approach is constrained by the editability of the chosen GAN and incurs significant cost due to the per-image optimization.
IP-Composer~\cite{dorfman2025ipcomposersemanticcompositionvisual} also explores a pretrained personalization encoder, leveraging CLIP image–text alignment for semantic composition. However, it does not address fine-grained facial editing or discover interpretable directions in the identity space.

\ifarxiv
    \vspace{-2pt}
\else
    \vspace{-7pt}
\fi
\section{Preliminaries: Q-Formers}
\vspace{-5pt}
\label{sec:qformers}
Introduced under different names, specifically as Querying Transformers (Q-Formers) in BLIP-2~\cite{li2023blip2bootstrappinglanguageimagepretraining} and the Perceiver Resampler in Flamingo~\cite{alayrac2022flamingovisuallanguagemodel}, these architectures act as intermediate Transformer modules designed to adapt a source token stream for a downstream model. For clarity, we adopt the Q-Former terminology in this work. 
A Q-Former maintains a small set of learnable query tokens that, via cross-attention, attend to a source sequence (e.g., tokens from a frozen image encoder) and produce a fixed number of output tokens.
Each output token summarizes the information most relevant to its learned query, yielding a compact and structured representation while leaving the source model frozen.

In text-to-image personalization, given one or more images of a subject $\{I_i\}_{i=1}^k$, the goal is to extract a representation $Z \in \mathbb{R}^{N \times C}$ , where $N$ and $C$ denote the number of tokens and hidden dimension of each token, that captures the subject’s identity and can condition the diffusion model during generation.
IP-Adapter~\cite{ye2023ipadaptertextcompatibleimage} follows this approach by first extracting features $F \in \mathbb{R}^{HW \times d}$ with a pretrained image encoder, mapping them to a set of tokens $Z$ through an adapter network, and injecting them into the diffusion model via decoupled cross-attention layers.
Q-Formers naturally serve as adapters for this task, using learnable queries $Q \in \mathbb{R}^{N \times C'}$ to distill information from the input image(s) into a fixed-length set of tokens $Z \in \mathbb{R}^{N \times C}$.
Later works~\cite{qian2025omniidholisticidentityrepresentation, patashnik2025nestedattentionsemanticawareattention,guo2024pulid, wang2024instantid, han2024faceadapterpretraineddiffusion}
extends this framework to face-based personalization by introducing a backbone encoder tailored for human identity and employing a Q-Former as the adapter. In the case of human personalization, we refer to the Q-Former’s output tokens $Z$ as identity tokens.

Previous face-based personalization methods employing a Q-Former as the adapter analyzed the internal structure of the identity tokens and observed that different learned queries attend to distinct facial regions, such as the eyes, mouth, ears, and eyeglasses~\cite{patashnik2025nestedattentionsemanticawareattention,qian2025omniidholisticidentityrepresentation}.
These findings suggest that Q-Former–based adapters learn a structured and partially disentangled latent space representing human facial semantics.
In this paper, we further characterize the organization of the identity tokens space and show that it enables fine grained, localized identity tuning.

\ifsigg
\begin{figure}
\else
\begin{figure}
\fi
    \centering
    \setlength{\tabcolsep}{0pt}
    \scriptsize{
    \begin{tabular}{ccccc}
         \includegraphics[width=0.2\linewidth]{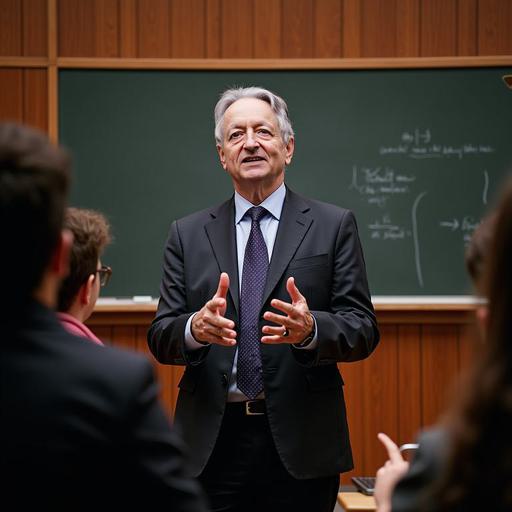} &
         \includegraphics[width=0.2\linewidth]{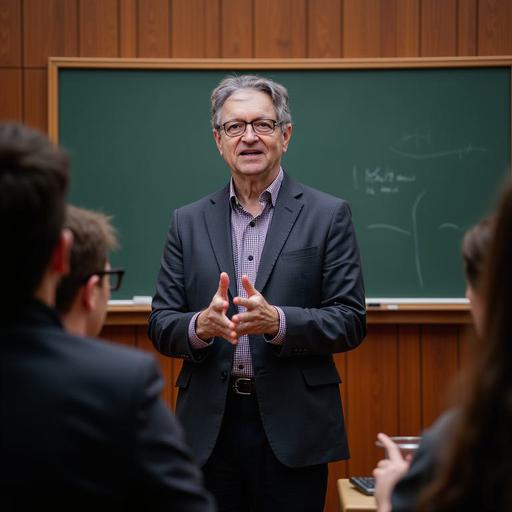} &
         \includegraphics[width=0.2\linewidth]{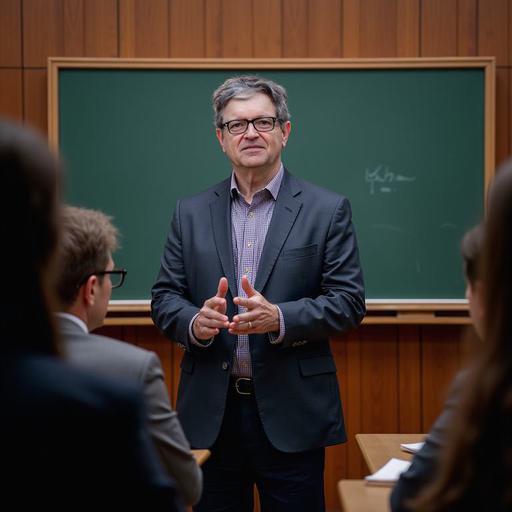} &
         \includegraphics[width=0.2\linewidth]{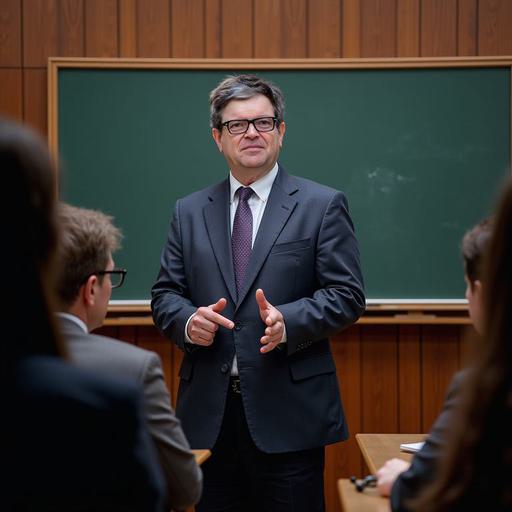} &
         \includegraphics[width=0.2\linewidth]{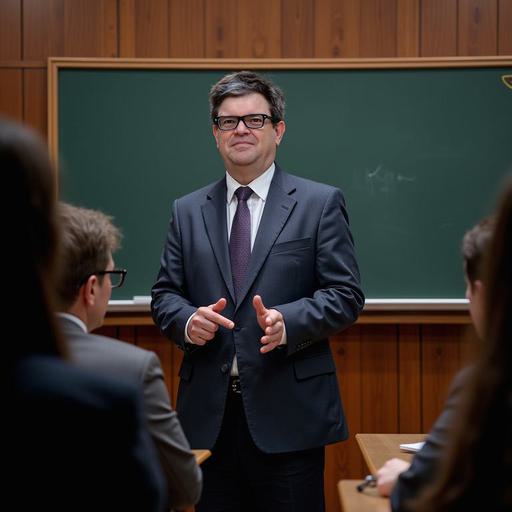} \\
         
         \includegraphics[width=0.2\linewidth]{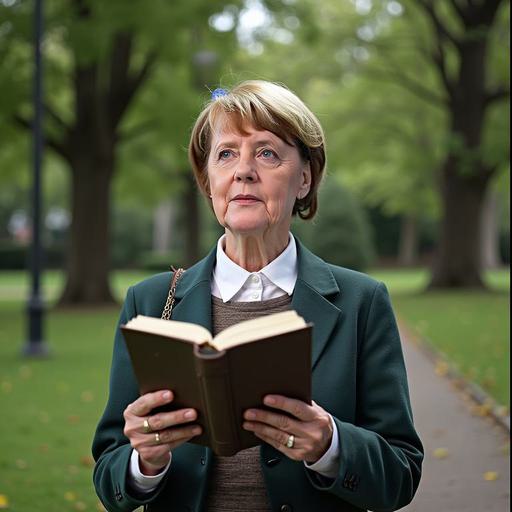} &
         \includegraphics[width=0.2\linewidth]{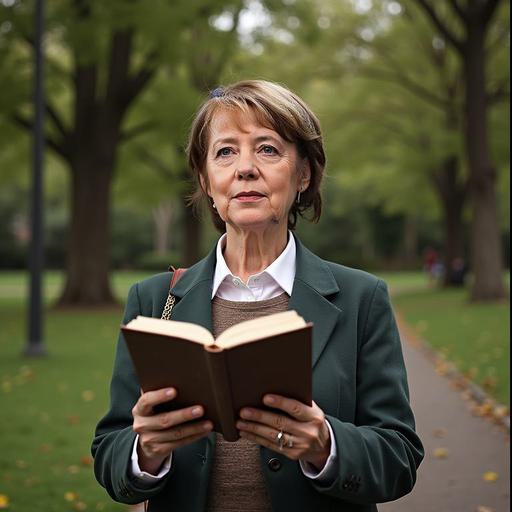} &
         \includegraphics[width=0.2\linewidth]{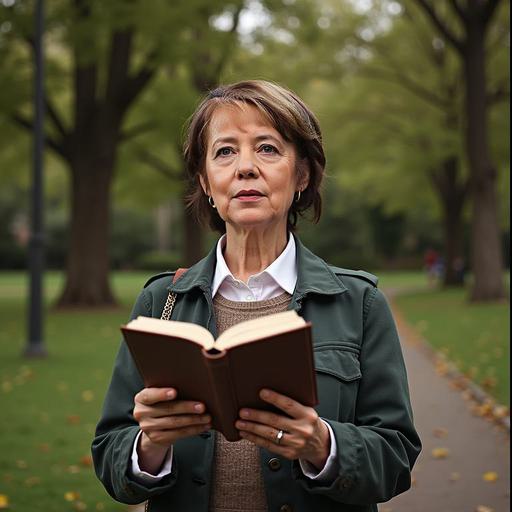} &
         \includegraphics[width=0.2\linewidth]{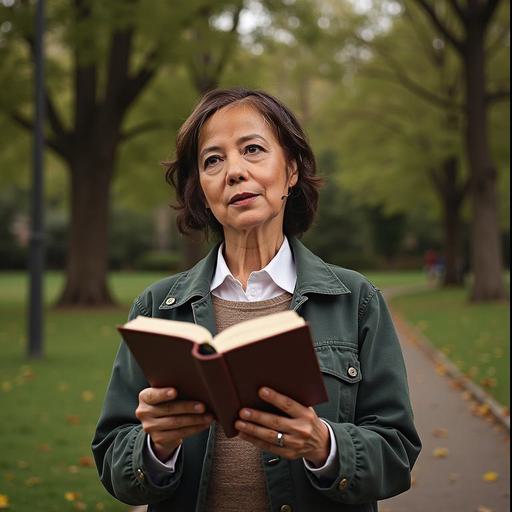} &
         \includegraphics[width=0.2\linewidth]{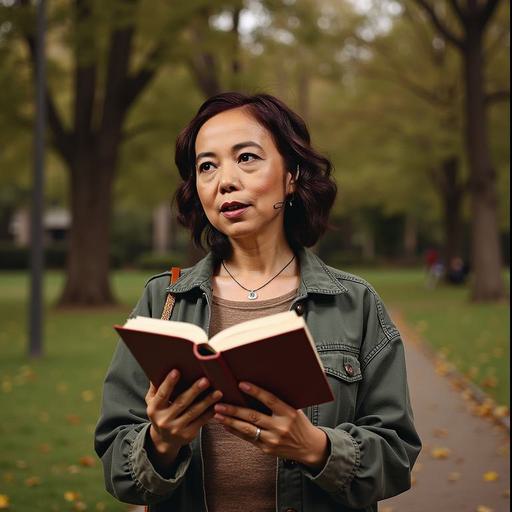} \\

         Identity A &&&& Identity B \\
    \end{tabular}
    }
    \vspace{-10pt}
    \caption{Identity interpolation. 
    The outer columns show two identities (A and B), where intermediate columns are generated by linearly interpolating between their identity embeddings and applying the mixed embedding to all tokens.
    }
    \vspace{-16pt}
    \label{fig:global_blend}

\ifsigg
\end{figure}
\else
\end{figure}
\fi

\begin{figure*}
    \centering
    \ifarxiv
        \vspace{-22pt}
    \else
        \vspace{-5pt}
    \fi
    \includegraphics[width=0.85\linewidth]{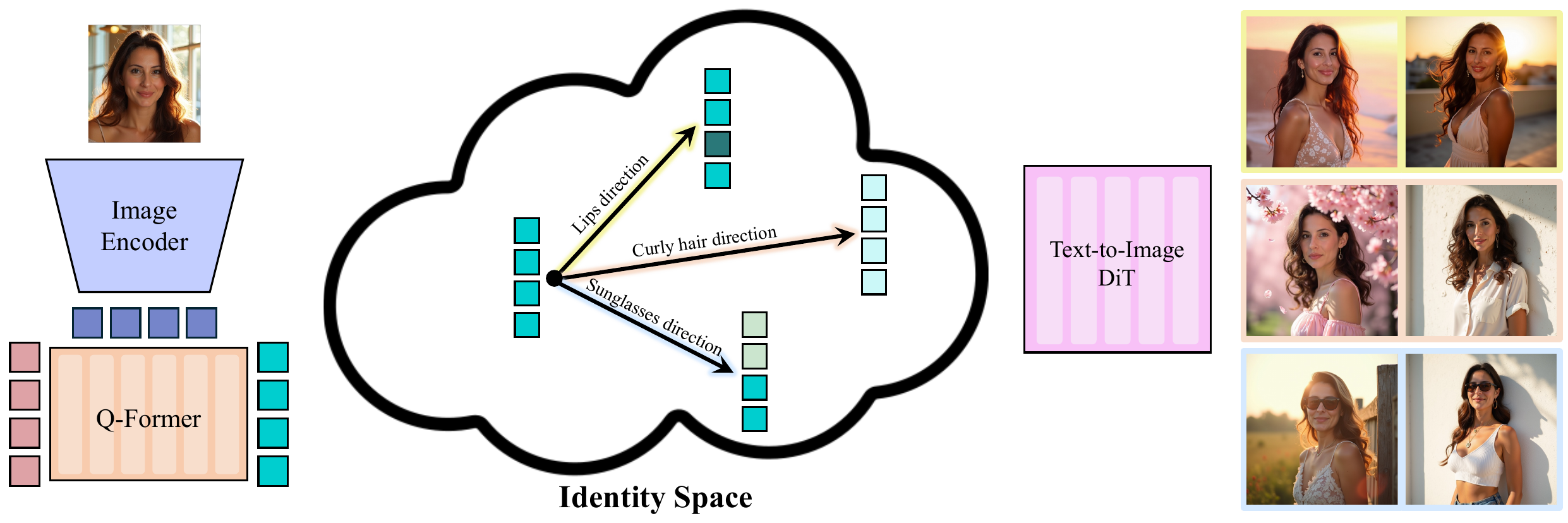}
    \ifarxiv
        \vspace{-8pt}
    \else
        \vspace{-12pt}
    \fi
    \caption{Overview of our identity tuning framework. An input image is encoded into a set of identity tokens, each capturing distinct aspects of the face. These tokens reside in an identity space, where meaningful directions (e.g., for eyes, age, or beard) can be discovered and traversed to achieve targeted edits. The modified identity tokens are then passed to a text-to-image model, which generates images of the edited identity in diverse contexts.}
    \label{fig:overview}
    \ifsigg
    \else
        \vspace{-17pt}
    \fi
\end{figure*}

\vspace{-3pt}
\section{Identity Tuning}
\vspace{-4pt}

In this section, we analyze the identity token space and introduce several methods for tuning the identity representation $Z \in \mathbb{R}^{N \times C}$. We observe that the manipulation of the identity representation can be applied to a single token $Z_n \in \mathbb{R}^{1 \times C}$, to all identity tokens $Z$, or to a selected subset of them $Z_S \in \mathbb{R}^{|S| \times C}$, where $S \subseteq \{1, ..., N\}$. To identify the subset of tokens $S$ most relevant to a given target attribute, we propose a dedicated token selection method. We then present several latent identity tuning techniques and demonstrate each of them using different token selections. 
An overview of our approach is shown in \cref{fig:overview}.

\subsection{\textbf{Global Identity Interpolation}}
\vspace{-4pt}
We begin by exploring controlled interpolation between two identities. Let $Z^{A}$ and $Z^{B}$ denote the identity representations of two subjects, and let $z^{A}$ and $z^{B}$ be the corresponding variables to be interpolated. We construct an interpolated representation:
\vspace{-5pt}
\textbf{\begin{equation}
z^{blend} = (1-\beta)\,z^{A} + \beta\,z^{B},
\label{eq:interpolation}
\end{equation}}
where $0 < \beta < 1$.
Setting $z = Z$ (global interpolation) produces a smooth transition of the identity from subject $A$ toward subject $B$, as illustrated in
\cref{fig:global_blend}.
Since the personalization encoder is not fully disentangled between identity and image composition, changes in layout or pose may appear across the endpoints.
However, our goal often requires fine-grained control over specific facial parts, such as interpolating only the eyes, rather than the entire face.

\ifarxiv
    \vspace{-2pt}
\else
    \vspace{-5pt}
\fi
\subsection{Token Selection}
\vspace{-4pt}
To isolate specific facial regions for editing, we must identify the relevant subset of tokens $S$. A straightforward approach to tuning the identity representation 
$Z$ is to treat it as a single high-dimensional vector, $\mathrm{vec}(Z) \in \mathbb{R}^{NC}$, and perform edits directly in this space, as demonstrated in the previous section with global interpolation. However, as shown earlier in
\cref{fig:attn_tokens},
individual tokens typically correspond to semantically localized regions in the image. Therefore, as we will demonstrate later, treating all tokens as a single vector and manipulating them jointly is generally effective for global identity edits, but not for localized edits.
Another option is to manipulate individual tokens $Z_n \in \mathbb{R}^{1 \times C}$, which can be regarded as vectors in $\mathbb{R}^{C}$. This results in highly localized changes that, as we will show, are often too limited for most tuning objectives.
The third option we explore is to manipulate a subset of tokens, $S \subseteq \{1, \ldots, N\}$, represented as $\mathrm{vec}(Z_S) \in \mathbb{R}^{|S|C}$. Next, we describe two methods for identifying $S$ given a target facial region to edit.

\ifarxiv
    \vspace{-10pt}
\else
    \vspace{-12pt}
\fi
\paragraph{Localized Attributes}
In the first method, we construct paired images that differ primarily in one facial part, and then measure the differences in their corresponding identity representations. We construct these pairs by applying simple, pixel-based localized image manipulations.
Specifically, given a facial part we aim to edit 
(e.g., lips) and a face image $I$, we crudely paste a patch (e.g., lips from another image) onto the corresponding region, resulting in a modified image $I_{\text{patch}}$. Examples of $(I, I_{\text{patch}})$ pairs are shown in 
\cref{fig:patch_editing_figure}.
For each pair $(I, I_{\text{patch}})$ that differs in a specific region, we compute the difference:
$
\Delta Z = Z(I_{\text{patch}}) - Z(I) \in \mathbb{R}^{N \times C}.
$
We quantify the contribution of each token $n$ to the given attribute by the magnitude of its difference $\|\Delta Z_n\|_2$, and construct a subset $S$ by selecting the top $k$ tokens for that pair. By aggregating just a small number of pairs (e.g., five) corresponding to the same facial part, we obtain a stable subset $S$ containing the tokens that most frequently appear among the top $k$ across all pairs. This process is performed once per facial region and is reused for different edits of that region across all subjects (the same subset $S$ is applied to any new identity). By directly manipulating specific facial regions, we enable control over parts of the face that are difficult to specify through text.

\ifsigg
\begin{figure}
\else
\begin{figure}
\fi
    \centering
    \ifarxiv
        \vspace{-7pt}
    \else
        \vspace{-10pt}
    \fi
    \scriptsize
    \setlength{\tabcolsep}{0pt}
        {
        \begin{tabular}{cc c cc}

        \includegraphics[width=0.24\linewidth]{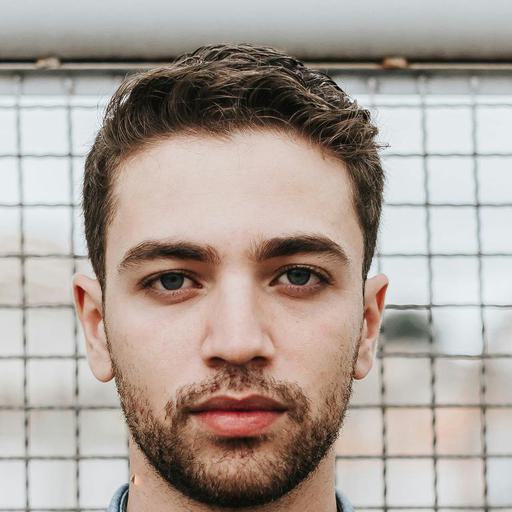} &
        \includegraphics[width=0.24\linewidth]{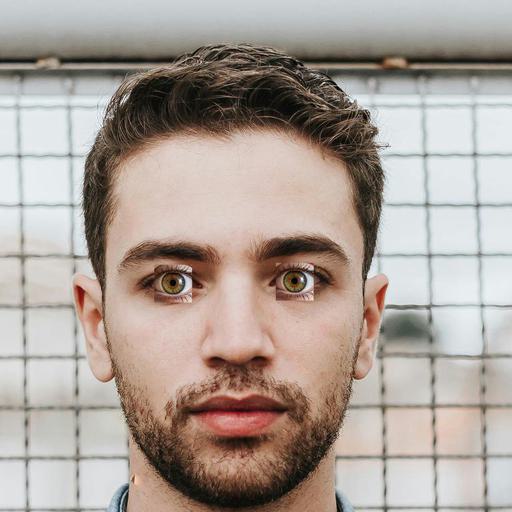} & { } &
        \includegraphics[width=0.24\linewidth]{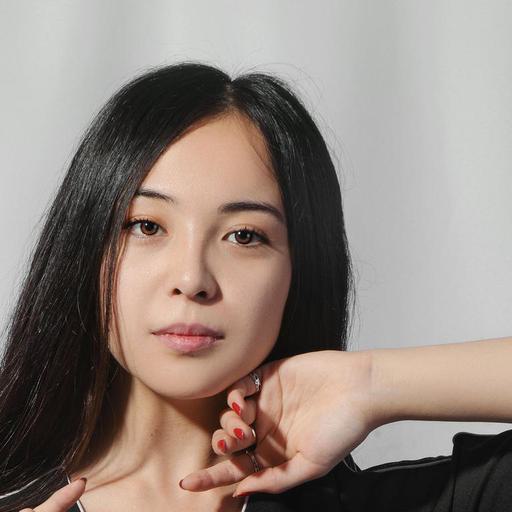} &
        \includegraphics[width=0.24\linewidth]{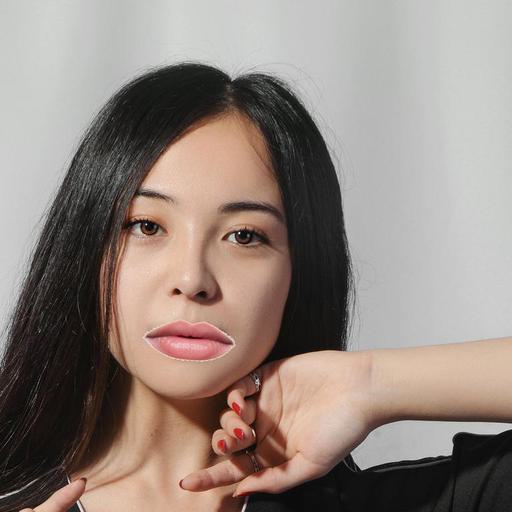} \\

        Original Image &
        Patched Eyes& { } &
        Original Image &
        Patched Lips

    \end{tabular}
    }
    \vspace{-7pt}
    \caption{
    We form $I_\text{patched}$ by pasting a new facial region onto $I$. Comparing their encoded identity representations reveals the tokens capturing the patched region's influence.
    }
    \label{fig:patch_editing_figure}
    \ifarxiv
        \vspace{-15pt}
    \else
        \vspace{-20pt}
    \fi
\ifsigg
\end{figure}
\else
\end{figure}
\fi

\begin{figure*} [t]
    \ifarxiv
        \vspace{-10pt}
    \else
        \vspace{-5pt}
    \fi
    \centering
    \setlength{\tabcolsep}{0.5pt}
    \renewcommand{\arraystretch}{0.3}
    \addtolength{\belowcaptionskip}{-5pt}
    {

    \hspace{-0.2cm}
    \begin{minipage}{0.5\textwidth}
        \centering
        \begin{tabular}{c c c c}

        \includegraphics[width=0.24\textwidth]{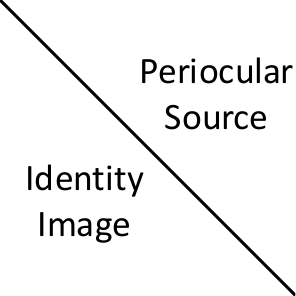} &
        \includegraphics[width=0.24\textwidth]{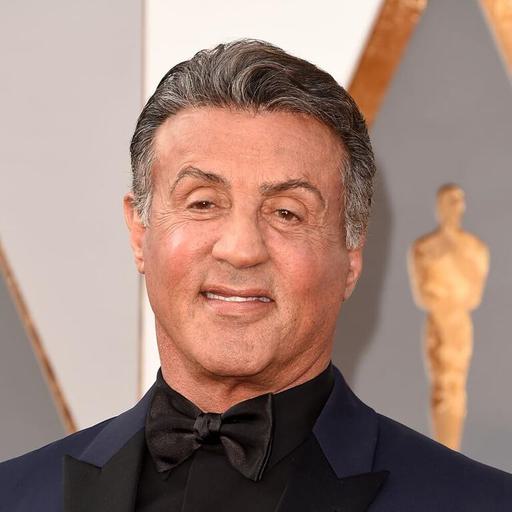} &
        \includegraphics[width=0.24\textwidth]{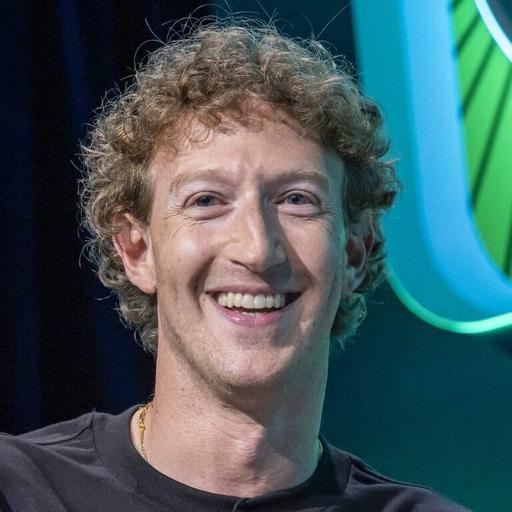} &
        \includegraphics[width=0.24\textwidth]{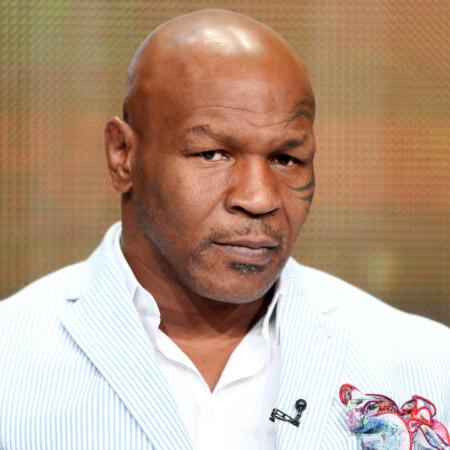} \\

        \includegraphics[width=0.24\textwidth]{images/blend/stallone001.jpg} &
        \includegraphics[width=0.24\textwidth]{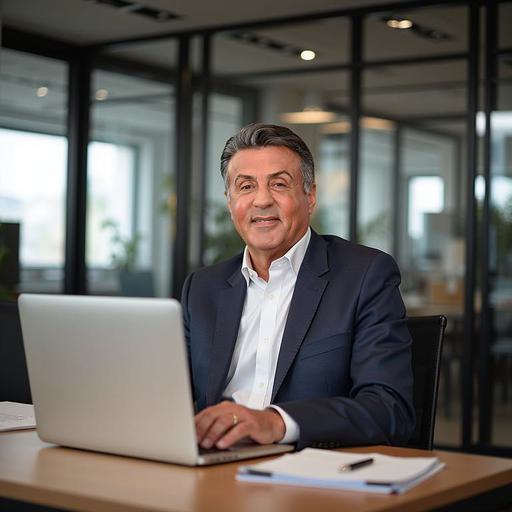} &
        \includegraphics[width=0.24\textwidth]{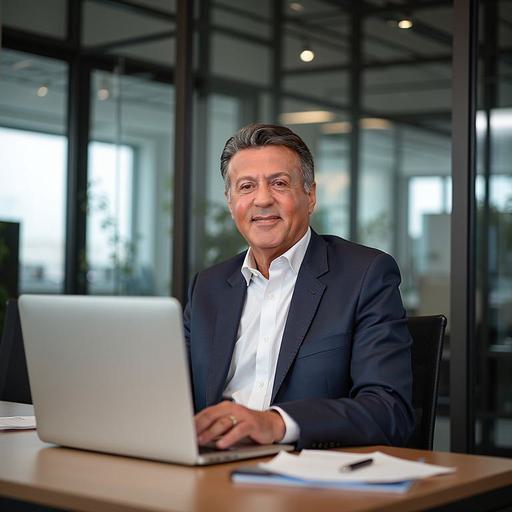} &
        \includegraphics[width=0.24\textwidth]{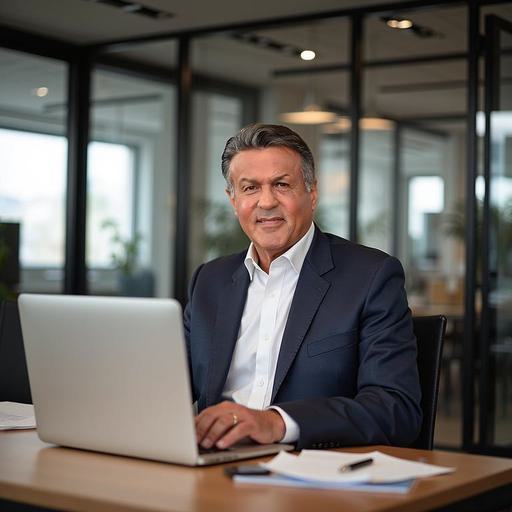} \\

        \includegraphics[width=0.24\textwidth]{images/blend/zuckerberg_man_001.jpg} &
        \includegraphics[width=0.24\textwidth]{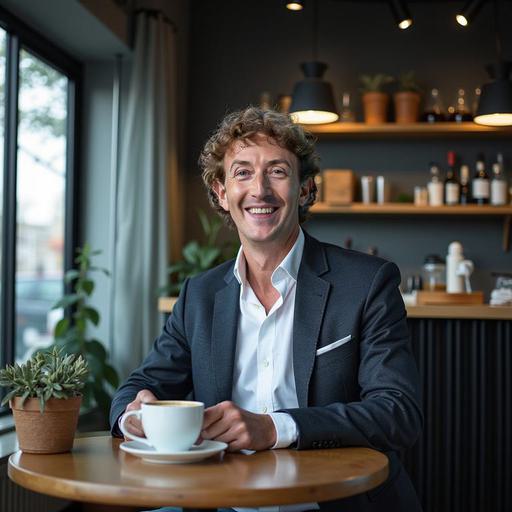} &
        \includegraphics[width=0.24\textwidth]{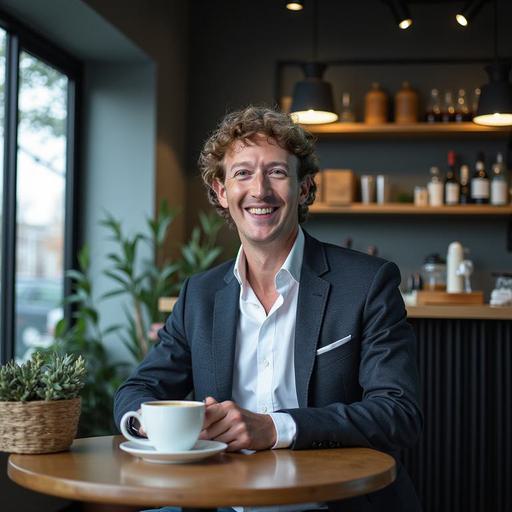} &
        \includegraphics[width=0.24\textwidth]{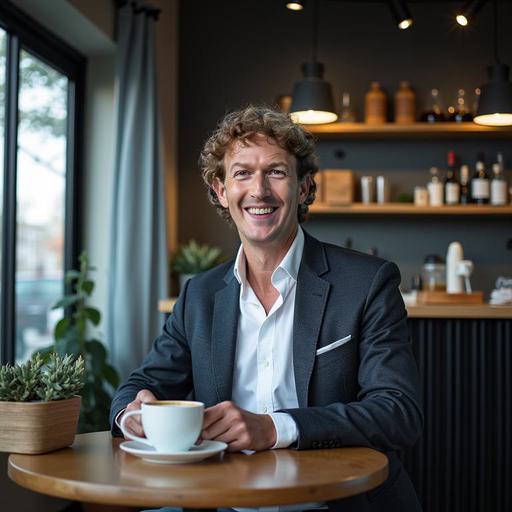} \\
        
        \includegraphics[width=0.24\textwidth]{images/blend/tyson01.jpg} &
        \includegraphics[width=0.24\textwidth]{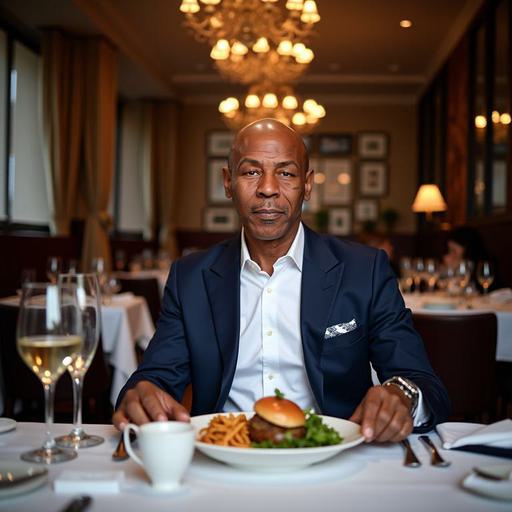} &
        \includegraphics[width=0.24\textwidth]{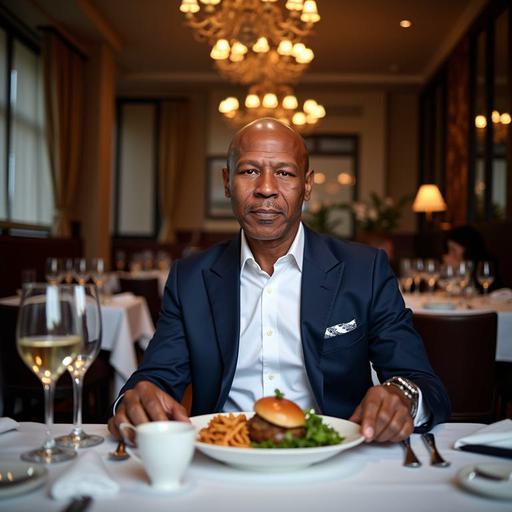} &
        \includegraphics[width=0.24\textwidth]{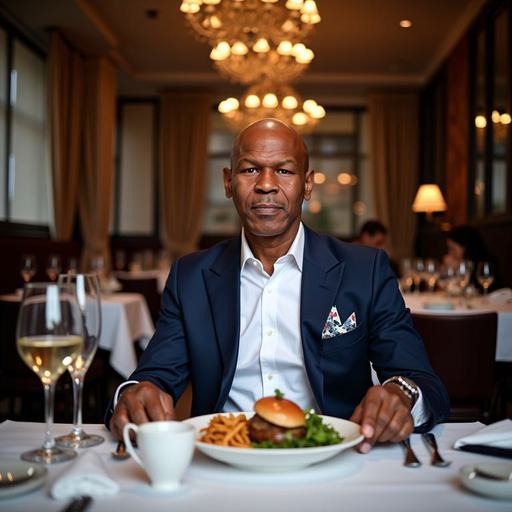} \\
        
        \end{tabular}
        
    \end{minipage}%
    \begin{minipage}{0.5\textwidth}
        \centering
        \begin{tabular}{c c c c}

        \includegraphics[width=0.24\textwidth]{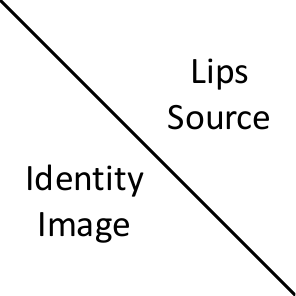} &
        \includegraphics[width=0.24\textwidth]{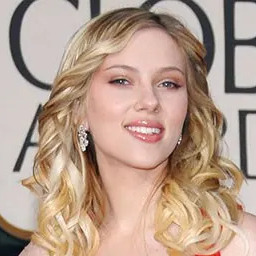} &
        \includegraphics[width=0.24\textwidth]{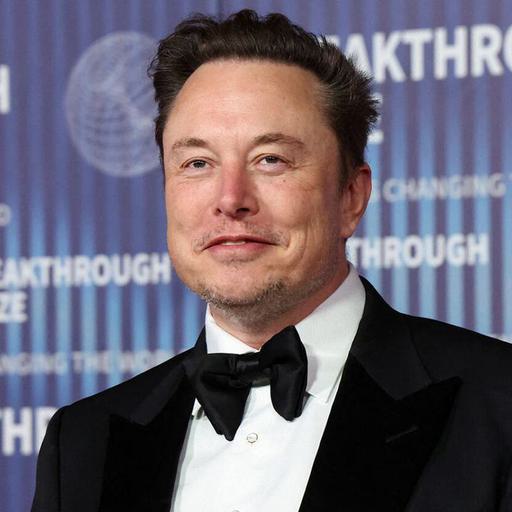} &
        \includegraphics[width=0.24\textwidth]{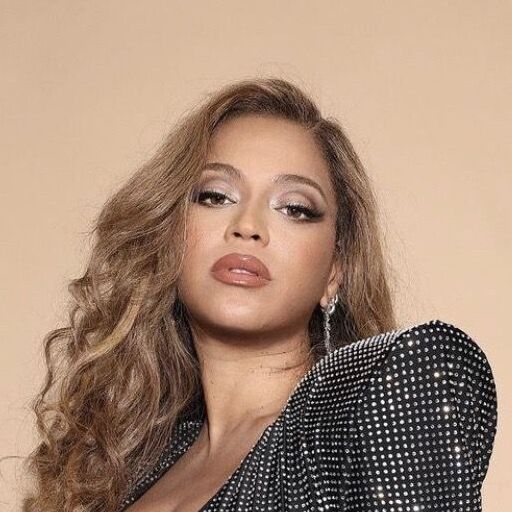} \\

        \includegraphics[width=0.24\textwidth]{images/blend/Scarlett-Johansson_00.woman.jpg} &
        \includegraphics[width=0.24\textwidth]{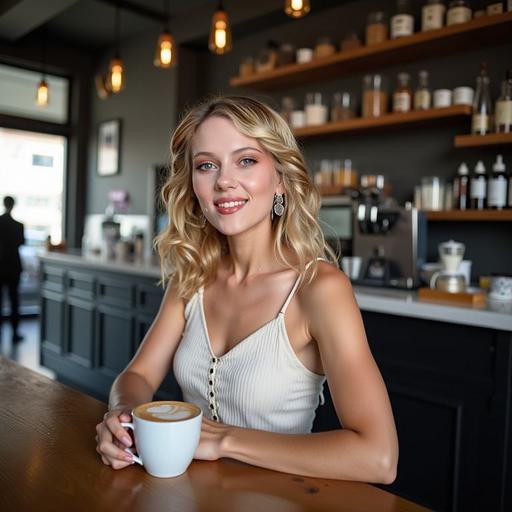} &
        \includegraphics[width=0.24\textwidth]{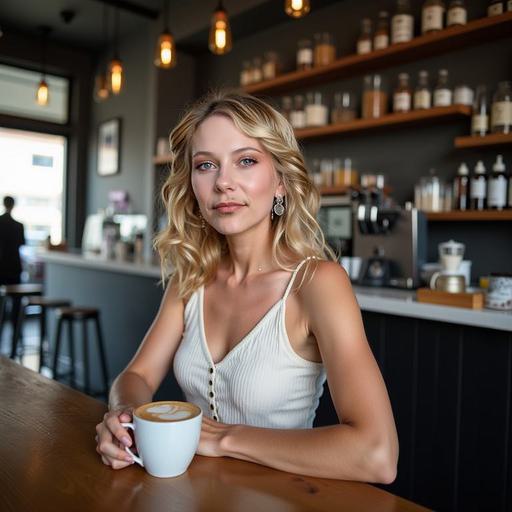} &
        \includegraphics[width=0.24\textwidth]{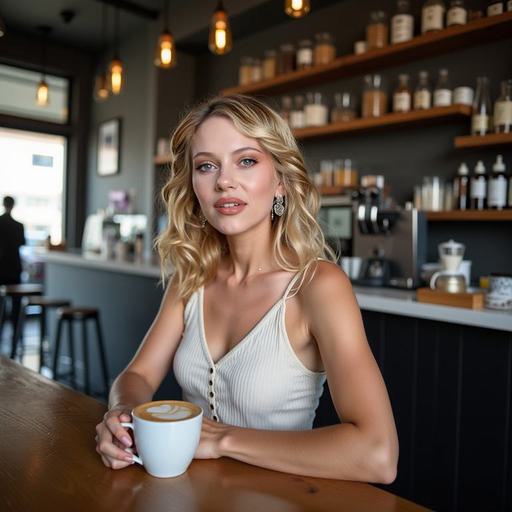} \\

        \includegraphics[width=0.24\textwidth]{images/blend/elon_man_04.jpeg} &
        \includegraphics[width=0.24\textwidth]{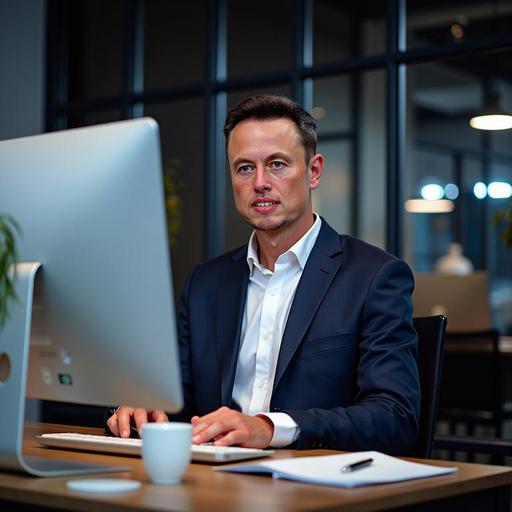} &
        \includegraphics[width=0.24\textwidth]{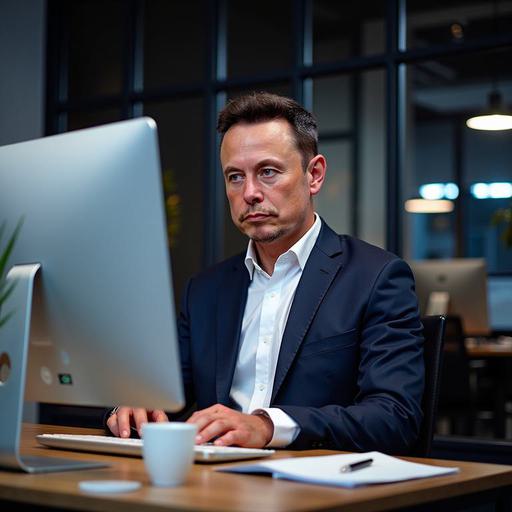} &
        \includegraphics[width=0.24\textwidth]{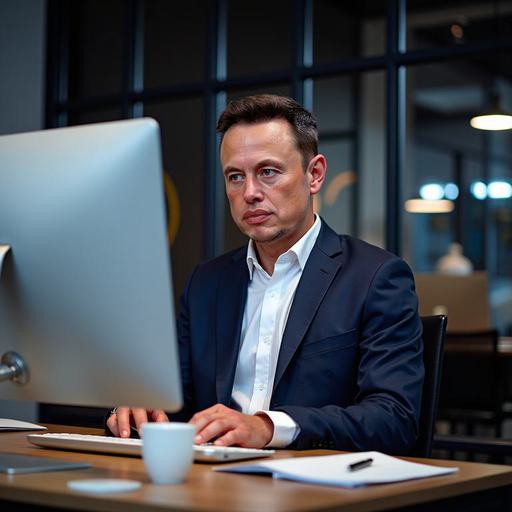} \\
        
        \includegraphics[width=0.24\textwidth]{images/blend/beyonce_woman_03.jpg} &
        \includegraphics[width=0.24\textwidth]{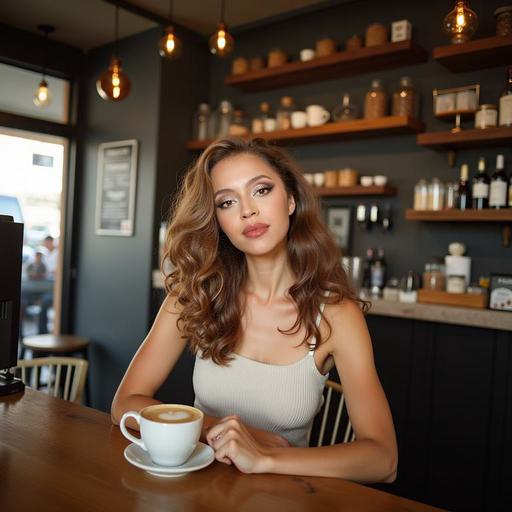} &
        \includegraphics[width=0.24\textwidth]{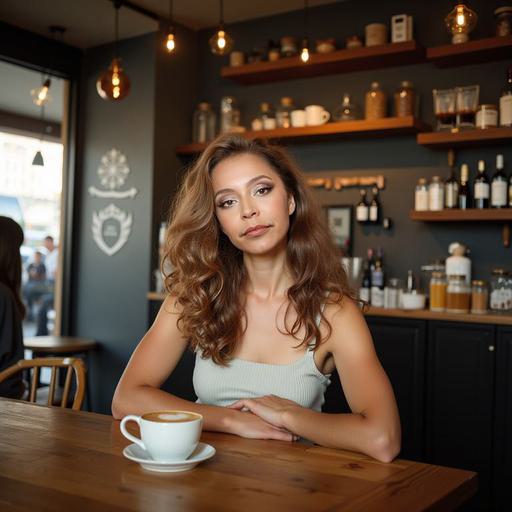} &
        \includegraphics[width=0.24\textwidth]{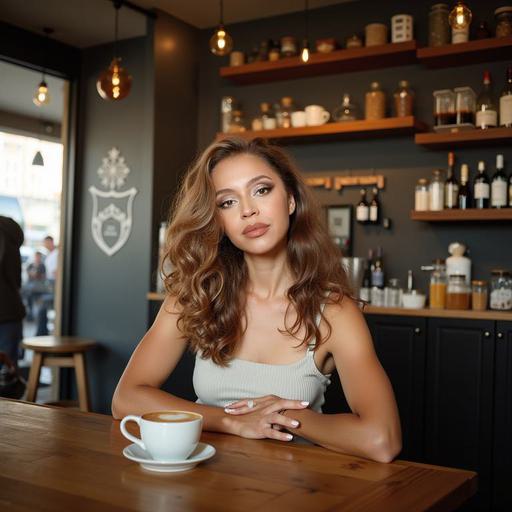} \\
        
        \end{tabular}
        
    \end{minipage}
    }
    \vspace{-10pt}
    \caption{
    Blending identities by part-level transfer. Our approach enables the transfer of a chosen facial component from a source identity to a target. In each grid, the target identity on the left receives a specific part from the source shown at the top. The left grid demonstrates periocular (eyelids+eyebrows) transfer, and the right grid demonstrates lip transfer. All other attributes and the background are preserved.
    }
    \label{fig:blending_results_grid}
    \ifarxiv
        \vspace{-5pt}
    \else
        \vspace{-14pt}
    \fi
\end{figure*}

\ifarxiv
    \vspace{-10pt}
\else
    \vspace{-13pt}
\fi
\paragraph{Global Attributes}
Beyond tokens that attend to localized facial regions, some tokens encode global appearance attributes (e.g., skin tone, freckles) that span multiple regions. These tokens cannot be isolated with localized patch edits, as modifying such attributes affects large portions of the face.
To identify such tokens, we leverage labeled data for specific global attributes. Given a global attribute $a$ and a labeled dataset $\{(I_i, y_i^a)\}_{i=1}^M$, we encode each image to the identity space $Z(I_i)\in\mathbb{R}^{N\times C}$. 
For each token index $n\in\{1,\ldots,N\}$, we train a linear SVM $f_{n,a}$ on the per-token features $\{Z_n(I_i)\}$ to predict the attribute label $y_i^a$. Let $m_{n,a}$ denote the validation accuracy of $f_{n,a}$. We then select the subset of token indices whose classifiers achieve validation accuracy above a threshold $\tau$:
$
A_a=\{\, n \mid m_{n,a}\ge \tau \,\}.
$
Intuitively, if a classifier trained on a single token reliably predicts attribute $a$, that token encodes information about $a$ and can thus be used to control it. This selection is performed once per attribute and reused for downstream edits.

\ifarxiv
    \vspace{-2pt}
\else
    \vspace{-7pt}
\fi
\subsection{\textbf{Localized Identity Transfer}}
\vspace{-5pt}
Having defined a method to isolate relevant tokens, we can now perform localized identity manipulations.
Setting $z = Z_S$ in
\cref{eq:interpolation}
(region transfer) localizes the transition to the facial part corresponding to subset $S$. We achieve this by replacing the tokens in $Z^{A}$ corresponding to the desired facial region with $z^{\text{blend}}$. This keeps the subject aligned with $A$ outside the selected region while gradually adopting the corresponding trait from $B$ within it.
\cref{fig:blending_results_grid}
presents results of transplanting the periocular region (eyebrows and eyelids) from different identities.

\begin{figure*}[t] 
\centering
\setlength{\tabcolsep}{1pt} %
\setlength\arrayrulewidth{0.8pt} %
\setlength\dashlinedash{1.6pt}   %
\setlength\dashlinegap{0.8pt}    %
\renewcommand{\arraystretch}{1.2} %
\scriptsize{
\begin{tabular}{c : c c c c}

{\setlength{\tabcolsep}{1pt}%
 \renewcommand{\arraystretch}{1.2}%
 \begin{tabular}{@{}c@{}}
   $\vcenter{\hbox{\includegraphics[width=0.15\linewidth]{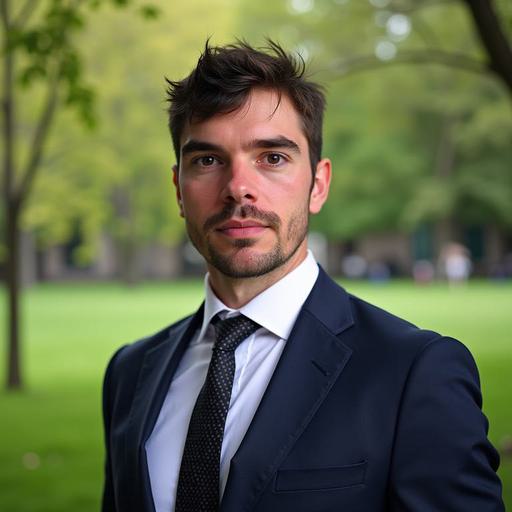}}}%
   $
 \end{tabular}
} & \hspace{1pt} &

{\setlength{\tabcolsep}{1pt}%
 \renewcommand{\arraystretch}{1.2}%
 \begin{tabular}{@{}c@{}}
   $\vcenter{\hbox{\includegraphics[width=0.15\linewidth]{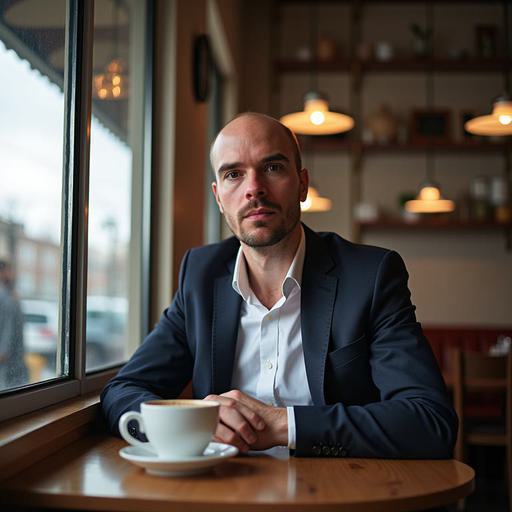}}}%
   \hspace{-2.25pt}%
   \vcenter{\hbox{
   \renewcommand{\arraystretch}{0}%
       \begin{tabular}{@{}c@{}}
           \includegraphics[width=0.075\linewidth]{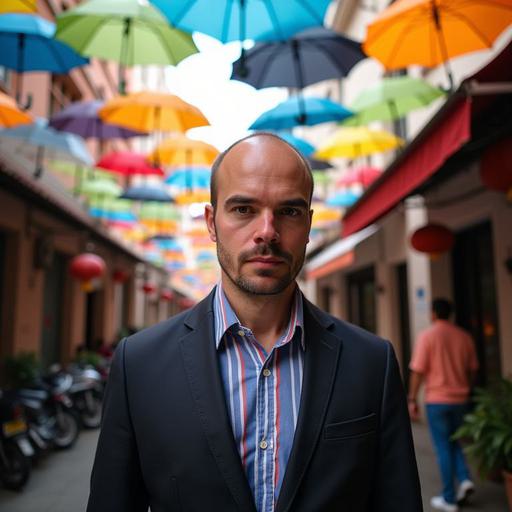} \\
           \includegraphics[width=0.075\linewidth]{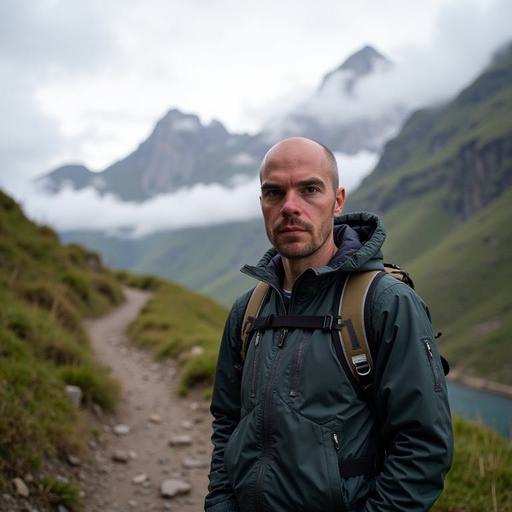}
       \end{tabular}
   }}
   $
 \end{tabular}
} &

{\setlength{\tabcolsep}{1pt}%
 \renewcommand{\arraystretch}{1.2}%
 \begin{tabular}{@{}c@{}}
   $\vcenter{\hbox{\includegraphics[width=0.15\linewidth]{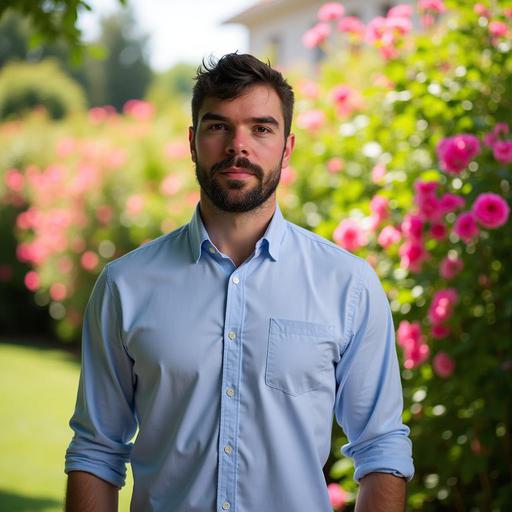}}}%
   \hspace{-2.25pt}%
   \vcenter{
   \hbox{
   \renewcommand{\arraystretch}{0}%
       \begin{tabular}{@{}c@{}}
           \includegraphics[width=0.075\linewidth]{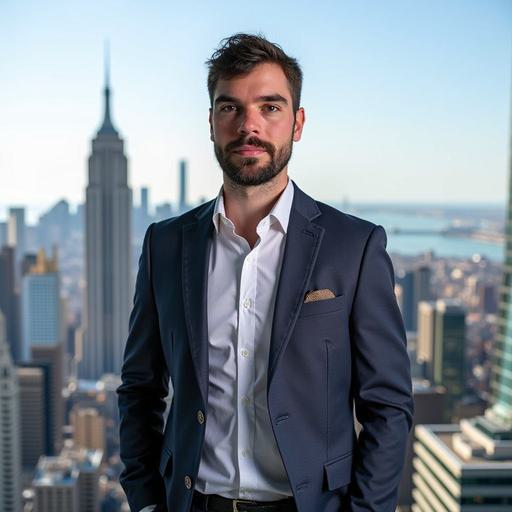} \\
           \includegraphics[width=0.075\linewidth]{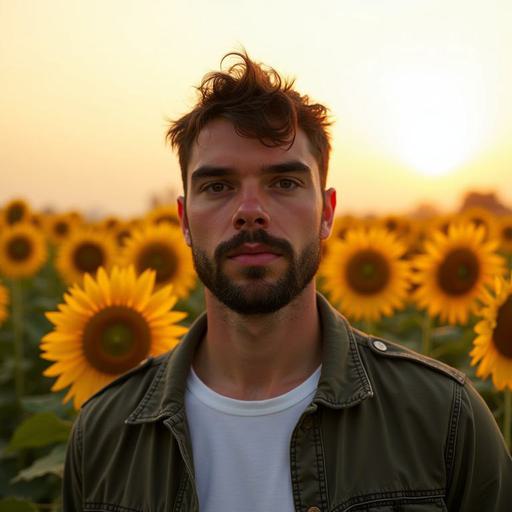}
       \end{tabular}
   }}
   $
 \end{tabular}
} &

{\setlength{\tabcolsep}{1pt}%
 \renewcommand{\arraystretch}{1.2}%
 \begin{tabular}{@{}c@{}}
   $\vcenter{\hbox{\includegraphics[width=0.15\linewidth]{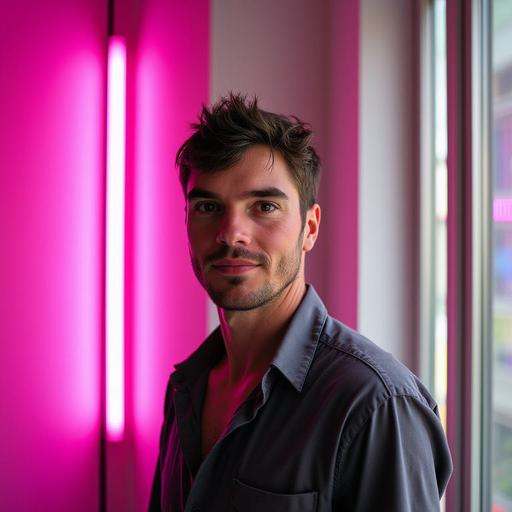}}}%
   \hspace{-2.25pt}%
   \vcenter{\hbox{
   \renewcommand{\arraystretch}{0}%
       \begin{tabular}{@{}c@{}}
           \includegraphics[width=0.075\linewidth]{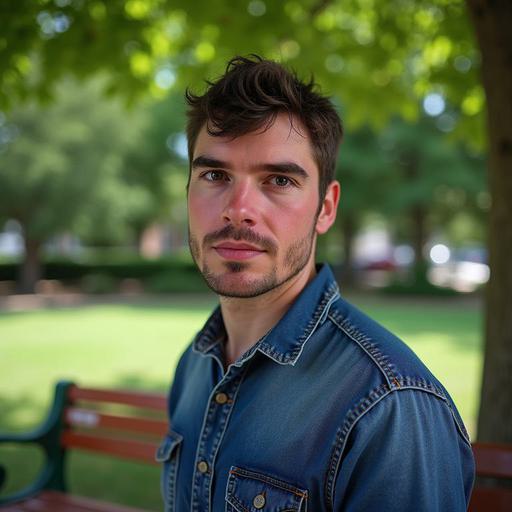} \\
           \includegraphics[width=0.075\linewidth]{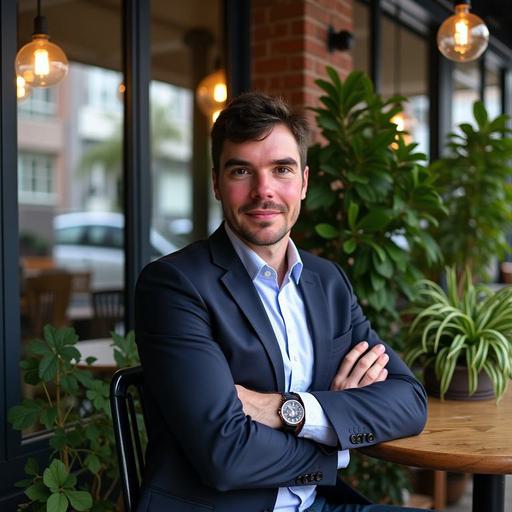}
       \end{tabular}
   }}
   $
 \end{tabular}
} \\ \\[-8pt]

{\setlength{\tabcolsep}{1pt}%
 \renewcommand{\arraystretch}{1.2}%
 \begin{tabular}{@{}c@{}}
   $\vcenter{\hbox{\includegraphics[width=0.15\linewidth]{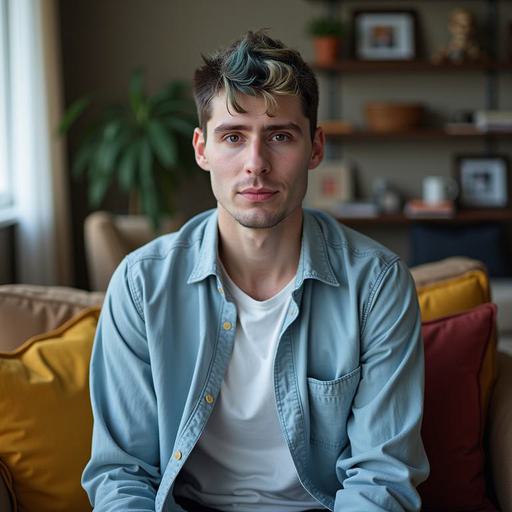}}}%
   $
 \end{tabular}
} &&

{\setlength{\tabcolsep}{1pt}%
 \renewcommand{\arraystretch}{1.2}%
 \begin{tabular}{@{}c@{}}
   $\vcenter{\hbox{\includegraphics[width=0.15\linewidth]{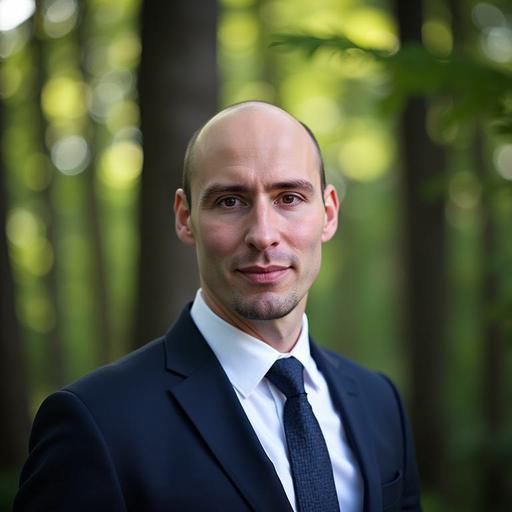}}}%
   \hspace{-2.25pt}%
   \vcenter{\hbox{
   \renewcommand{\arraystretch}{0}%
       \begin{tabular}{@{}c@{}}
           \includegraphics[width=0.075\linewidth]{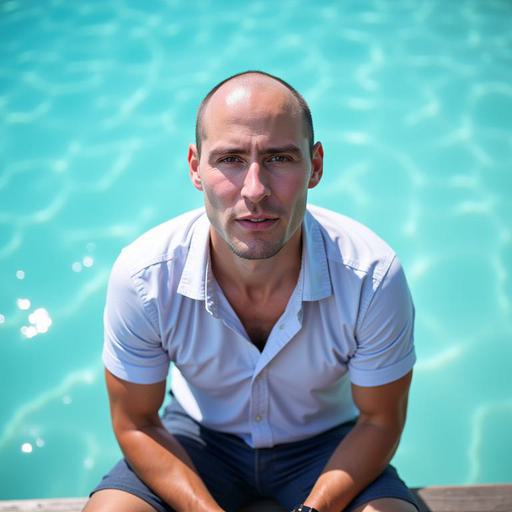} \\
           \includegraphics[width=0.075\linewidth]{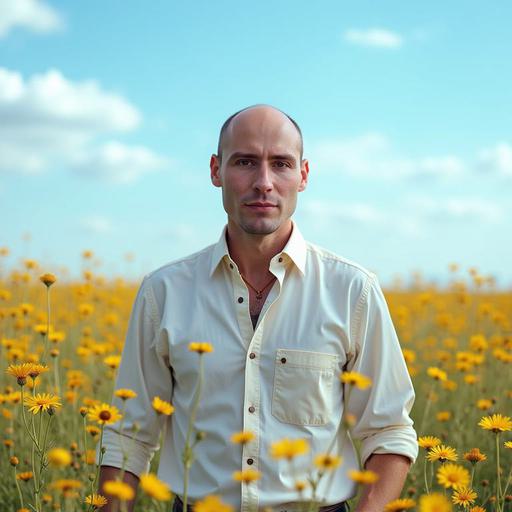}
       \end{tabular}
   }}
   $
 \end{tabular}
} &

{\setlength{\tabcolsep}{1pt}%
 \renewcommand{\arraystretch}{1.2}%
 \begin{tabular}{@{}c@{}}
   $\vcenter{\hbox{\includegraphics[width=0.15\linewidth]{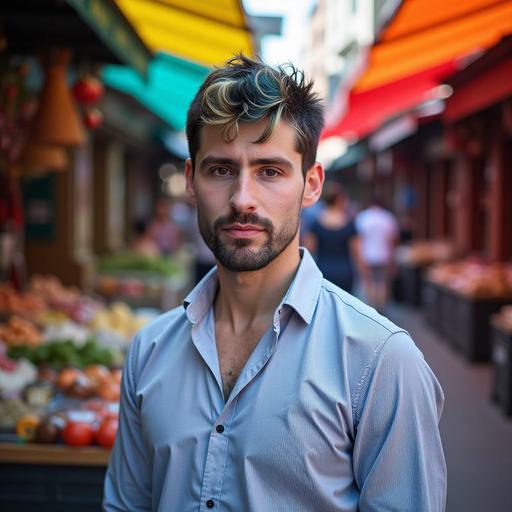}}}%
   \hspace{-2.25pt}%
   \vcenter{\hbox{
   \renewcommand{\arraystretch}{0}%
       \begin{tabular}{@{}c@{}}
           \includegraphics[width=0.075\linewidth]{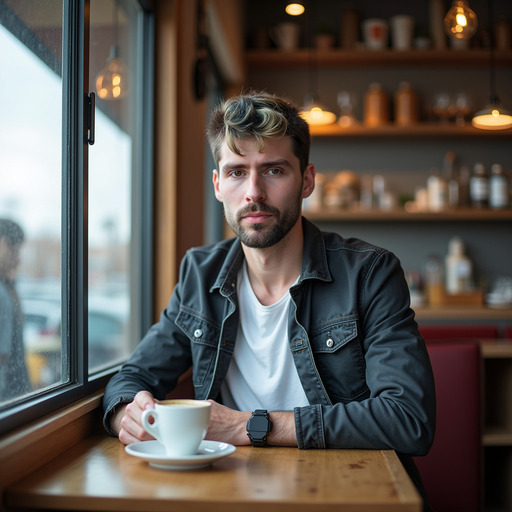} \\
           \includegraphics[width=0.075\linewidth]{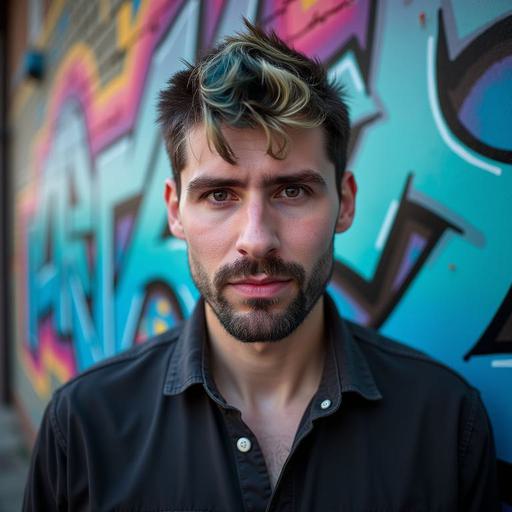}
       \end{tabular}
   }}
   $
 \end{tabular}
} &

{\setlength{\tabcolsep}{1pt}%
 \renewcommand{\arraystretch}{1.2}%
 \begin{tabular}{@{}c@{}}
   $\vcenter{\hbox{\includegraphics[width=0.15\linewidth]{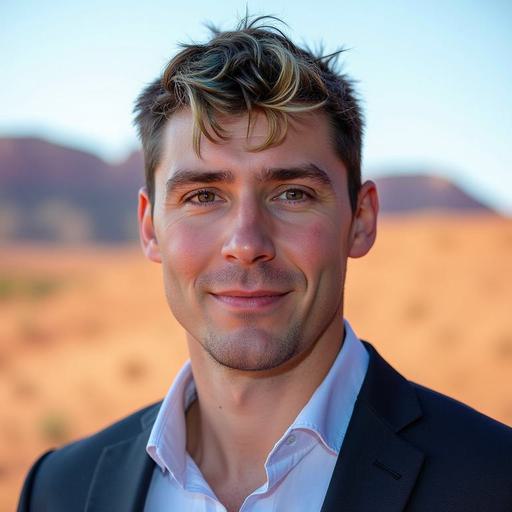}}}%
   \hspace{-2.25pt}%
   \vcenter{\hbox{
   \renewcommand{\arraystretch}{0}%
       \begin{tabular}{@{}c@{}}
           \includegraphics[width=0.075\linewidth]{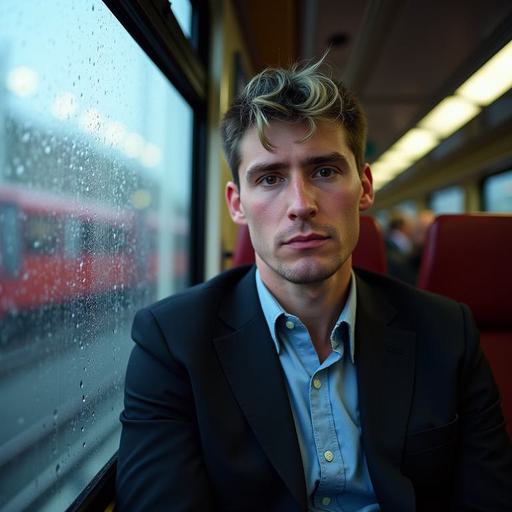} \\
           \includegraphics[width=0.075\linewidth]{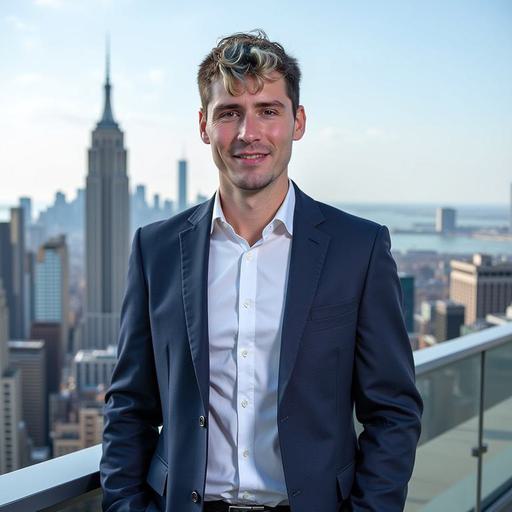}
       \end{tabular}
   }}
   $
 \end{tabular}
} \\ \\[-8pt]

Original Identity && Bald & Beard & Rosy Cheeks \\

\end{tabular}
}
\vspace{-6pt}
\caption{
Each row shows a different source identity (leftmost column) and its edits along three directions: bald, beard, and rosy cheeks. The images in each column are generated by adding the same direction found by SVM. For each edit, we show the main result with two alternative prompts stacked vertically to the right.
}
\label{fig:directional_edits}
\vspace{-18pt}
\end{figure*}

\vspace{-2pt}
\subsection{\textbf{Linear Token Manipulation}}
\vspace{-2pt}
Beyond interpolation, we introduce a set of methods for manipulating the identity tokens via linear directions in the latent space. To apply these linear operations, we treat the target representation as a flattened vector. Given an image $I$, we denote by $z(I) \in \mathbb{R}^M$ its corresponding identity representation vector to be edited. Depending on the desired level of granularity, $z$ can represent the entire identity $\mathrm{vec}(Z)$ ($M=N\cdot C$), an individual token $Z_n$ ($M=C$), or a subset of tokens $\mathrm{vec}(Z_S)$ ($M=|S|\cdot C$).

\subsubsection{\textbf{Supervised Linear Directions}}
\label{subsec:supervised_method}
We first learn supervised directions associated with specific facial attributes. As before, we begin by extracting identity representations from the images in our dataset, resulting in the set $\{z(I)\}_{I \in \mathcal{D}}$. 
Given binary attribute labels $y(I)\in\{-1,+1\}$ (either from annotated datasets or predicted by an off-the-shelf classifier), we estimate a linear direction in the latent space that corresponds to the given attribute using two complementary ways.

First, we use a mean-difference estimator. 
We define $\mu_{+}$ and $\mu_{-}$ as the mean identity representations of the positive and negative samples,
$\mu_{+} = 1/{|\mathcal{D}_+|} \sum_{I \in \mathcal{D}_+} z(I), 
\mu_{-} = 1/{|\mathcal{D}_-|} \sum_{I \in \mathcal{D}_-} z(I),$
where $\mathcal{D}_+ = \{\, I \in \mathcal{D} \mid y(I) = 1 \,\}$ and $\mathcal{D}_-$ is defined similarly.
The editing direction is then: $d_\text{attr} = \mu_{+}-\mu_{-}$. 

Second, we
train a linear SVM on $\{z(I),y(I)\}$ to obtain a weight vector $w$ and bias $b$. The learned hyperplane $w^\top x + b = 0$ separates images with and without the attribute. This formulation is inspired by GAN-based latent space editing methods, where linear separators are learned for attributes and the hyperplane normal is used as an edit direction~\cite{shen2020interpreting}.
We use the unit-norm normal of this hyperplane as an edit direction in the token space,
$
v_{\text{svm}} \;=\; \frac{w}{\|w\|_2} \in \mathbb{R}^{M},
$
where $w$ is the hyperplane normal, and choose its sign so that a small positive step increases the classifier score on a held out set.

Given this direction and a chosen scale, we apply edits in a norm-preserving manner to approximately preserve distributional scale. Let $z \in \mathbb{R}^{M}$ be the vector to edit, $\bar z = \frac{1}{|\mathcal{D}|} \sum_{I \in \mathcal{D}} z(I)$ its dataset mean, and $v$ a unit-normalized direction. For a scale $\alpha \in \mathbb{R}$, we define:
\vspace{-3pt}
\begin{equation}
e = (z - \bar z) + \alpha \cdot v,
\qquad
z' = \bar z + e \cdot {|z - \bar z|_2}/{|e|_2}.
\label{eq:norm_update}
\end{equation}
This operation centers the vector, adds a scaled direction, and rescales to preserve the original distance from the mean.

\cref{fig:directional_edits}
presents qualitative results obtained using both supervised approaches, the mean-difference and SVM-based estimators. We observe that the SVM-based method is more effective for fine-grained edits, such as the rosy cheeks attribute, whereas the mean-difference estimator performs better for more pronounced modifications, such as bald or beard. This difference can be explained by the nature of the learned directions: for substantial changes, the SVM classifier may rely on a subset of highly discriminative dimensions rather than all dimensions relevant to the attribute, effectively learning shortcuts that do not generalize to strong global edits. In contrast, the mean-difference estimator captures a broader, more global direction but lacks the precision required for subtle, fine-grained attributes.

\subsubsection{Unsupervised Directions}
\label{subsec:pca_method}

In addition to supervised directions, we also explore unsupervised manipulation of identity tokens by identifying linear directions in the identity space. Traversing along such directions enables fine-grained control over the strength of the edit.
Following prior works that discover meaningful edit directions in GAN latent spaces using Principal Component Analysis (PCA)~\cite{harkonen2020ganspace, shen2020closedform}, we apply a similar approach to our identity representation.

Specifically, given a dataset $\mathcal{D}$ of images, we extract for each image $I \in \mathcal{D}$ its identity representation $z(I)$. We then perform PCA on the set of vectors $\{z(I)\}_{I \in \mathcal{D}}$, obtaining a set of principal directions.
Given a PCA direction $v$, we apply the same norm-preserving update described in
\cref{eq:norm_update}.

\begin{figure*}[t]
    \centering
    \vspace{-7pt}
    \ifarxiv
        \vspace{-18pt}
    \else
        \vspace{-7pt}
    \fi

    \setlength{\tabcolsep}{1pt}
    \setlength\arrayrulewidth{0.8pt} %
    \setlength\dashlinedash{1.6pt}   %
    \setlength\dashlinegap{0.8pt}    %

    \scriptsize{
    \begin{tabular}{c c@{}c c@{}c c@{}c}
        & \multicolumn{2}{c}{Per token PCA}   %
        & \multicolumn{2}{c}{Global PCA}      %
        & \multicolumn{2}{c}{Group PCA} \\     %

        {\setlength{\tabcolsep}{0pt}%
         \renewcommand{\arraystretch}{0}%
         \begin{tabular}{cc}
            \multicolumn{2}{c}{\includegraphics[width=0.171\linewidth]{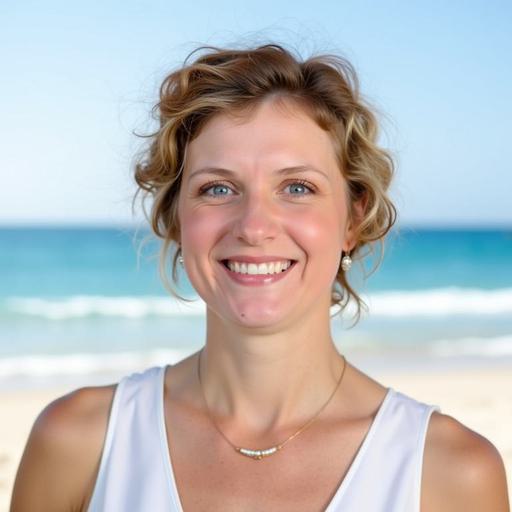}} \\
            &
         \end{tabular}
        } &
        {\setlength{\tabcolsep}{0pt}%
         \renewcommand{\arraystretch}{0}%
         \begin{tabular}{ccc}
            \multicolumn{3}{c}{\includegraphics[clip, viewport=128bp 192bp 896bp 960bp, width=0.128\linewidth]{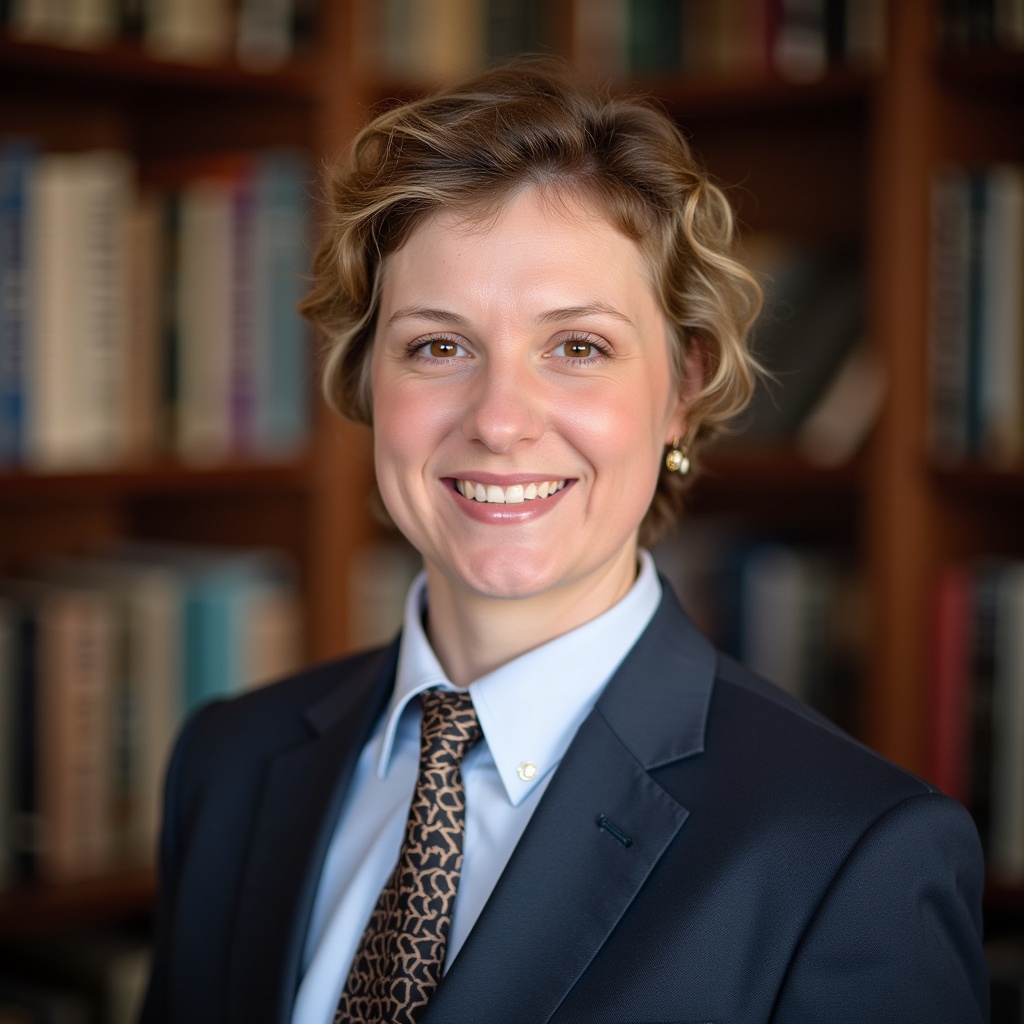}} \\
            \includegraphics[clip, viewport=128bp 192bp 896bp 960bp, width=0.043\linewidth]{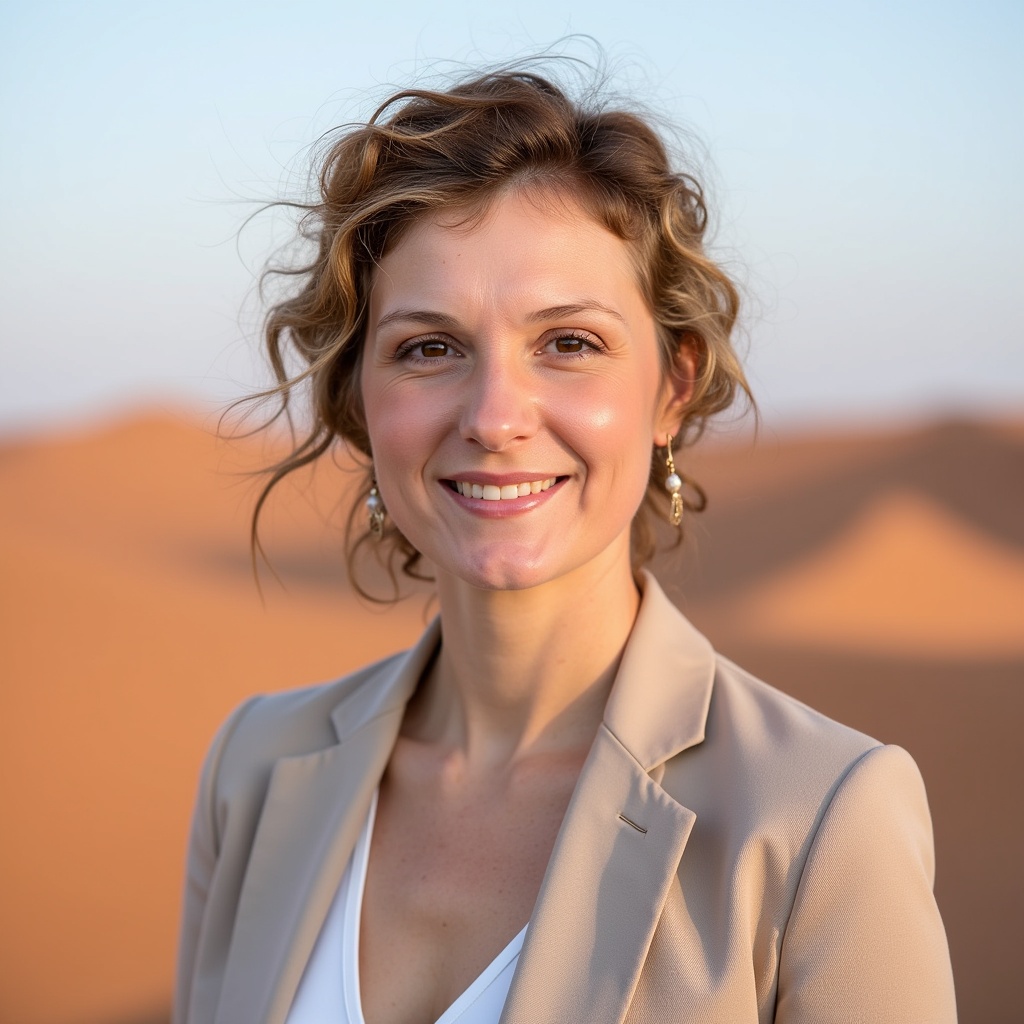} &
            \includegraphics[clip, viewport=128bp 192bp 896bp 960bp, width=0.043\linewidth]{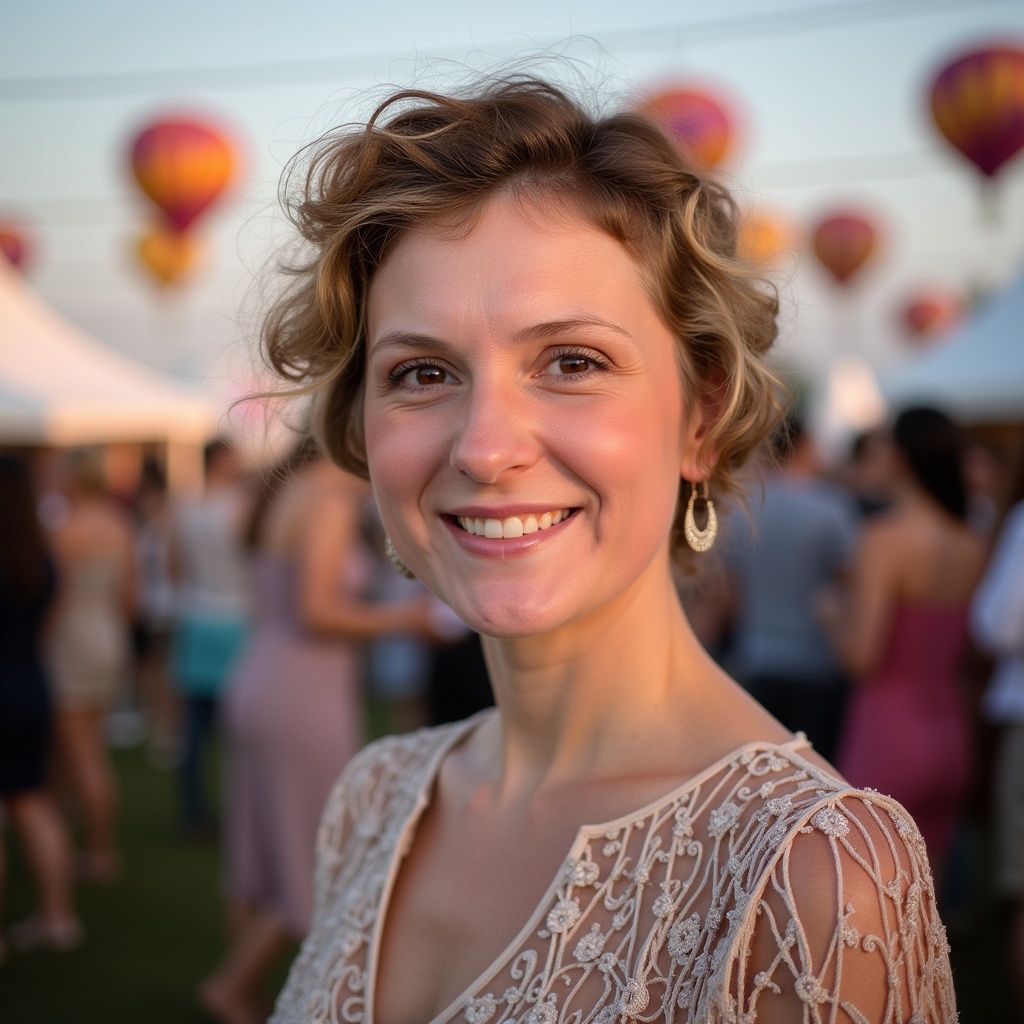} &
            \includegraphics[clip, viewport=128bp 192bp 896bp 960bp, width=0.043\linewidth]{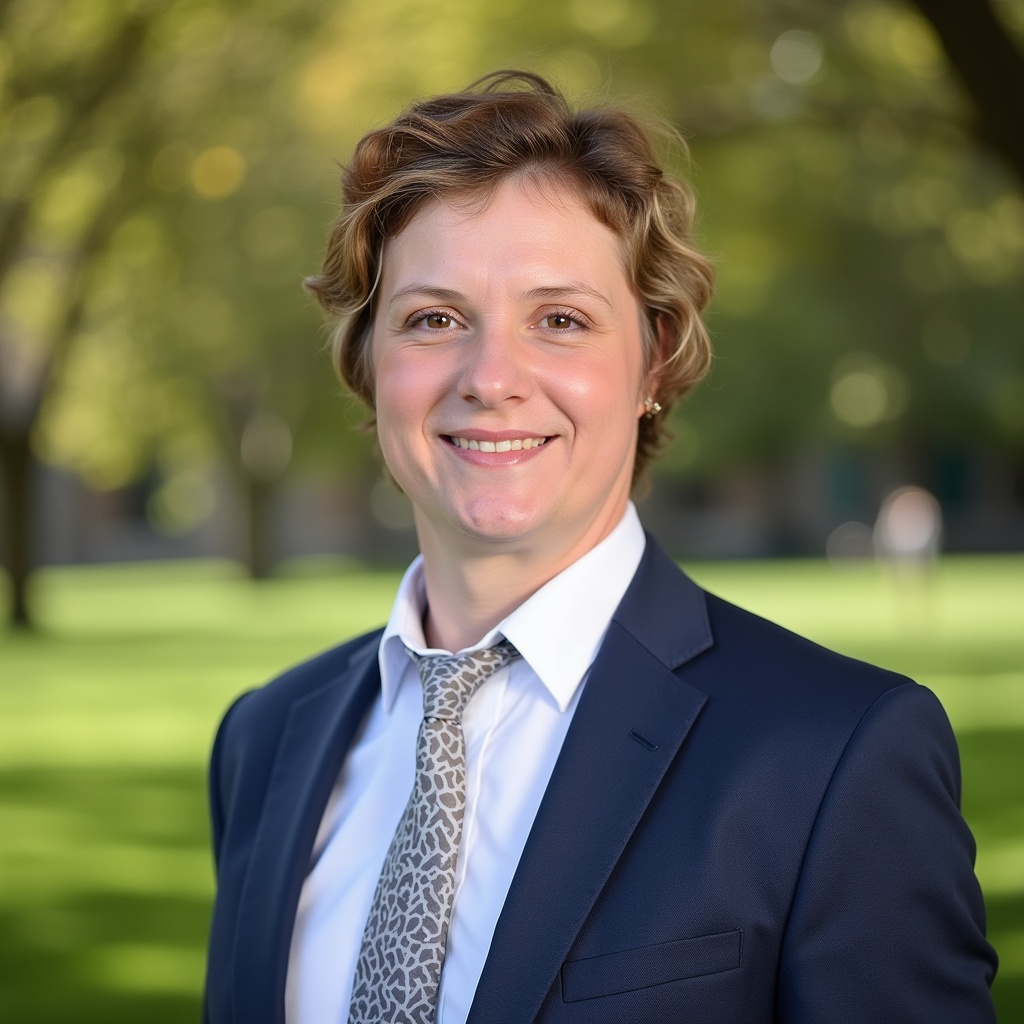}
         \end{tabular}
        } &
        {\setlength{\tabcolsep}{0pt}%
         \renewcommand{\arraystretch}{0}%
         \begin{tabular}{ccc}
            \multicolumn{3}{c}{\includegraphics[clip, viewport=128bp 192bp 896bp 960bp, width=0.128\linewidth]{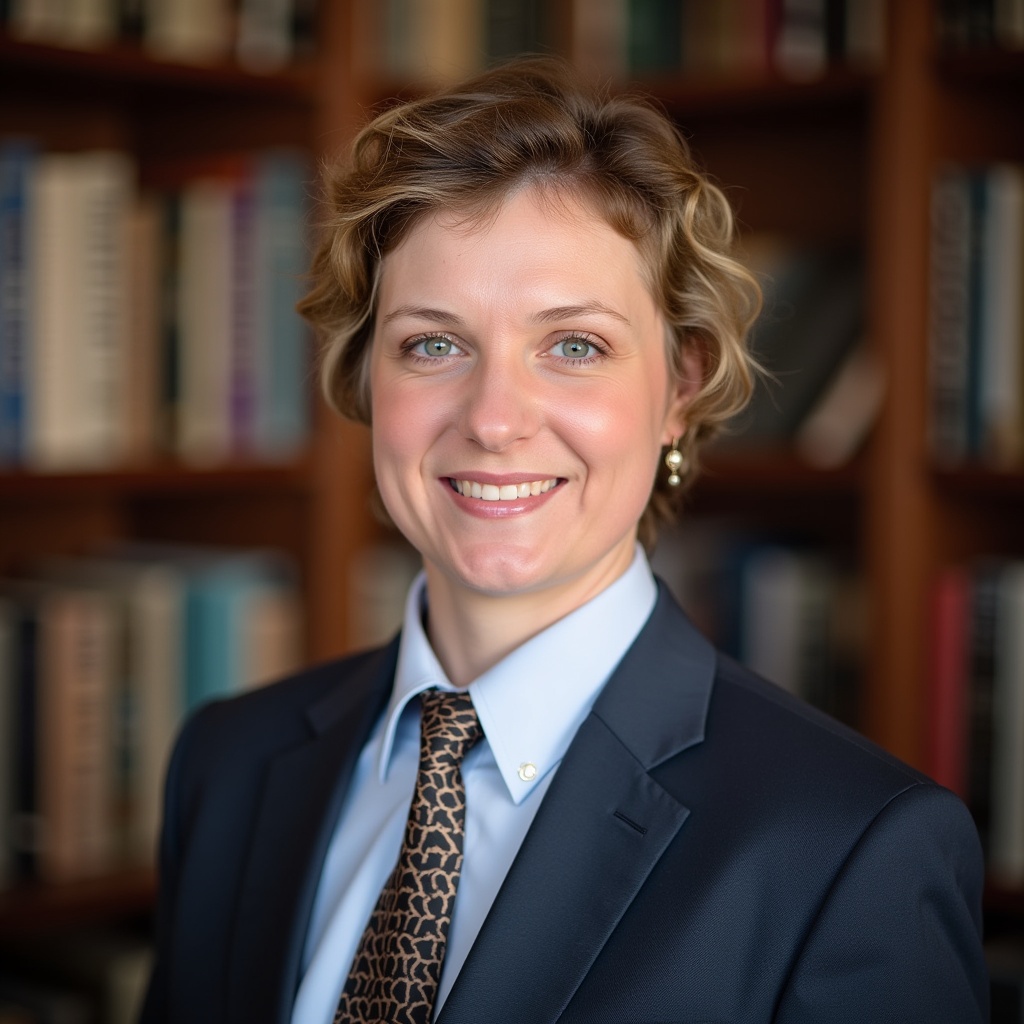}} \\
            \includegraphics[clip, viewport=128bp 192bp 896bp 960bp, width=0.043\linewidth]{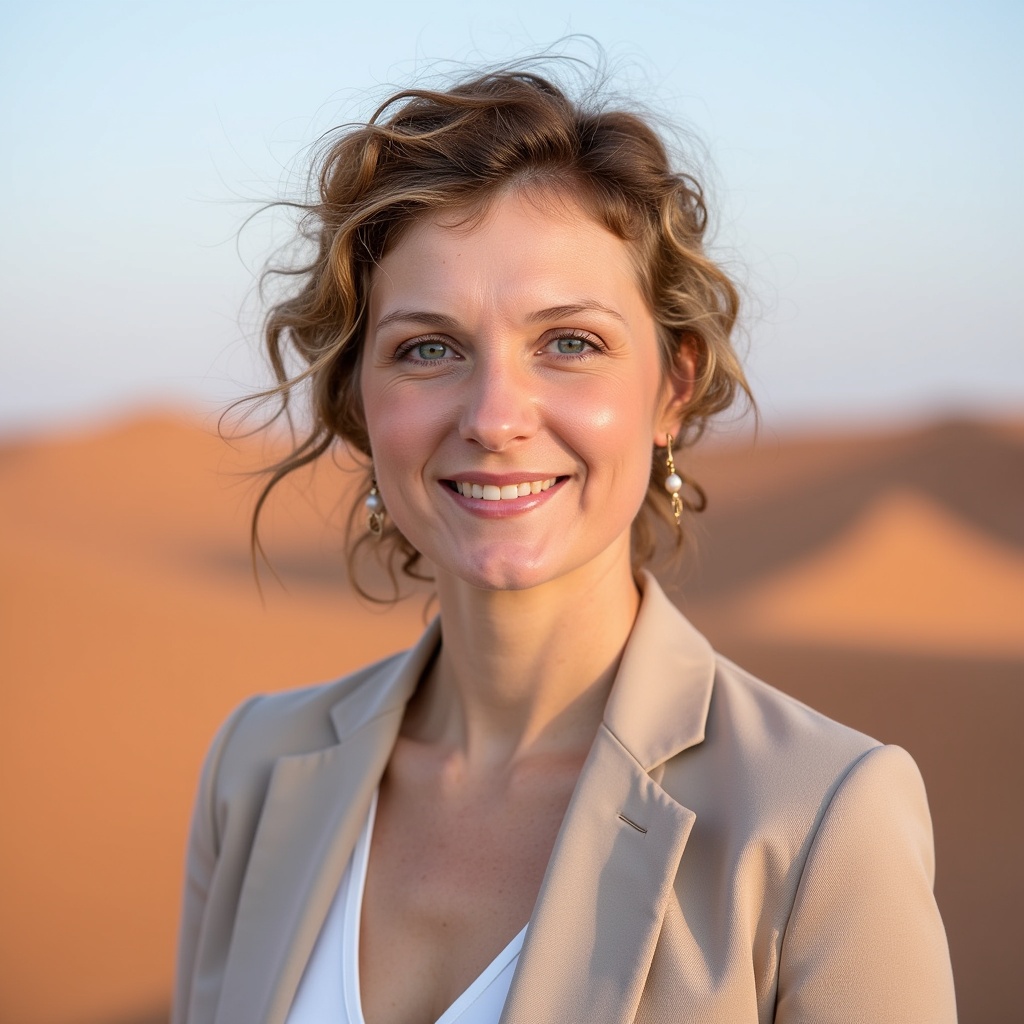} &
            \includegraphics[clip, viewport=128bp 192bp 896bp 960bp, width=0.043\linewidth]{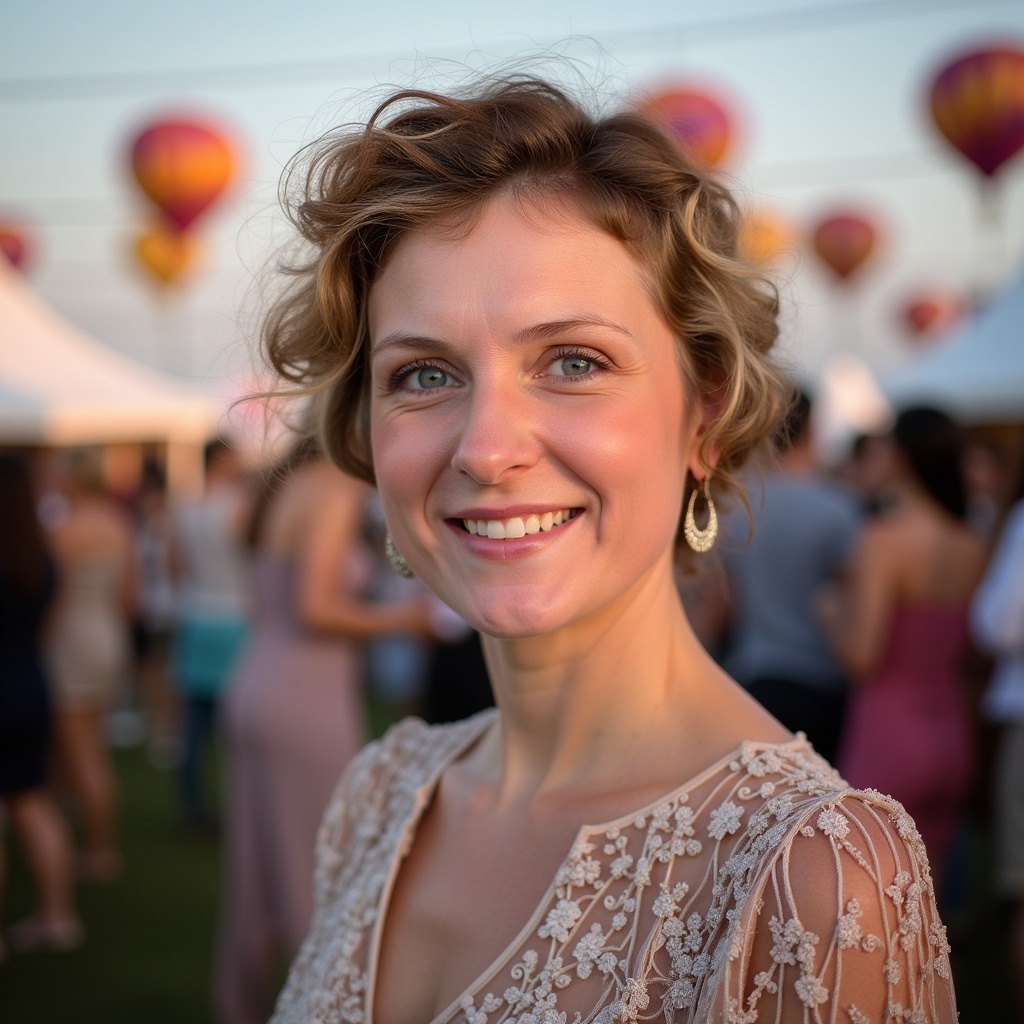} &
            \includegraphics[clip, viewport=128bp 192bp 896bp 960bp, width=0.043\linewidth]{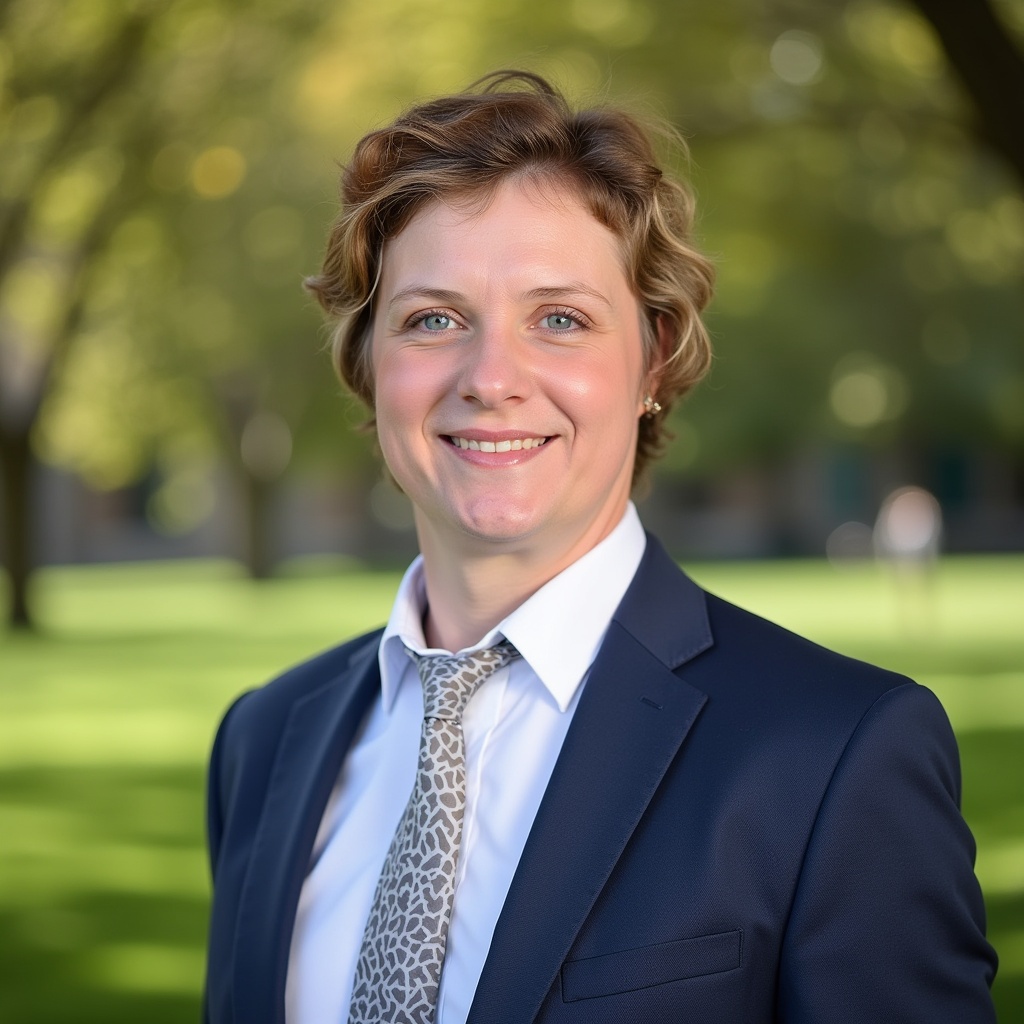}
         \end{tabular}
        } &
        {\setlength{\tabcolsep}{0pt}%
         \renewcommand{\arraystretch}{0}%
         \begin{tabular}{ccc}
            \multicolumn{3}{c}{\includegraphics[clip, viewport=128bp 192bp 896bp 960bp, width=0.128\linewidth]{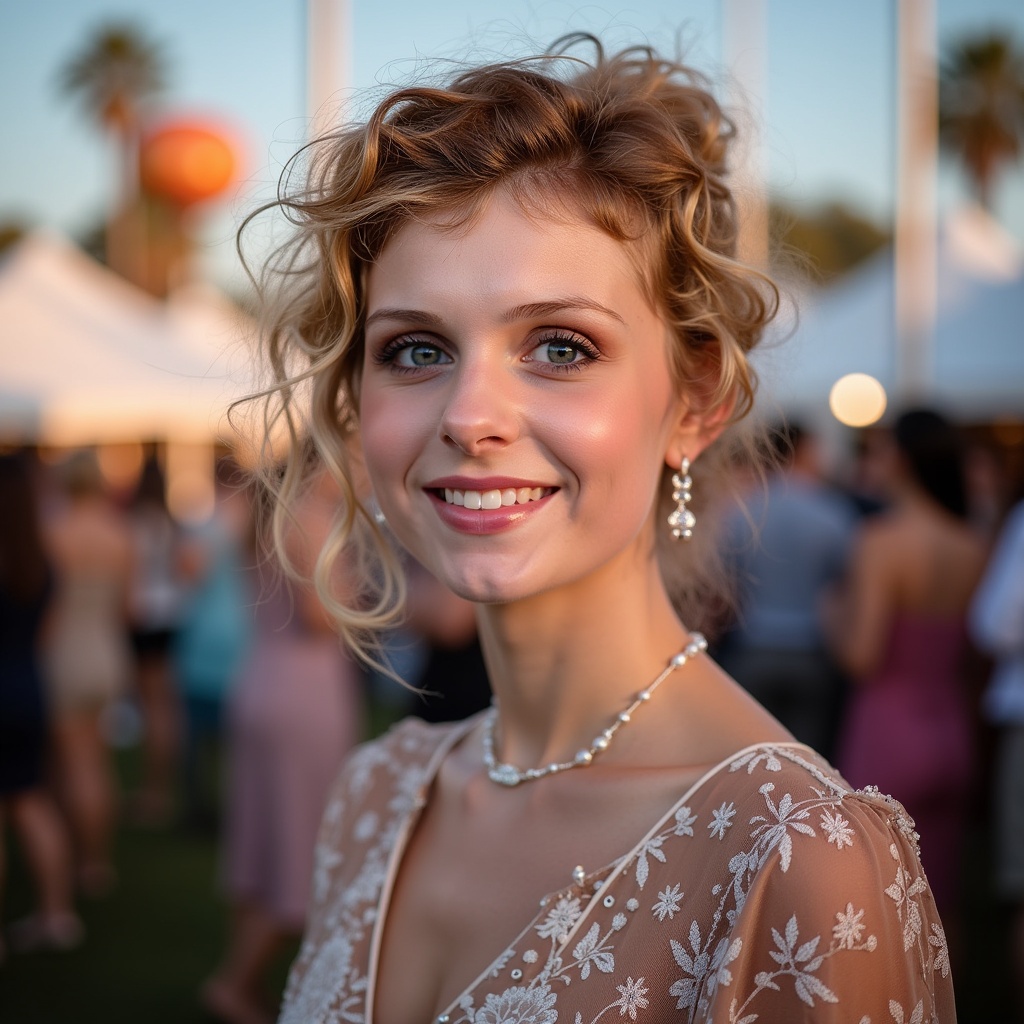}} \\
            \includegraphics[clip, viewport=128bp 192bp 896bp 960bp, width=0.043\linewidth]{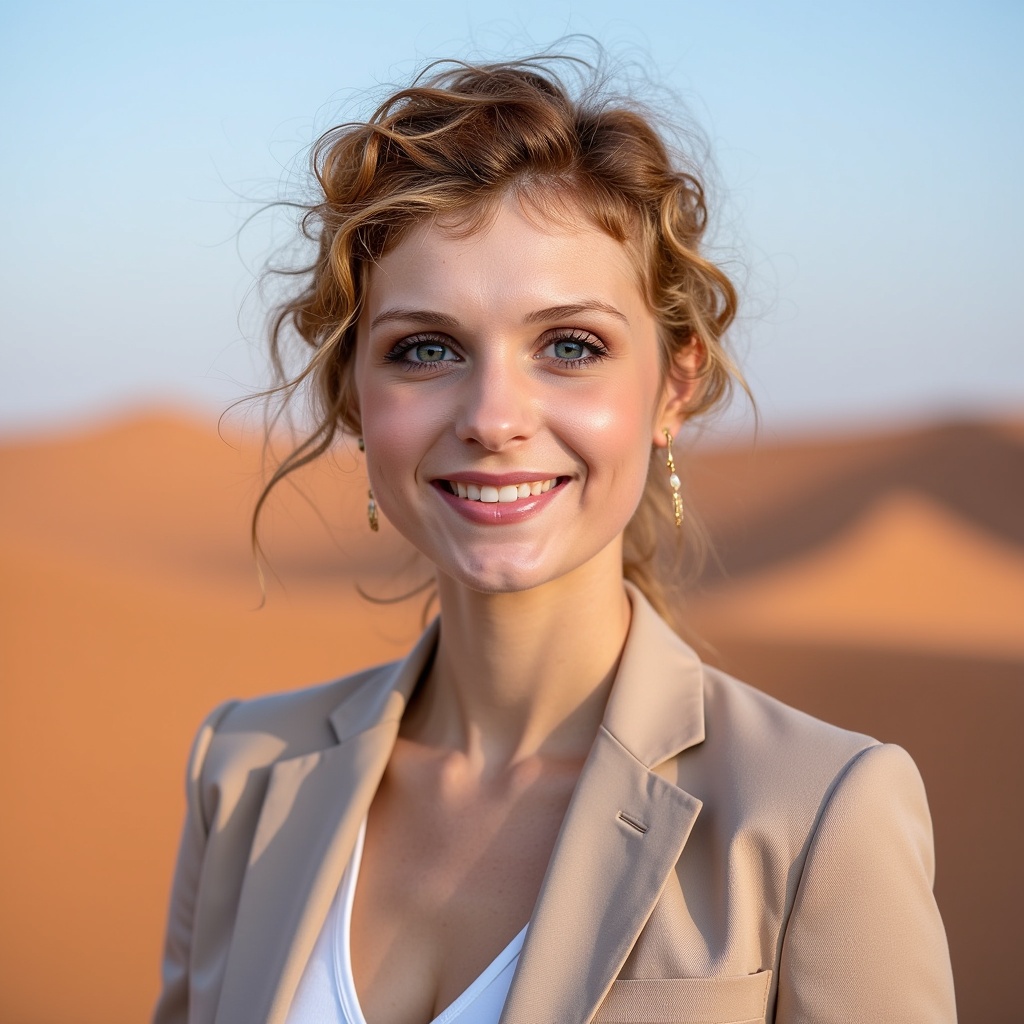} &
            \includegraphics[clip, viewport=128bp 192bp 896bp 960bp, width=0.043\linewidth]{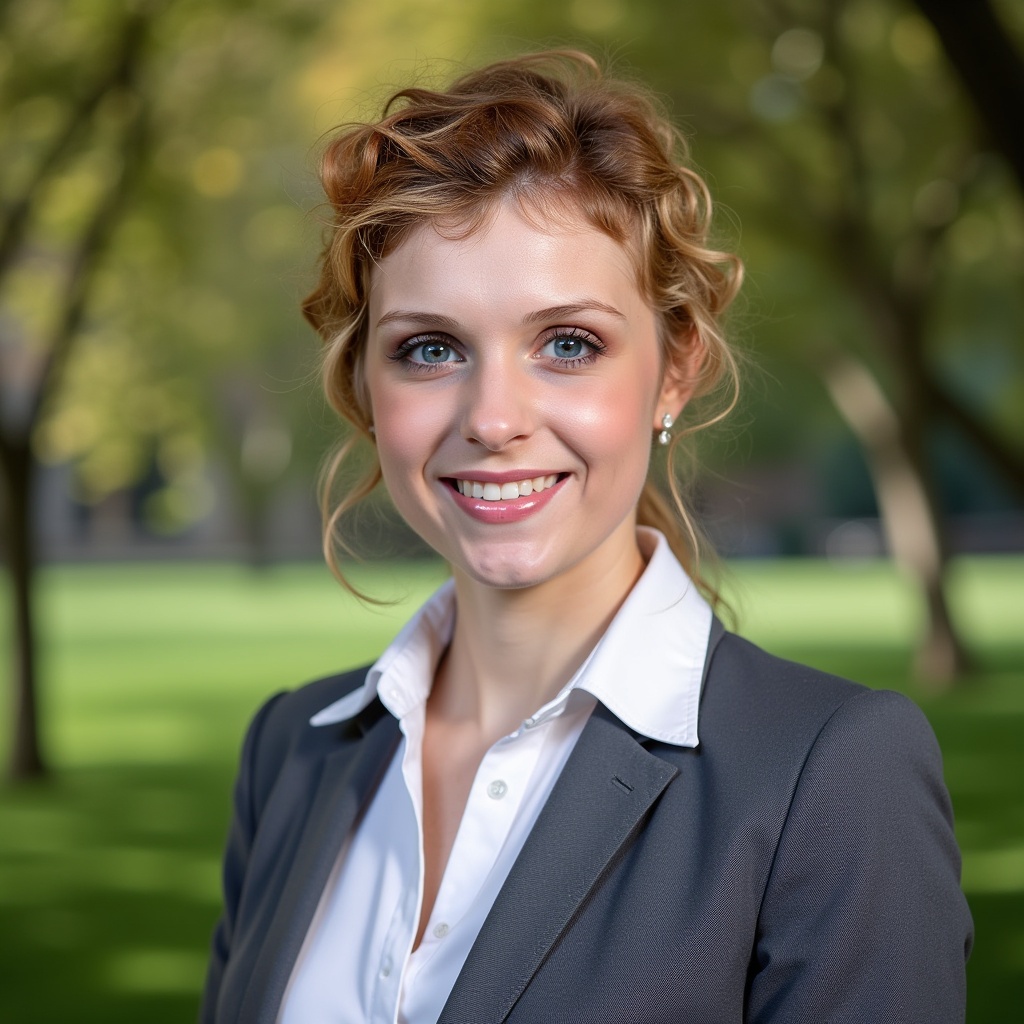} &
            \includegraphics[clip, viewport=128bp 192bp 896bp 960bp, width=0.043\linewidth]{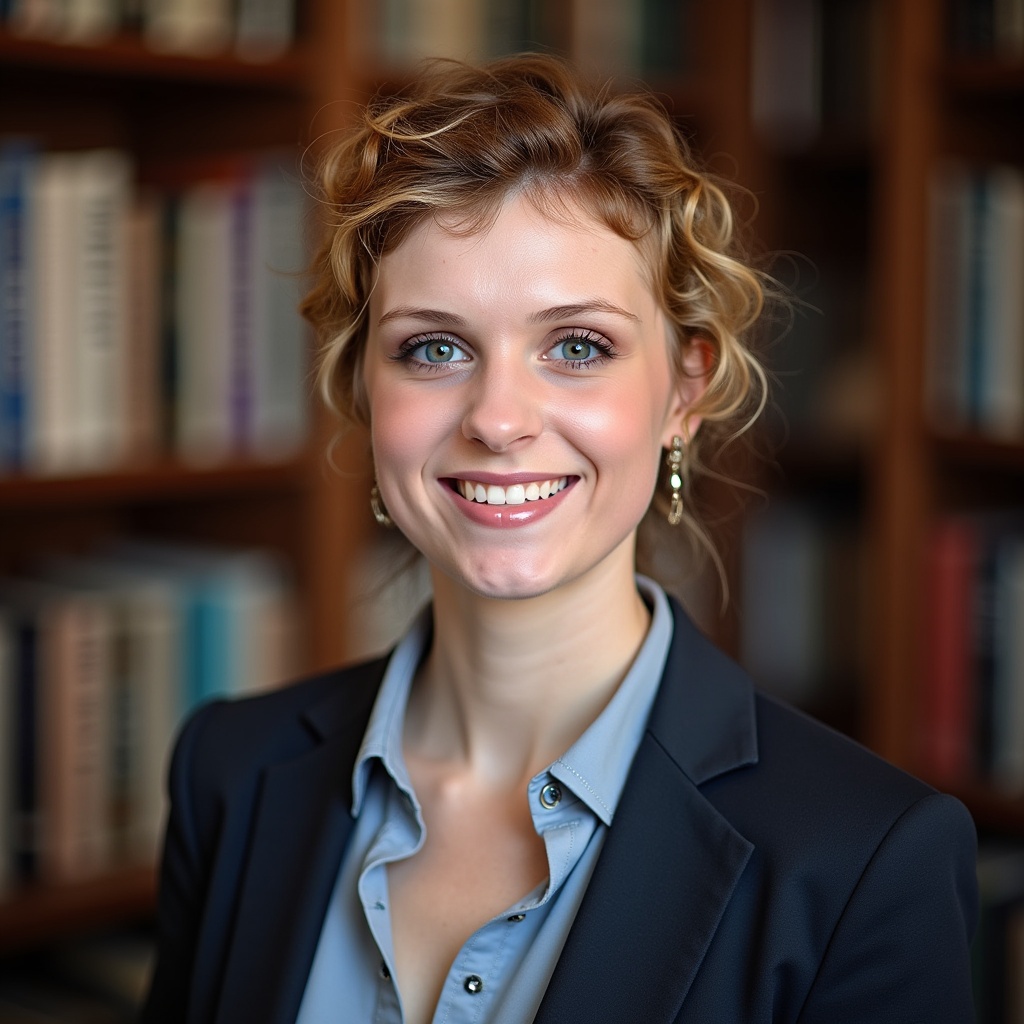}
         \end{tabular}
        } &
        {\setlength{\tabcolsep}{0pt}%
         \renewcommand{\arraystretch}{0}%
         \begin{tabular}{ccc}
            \multicolumn{3}{c}{\includegraphics[clip, viewport=128bp 192bp 896bp 960bp, width=0.128\linewidth]{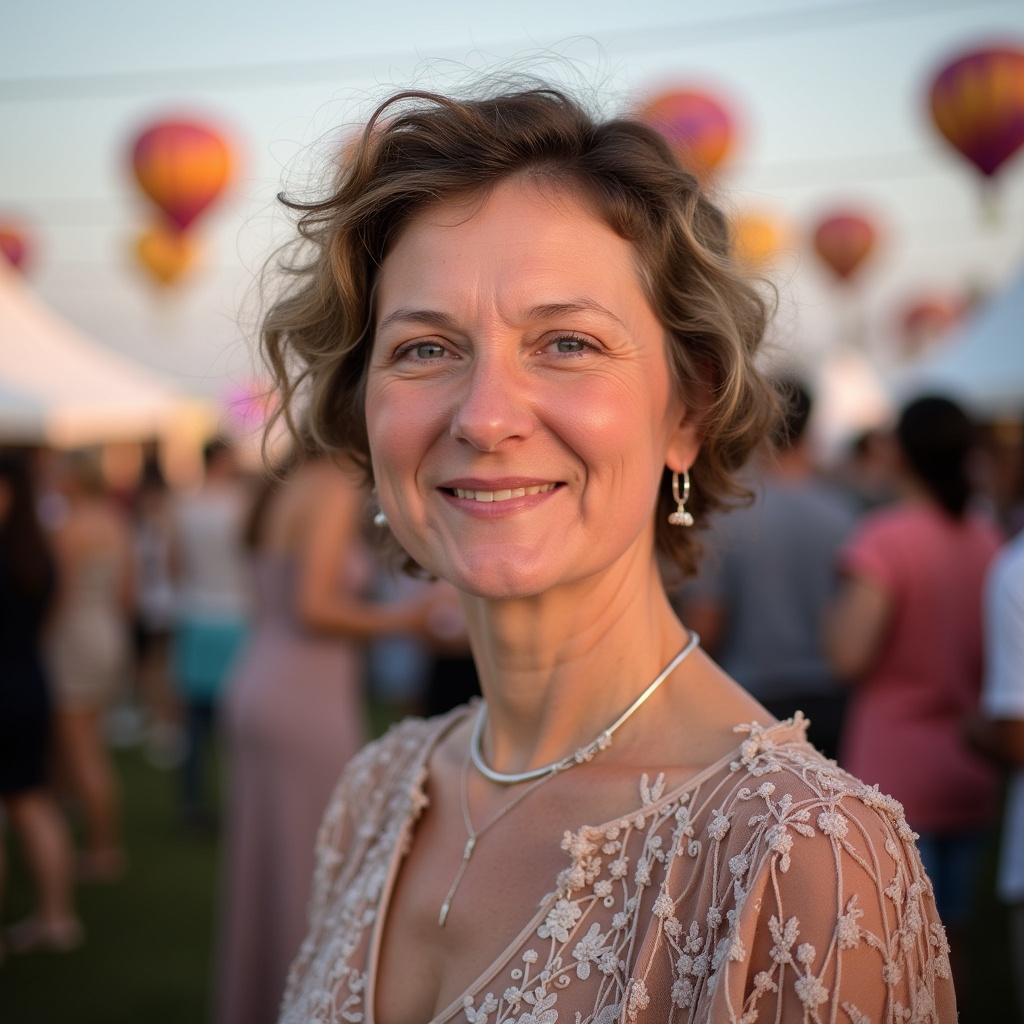}} \\
            \includegraphics[clip, viewport=128bp 192bp 896bp 960bp, width=0.043\linewidth]{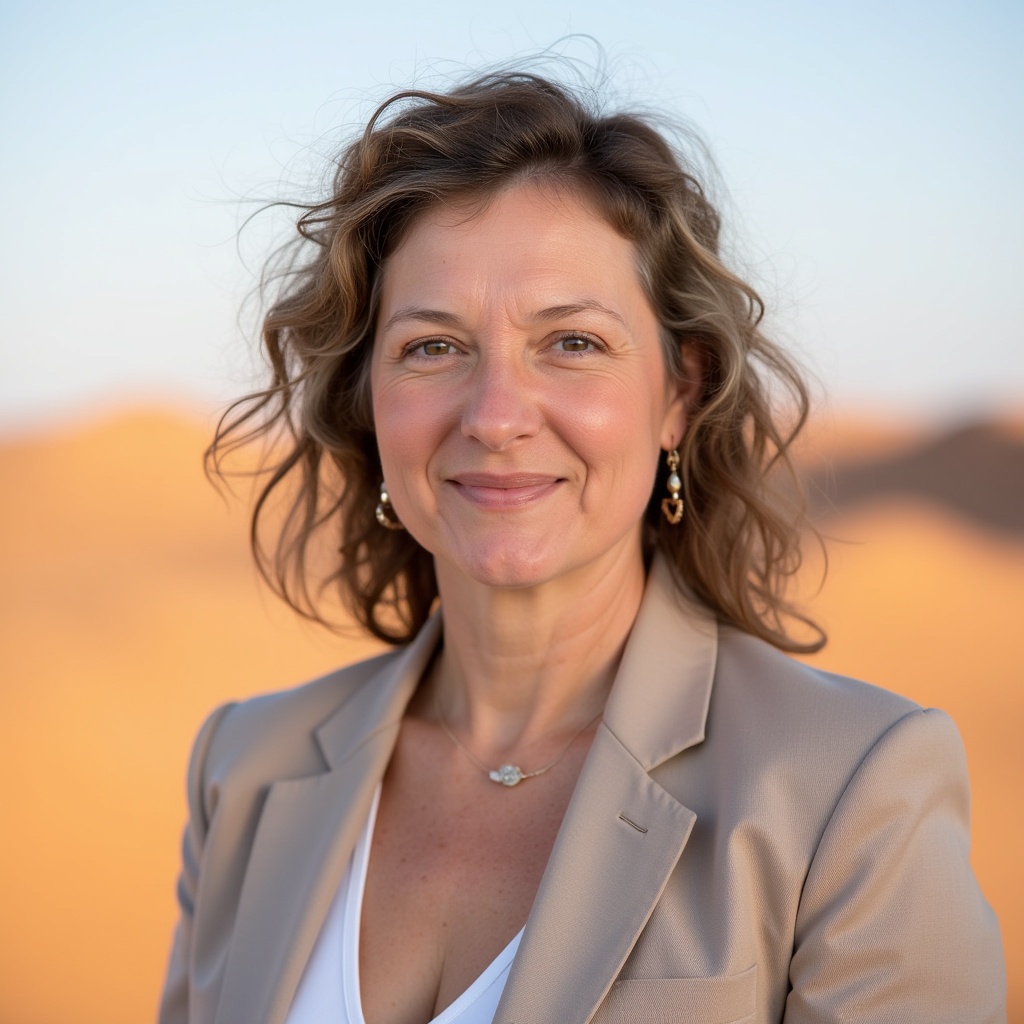} &
            \includegraphics[clip, viewport=128bp 192bp 896bp 960bp, width=0.043\linewidth]{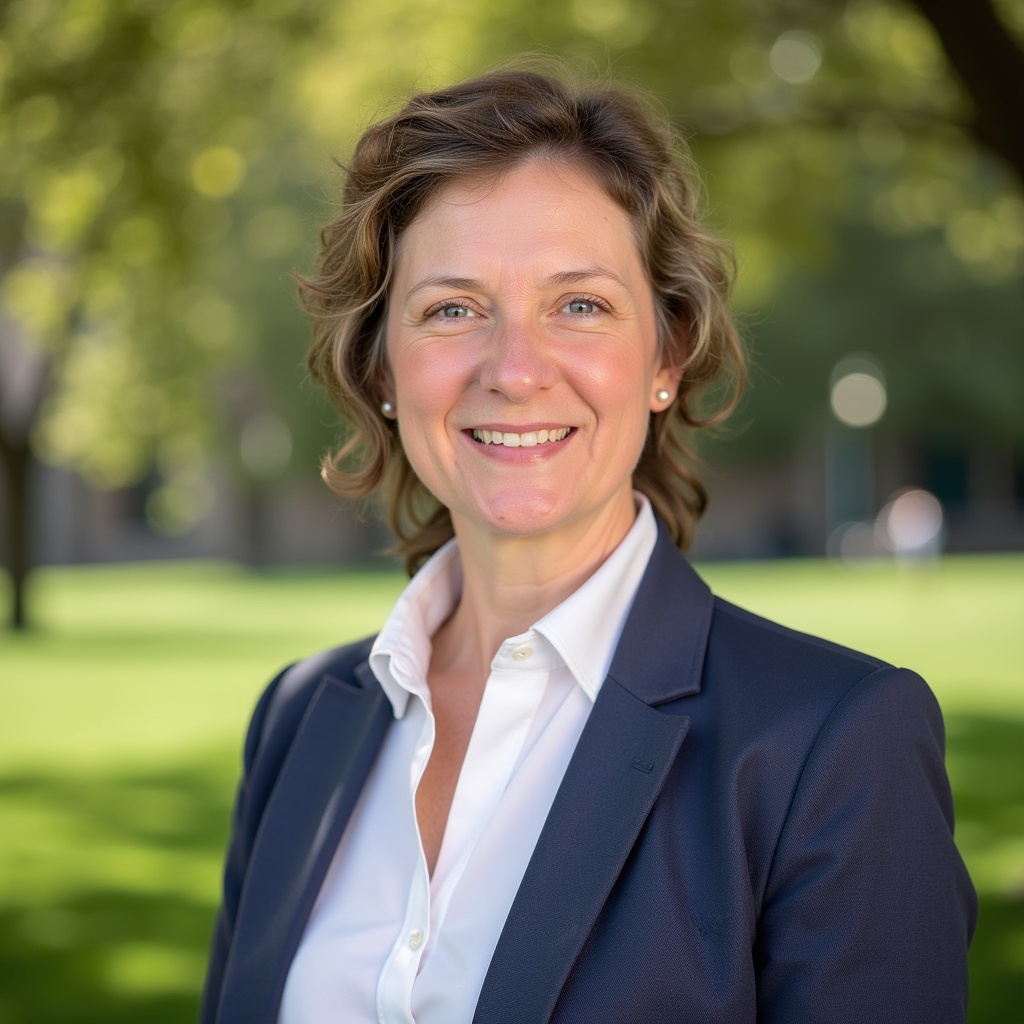} &
            \includegraphics[clip, viewport=128bp 192bp 896bp 960bp, width=0.043\linewidth]{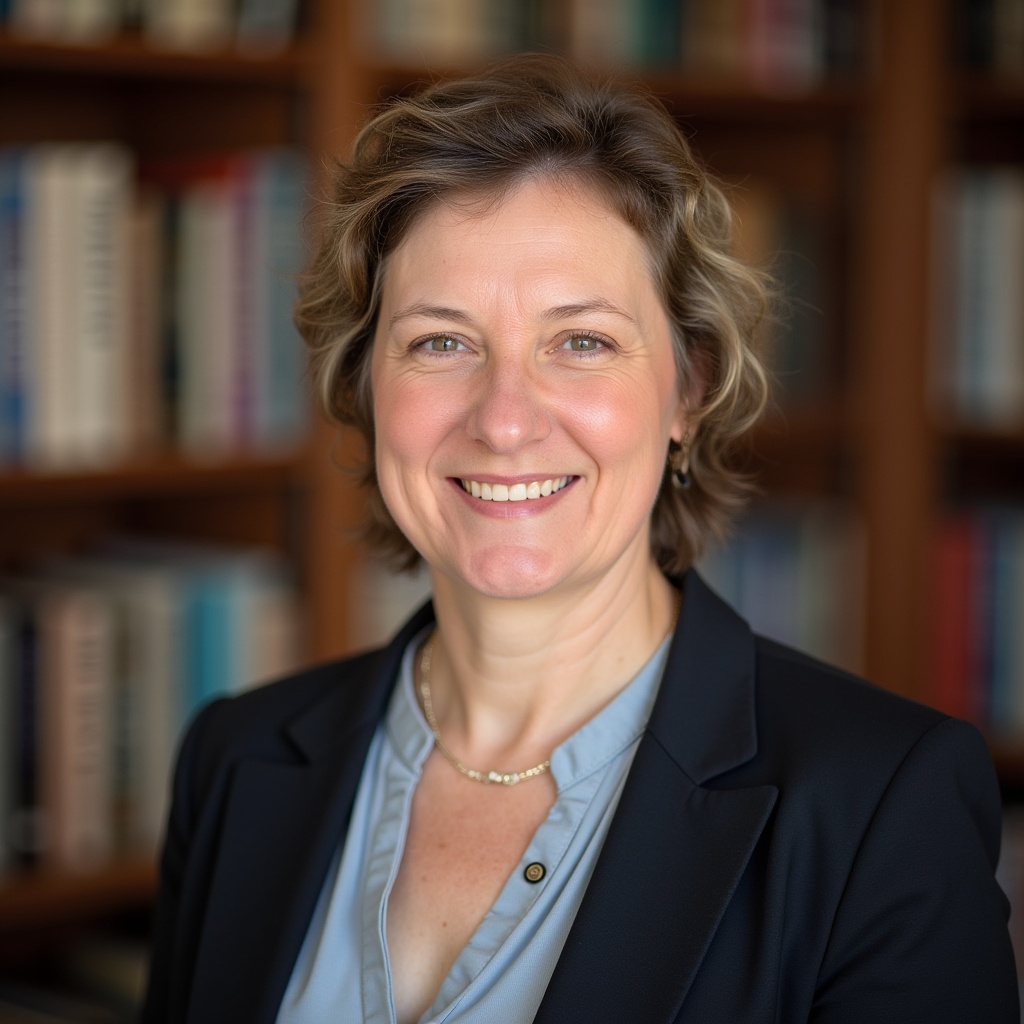}
         \end{tabular}
        } &
        {\setlength{\tabcolsep}{0pt}%
         \renewcommand{\arraystretch}{0}%
         \begin{tabular}{ccc}
            \multicolumn{3}{c}{\includegraphics[clip, viewport=128bp 192bp 896bp 960bp, width=0.128\linewidth]{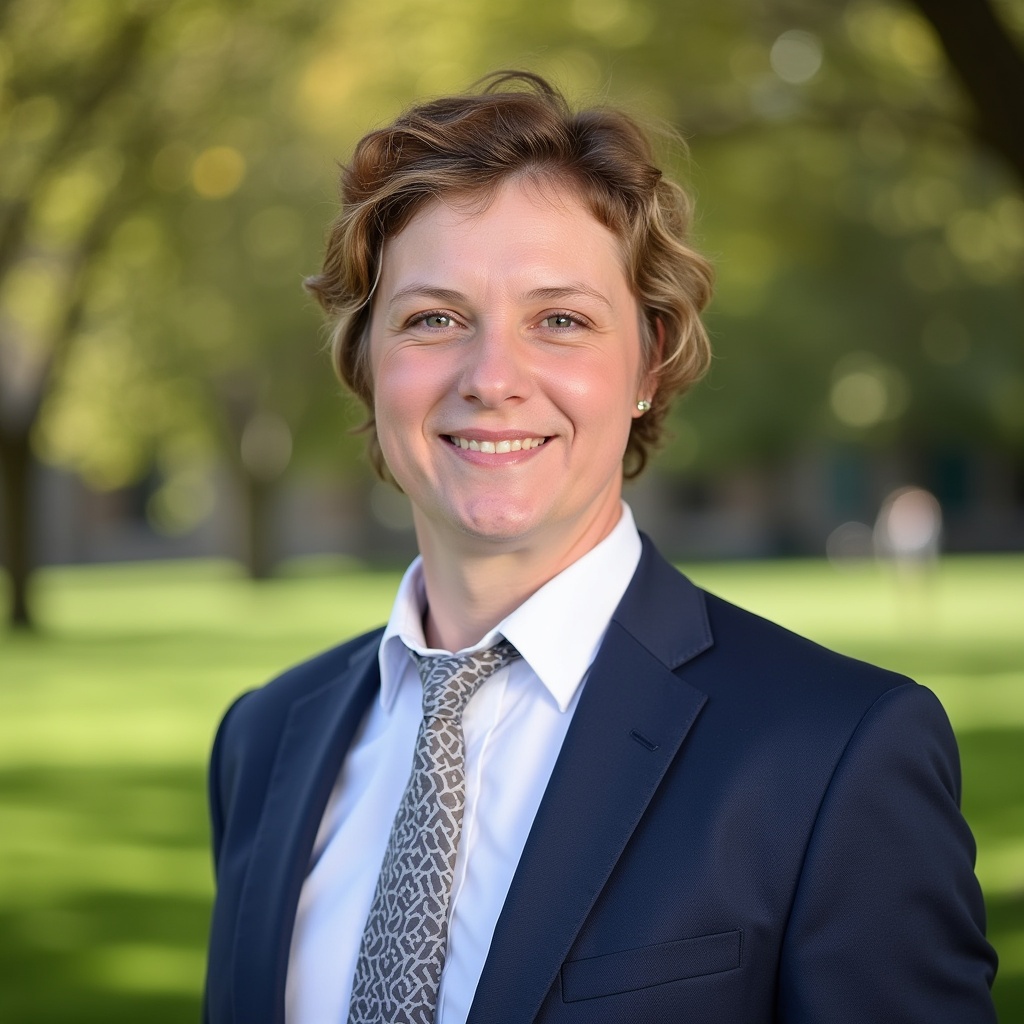}} \\
            \includegraphics[clip, viewport=128bp 192bp 896bp 960bp, width=0.043\linewidth]{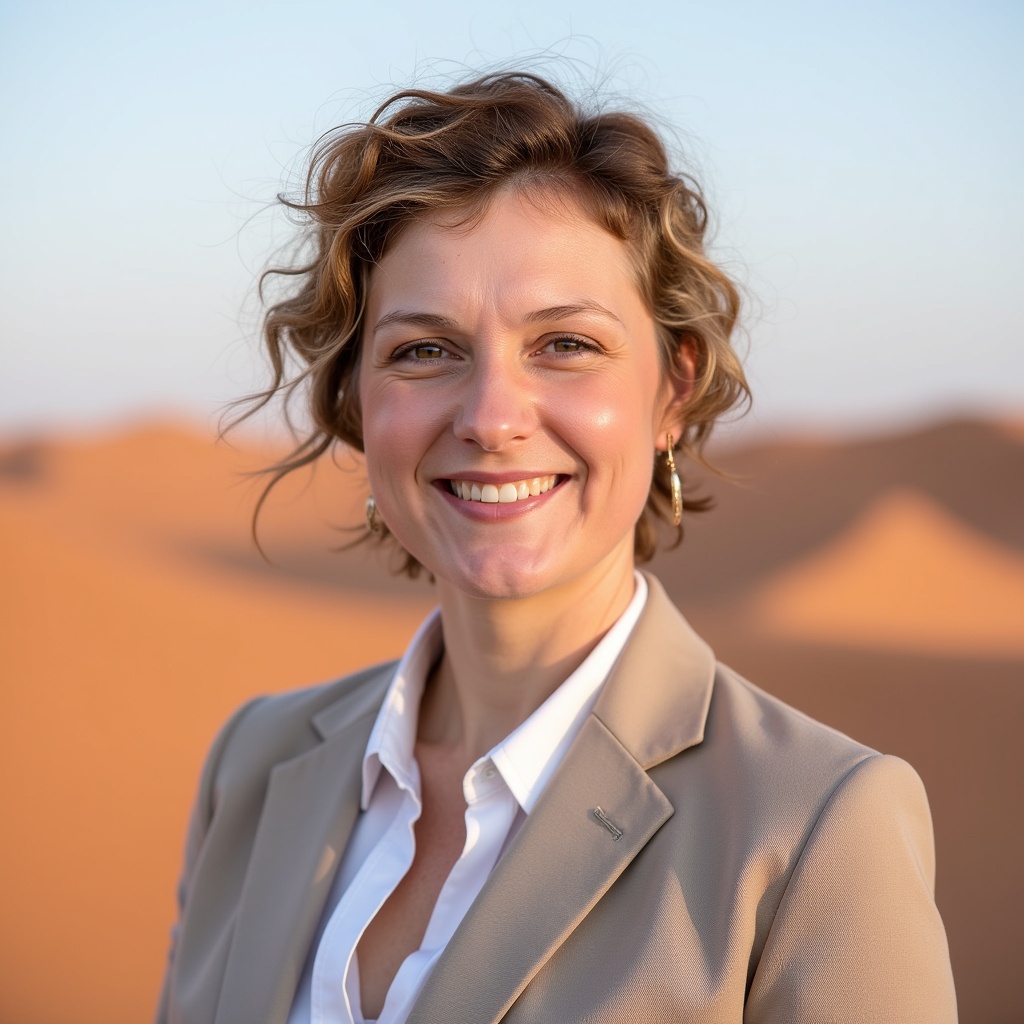} &
            \includegraphics[clip, viewport=128bp 192bp 896bp 960bp, width=0.043\linewidth]{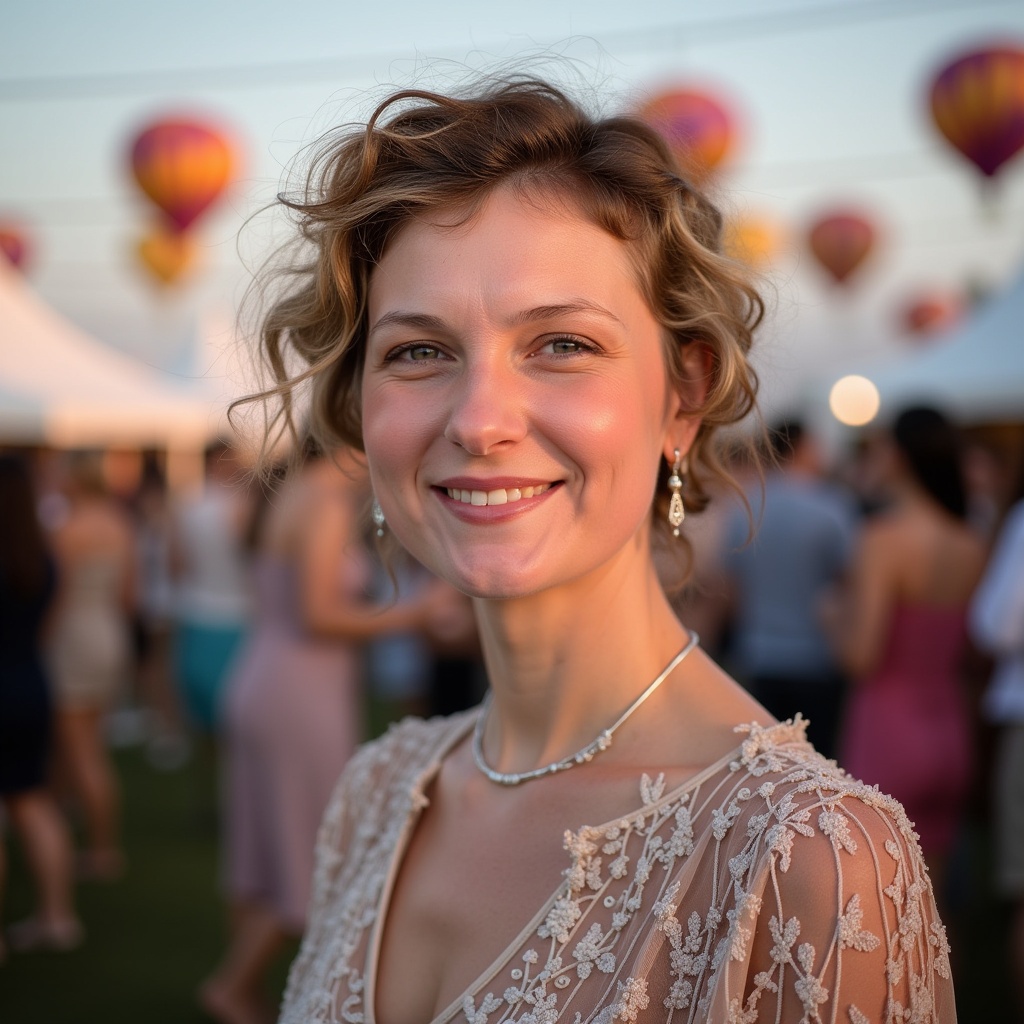} &
            \includegraphics[clip, viewport=128bp 192bp 896bp 960bp, width=0.043\linewidth]{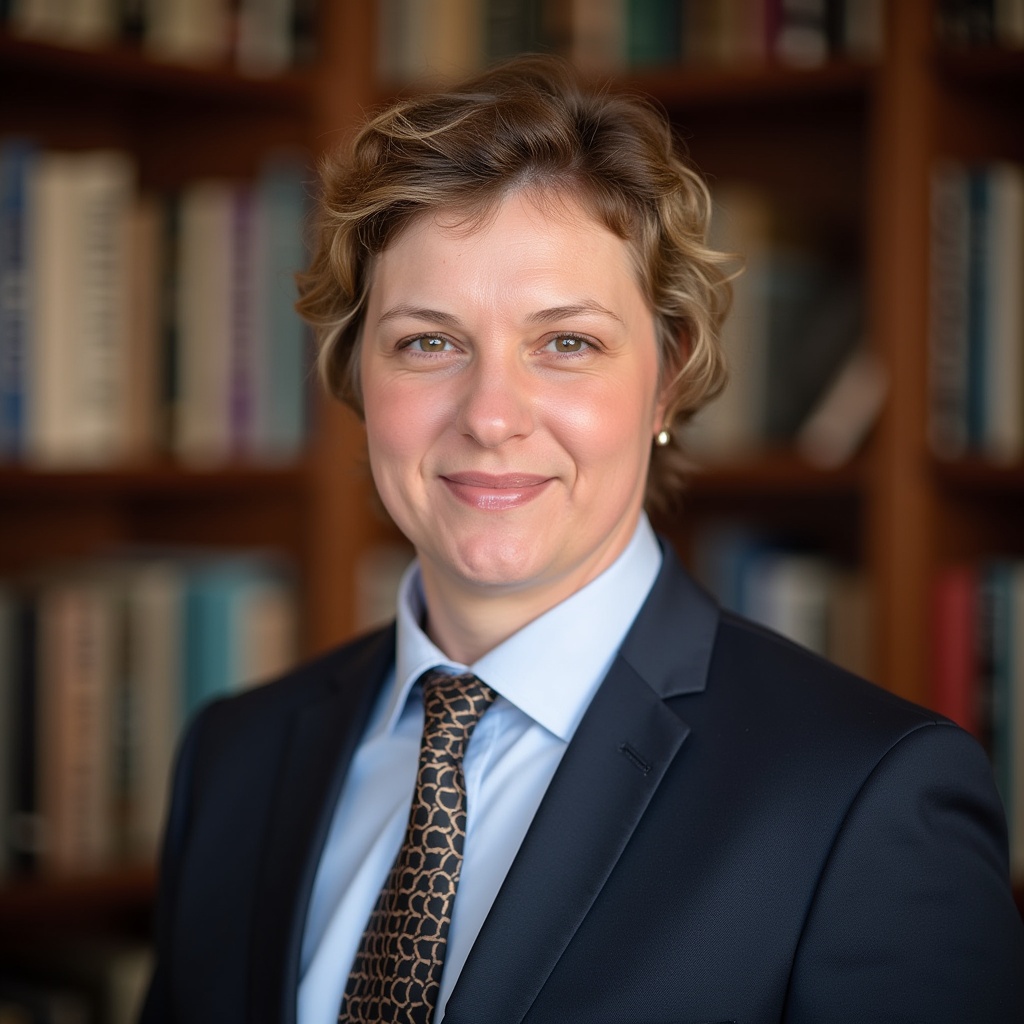}
         \end{tabular}
        } &
        {\setlength{\tabcolsep}{0pt}%
         \renewcommand{\arraystretch}{0}%
         \begin{tabular}{ccc}
            \multicolumn{3}{c}{\includegraphics[clip, viewport=128bp 192bp 896bp 960bp, width=0.128\linewidth]{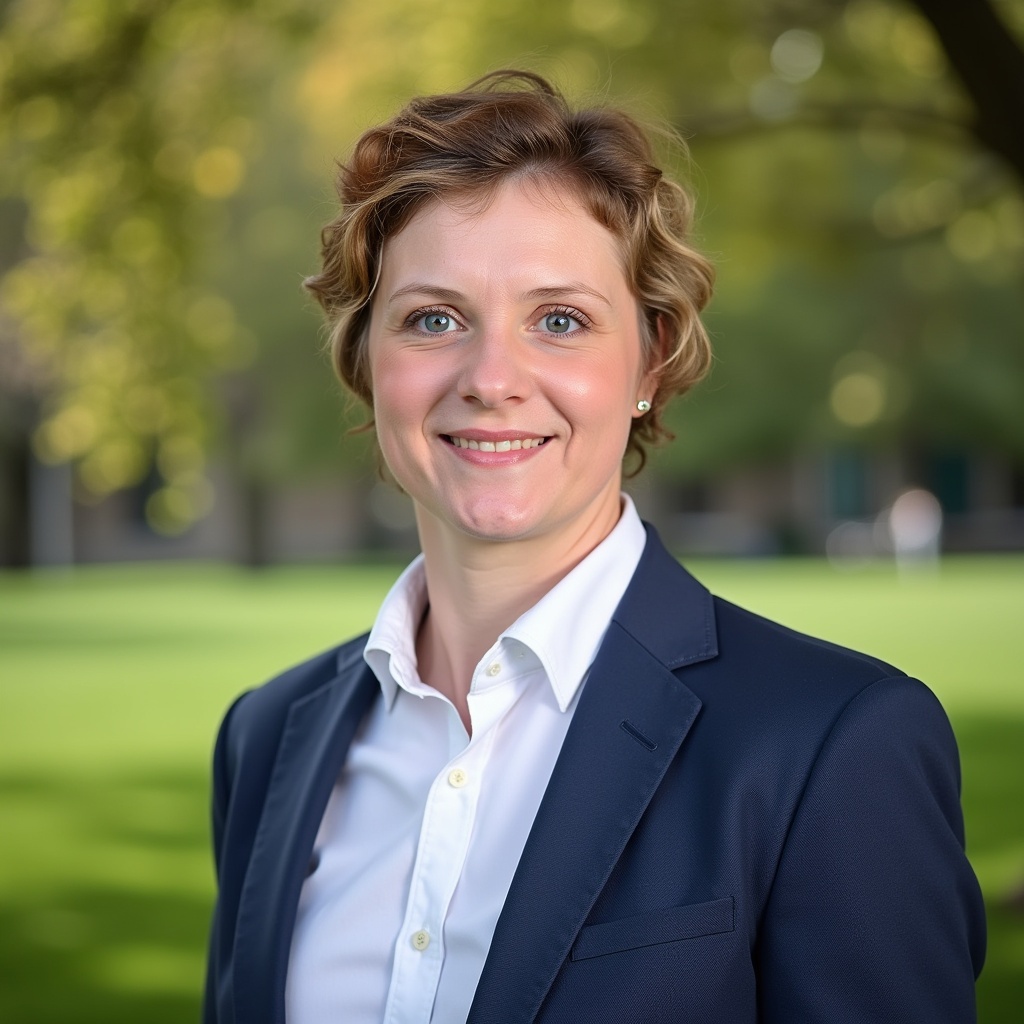}} \\
            \includegraphics[clip, viewport=128bp 192bp 896bp 960bp, width=0.043\linewidth]{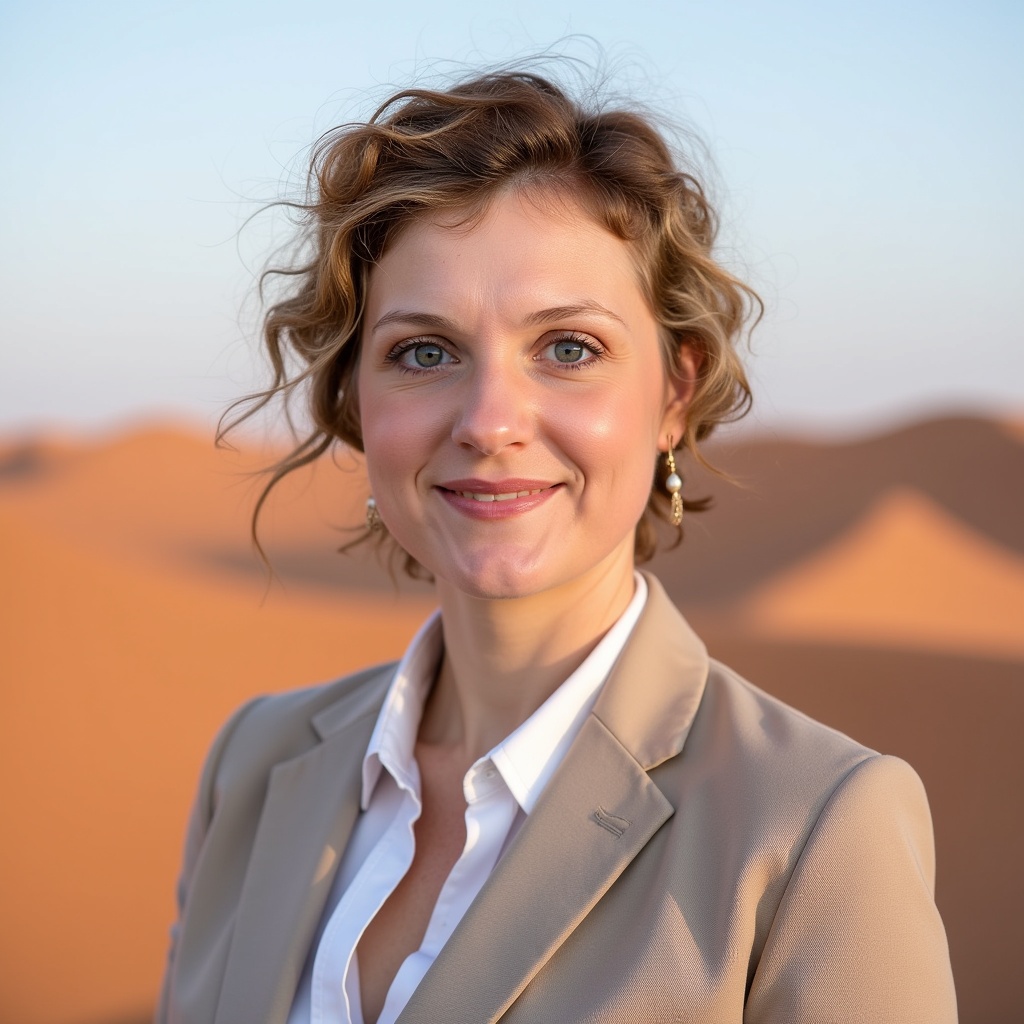} &
            \includegraphics[clip, viewport=128bp 192bp 896bp 960bp, width=0.043\linewidth]{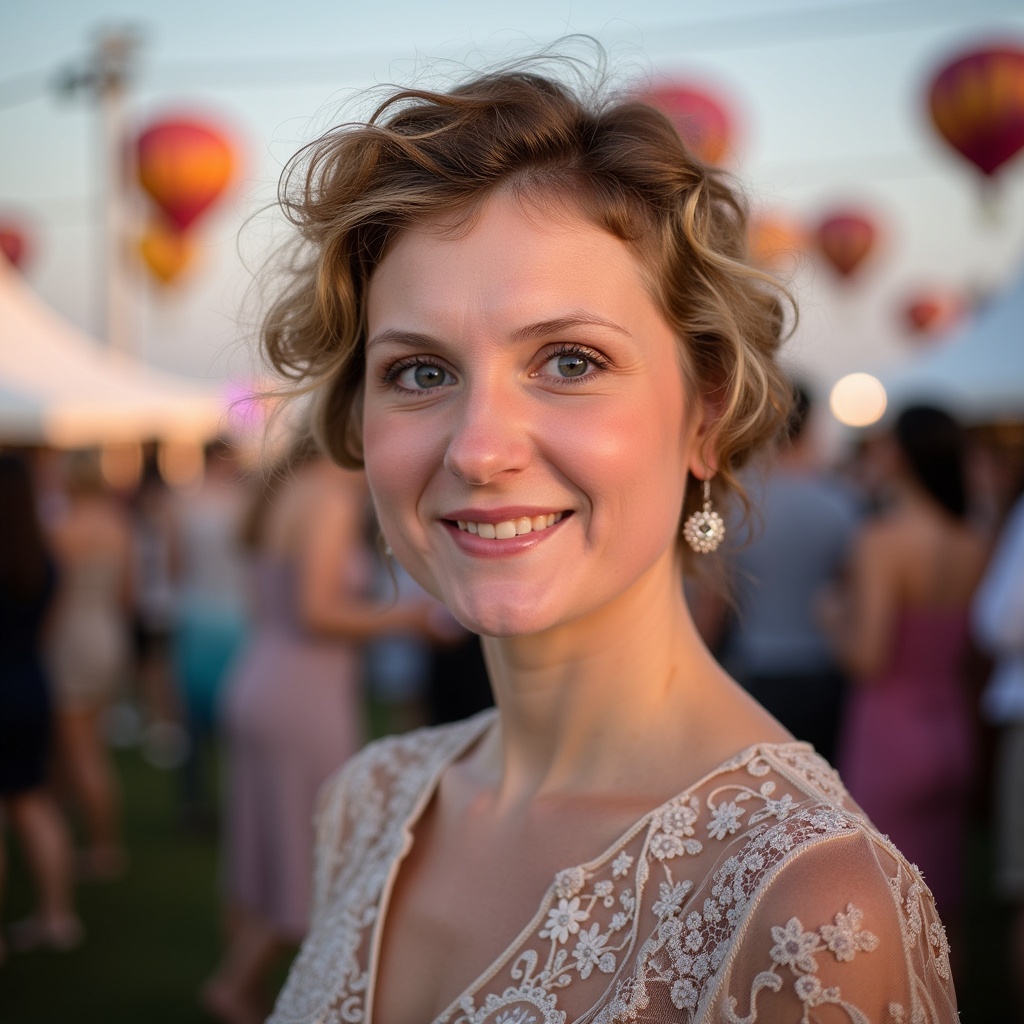} &
            \includegraphics[clip, viewport=128bp 192bp 896bp 960bp, width=0.043\linewidth]{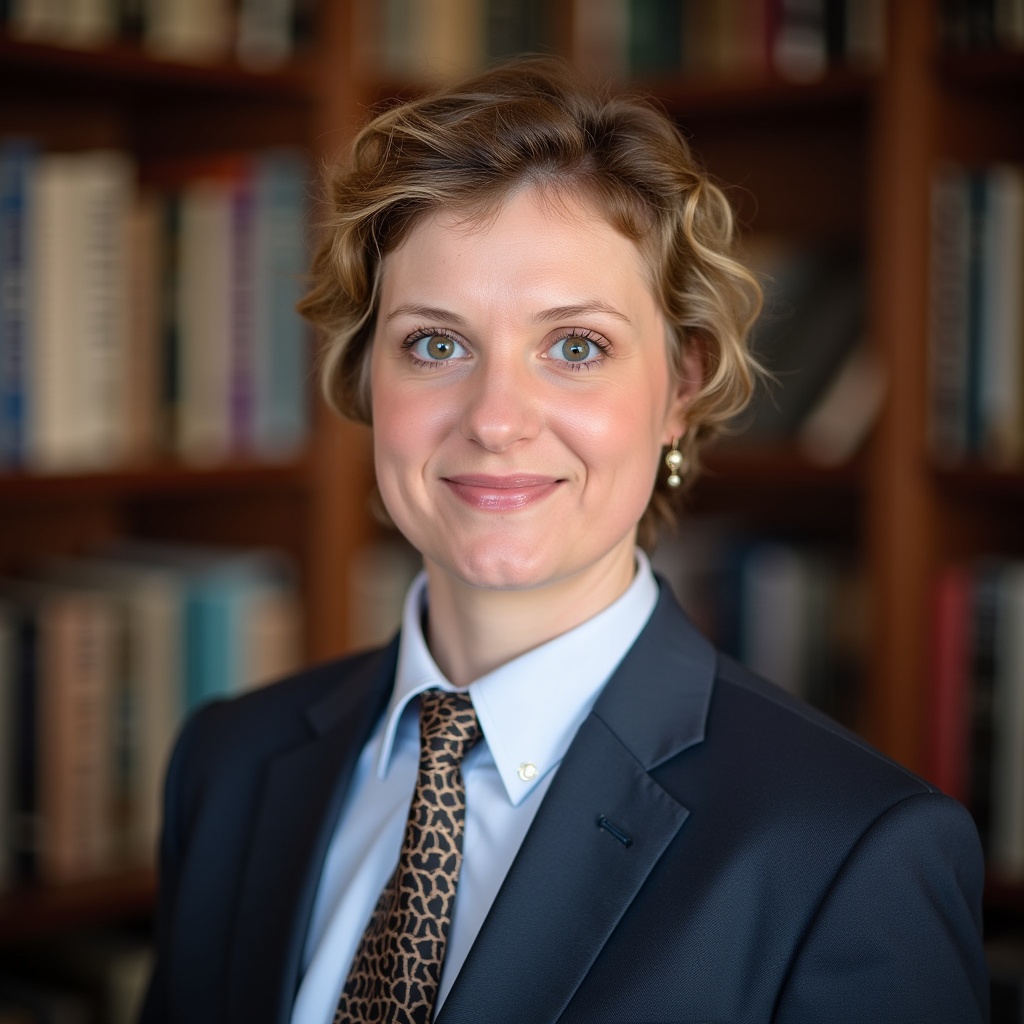}
         \end{tabular}
        } \\
        
        Original Identity &
        \multicolumn{2}{c}{$\xleftarrow{\hspace{1.5em}}$ - eye color + $\xrightarrow{\hspace{1.5em}}$} &
        \multicolumn{2}{c}{$\xleftarrow{\hspace{2.75em}}$ - age + $\xrightarrow{\hspace{2.75em}}$} &
        \multicolumn{2}{c}{$\xleftarrow{\hspace{0.7em}}$ - eye openness + $\xrightarrow{\hspace{0.7em}}$} \\

        {\setlength{\tabcolsep}{0pt}%
         \renewcommand{\arraystretch}{0}%
         \begin{tabular}{cc}
            \multicolumn{2}{c}{\includegraphics[width=0.171\linewidth]{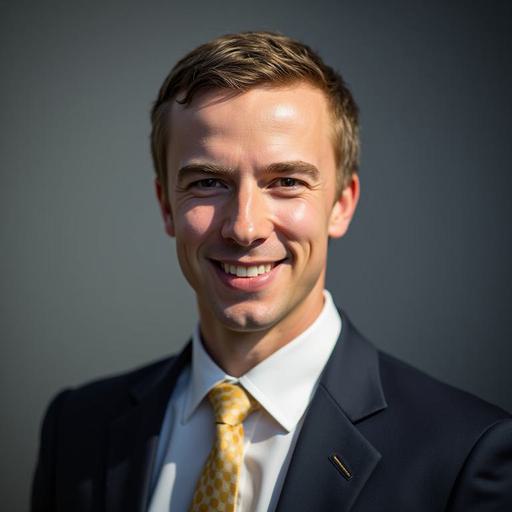}} \\
            &
         \end{tabular}
        } &
        {\setlength{\tabcolsep}{0pt}%
         \renewcommand{\arraystretch}{0}%
         \begin{tabular}{ccc}
            \multicolumn{3}{c}{\includegraphics[clip, viewport=128bp 192bp 896bp 960bp, width=0.128\linewidth]{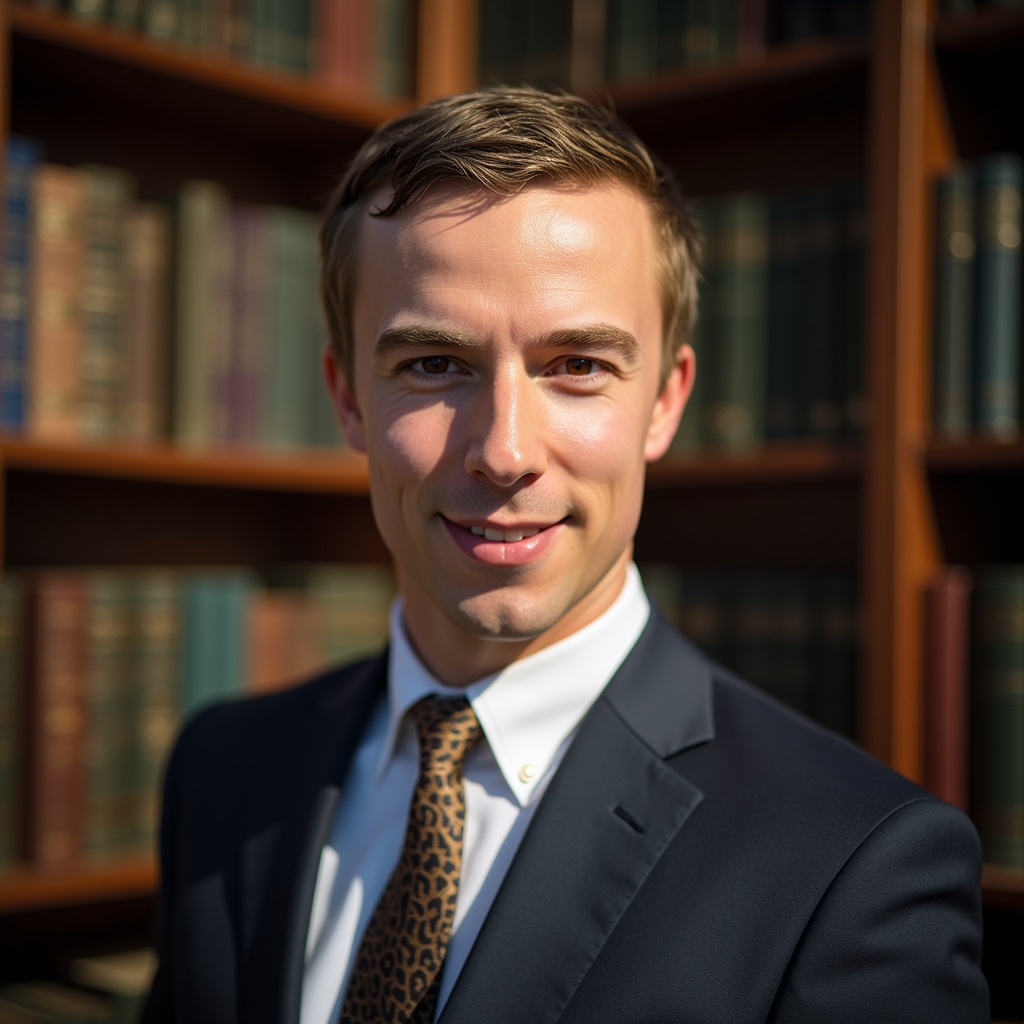}} \\
            \includegraphics[clip, viewport=128bp 192bp 896bp 960bp, width=0.043\linewidth]{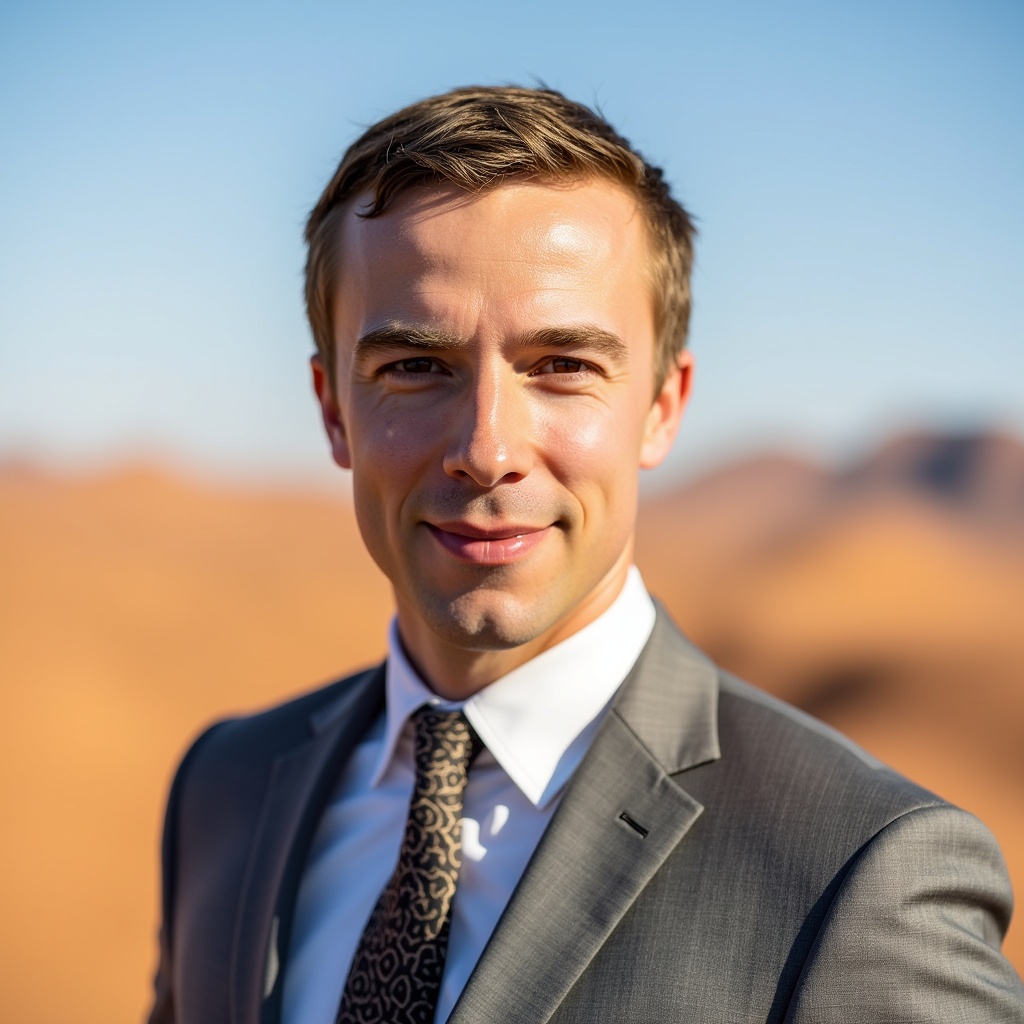} &
            \includegraphics[clip, viewport=128bp 192bp 896bp 960bp, width=0.043\linewidth]{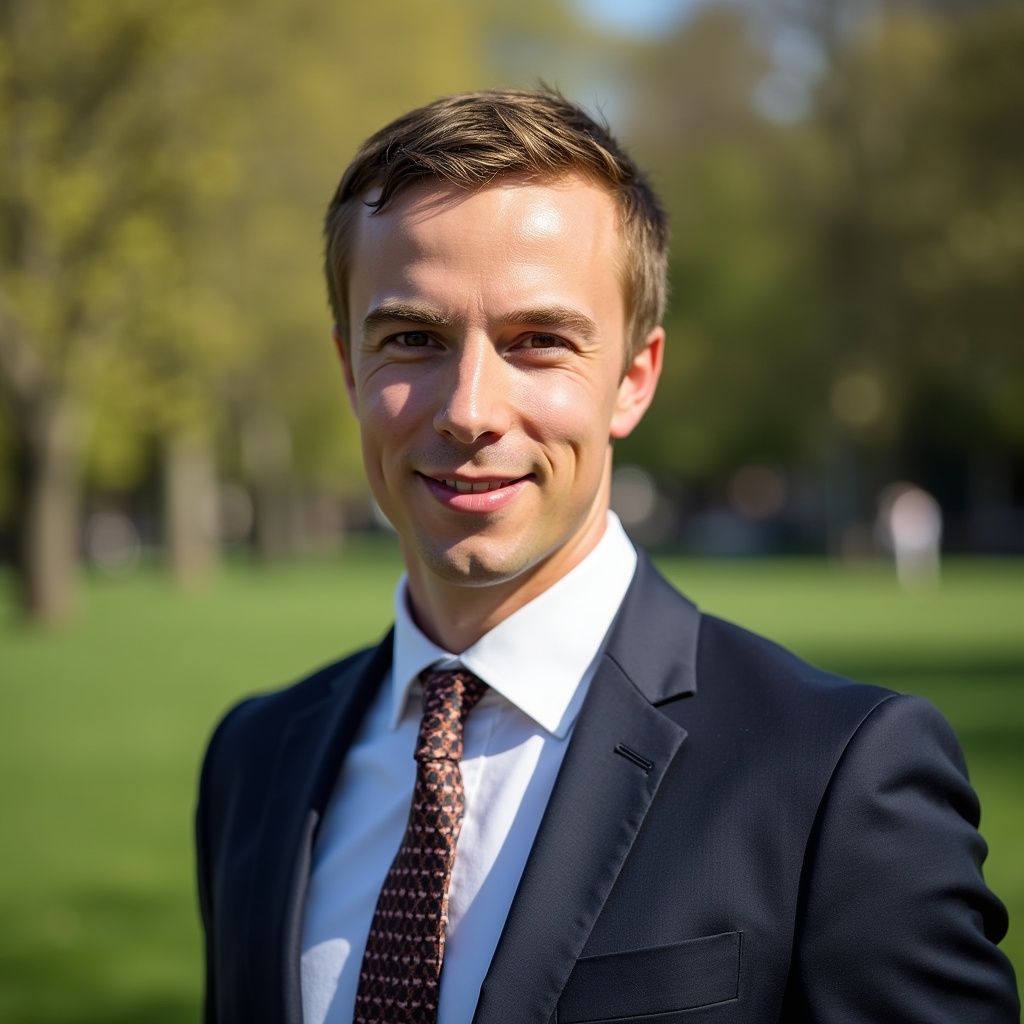} &
            \includegraphics[clip, viewport=128bp 192bp 896bp 960bp, width=0.043\linewidth]{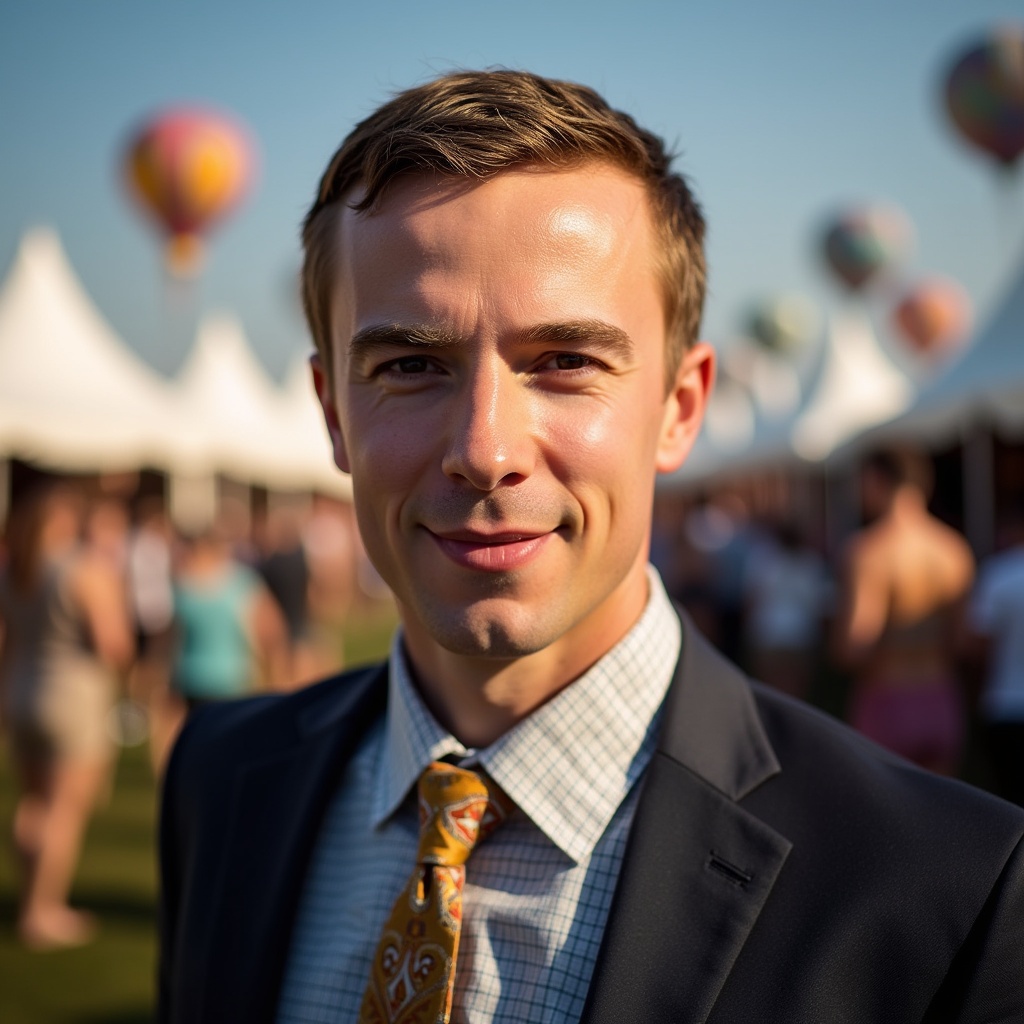}
         \end{tabular}
        } &
        {\setlength{\tabcolsep}{0pt}%
         \renewcommand{\arraystretch}{0}%
         \begin{tabular}{ccc}
            \multicolumn{3}{c}{\includegraphics[clip, viewport=128bp 192bp 896bp 960bp, width=0.128\linewidth]{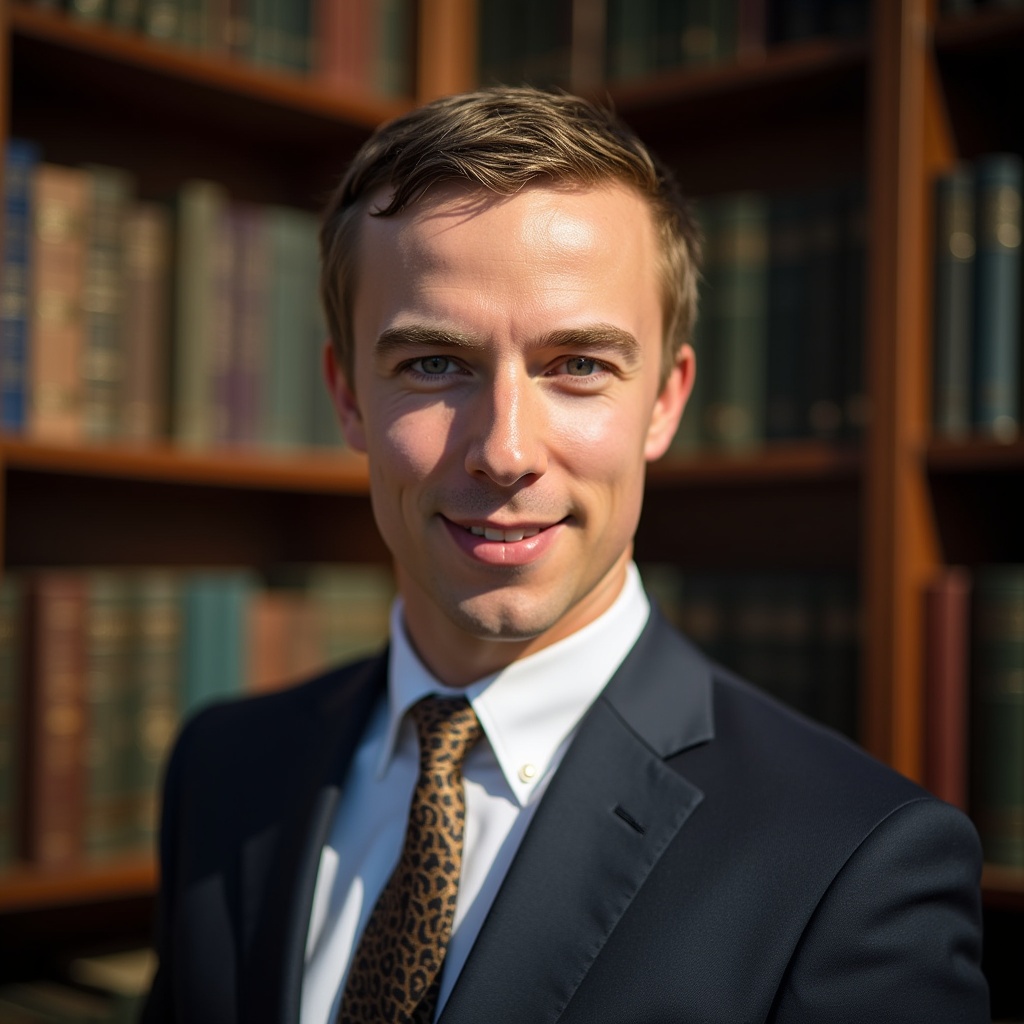}} \\
            \includegraphics[clip, viewport=128bp 192bp 896bp 960bp, width=0.043\linewidth]{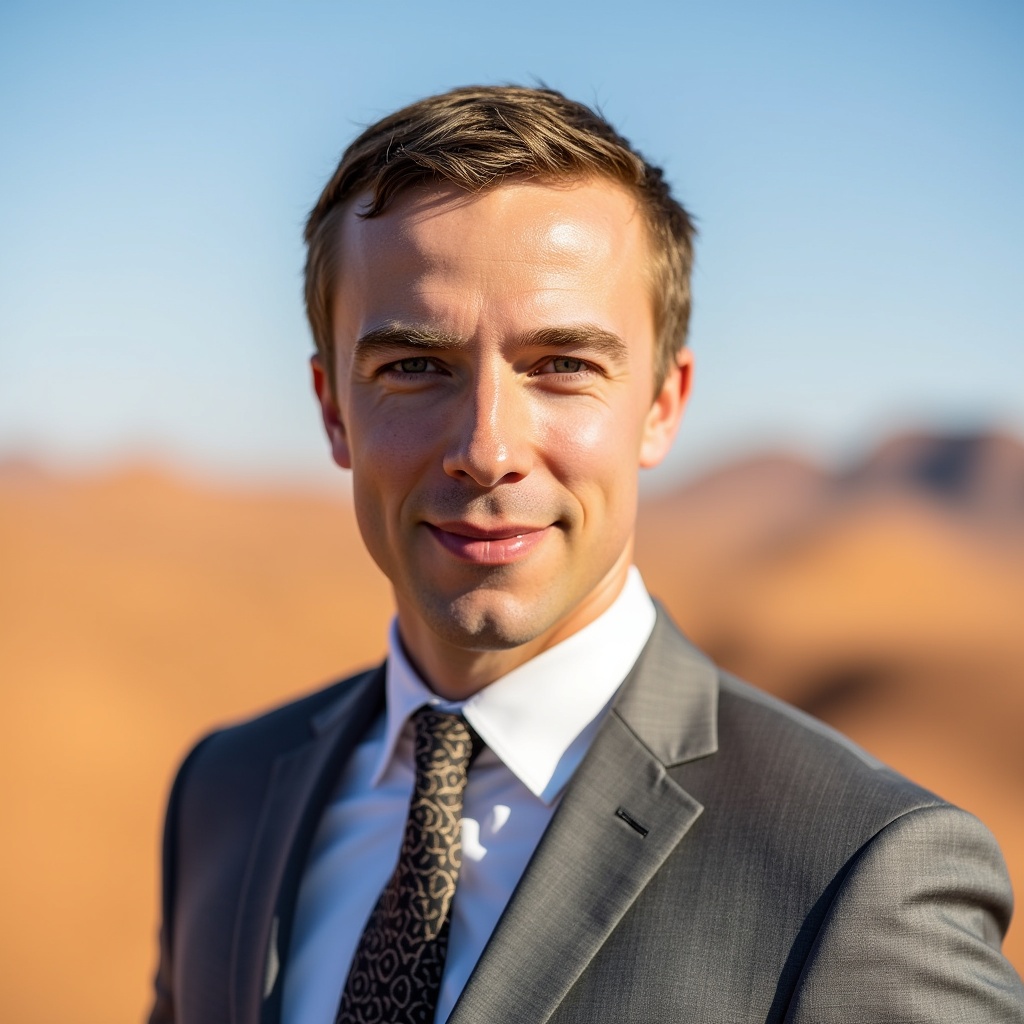} &
            \includegraphics[clip, viewport=128bp 192bp 896bp 960bp, width=0.043\linewidth]{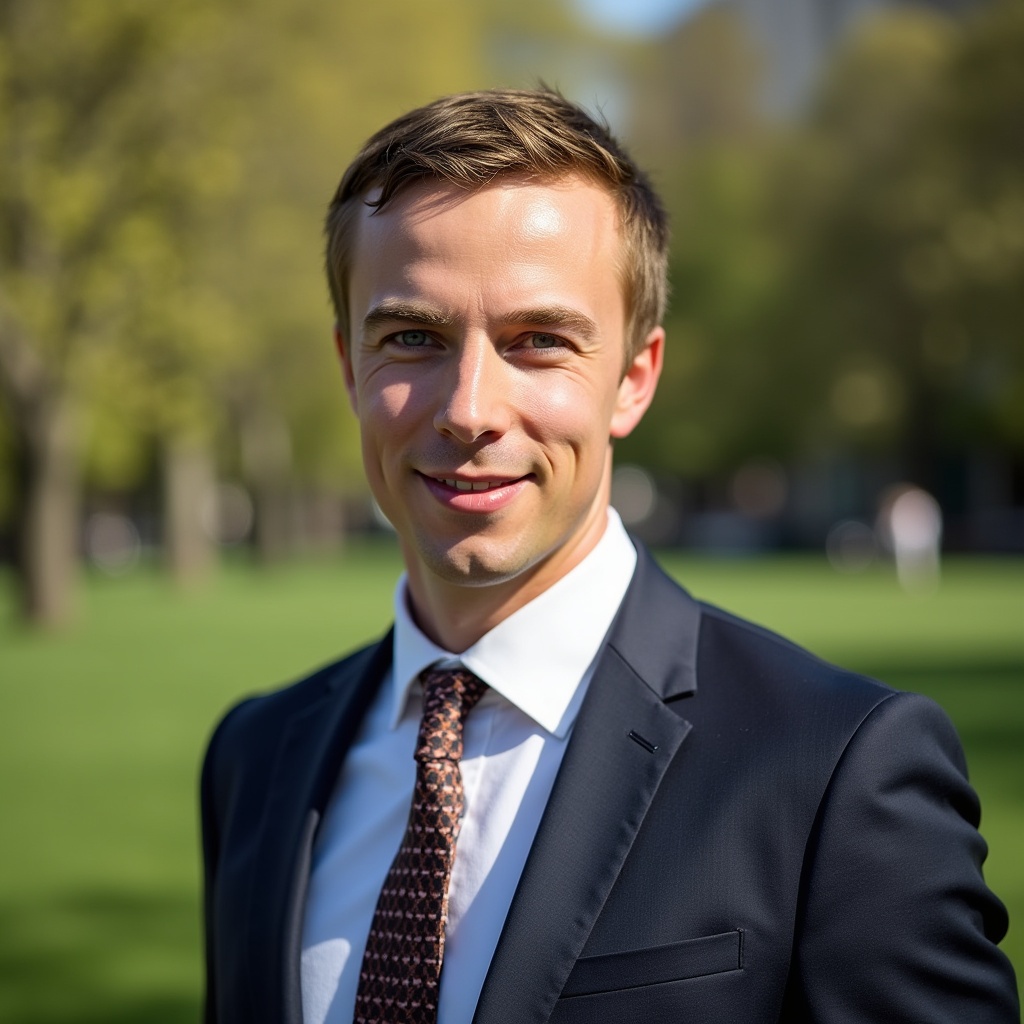} &
            \includegraphics[clip, viewport=128bp 192bp 896bp 960bp, width=0.043\linewidth]{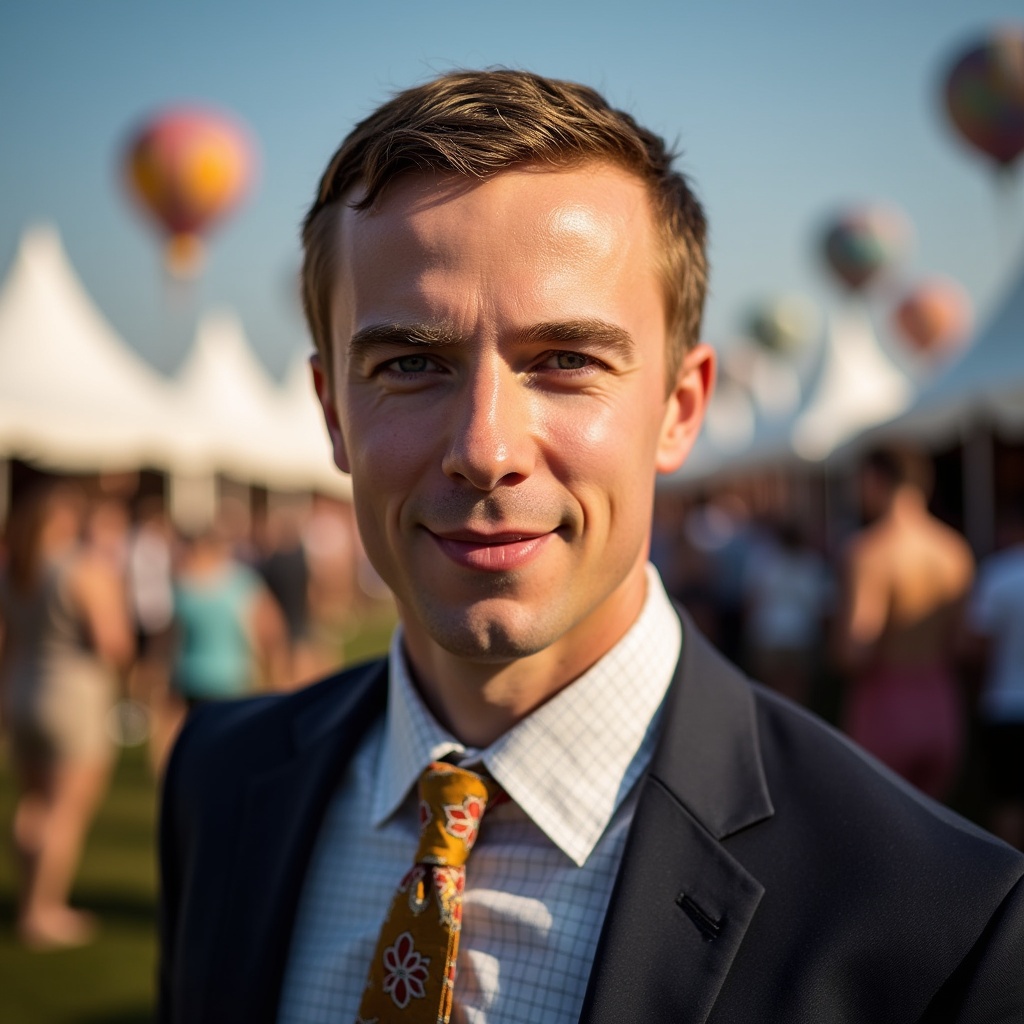}
         \end{tabular}
        } &
        {\setlength{\tabcolsep}{0pt}%
         \renewcommand{\arraystretch}{0}%
         \begin{tabular}{ccc}
            \multicolumn{3}{c}{\includegraphics[clip, viewport=128bp 192bp 896bp 960bp, width=0.128\linewidth]{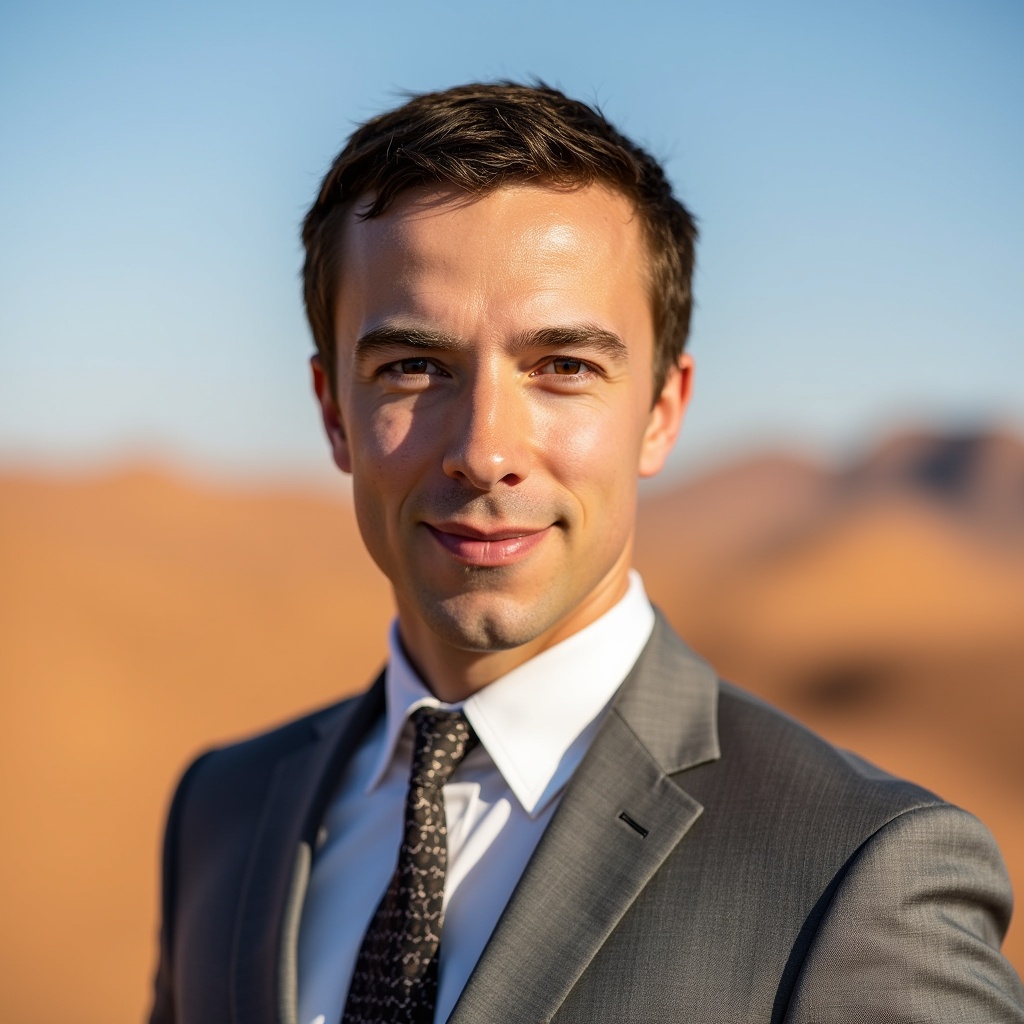}} \\
            \includegraphics[clip, viewport=128bp 192bp 896bp 960bp, width=0.043\linewidth]{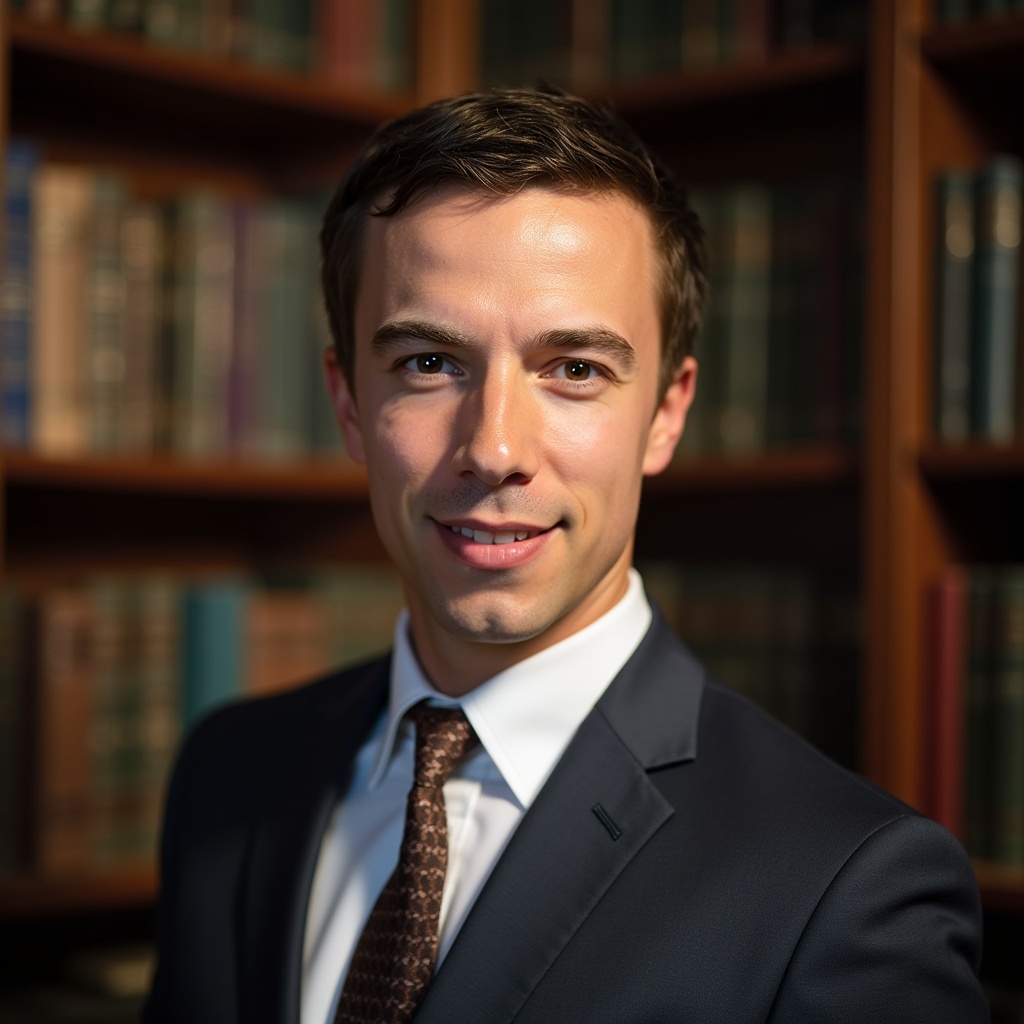} &
            \includegraphics[clip, viewport=128bp 192bp 896bp 960bp, width=0.043\linewidth]{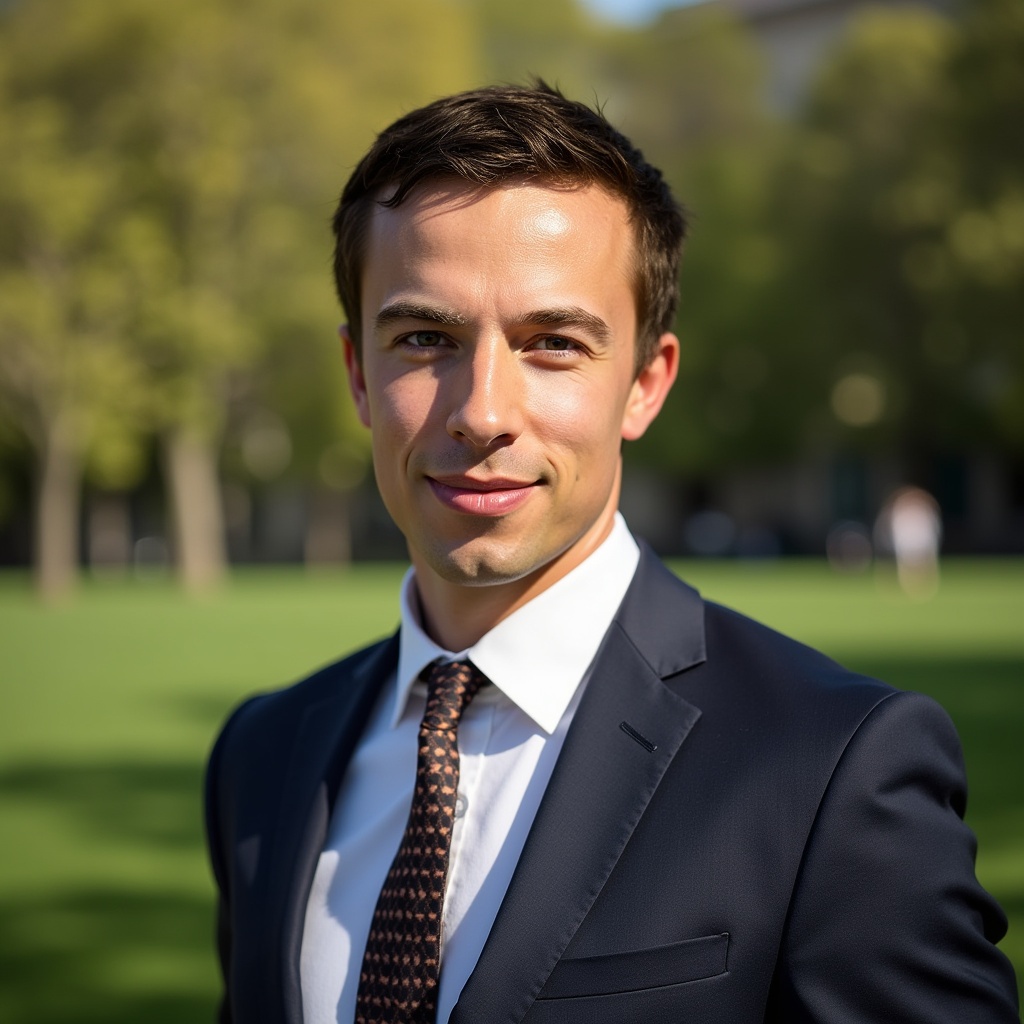} &
            \includegraphics[clip, viewport=128bp 192bp 896bp 960bp, width=0.043\linewidth]{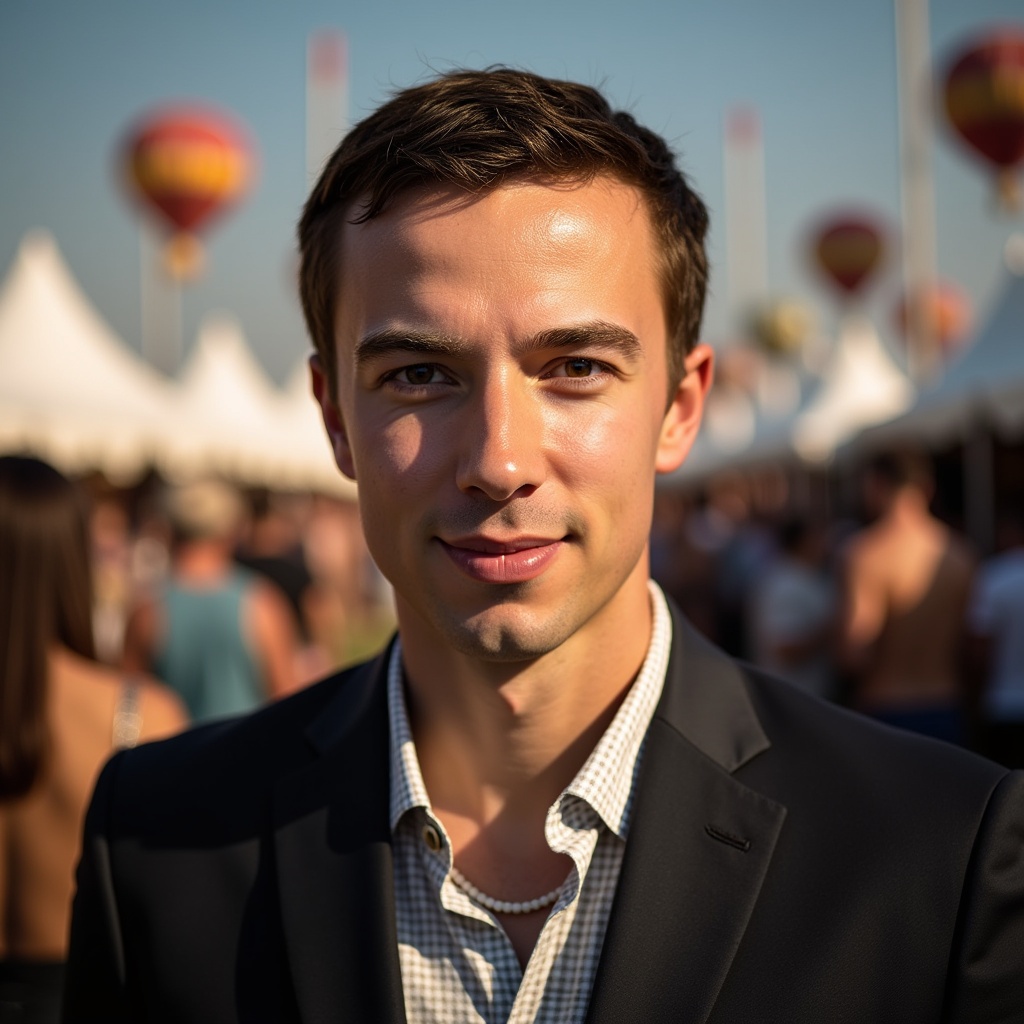}
         \end{tabular}
        } &
        {\setlength{\tabcolsep}{0pt}%
         \renewcommand{\arraystretch}{0}%
         \begin{tabular}{ccc}
            \multicolumn{3}{c}{\includegraphics[clip, viewport=128bp 192bp 896bp 960bp, width=0.128\linewidth]{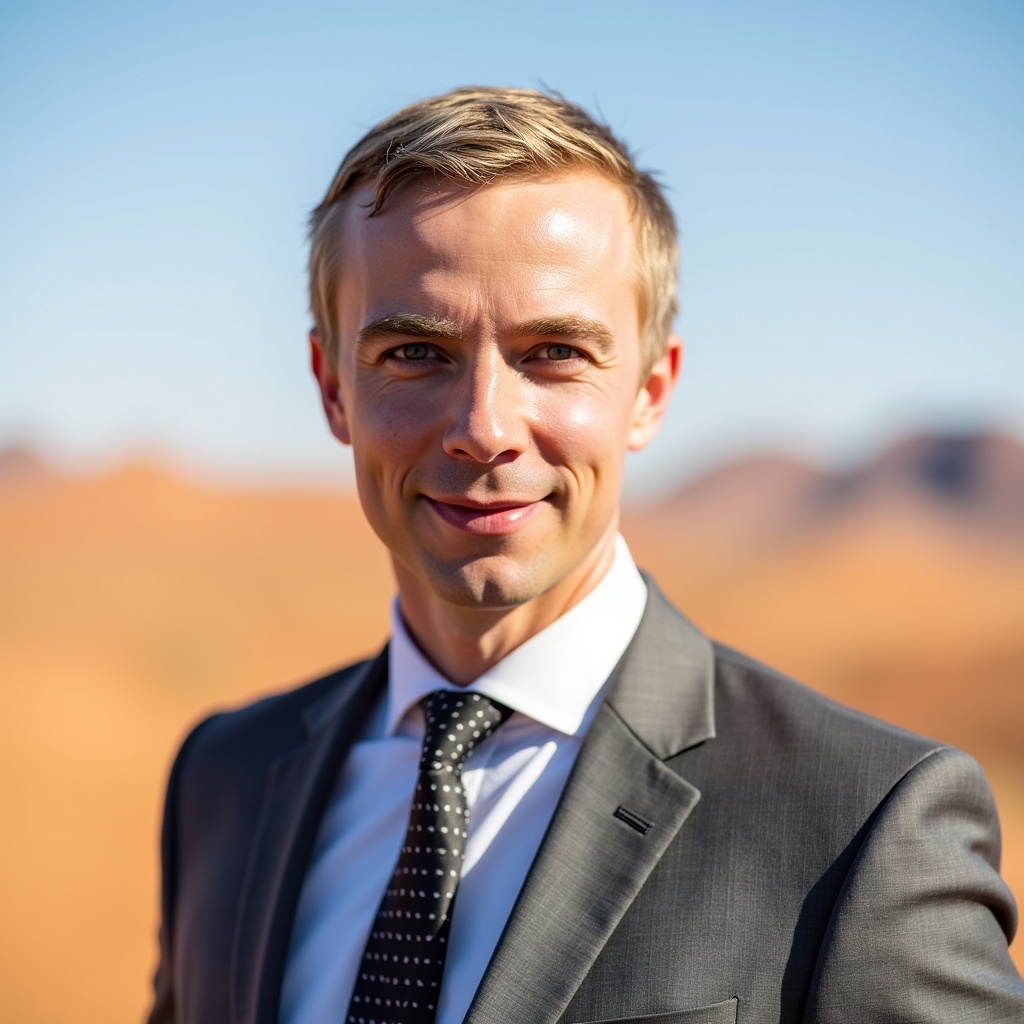}} \\
            \includegraphics[clip, viewport=128bp 192bp 896bp 960bp, width=0.043\linewidth]{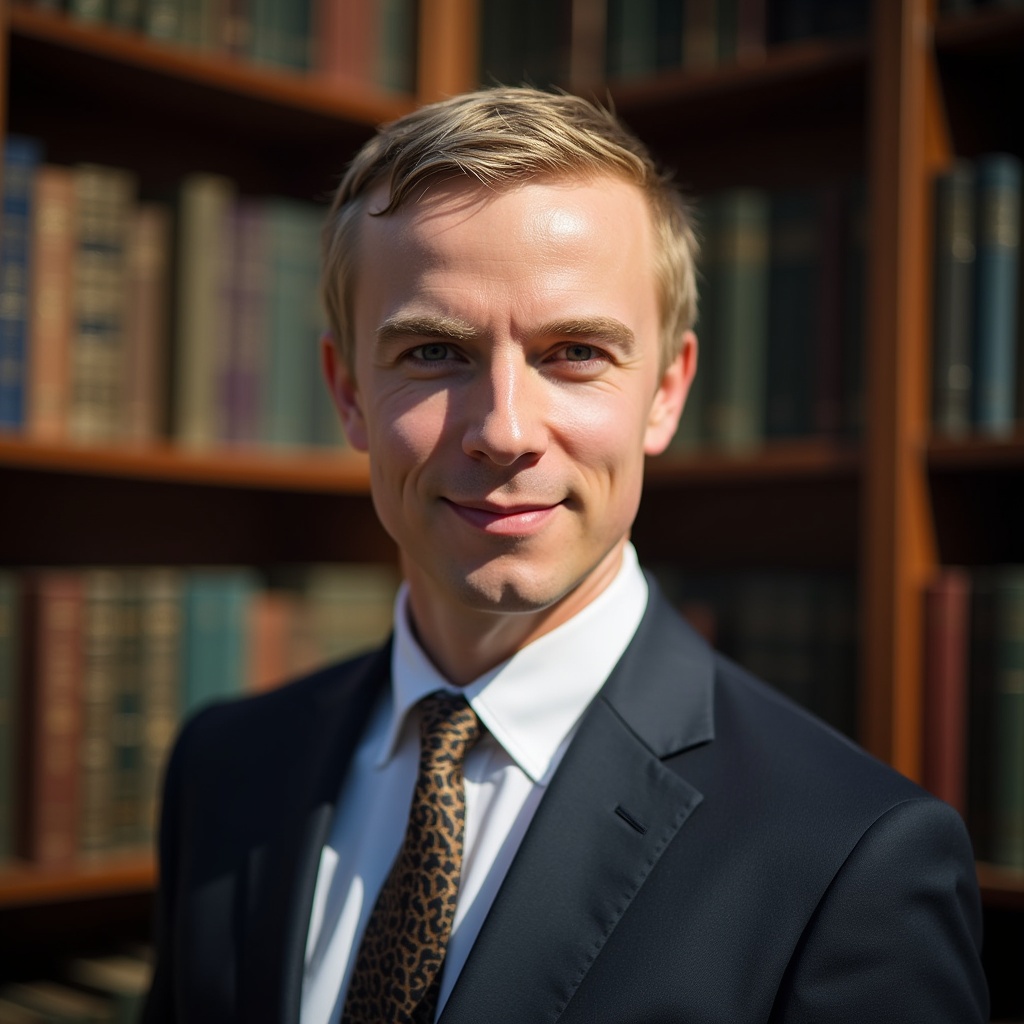} &
            \includegraphics[clip, viewport=128bp 192bp 896bp 960bp, width=0.043\linewidth]{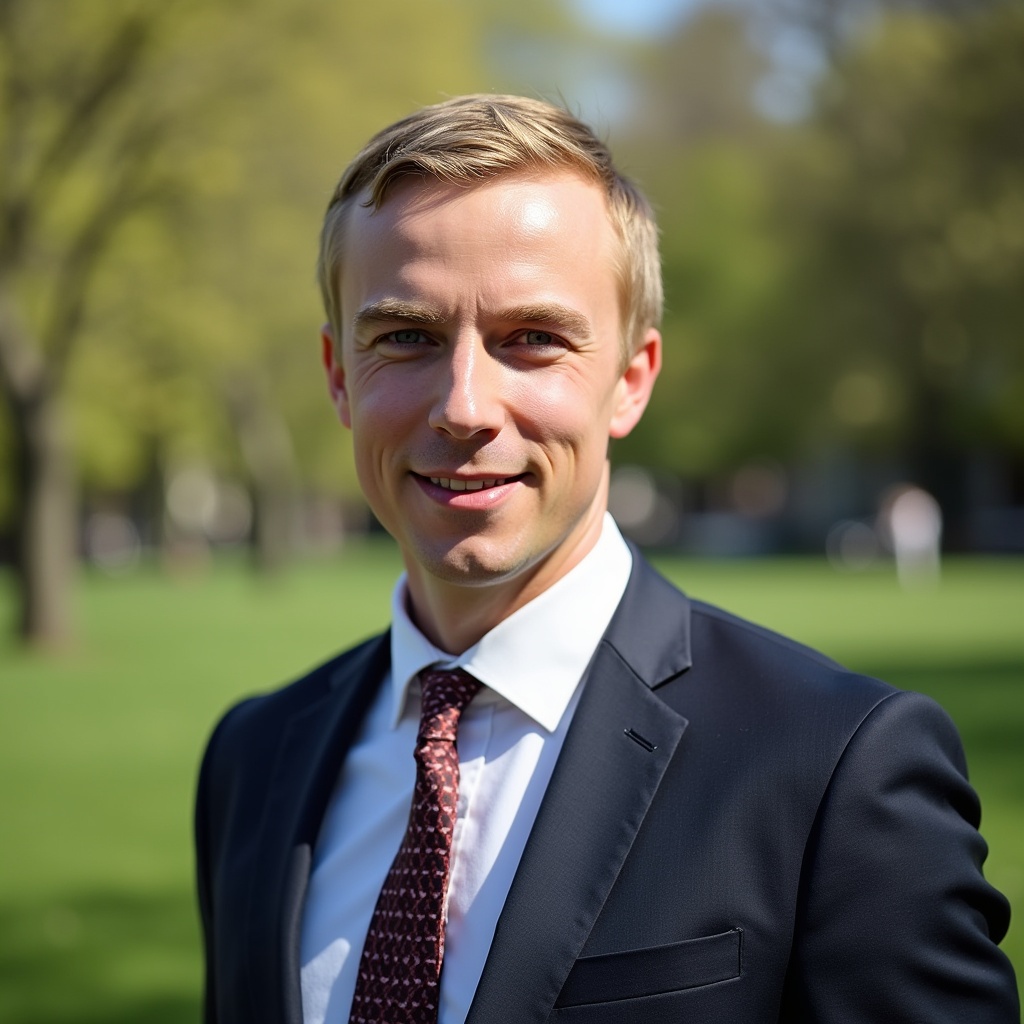} &
            \includegraphics[clip, viewport=128bp 192bp 896bp 960bp, width=0.043\linewidth]{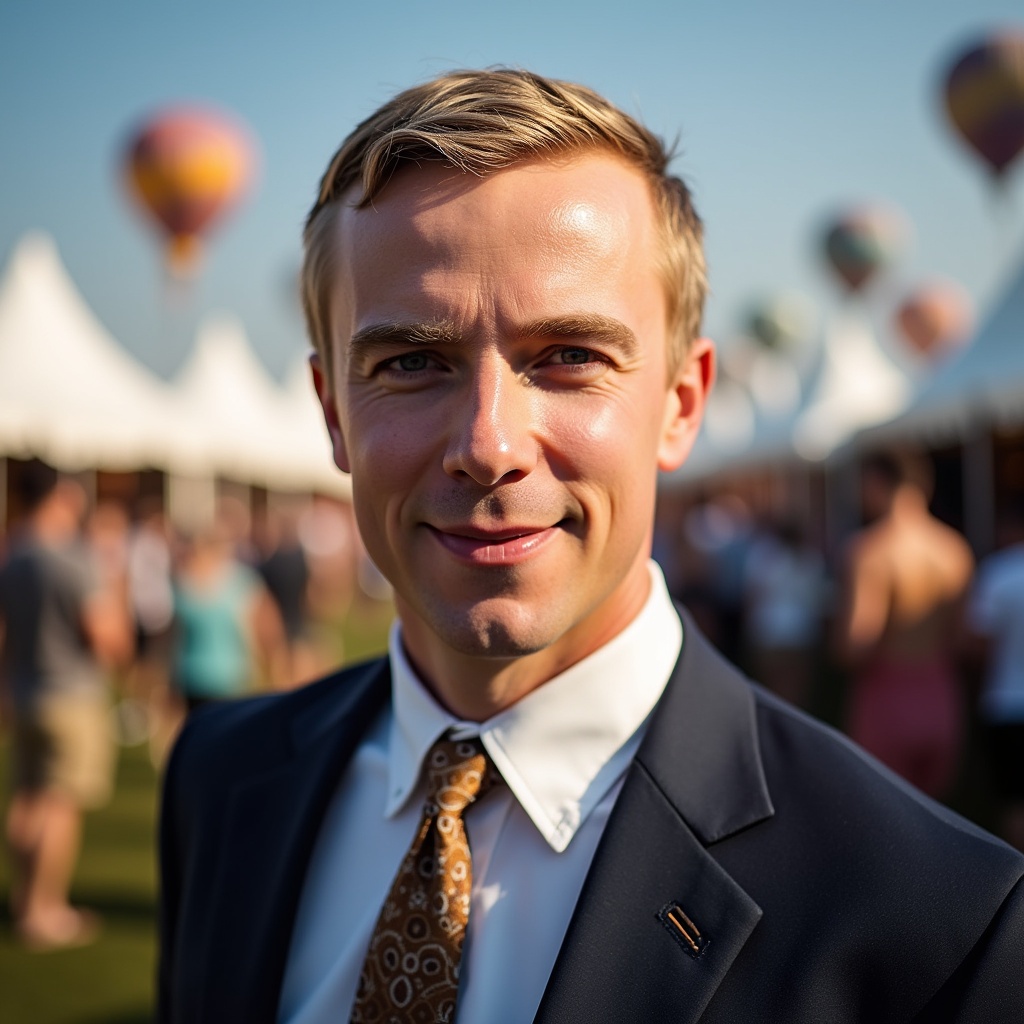}
         \end{tabular}
        } &
        {\setlength{\tabcolsep}{0pt}%
         \renewcommand{\arraystretch}{0}%
         \begin{tabular}{ccc}
            \multicolumn{3}{c}{\includegraphics[clip, viewport=128bp 192bp 896bp 960bp, width=0.128\linewidth]{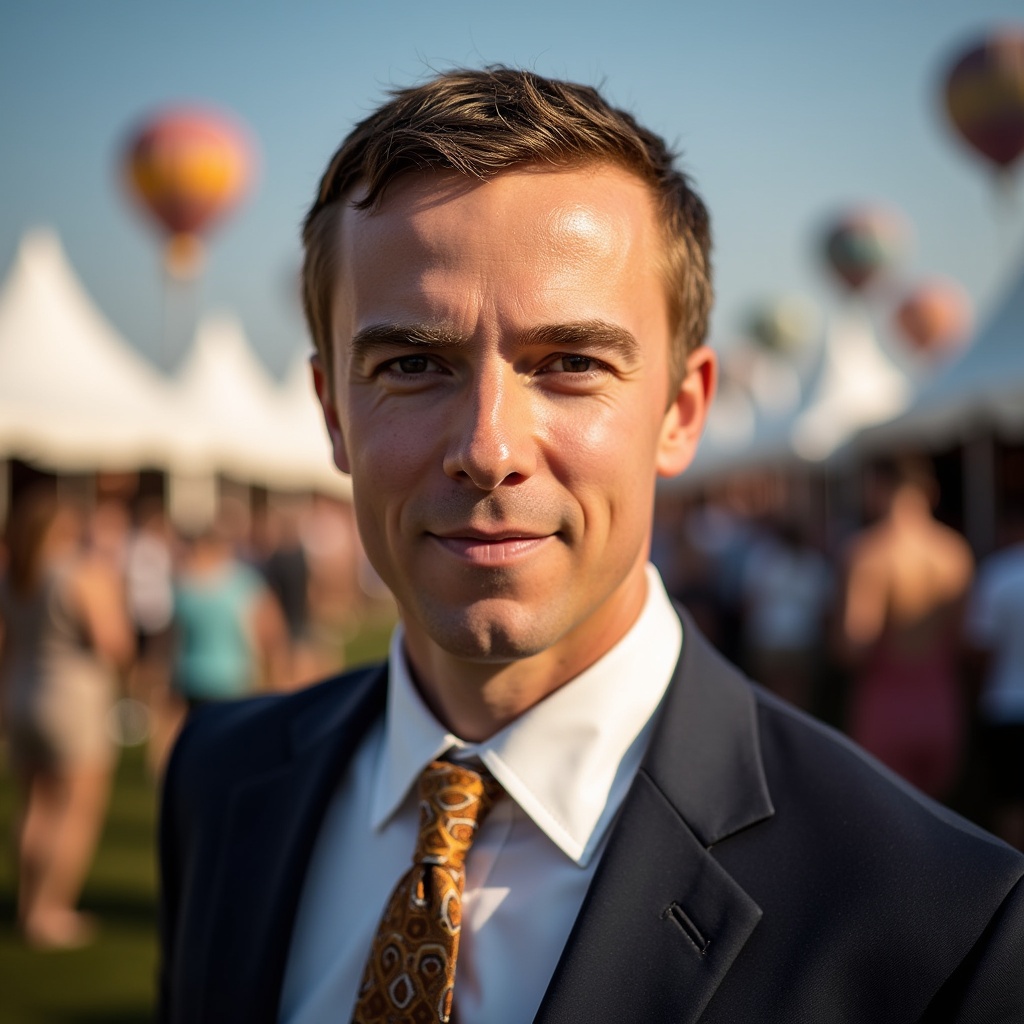}} \\
            \includegraphics[clip, viewport=128bp 192bp 896bp 960bp, width=0.043\linewidth]{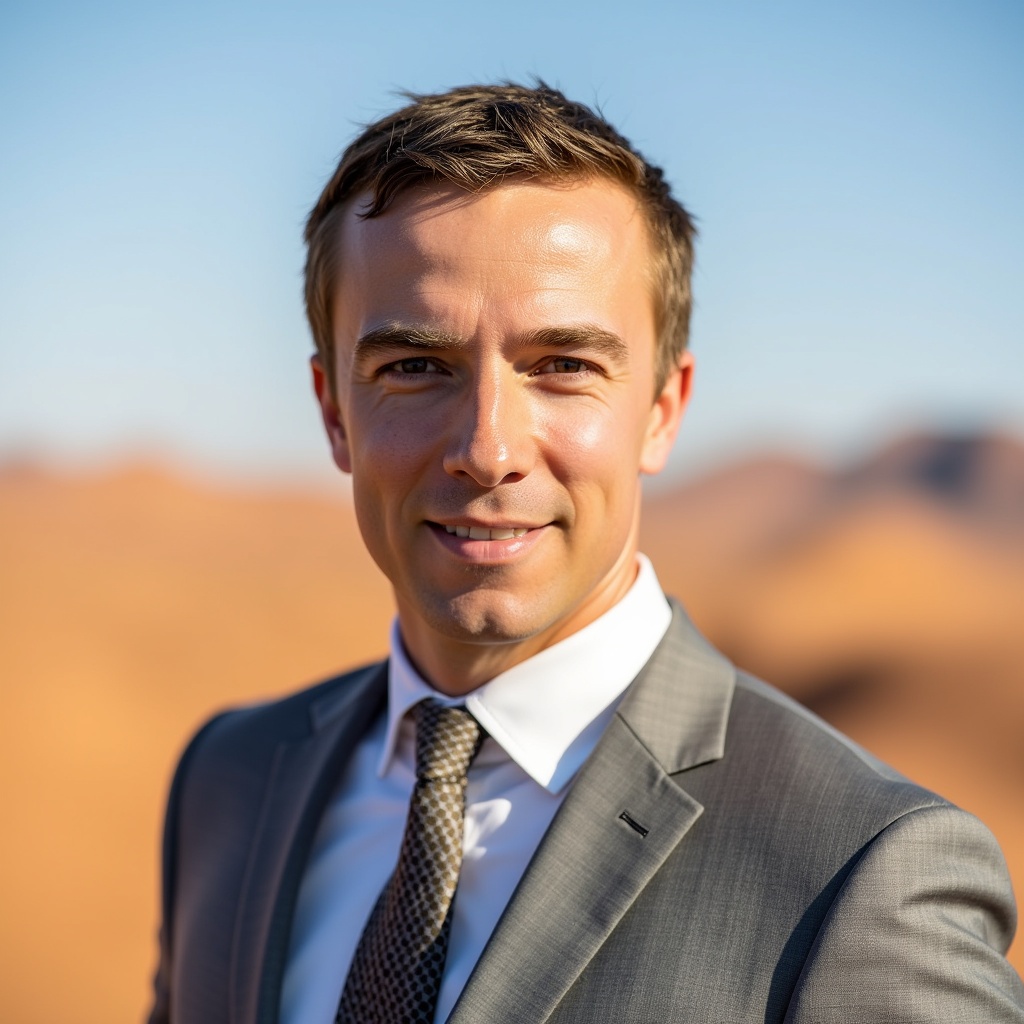} &
            \includegraphics[clip, viewport=128bp 192bp 896bp 960bp, width=0.043\linewidth]{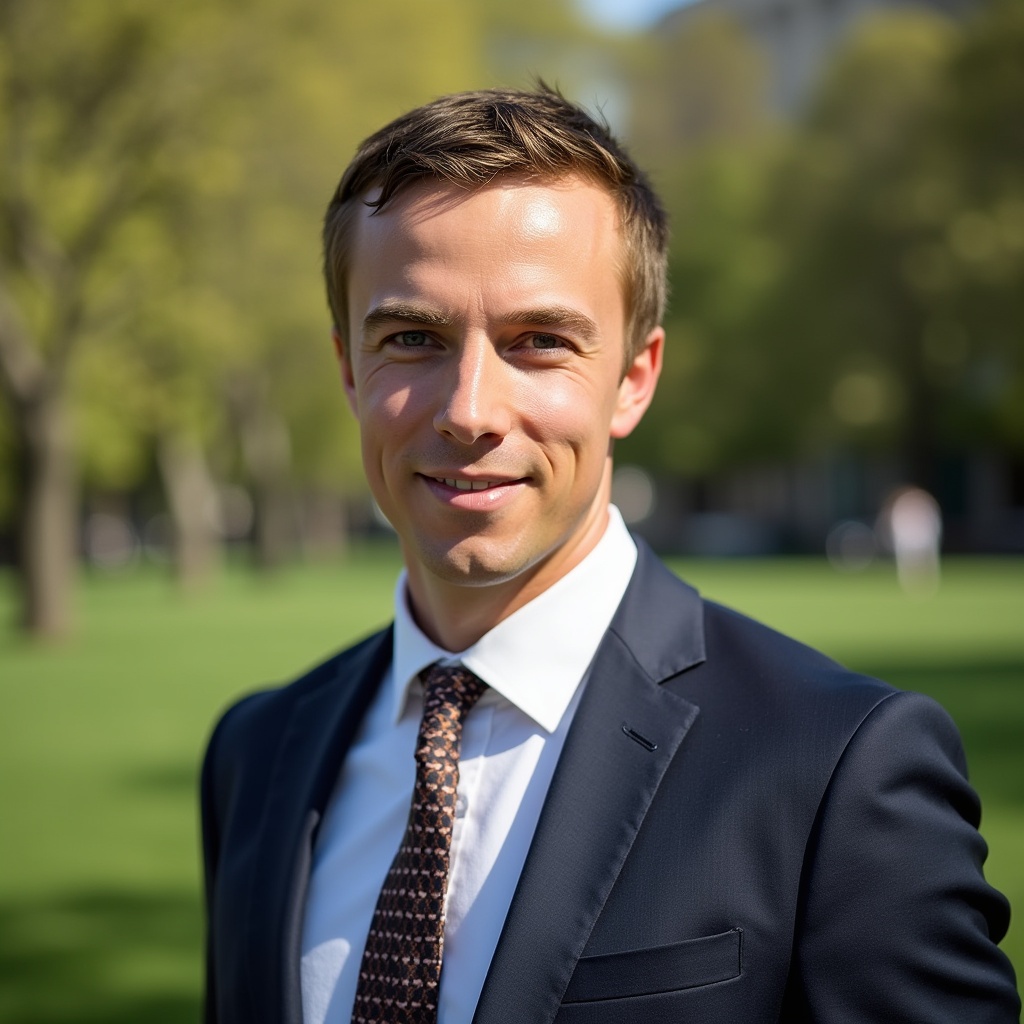} &
            \includegraphics[clip, viewport=128bp 192bp 896bp 960bp, width=0.043\linewidth]{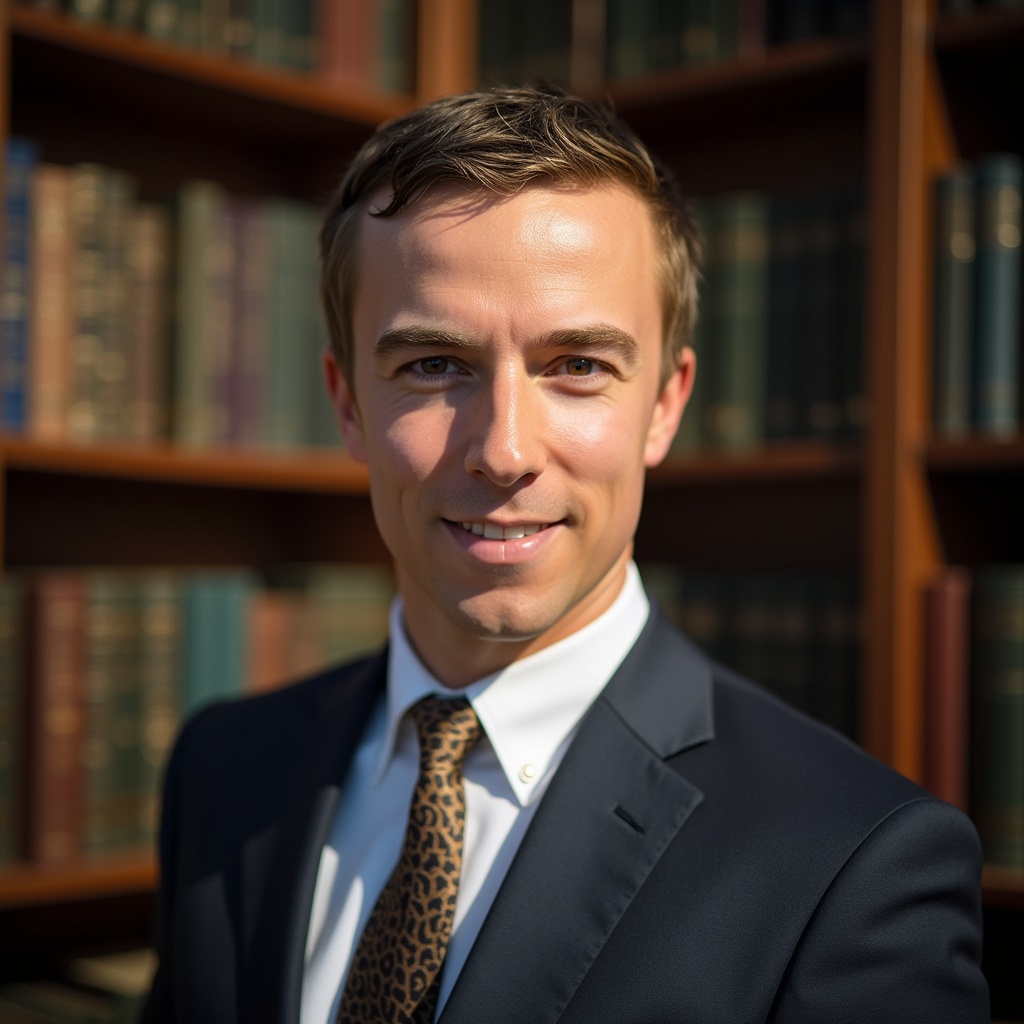}
         \end{tabular}
        } &
        {\setlength{\tabcolsep}{0pt}%
         \renewcommand{\arraystretch}{0}%
         \begin{tabular}{ccc}
            \multicolumn{3}{c}{\includegraphics[clip, viewport=128bp 192bp 896bp 960bp, width=0.128\linewidth]{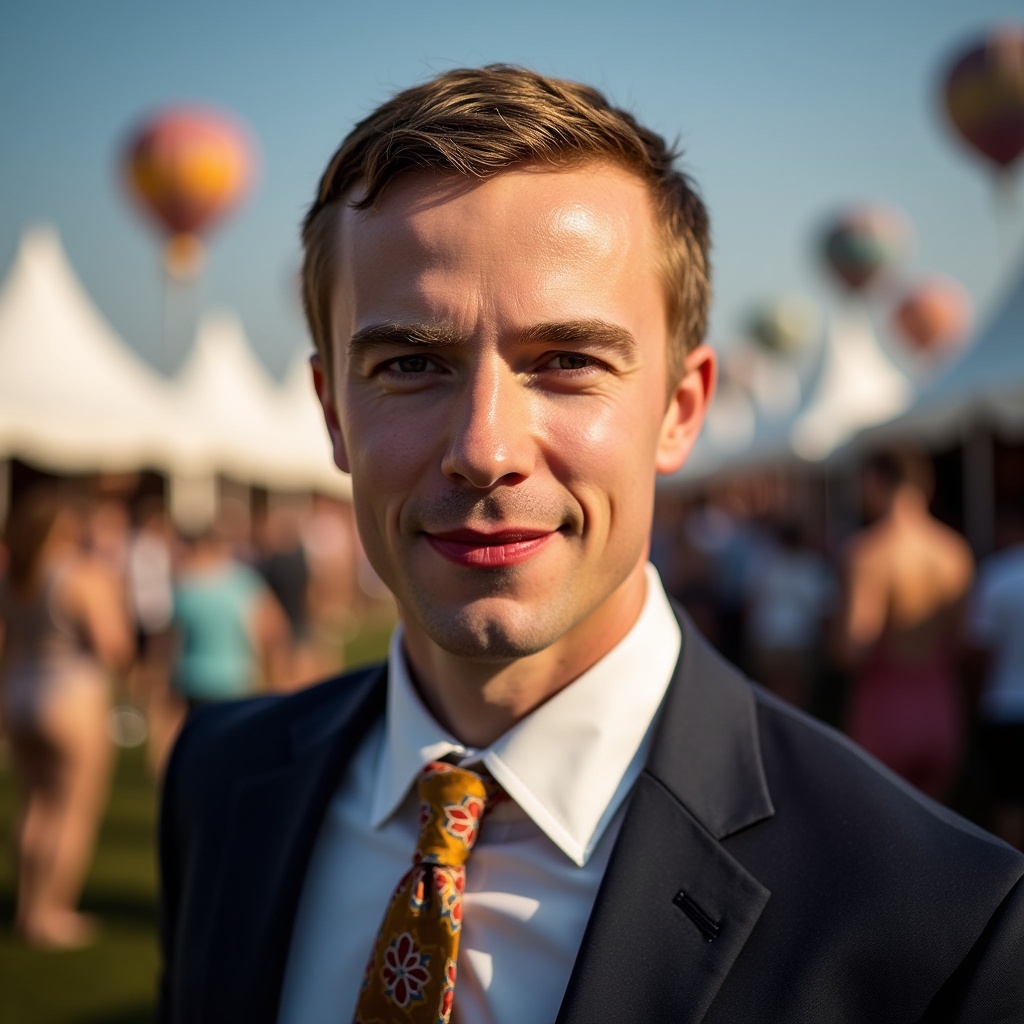}} \\
            \includegraphics[clip, viewport=128bp 192bp 896bp 960bp, width=0.043\linewidth]{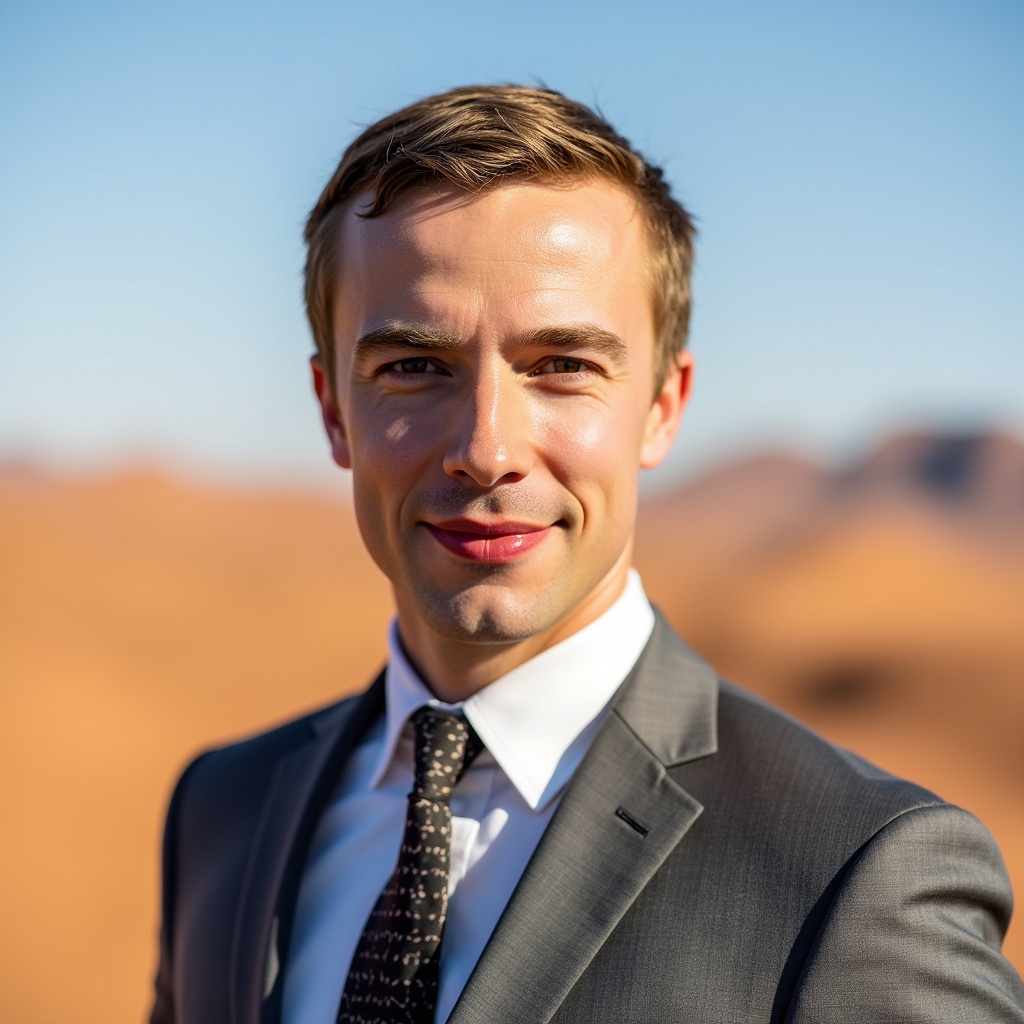} &
            \includegraphics[clip, viewport=128bp 192bp 896bp 960bp, width=0.043\linewidth]{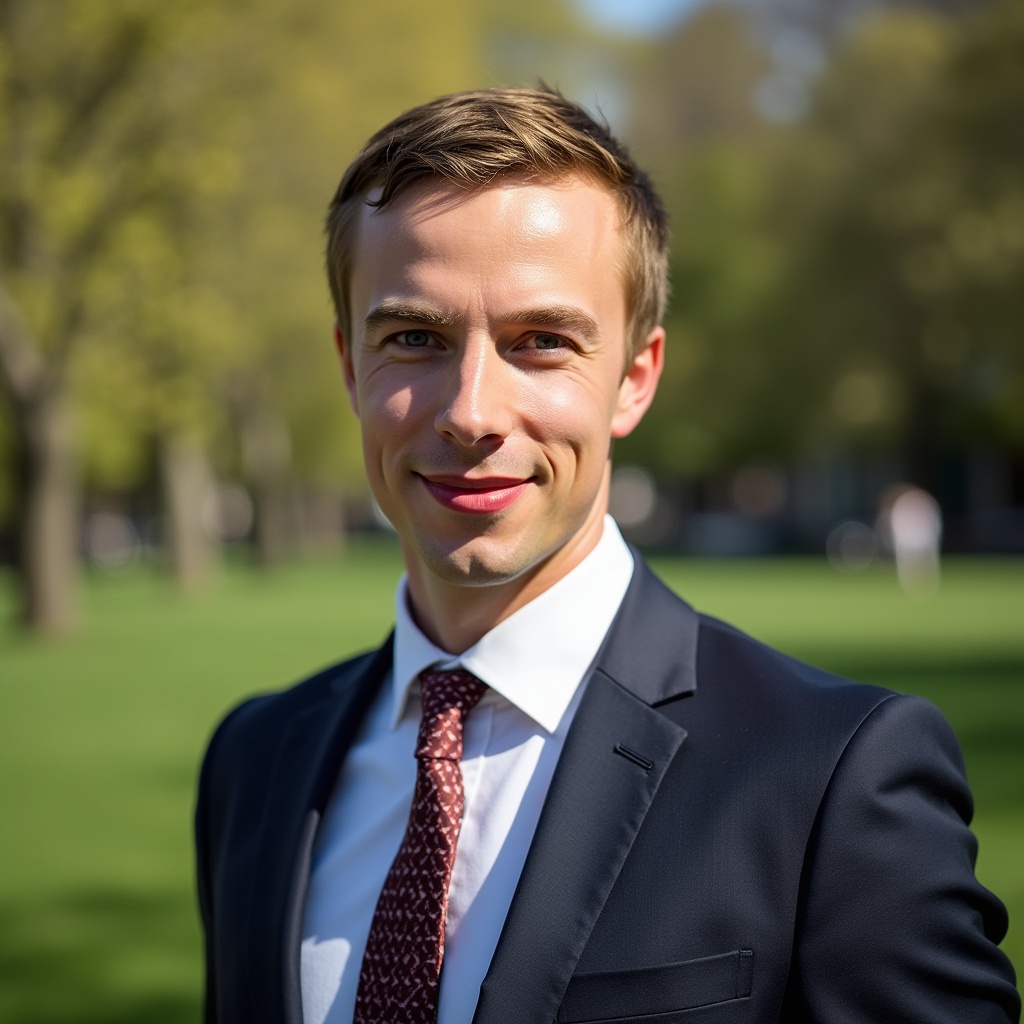} &
            \includegraphics[clip, viewport=128bp 192bp 896bp 960bp, width=0.043\linewidth]{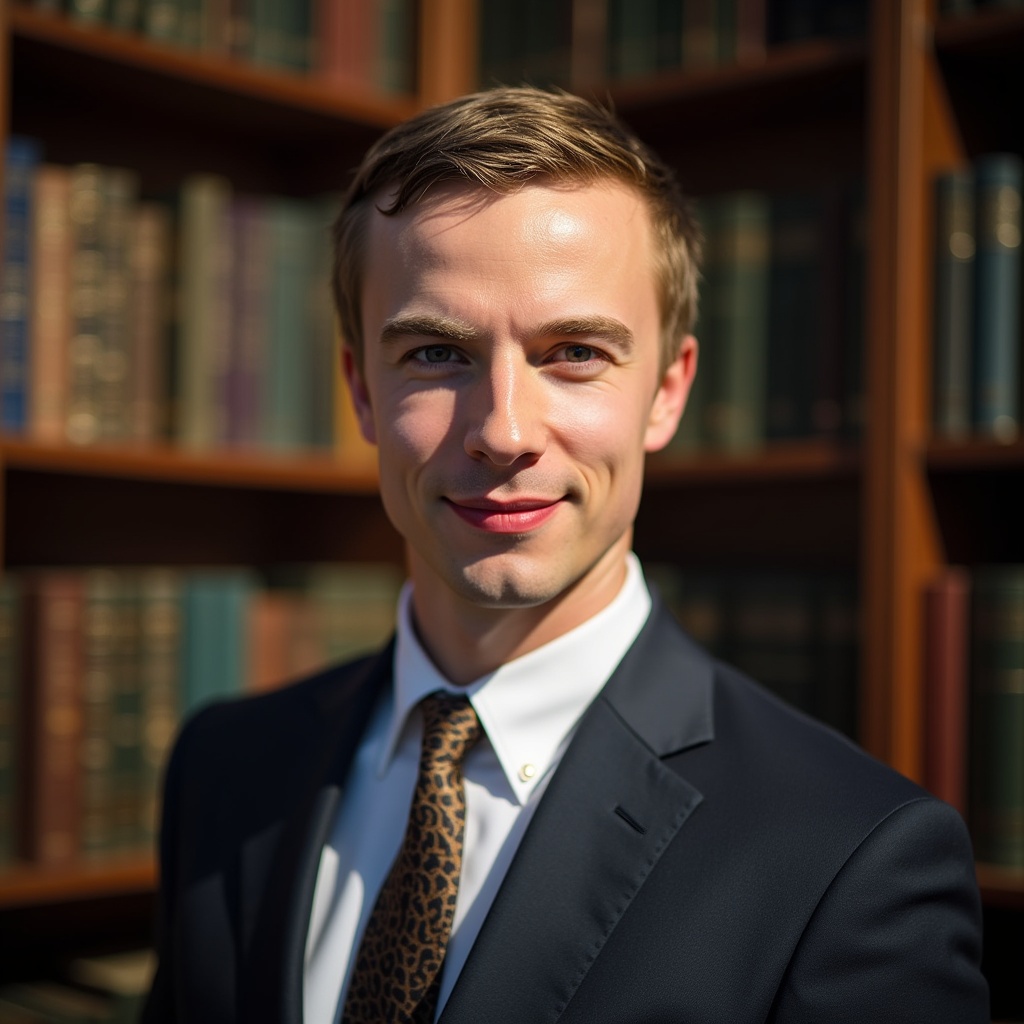}
         \end{tabular}
        } \\
        
        Original Identity &
        \multicolumn{2}{c}{$\xleftarrow{\hspace{1.5em}}$ - eye color + $\xrightarrow{\hspace{1.5em}}$} &
        \multicolumn{2}{c}{$\xleftarrow{\hspace{1.5em}}$ - hair color + $\xrightarrow{\hspace{1.5em}}$} &
        \multicolumn{2}{c}{$\xleftarrow{\hspace{1.5em}}$ - lips color + $\xrightarrow{\hspace{1.5em}}$} \\
    \end{tabular}
    }
    \vspace{-8pt}
    \caption{PCA-based controls across three regimes.
    Depending on the section, we vary a single principal direction for a single token, across all tokens, or within a targeted token group. The arrow labels name the observed attribute (eye color, age, eye openness, lip color). For every direction we display identity edits from negative to positive, and for each edited setting we render four prompts to show that the subject identity stays consistent while only the targeted attribute changes.
    }
    \label{fig:pca_table}
    \ifarxiv
        \vspace{-14pt}
    \else
        \vspace{-18pt}
    \fi
\end{figure*}

We show qualitative results of this method in
\cref{fig:pca_table}.
Setting $z=\mathrm{vec}(Z)$ (Global PCA) captures correlations across tokens and produces coherent global edits, for example smooth changes in apparent age, but the resulting directions often entangle multiple facial parts and therefore provide limited locality and limited fine-grained control.

Setting $z=Z_n$ (Per token PCA) often reveals tokens that specialize to particular facial regions and thus provides locality and interpretability, e.g., a token that controls eye color. However, controlling most individual tokens has only a minor effect, suggesting information is distributed across several tokens and it is not sufficient to control only one.

By setting $z = \mathrm{vec}(Z_S)$ and computing $S$ using our token selection method for the eye region, we achieve smooth and localized edits of the eyes, enabling control over their openness. Similarly, by selecting the tokens relevant to the lips, we can control their color.

\vspace{-5pt}
\section{Experiments}
\vspace{-5pt}
\label{sec:experiments}

In this work, we utilize Omni-ID~\cite{qian2025omniidholisticidentityrepresentation} and PuLID~\cite{guo2024pulid} as encoders for the Flux.dev~\cite{flux} text-to-image model
This entire framework is built upon the IP-Adapter architecture~\cite{ye2023ipadaptertextcompatibleimage}.
To compute PCA over identity tokens, we use the FFHQ dataset~\cite{karras2019style}, which contains 70{,}000 high-quality face images. For supervised attribute directions, we use the CelebA dataset~\cite{liu2015faceattributes}, which provides 202{,}599 images annotated with 40 binary attribute labels.
\ifarxiv
    Additional implementation details in \cref{sec:additional_imp_details}.
\else
    Additional implementation details in the Supplementary Materials.
\fi

\ifarxiv
    \vspace{-2pt}
\else
    \vspace{-5pt}
\fi
\subsection{Qualitative Results}
\ifarxiv
    \vspace{-3pt}
\else
    \vspace{-5pt}
\fi

We present qualitative results in Figures~\ref{fig:teaser}, \ref{fig:blending_results_grid}, \ref{fig:directional_edits}, and \ref{fig:pca_table}.
\cref{fig:teaser}
illustrates identity tuning: given a source image, our approach enables iterative tuning of the identity, highlighting the advantage of operating directly on the identity representation. The tuned identity produces consistent generations across diverse prompts, and the figure includes edits that are difficult to express via text, such as modifying the shape of the nose.
\cref{fig:directional_edits}
presents supervised attribute-aligned directions, which similarly transfer across different identities.
more results in
\ifarxiv
    \cref{sec:more_results_sup}.
\else
    The Supplementary Materials.
\fi
\cref{fig:blending_results_grid}
illustrates identity tuning by injecting facial parts from another person.
Identity tuning applies an edit direction in the identity latent space, scaled by a user-selected intensity.
\ifarxiv
    See \cref{sec:continues_tuning_sup} and \cref{sec:patch_editing_app_supp}
\else
    See the Supplementary Materials
\fi
for continuous tuning and further applications of patch editing beyond paired-data creation, such as inserting specific visual elements like a beard style.

\vspace{-7pt}

\vspace{-5pt}
\paragraph{Ablation on Token Granularity}

\cref{fig:pca_table} evaluates edit granularity across three control levels: a single token, a targeted token subset, and the global token space. Manipulating a single token yields a localized but minimal effect, useful mainly for isolated changes like eye color. Conversely, applying directions globally produces pronounced changes but loses fine-grained control over specific facial regions. Thus, most attributes require modifying a selected token subset. This demonstrates that tailoring token granularity to the target attribute is essential for precise, localized edits.

\ifarxiv
    \vspace{-2pt}
\else
    \vspace{-5pt}
\fi
\subsection{Comparisons}
\ifarxiv
    \vspace{-2pt}
\else
    \vspace{-5pt}
\fi

Identity tuning is largely overlooked in the existing literature, meaning few methods are directly applicable to this task. The most relevant works are Precise Control~\cite{rishubh2024precisecontrol} and Weights2Weights~\cite{dravid2024interpretingweightspacecustomized}, for which we use the authors’ official implementations and provided editing directions.
In addition, we compare against the state-of-the-art \textit{image editing} model, Flux Kontext~\cite{labs2025flux1kontextflowmatching}.
Since this model does not natively support consistent identity tuning, we adapt it in two ways. First (Kontext Direct), we input the person’s reference image and use a single prompt that combines both the desired modification (e.g., “add a beard”) and the target scene. Second, we use a two-step procedure (Kontext Sequential). We first apply the edit to the reference image, and then use that edited image as the new reference to generate the final result using the target prompt.

Our approach supports a broad range of localized and global identity edits, including patch-based changes, token directions, interpolations, and unsupervised edit directions that cannot be expressed in text. However, not all baselines support all of these edit types. 
To ensure a fair comparison, we evaluate each method only on the edits it can perform.
Our identity tuning evaluation focuses on three aspects: (i) the ability of each method to perform the required edit, (ii) the identity consistency of the tuned identity across different generations, and (iii) the prompt adherence.

\begin{figure}[t]
    \centering
    \scriptsize
    \ifarxiv
        \vspace{0pt}
    \else
        \vspace{-11pt}
    \fi
    {
    \centering
    \scriptsize
    \setlength{\tabcolsep}{0.5pt}
    \renewcommand{\arraystretch}{0.3}
    \addtolength{\belowcaptionskip}{-5pt}
    {
        \centering
        \begin{tabular}{c c c c c}
        &
        \multicolumn{2}{c}{
            \begin{tabular}{c}
                \textbf{Original} \\
                \textbf{Identity}:\,
            \end{tabular}
            \raisebox{-0.45\height}{%
                \includegraphics[width=0.15\linewidth]{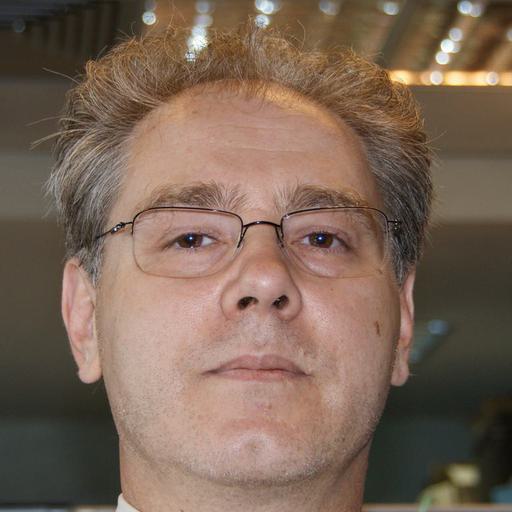}%
            }
        } &
        \multicolumn{2}{c}{\textbf{Edit}: \begin{tabular}{c} Add \\ Beard \end{tabular}} \\ \\

        \raisebox{5pt}{
            \rotatebox{90}{
            \begin{tabular}{c} PreciseControl  \end{tabular}
            }
        } &
        \includegraphics[width=0.23\linewidth]{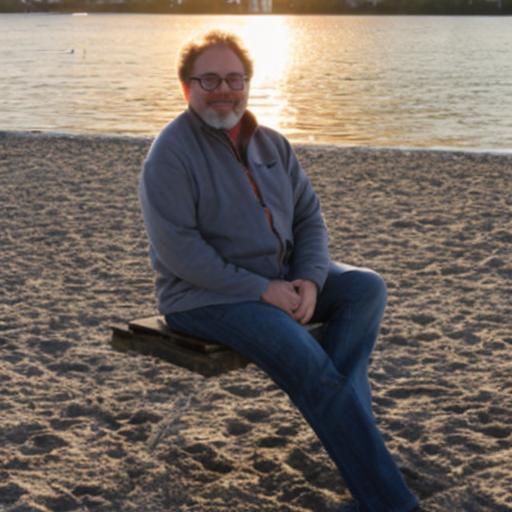} &
        \includegraphics[width=0.23\linewidth]{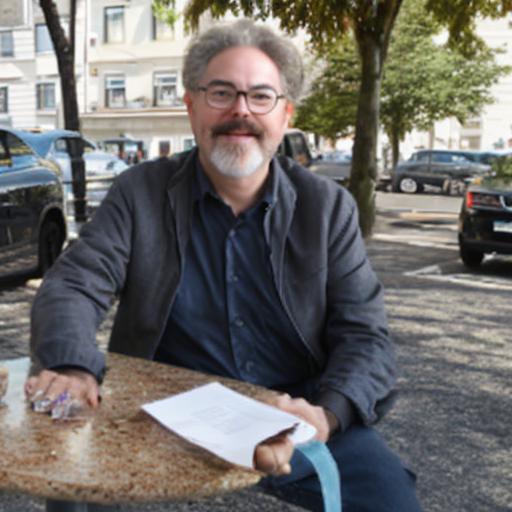} &
        \includegraphics[width=0.23\linewidth]{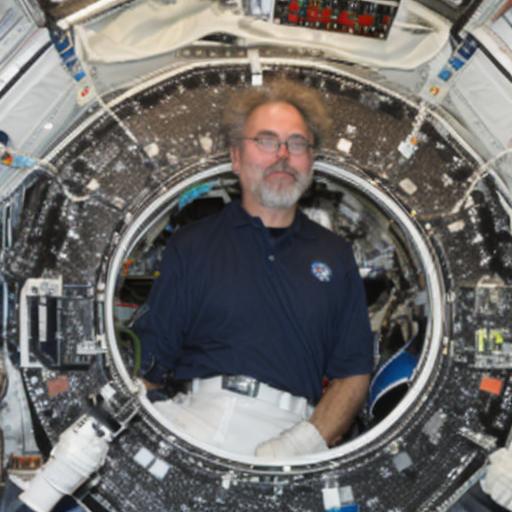} &
        \includegraphics[width=0.23\linewidth]{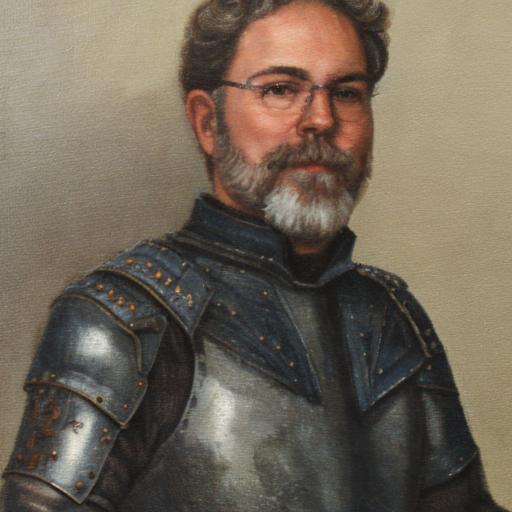} \\

        \raisebox{18pt}{
            \rotatebox{90}{
            \begin{tabular}{c} W2W \end{tabular}
            }
        } &
        \includegraphics[width=0.23\linewidth]{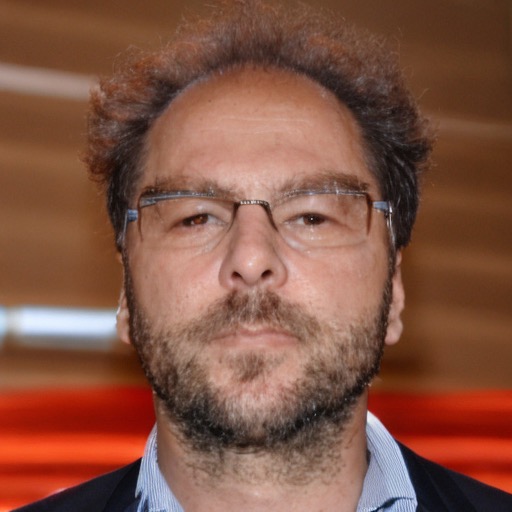} &
        \includegraphics[width=0.23\linewidth]{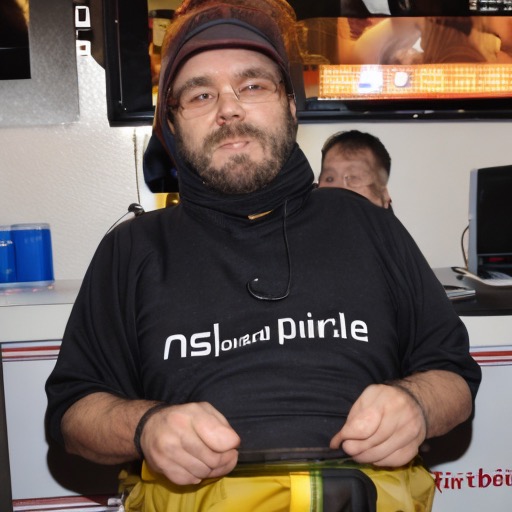} &
        \includegraphics[width=0.23\linewidth]{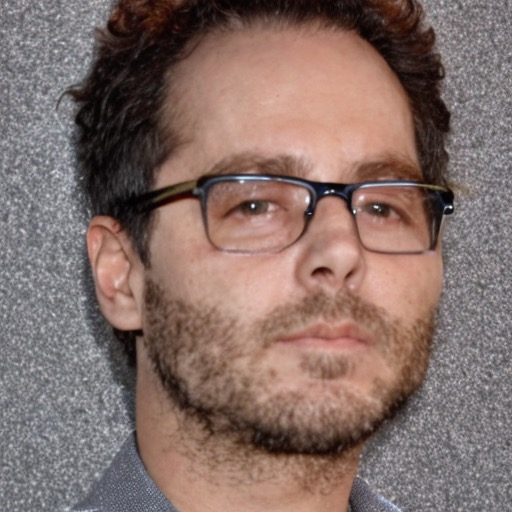} &
        \includegraphics[width=0.23\linewidth]{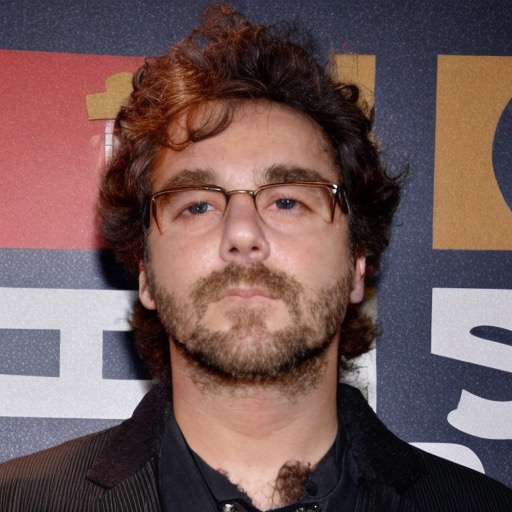} \\

        \raisebox{18pt}{
            \rotatebox{90}{
            \begin{tabular}{c} Ours  \end{tabular}
            }
        } &
        \includegraphics[width=0.23\linewidth]{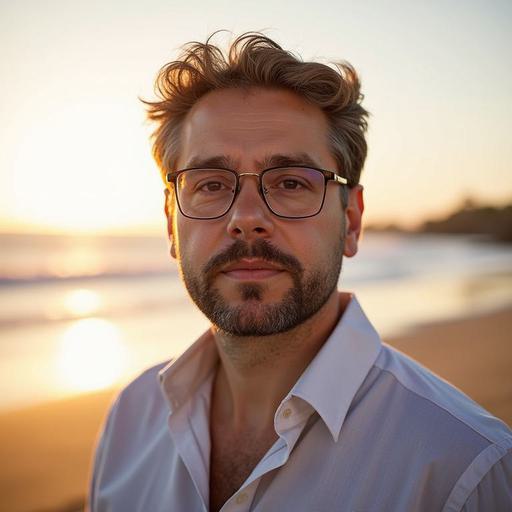} &
        \includegraphics[width=0.23\linewidth]{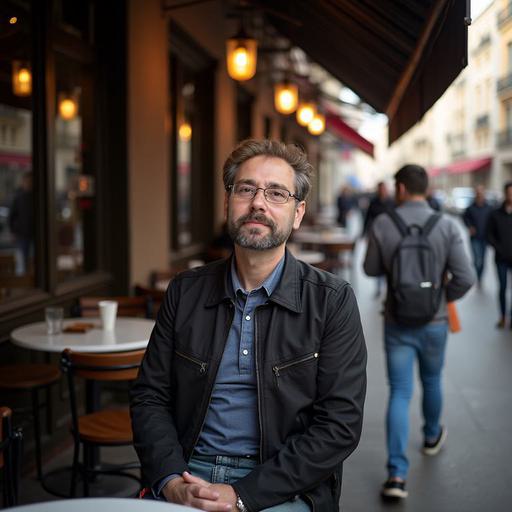} &
        \includegraphics[width=0.23\linewidth]{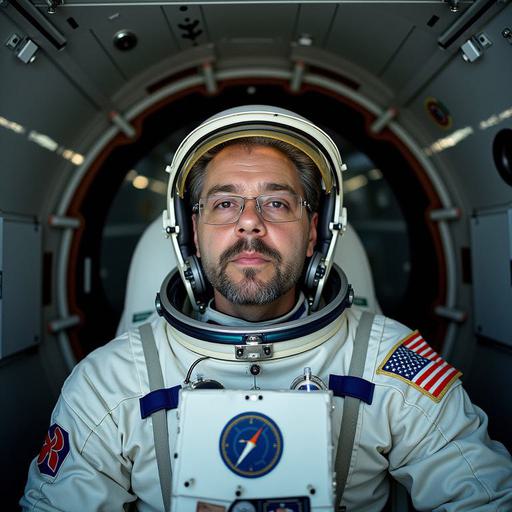} &
        \includegraphics[width=0.23\linewidth]{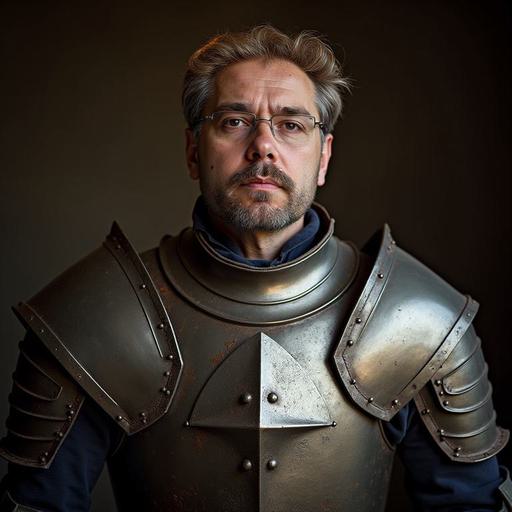} \\ %

        \end{tabular}
    }
    \vspace{-7pt}
    \caption{
    Qualitative comparison with identity-tuning methods. Each row shows an identity edit of one attribute, with four prompts as columns. Our approach consistently applies the target attribute across diverse prompts while preserving a coherent edited identity.
    }
    \label{fig:qual_compare_figure}
}

    \ifarxiv
        \vspace{-15pt}
    \else
        \vspace{-15pt}
    \fi
\end{figure}

\begin{figure}[t]
    \centering
    \scriptsize
    \vspace{-5pt}
    {
    \centering
    \scriptsize
    \setlength{\tabcolsep}{0.5pt}
    \renewcommand{\arraystretch}{0.3}
    \addtolength{\belowcaptionskip}{-5pt}
    {

    \begin{tabular}{c c c c c}
    &
    \multicolumn{2}{c}{
        \begin{tabular}{c}
            \textbf{Original} \\
            \textbf{Identity}:\,
        \end{tabular}
        \raisebox{-0.45\height}{%
            \includegraphics[width=0.15\linewidth]{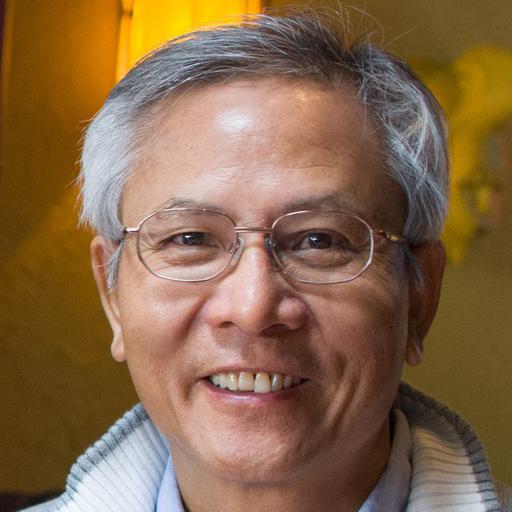}%
        }
    } &
    \multicolumn{2}{c}{
        \begin{tabular}{c}
            \textbf{Original} \\
            \textbf{Identity}:\,
        \end{tabular}
        \raisebox{-0.45\height}{%
            \includegraphics[width=0.15\linewidth]{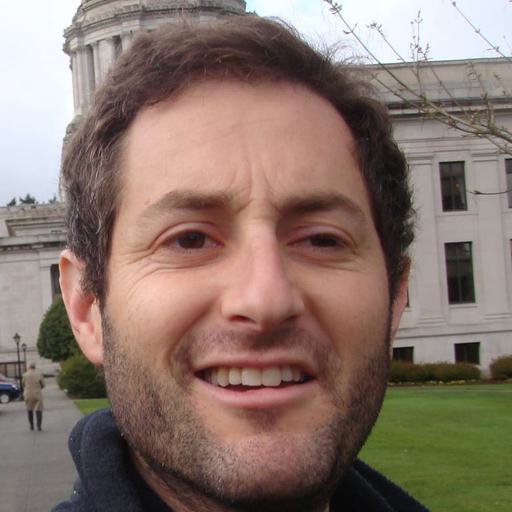}%
        }
    } \\ \\
    
    \raisebox{3pt}{
        \rotatebox{90}{
        \begin{tabular}{c} Kont. Direct \end{tabular}
        }
    } &
    \includegraphics[width=0.23\linewidth]{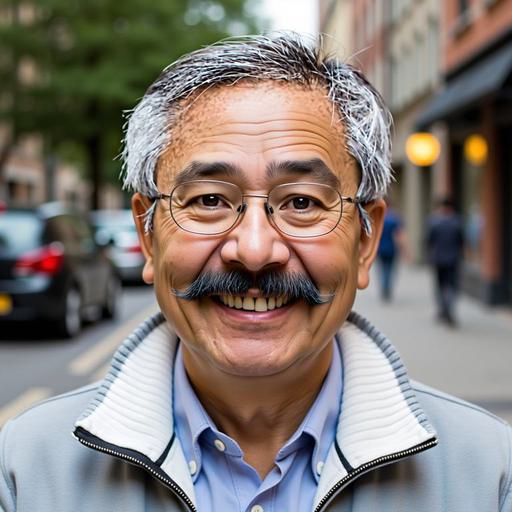} &
    \includegraphics[width=0.23\linewidth]{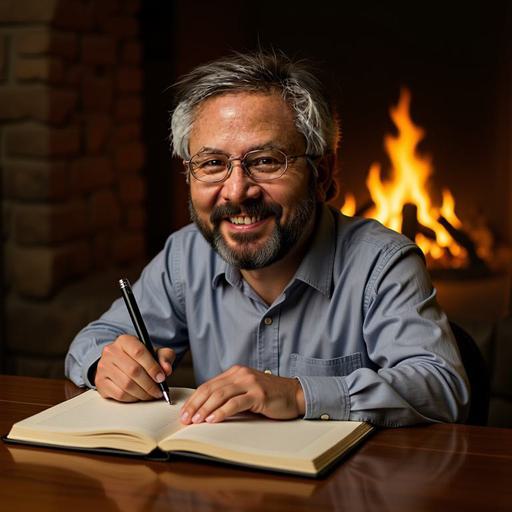} &
    \includegraphics[width=0.23\linewidth]{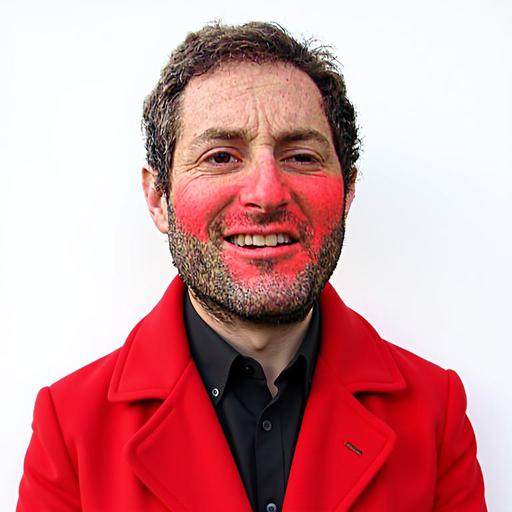} &
    \includegraphics[width=0.23\linewidth]{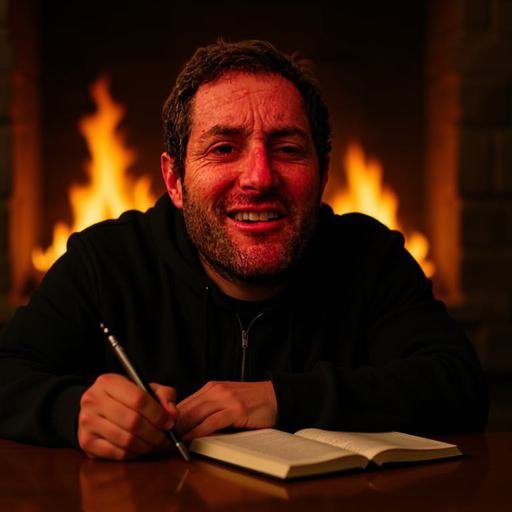} \\
    
    \raisebox{4pt}{
        \rotatebox{90}{
        \begin{tabular}{c} Kont. Seq. \end{tabular}
        }
    } &
    \includegraphics[width=0.23\linewidth]{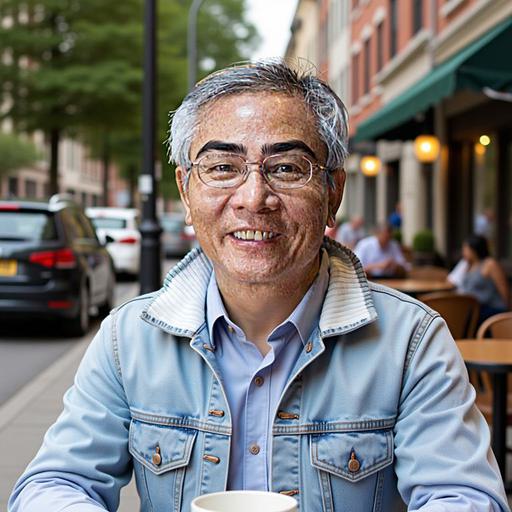} &
    \includegraphics[width=0.23\linewidth]{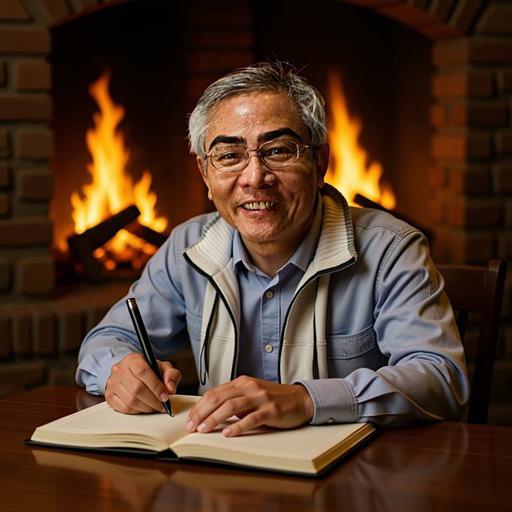} &
    \includegraphics[width=0.23\linewidth]{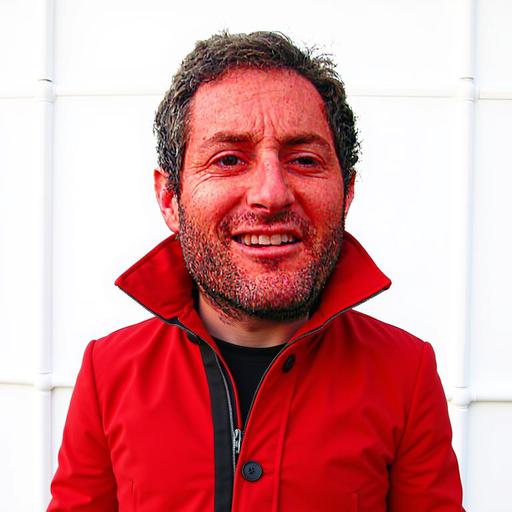} &
    \includegraphics[width=0.23\linewidth]{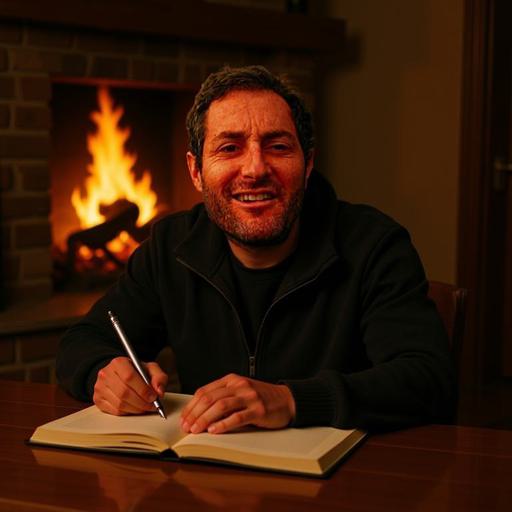} \\

    \raisebox{12pt}{
        \rotatebox{90}{
        \begin{tabular}{c} Ours  \end{tabular}
        }
    } &
    \includegraphics[clip, viewport=256bp 300bp 768bp 812bp,width=0.23\linewidth]{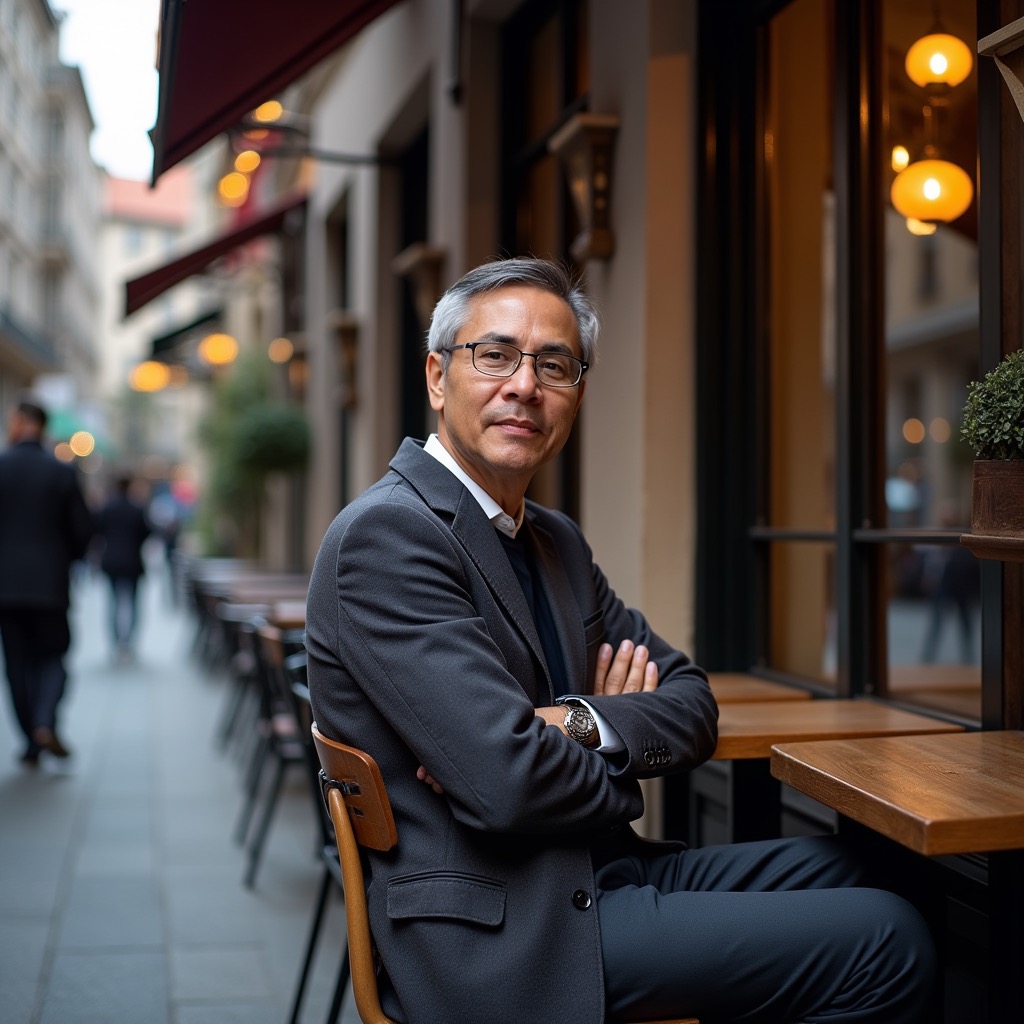} &
    \includegraphics[clip, viewport=200bp 200bp 824bp 824bp,width=0.23\linewidth]{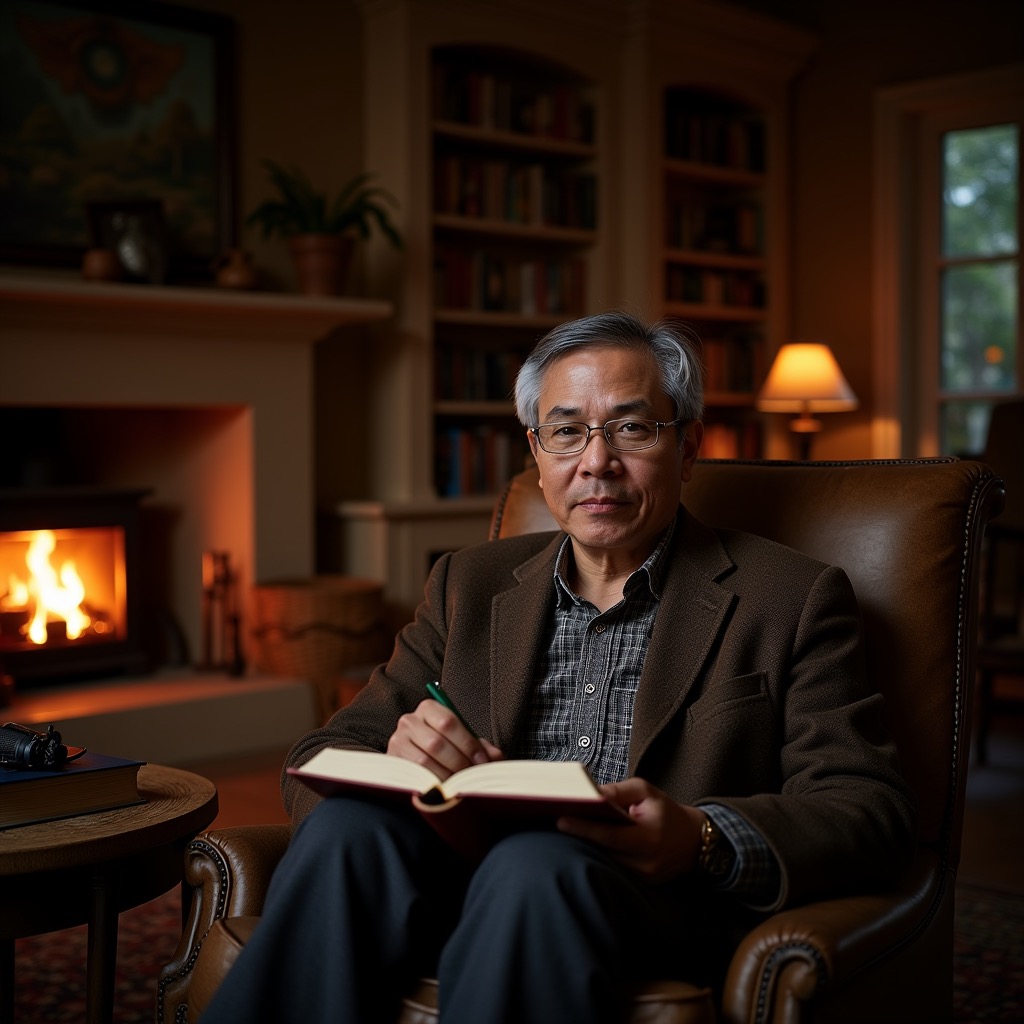} &
    \includegraphics[clip, viewport=256bp 256bp 768bp 768bp, width=0.23\linewidth]{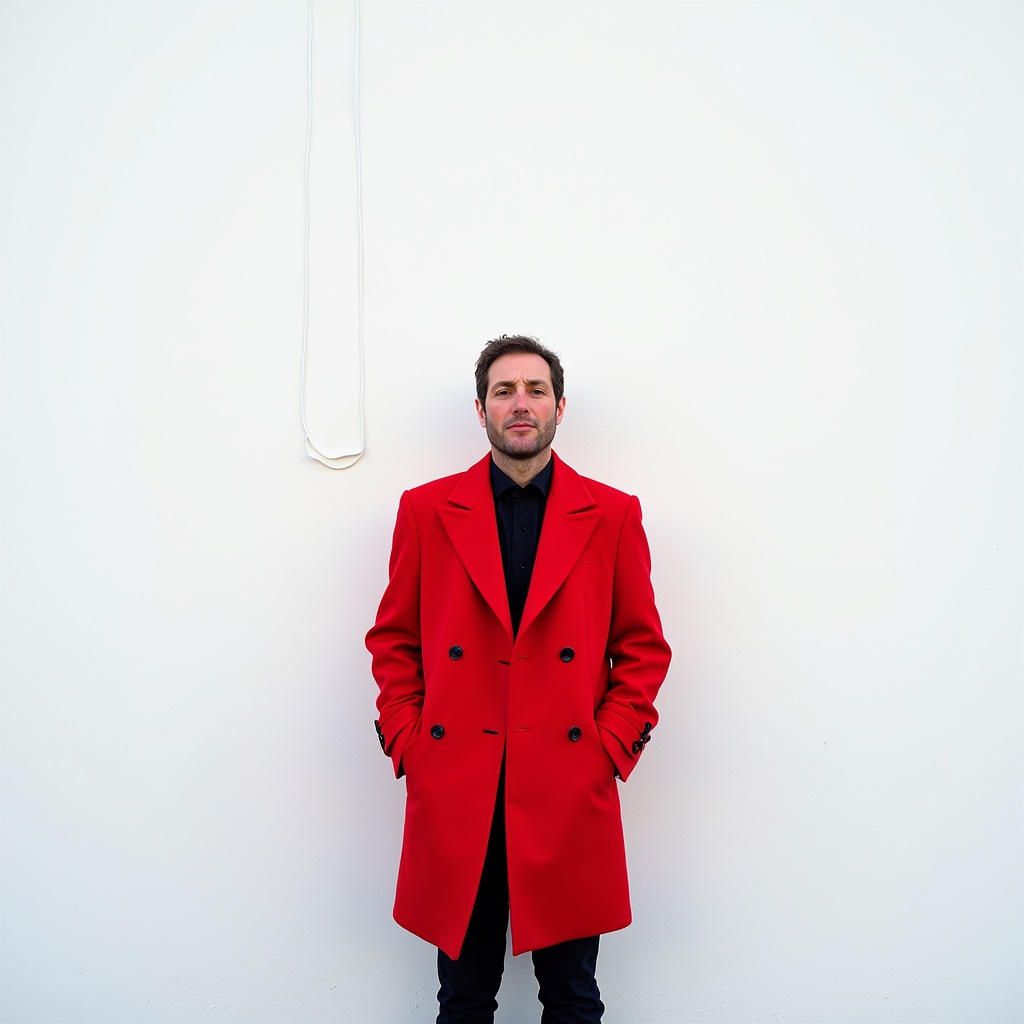} &
    \includegraphics[width=0.23\linewidth]{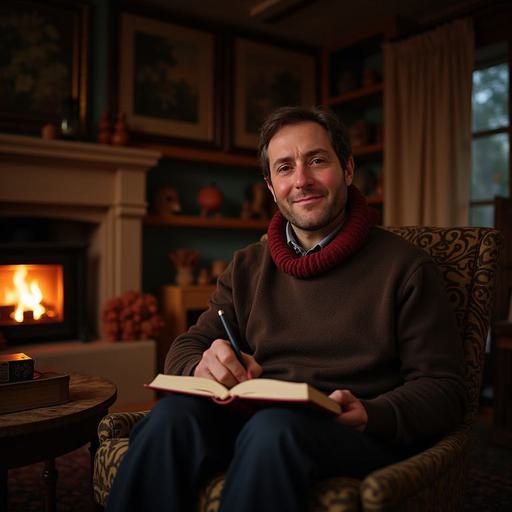} \\

    {} &
    \multicolumn{2}{c}{+Bushy Eyebrows} &
    \multicolumn{2}{c}{+Rosy Cheeks} \\
    \end{tabular}
    }
    \vspace{-7pt}
    \caption{
    Qualitative comparison of fine-grained identity.
    While FluxKontext struggles with precise local control and overall consistency, our approach achieves accurate, fine-grained, region-specific identity edits with stronger consistency.
    }
    \label{fig:qual_advcompare_figure}
}

    \vspace{-12pt}
\end{figure}

\ifarxiv
    \vspace{-8pt}
\else
    \vspace{-8pt}
\fi
\paragraph{Qualitative Comparisons}
We begin with a comparison to methods that support identity-tuning.
\cref{fig:qual_compare_figure}
presents results for PreciseControl and Weights2Weights (W2W). For each method, we generate four images under diverse prompts to evaluate identity consistency across prompts. While these methods often apply the requested edit, they do not reliably preserve the same person across prompts. In addition, W2W frequently struggles to follow the input text prompt. Additional examples and results, including those using both Flux Kontext baselines, are provided in
\ifarxiv
    \cref{sec:comparisons_sup}.
\else
    the Supplementary Materials.
\fi
Since Flux Kontext supports a broader range of edits than the identity-tuning baselines, we also compare it on more challenging, fine-grained edits shown in
\cref{fig:qual_advcompare_figure}.
Both Kontext variants (Direct and Sequential) struggle to apply the modifications locally and often introduce global artifacts. In contrast, our method produces localized, semantically aligned changes while maintaining a coherent identity across prompts.

\ifarxiv
    \vspace{-1pt}
\else
    \vspace{-15pt}
\fi
\paragraph{Quantitative Comparisons}

To evaluate identity consistency, edit adherence, and prompt adherence, we construct a custom benchmark.
It spans basic edits and advanced edits requiring finer control (unsupported by PreciseControl and partially supported by W2W), yielding over 3,000 generated images per method.
\ifarxiv
    See \cref{sec:comparisons_sup}
\else
    See Supplementary Materials
\fi
for details.
We evaluatealong three axes.
\textit{Identity consistency}: For each identity-edit pair, we compute ArcFace~\cite{Deng_2022} embeddings for the 13 generated images, measure their mean pairwise cosine similarity, and average over all pairs.
\textit{Edit adherence}: Following~\cite{lin2024evaluatingtexttovisualgenerationimagetotext}, a VQA-based judge scores adherence to the requested edit.
\textit{Prompt adherence}: Using the same protocol, we score alignment with the target prompt.

\cref{tb:quantitative}
reports results for both edit types. For Precise Control and Weights2Weights, which support identity tuning, our approach achieves substantially higher identity consistency, higher edit adherence, and higher prompt adherence.
For Flux Kontext Direct, although edit adherence is comparable, our method attains significantly higher identity consistency. Flux Kontext Sequential achieves identity consistency close to ours but underperforms in edit adherence, especially on advanced edits requiring fine-grained control. Both baselines also obtain lower prompt-adherence scores.
Finally, while the metrics indicate that Kontext adheres to the requested edits, in practice it introduces many artifacts that the metrics do not capture, as shown in
\cref{fig:qual_advcompare_figure}.

\begin{table}
    \centering
    \footnotesize{
        \centering
        \setlength{\tabcolsep}{2.5pt}
        \caption{
            Quantitative Comparison. We assess ID consistency, edit and prompt adherence for methods on basic and advanced edits.
        } 
        \label{tb:quantitative}
        \vspace{-7pt}
        \begin{tabular}{l | c c c | c c c } 
            \toprule
            & \multicolumn{3}{c}{Basic Edits} & \multicolumn{3}{c}{Advanced Edits} \\
            \midrule
            & ID & Edit & Prompt &
            ID & Edit & Prompt \\
            & Consis. $\uparrow$& Adher. $\uparrow$ & Adher. $\uparrow$ &
            Consis. $\uparrow$& Adher. $\uparrow$ & Adher. $\uparrow$ \\
            \midrule
            PreciseControl & 0.25 & 0.66 & 0.75 & N/A & N/A & N/A \\
            W2W & 0.31 & 0.64 & 0.27 & 0.32 & 0.71 & 0.22 \\
            Kontext Direct & 0.37 & \textbf{0.81} & 0.85  & 0.39 & \textbf{0.78} & 0.81 \\
            Kontext Seq. & \textbf{0.51} & 0.74 & 0.82  & \textbf{0.49} & 0.71 & 0.80 \\
            \textbf{Ours} (OmniID) & 0.47 & \underline{0.79} & \textbf{0.91} & 0.47 & \textbf{0.78} & \textbf{0.89} \\
            \textbf{Ours} (PuLID) & \underline{0.49} & \textbf{0.81} & \underline{0.88}& \underline{0.48} & \underline{0.77} & \underline{0.86} \\
            \bottomrule
        \end{tabular}
    }
\end{table}

\begin{table}
    \footnotesize
    \vspace{-11pt}
    \centering
    \caption{
        User Study. Pairwise win rate of our approach vs. others.
    }
    \vspace{-8pt}
    \begin{tabular}{lccc}
        \toprule
        & W2W & Kont. Direct & Kont. Seq. \\
        \midrule
        ID Preservation   & $78\%$ & $75\%$ & $70\%$ \\
        ID Coherent       & $85\%$ & $82\%$ & $76\%$ \\
        Edit Adher.          & $94\%$ & $78\%$ & $83\%$ \\
        Prompt Adher.        & $96\%$ & $88\%$ & $85\%$ \\
        Overall Prefer.      & $94\%$ & $85\%$ & $83\%$ \\
        \bottomrule
    \end{tabular}
    \label{tb:user_study}
    \vspace{-13pt}
\end{table}

\vspace{-10pt}
\paragraph{User Study}
To complement our quantitative analysis, we conducted a user study comparing our method with baselines that support advanced edits.
In a pairwise comparison, participants saw (i) a source image of a person, (ii) a textual edit instruction, and (iii) edited results from our method and a competing method. For each method, we displayed two results generated with different prompts. Participants provided pairwise preferences on five criteria: Identity Preservation, Identity Consistency, Edit Adherence, Prompt Adherence, and Overall Preference.
\ifarxiv
    See the question example in \cref{sec:comparisons_sup}.
\else
    See the question example in the supplementary materials.
\fi
In total, we collected 250 pairwise judgments.
As summarized in
\cref{tb:user_study},
our method was preferred over all baselines across all categories.
These results indicate that users judged our approach to better preserve the original identity, maintain consistency across prompts, and adhere to both the specified edit and the target prompt.
That indicates the superiority of our approach for the task of identity tuning. 

\vspace{-5pt}
\section{Conclusions}
\vspace{-3pt}
\label{sec:conclusions}
In this work, we explored the latent identity space of a Q-Former–based personalization encoder and introduced a framework for fine-grained, localized, continuous identity tuning. We showed that the identity tokens form a structured, spatially meaningful space capturing distinct facial attributes. Leveraging this structure, we identified region-relevant tokens and extracted editing directions within their subspaces, enabling precise, region-aware manipulation.

While prior work focused on coarse identity preservation, our framework enables subtle, continuous adjustments, such as eye openness, freckle texture, and facial-hair density. Our results demonstrate that careful exploration of the identity space enable these nuanced edits while maintaining consistency across generated images.

Overall, our findings reveal that the identity token space is both interpretable and controllable at fine facial granularity, allowing users to refine delicate aspects of visual identity. Future work may pursue perceptually aligned metrics for subtle edits and interactive workflows that support precise user-guided adjustments.

\begin{acks}
We would like to thank Omer Dahari, Etai Sella, Shelly Golan, Sara Dorfman, Kfir Aberman, and Jackson Wang for their early feedback and insightful discussions.
This research was supported in part by the Israel Science Foundation (grants no. 2492/20 and 1473/24), Len Blavatnik, and the Blavatnik Family Foundation. We also thank NVIDIA for their generous support through the NVIDIA Academic Grant Program, which provided GPU hours via Brev for this research.
\end{acks}

{
    \small
    \bibliographystyle{ieeenat_fullname}
    \bibliography{main}
}

\clearpage
\begin{appendices}
    
\twocolumn[{%
    \renewcommand\twocolumn[1][]{#1}%
    \ifarxiv
    \else
        \maketitle
    \fi
    \begin{center}
        \vspace{-27pt}
        \includegraphics[width=1.0\textwidth]{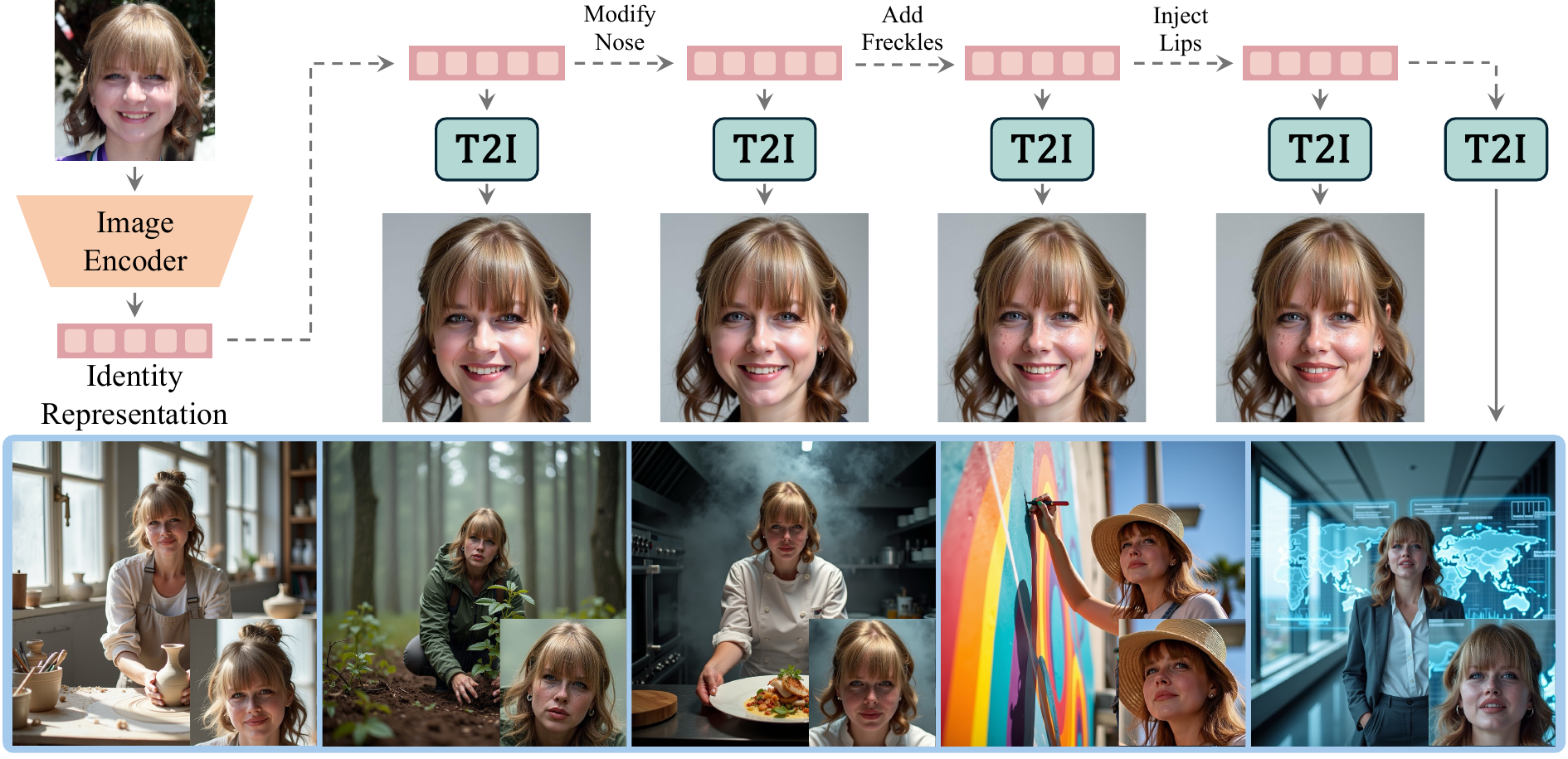}
        \vspace{-20pt}
        \captionof{figure}{
        We present methods for directly tuning the identity tokens of a personalization encoder, enabling fine-grained control of facial attributes, for example modifying the nose, adding freckles, or a beard (top).
        The edited identity can then be used across diverse prompts to generate the same tuned subject consistently in new scenes (bottom).
        }
    \vspace{-5pt}
    \label{fig:sup_teaser}
    \end{center}
}]

\section*{\LARGE{Appendix}}

\section{Additional Implementation Details}
\label{sec:additional_imp_details}

\begin{figure*}[t]
\centering
\setlength{\tabcolsep}{1pt} %
\setlength\arrayrulewidth{0.8pt} %
\setlength\dashlinedash{1.6pt}   %
\setlength\dashlinegap{0.8pt}    %
\renewcommand{\arraystretch}{1.2} %
\scriptsize{
\begin{tabular}{c : c c c c}

    {\setlength{\tabcolsep}{1pt}%
     \renewcommand{\arraystretch}{1.2}%
     \begin{tabular}{@{}c@{}}
       $\vcenter{\hbox{\includegraphics[width=0.15\linewidth]{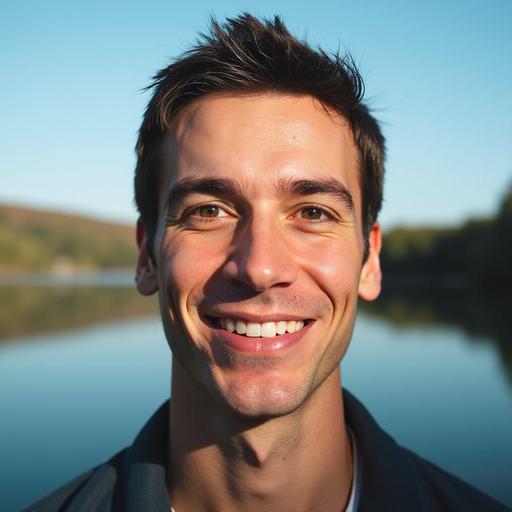}}}%
       $
     \end{tabular}
    } &&
    
    {\setlength{\tabcolsep}{1pt}%
     \renewcommand{\arraystretch}{1.2}%
     \begin{tabular}{@{}c@{}}
       $\vcenter{\hbox{\includegraphics[width=0.15\linewidth]{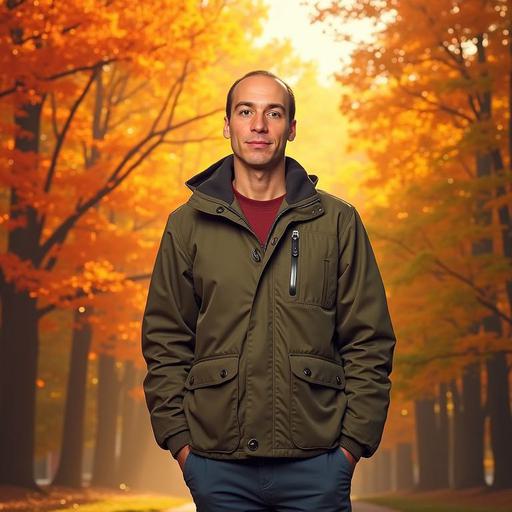}}}%
       \hspace{-2.25pt}%
       \vcenter{\hbox{
       \renewcommand{\arraystretch}{0}%
           \begin{tabular}{@{}c@{}}
               \includegraphics[width=0.075\linewidth]{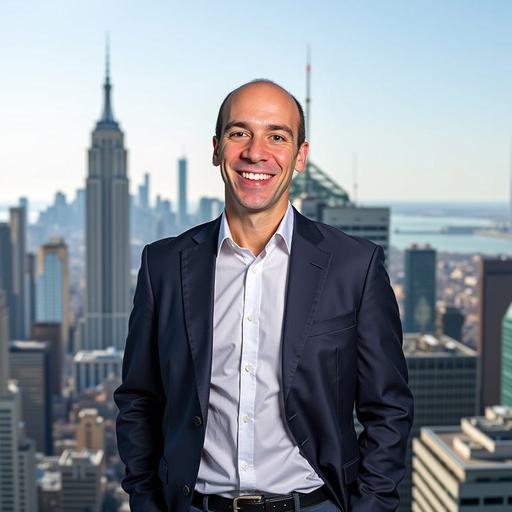} \\
               \includegraphics[width=0.075\linewidth]{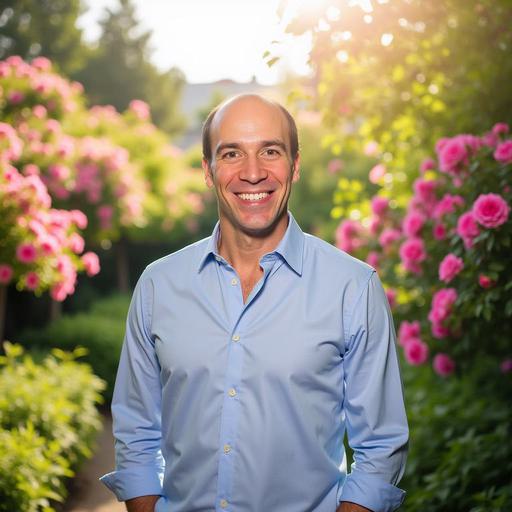}
           \end{tabular}
       }}
       $
     \end{tabular}
    } &
    
    {\setlength{\tabcolsep}{1pt}%
     \renewcommand{\arraystretch}{1.2}%
     \begin{tabular}{@{}c@{}}
       $\vcenter{\hbox{\includegraphics[width=0.15\linewidth]{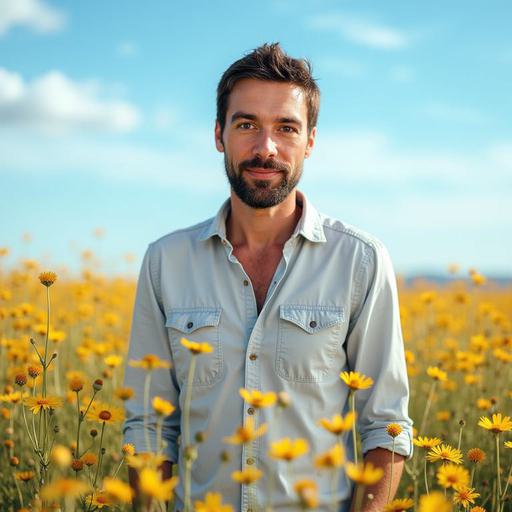}}}%
       \hspace{-2.25pt}%
       \vcenter{\hbox{
       \renewcommand{\arraystretch}{0}%
           \begin{tabular}{@{}c@{}}
               \includegraphics[width=0.075\linewidth]{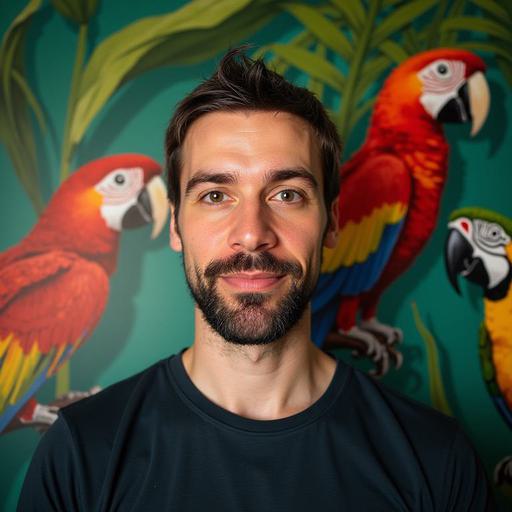} \\
               \includegraphics[width=0.075\linewidth]{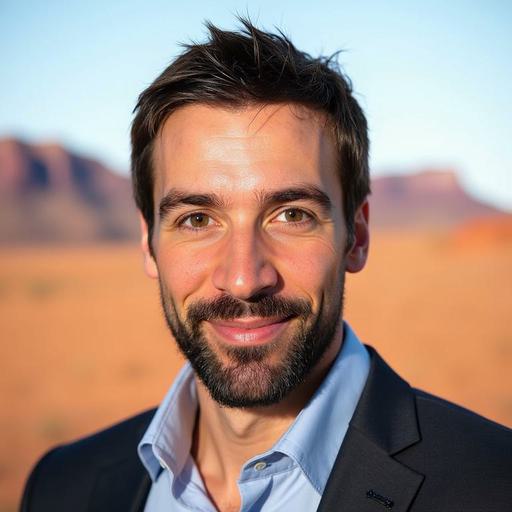}
           \end{tabular}
       }}
       $
     \end{tabular}
    } &
    
    {\setlength{\tabcolsep}{1pt}%
     \renewcommand{\arraystretch}{1.2}%
     \begin{tabular}{@{}c@{}}
       $\vcenter{\hbox{\includegraphics[width=0.15\linewidth]{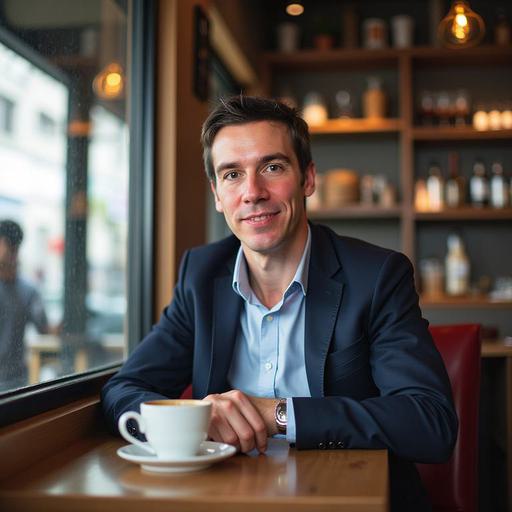}}}%
       \hspace{-2.25pt}%
       \vcenter{\hbox{
       \renewcommand{\arraystretch}{0}%
           \begin{tabular}{@{}c@{}}
               \includegraphics[width=0.075\linewidth]{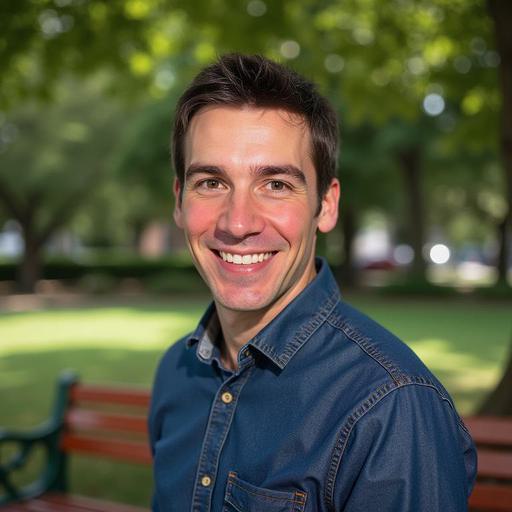} \\
               \includegraphics[width=0.075\linewidth]{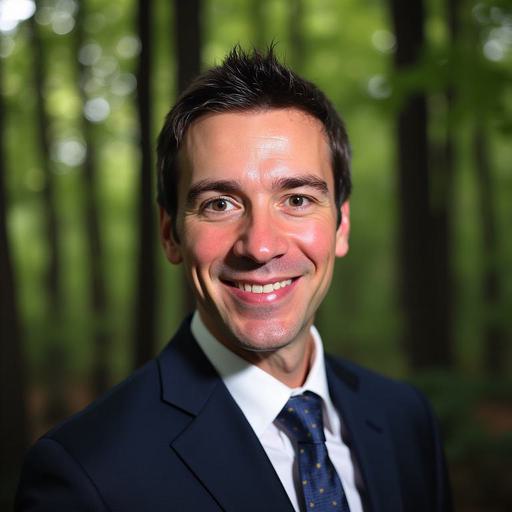}
           \end{tabular}
       }}
       $
     \end{tabular}
    } \\ \\[-11pt]

    Original Identity && Bald & Beard & Rosy Cheeks \\

    {\setlength{\tabcolsep}{1pt}%
     \renewcommand{\arraystretch}{1.2}%
     \begin{tabular}{@{}c@{}}
       $\vcenter{\hbox{\includegraphics[width=0.15\linewidth]{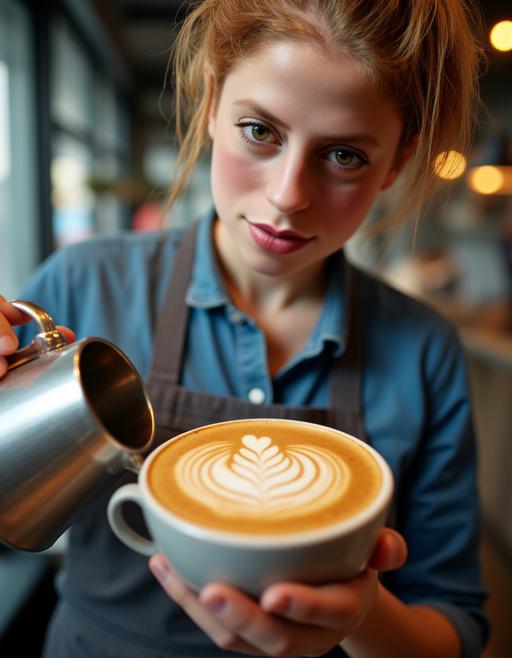}}}%
       $
     \end{tabular}
    } &&
    
    {\setlength{\tabcolsep}{1pt}%
     \renewcommand{\arraystretch}{1.2}%
     \begin{tabular}{@{}c@{}}
       $\vcenter{\hbox{\includegraphics[width=0.15\linewidth]{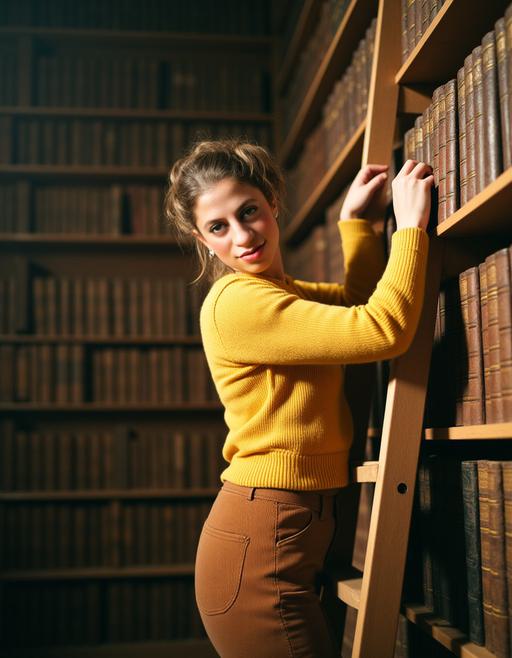}}}%
       \hspace{-2.25pt}%
       \vcenter{\hbox{
       \renewcommand{\arraystretch}{0}%
           \begin{tabular}{@{}c@{}}
               \includegraphics[width=0.075\linewidth]{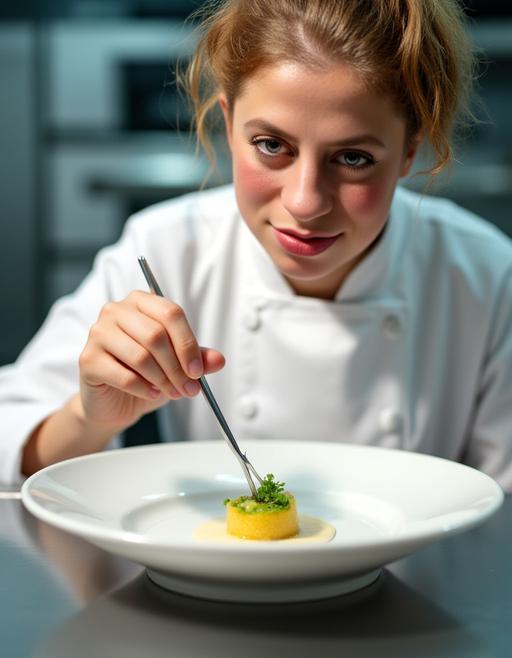} \\
               \includegraphics[width=0.075\linewidth]{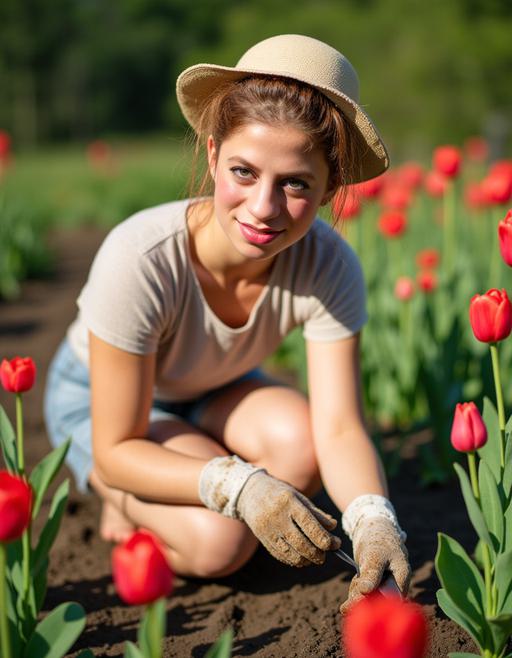}
           \end{tabular}
       }}
       $
     \end{tabular}
    } &
    
    {\setlength{\tabcolsep}{1pt}%
     \renewcommand{\arraystretch}{1.2}%
     \begin{tabular}{@{}c@{}}
       $\vcenter{\hbox{\includegraphics[width=0.15\linewidth]{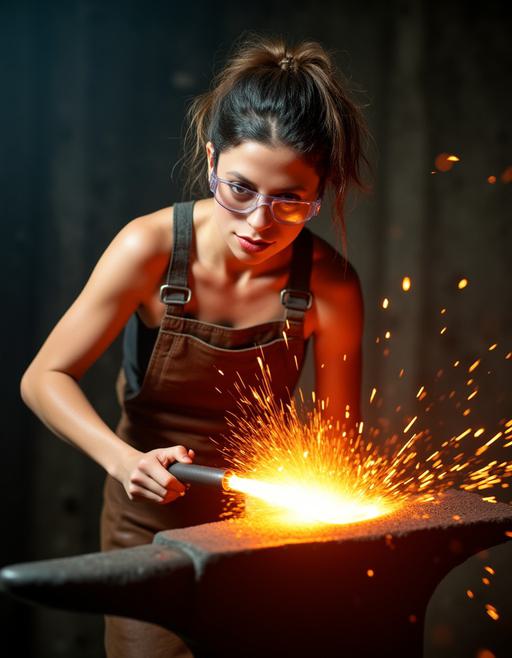}}}%
       \hspace{-2.25pt}%
       \vcenter{\hbox{
       \renewcommand{\arraystretch}{0}%
           \begin{tabular}{@{}c@{}}
               \includegraphics[width=0.075\linewidth]{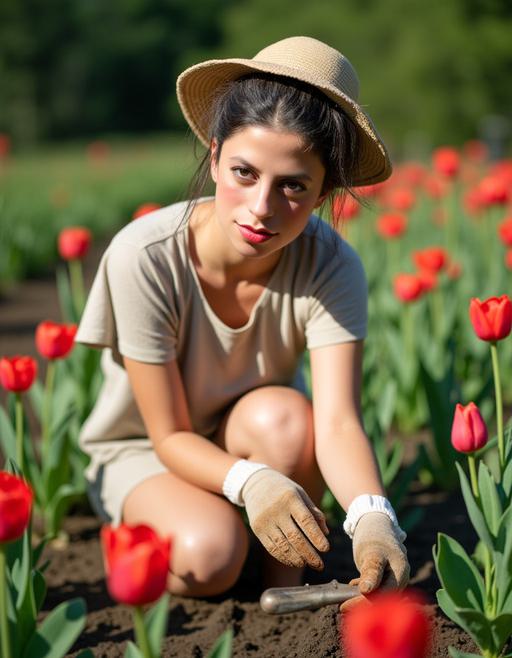} \\
               \includegraphics[width=0.075\linewidth]{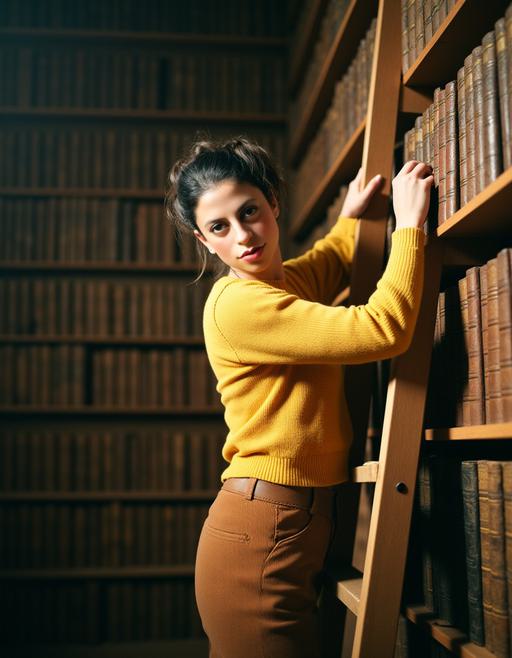}
           \end{tabular}
       }}
       $
     \end{tabular}
    } &
    
    {\setlength{\tabcolsep}{1pt}%
     \renewcommand{\arraystretch}{1.2}%
     \begin{tabular}{@{}c@{}}
       $\vcenter{\hbox{\includegraphics[width=0.15\linewidth]{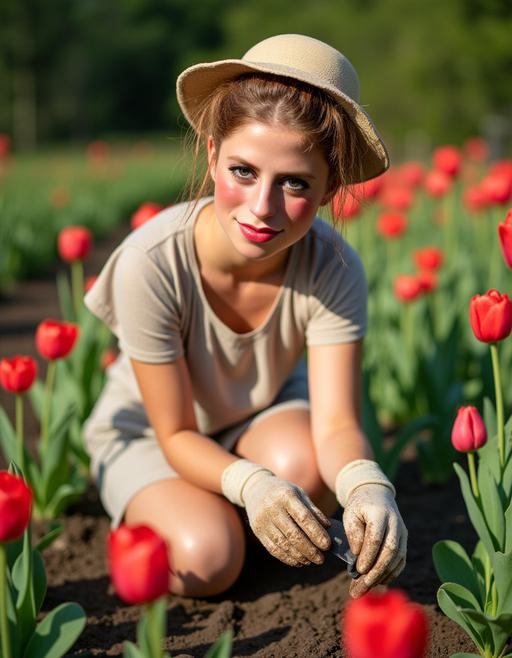}}}%
       \hspace{-2.25pt}%
       \vcenter{\hbox{
       \renewcommand{\arraystretch}{0}%
           \begin{tabular}{@{}c@{}}
               \includegraphics[width=0.075\linewidth]{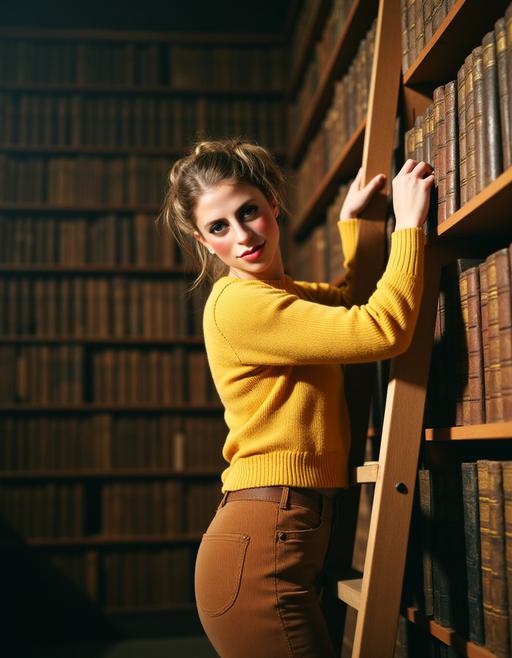} \\
               \includegraphics[width=0.075\linewidth]{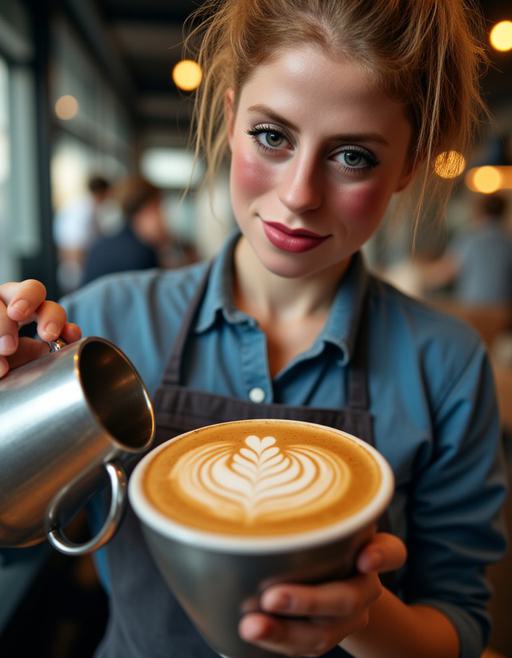}
           \end{tabular}
       }}
       $
     \end{tabular}
    } \\ \\[-11pt]

    Original Identity && Bigger Nose & Darker Hair & Rosy Cheeks \\

    {\setlength{\tabcolsep}{1pt}%
     \renewcommand{\arraystretch}{1.2}%
     \begin{tabular}{@{}c@{}}
       $\vcenter{\hbox{\includegraphics[width=0.15\linewidth]{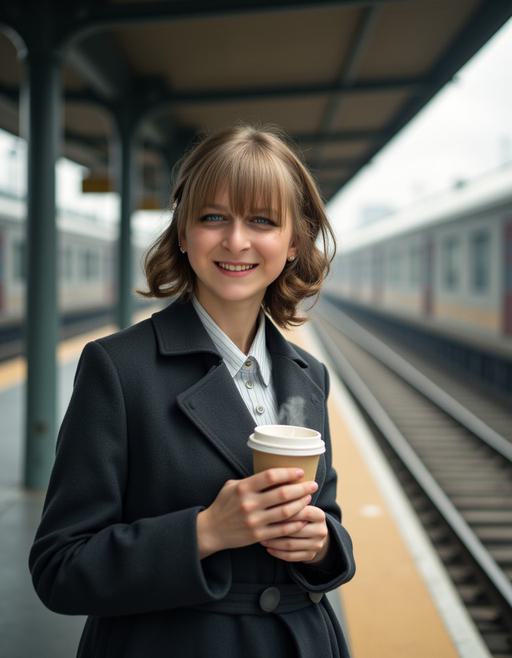}}}%
       $
     \end{tabular}
    } &&
    
    {\setlength{\tabcolsep}{1pt}%
     \renewcommand{\arraystretch}{1.2}%
     \begin{tabular}{@{}c@{}}
       $\vcenter{\hbox{\includegraphics[width=0.15\linewidth]{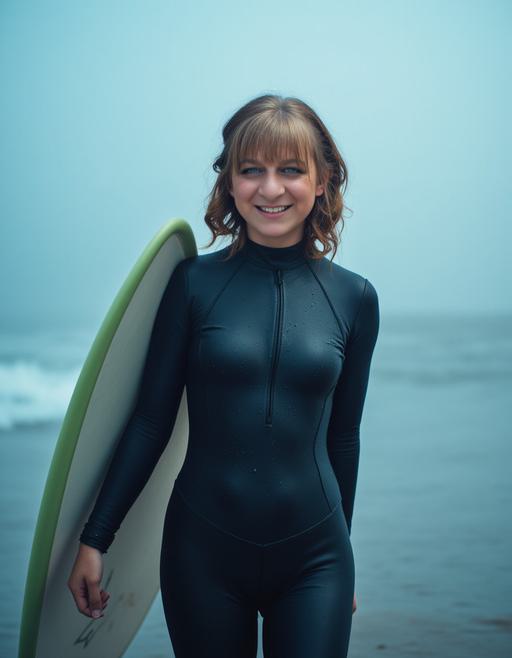}}}%
       \hspace{-2.25pt}%
       \vcenter{\hbox{
       \renewcommand{\arraystretch}{0}%
           \begin{tabular}{@{}c@{}}
               \includegraphics[width=0.075\linewidth]{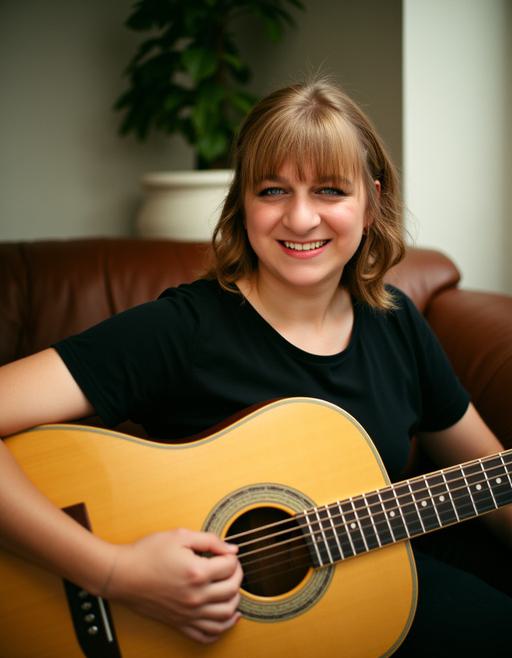} \\
               \includegraphics[width=0.075\linewidth]{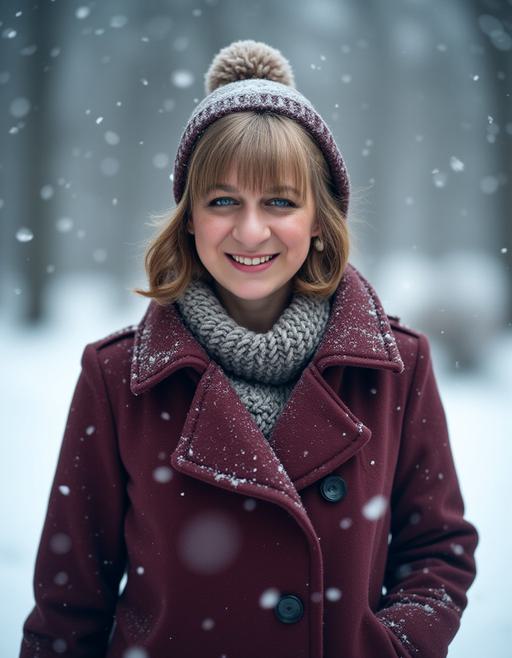}
           \end{tabular}
       }}
       $
     \end{tabular}
    } &
    
    {\setlength{\tabcolsep}{1pt}%
     \renewcommand{\arraystretch}{1.2}%
     \begin{tabular}{@{}c@{}}
       $\vcenter{\hbox{\includegraphics[width=0.15\linewidth]{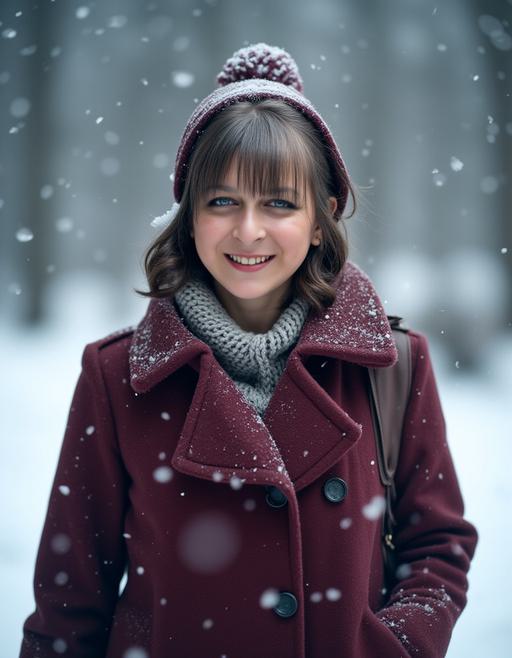}}}%
       \hspace{-2.25pt}%
       \vcenter{\hbox{
       \renewcommand{\arraystretch}{0}%
           \begin{tabular}{@{}c@{}}
               \includegraphics[width=0.075\linewidth]{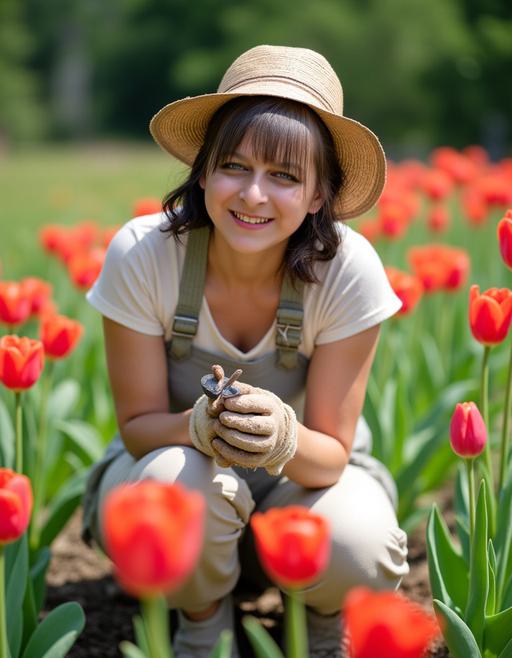} \\
               \includegraphics[width=0.075\linewidth]{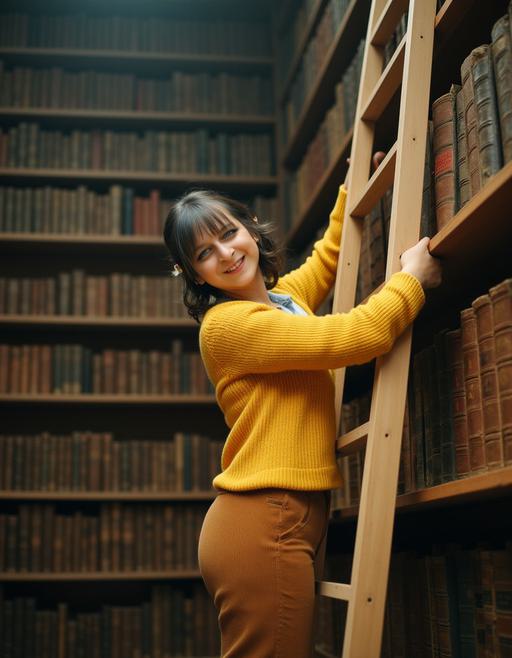}
           \end{tabular}
       }}
       $
     \end{tabular}
    } &
    
    {\setlength{\tabcolsep}{1pt}%
     \renewcommand{\arraystretch}{1.2}%
     \begin{tabular}{@{}c@{}}
       $\vcenter{\hbox{\includegraphics[width=0.15\linewidth]{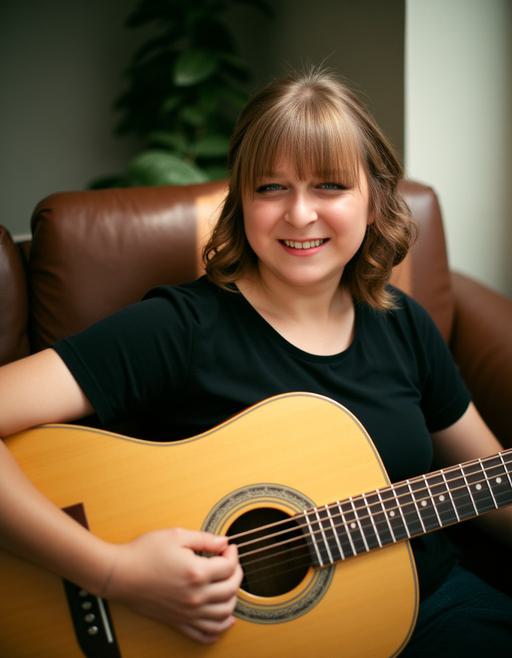}}}%
       \hspace{-2.25pt}%
       \vcenter{\hbox{
       \renewcommand{\arraystretch}{0}%
           \begin{tabular}{@{}c@{}}
               \includegraphics[width=0.075\linewidth]{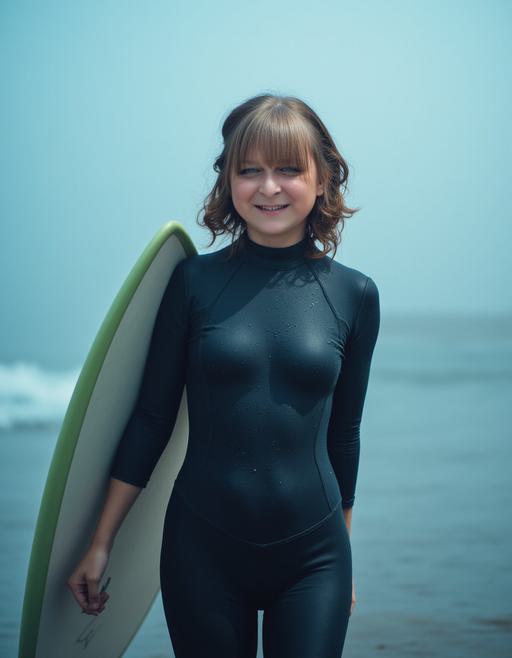} \\
               \includegraphics[width=0.075\linewidth]{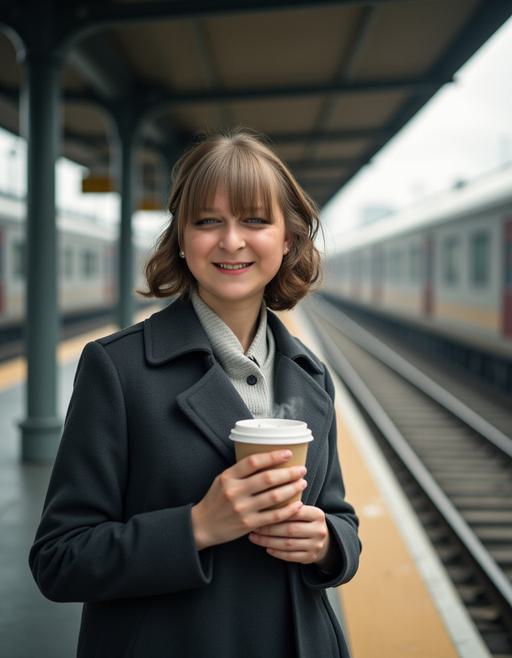}
           \end{tabular}
       }}
       $
     \end{tabular}
    } \\ \\[-11pt]

    Original Identity && Bigger Nose & Darker Hair & Chubby \\

\end{tabular}
}
\caption{
Each row shows a different source identity (leftmost column) and its edits along different directions. The images in each column are generated by adding the same direction found by SVM. For each edit, we show the main result with two alternative prompts stacked vertically to the right.
}
\label{fig:supp_directional_edits}
\end{figure*}

\paragraph{Token Selection}
We construct the token groups $S$ as follows.
For localized facial attributes, we create 5-10 $(I, I_{\text{patch}})$ pairs per attribute and apply the localized attribute selection method. For each pair, we select the top $k = 20$ tokens ranked by $\lVert \Delta Z_n \rVert_2$. The final set $S$ for each attribute consists of the tokens that appear most frequently in these top-$k$ selections across pairs.
To select tokens for global attributes, we train one linear SVM per token and retain those whose validation accuracy exceeds a threshold $\tau = 0.7$.
The selection process is computed once per region or attribute and reused for all subsequent edits.

\paragraph{PCA over identity tokens}
We encode the entire FFHQ dataset with the Omni-ID encoder to obtain token sequences $Z$. For a chosen set of token indices $S$, we concatenate the corresponding token embeddings for each image into a single vector $x_S = \mathrm{vec}(Z_S)$. We center each coordinate across the dataset and fit scikit learn \texttt{IncrementalPCA} using default parameters. On our hardware the fit takes about 10 minutes on a CPU. To identify salient components and their roles, we apply the top principal direction to a small set of example identities with a scale of 100 and inspect the outcomes. We then evaluate the effect of each component either manually or with a vision and language model, retaining those that produce meaningful identity tuning, including attributes that are not easily described in words.

\paragraph{Supervised attribute directions via linear SVM}
We encode the CelebA dataset to obtain $Z$ for all images. For each target attribute and token set $S$, we form $x_S = \mathrm{vec}(Z_S)$, balance the training data to have equal positive and negative examples using the CelebA binary labels, and train a linear classifier with hinge loss using the scikit learn \texttt{SGDClassifier} with default parameters. In our runs this training completes in about 10 minutes on a CPU and can be accelerated with a GPU.

\section{Additional Experiments}

\subsection{Identity Tuning Resutls}
\label{sec:more_results_sup}
\cref{fig:sup_teaser}
illustrates identity tuning: given a source image, our approach enables iterative tuning of the identity, highlighting the advantage of operating directly on the identity representation. The tuned identity produces consistent generations across diverse prompts, and the figure includes edits that are difficult to express via text, such as modifying the shape of the nose.

\cref{fig:supp_directional_edits}
provides additional examples of supervised directional editing applied to facial identities. These results demonstrate that the modifications remain consistent across various text prompts while being strictly localized to the target attribute.

\begin{figure}
    \centering
    \vspace{-10pt}
    \setlength{\tabcolsep}{0pt}
    \begin{tabular}{ccccc}         
         \includegraphics[width=0.2\linewidth]{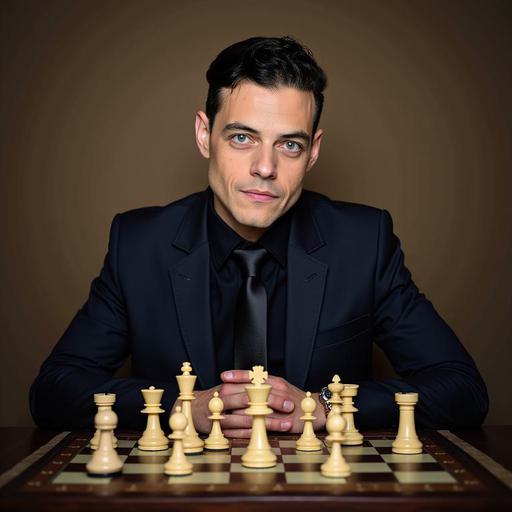} &
         \includegraphics[width=0.2\linewidth]{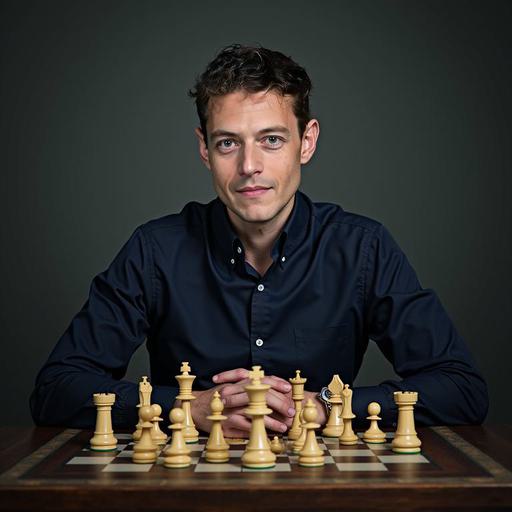} &
         \includegraphics[width=0.2\linewidth]{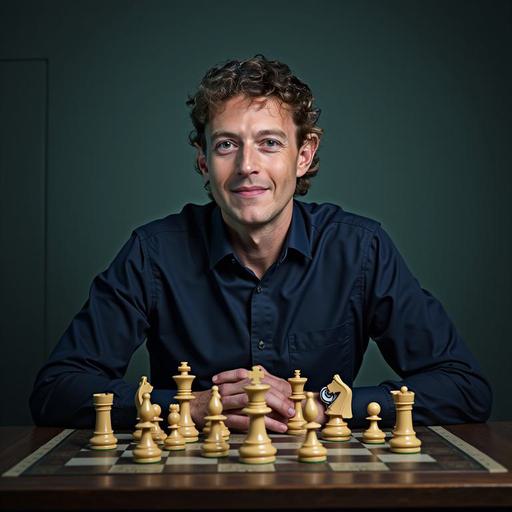} &
         \includegraphics[width=0.2\linewidth]{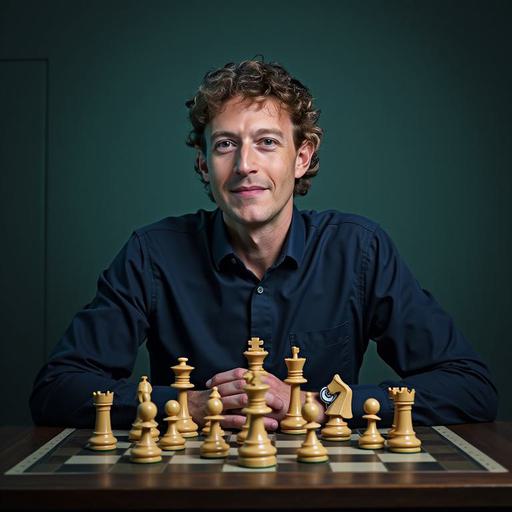} &
         \includegraphics[width=0.2\linewidth]{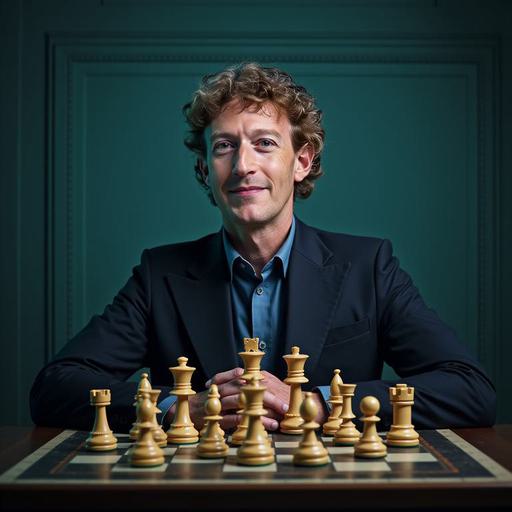} \\
         
         \includegraphics[width=0.2\linewidth]{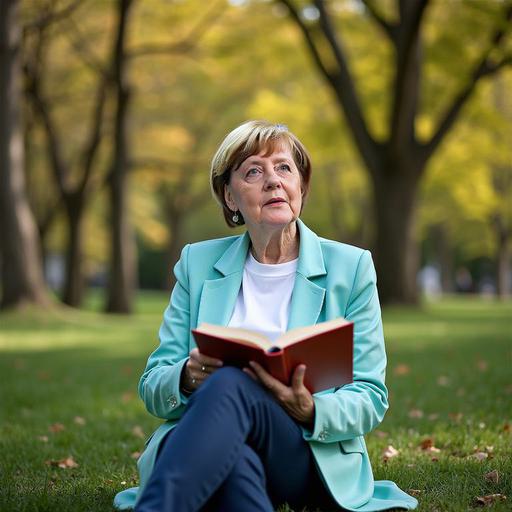} &
         \includegraphics[width=0.2\linewidth]{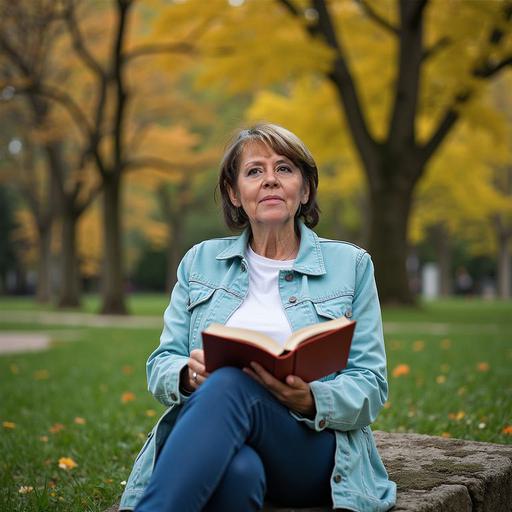} &
         \includegraphics[width=0.2\linewidth]{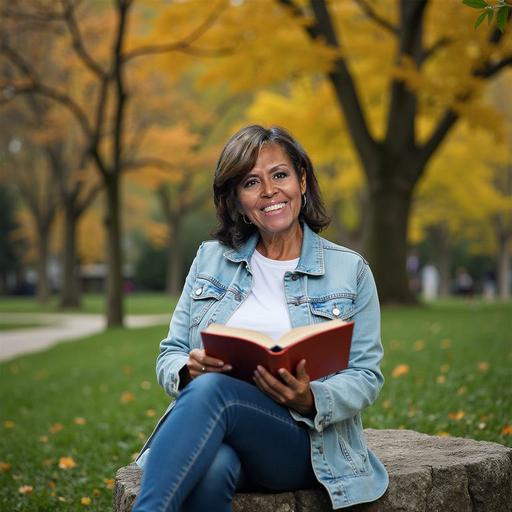} &
         \includegraphics[width=0.2\linewidth]{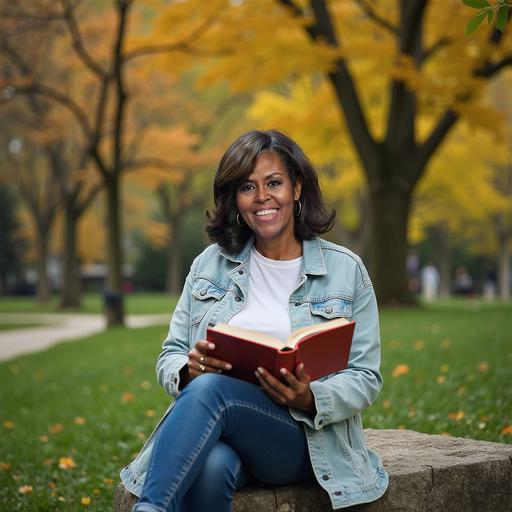} &
         \includegraphics[width=0.2\linewidth]{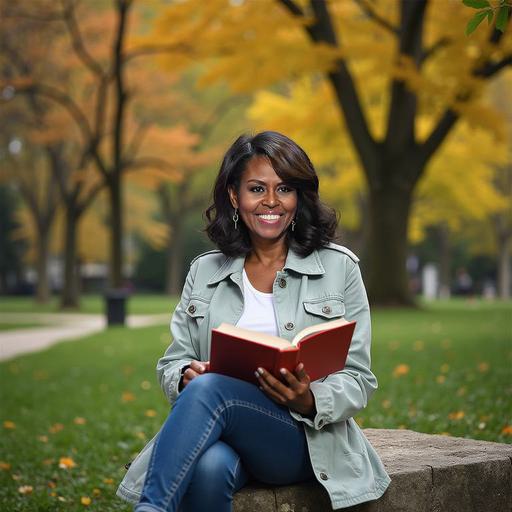} \\
         
         Identity A &&&& Identity B \\
    \end{tabular}
    \vspace{-8pt}
    \caption{Continuous identity space via interpolation. The leftmost and rightmost columns use the encoding of two different identities (A and B). Intermediate columns are generated by linearly interpolating between their identity embeddings and applying the mixed embedding to all tokens. The identity changes gradually without abrupt jumps, illustrating the continuity of the learned identity representation across varied scenes.}
    \label{fig:sup_global_blend}
    \vspace{-15pt}
\end{figure}

\subsection{Continuous Tuning}
\label{sec:continues_tuning_sup}
\begin{figure}[t]
    \centering
    \small
    
    \setlength{\tabcolsep}{0pt}
    \begin{tabular}{@{}c@{}c@{}c@{}c@{}c@{}c@{}c@{}}

        {\setlength{\tabcolsep}{0pt}%
         \renewcommand{\arraystretch}{0}%
         \begin{tabular}{@{}c@{}c@{}c@{}}
            \multicolumn{3}{@{}c@{}}{\includegraphics[width=0.141\linewidth]{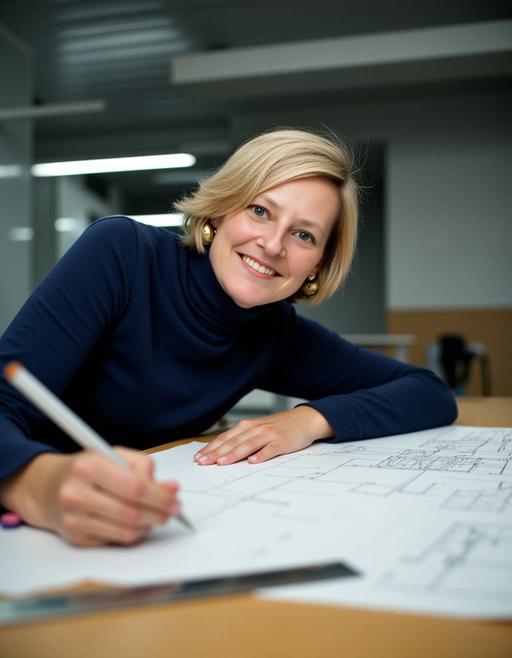}} \\
            \includegraphics[width=0.047\linewidth]{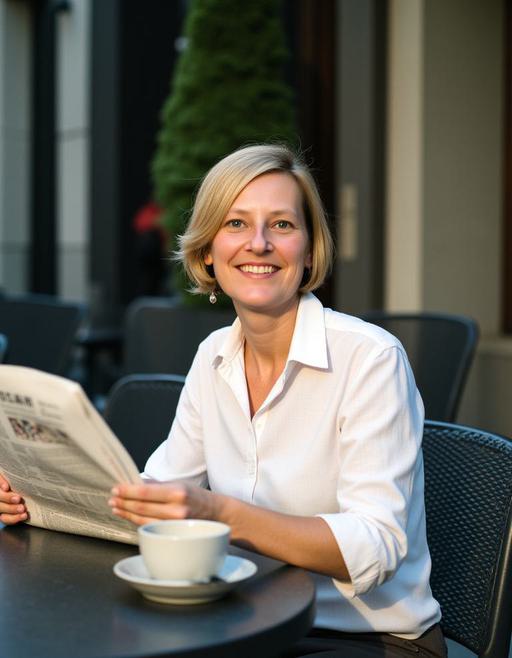} &
            \includegraphics[width=0.047\linewidth]{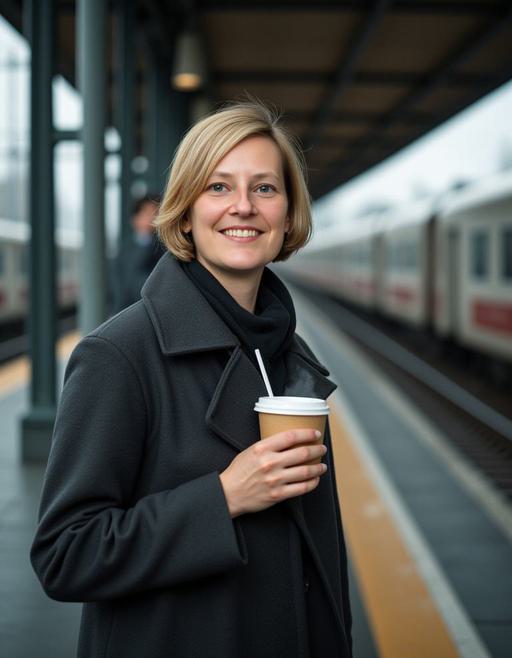} &
            \includegraphics[width=0.047\linewidth]{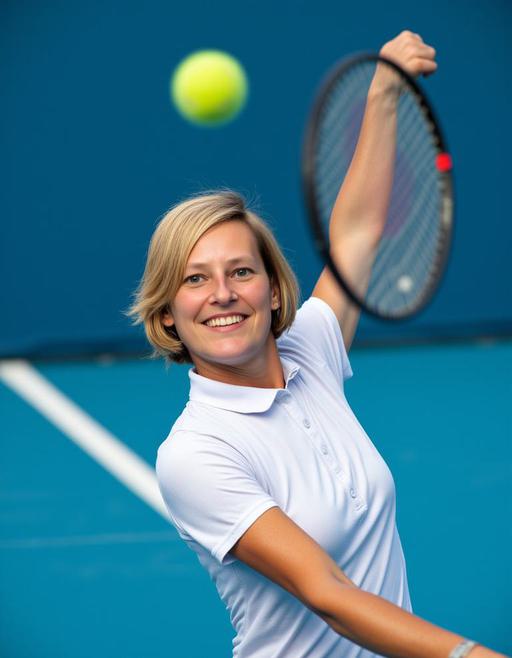}
         \end{tabular}%
        } &
        {\setlength{\tabcolsep}{0pt}%
         \renewcommand{\arraystretch}{0}%
         \begin{tabular}{@{}c@{}c@{}c@{}}
            \multicolumn{3}{@{}c@{}}{\includegraphics[width=0.141\linewidth]{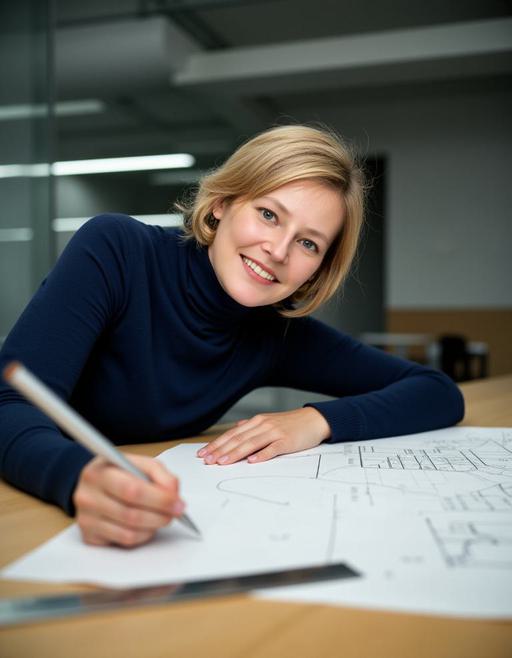}} \\
            \includegraphics[width=0.047\linewidth]{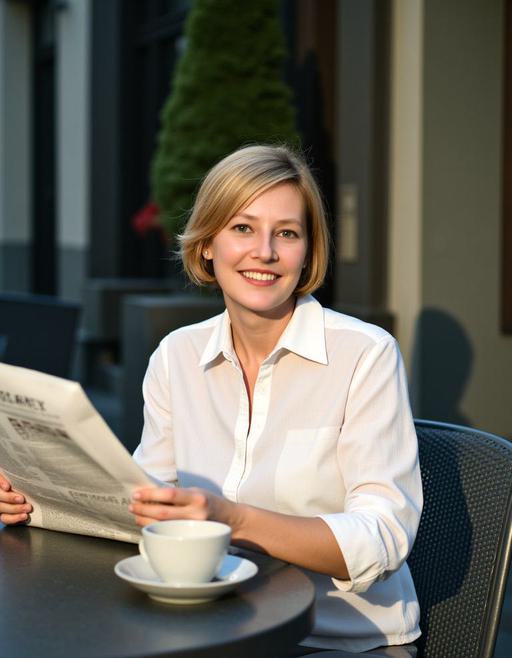} &
            \includegraphics[width=0.047\linewidth]{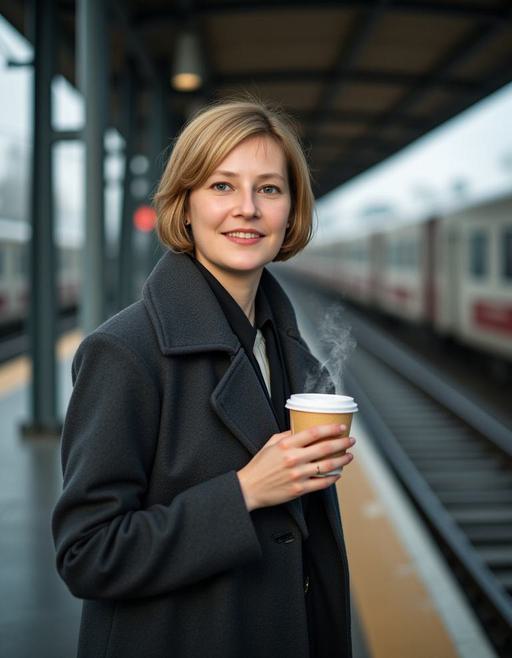} &
            \includegraphics[width=0.047\linewidth]{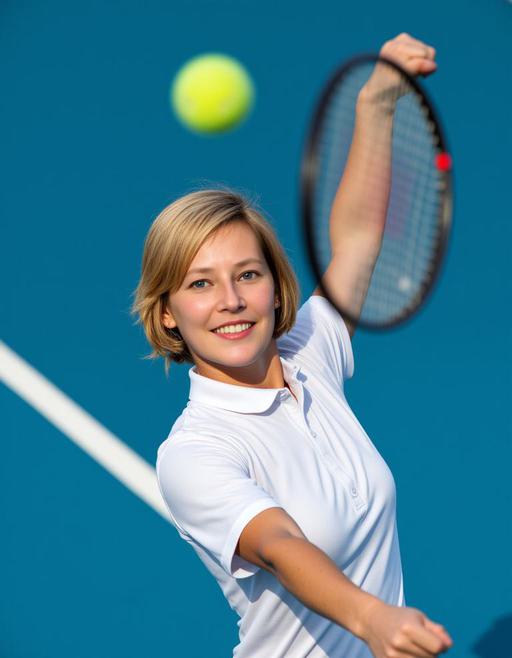}
         \end{tabular}%
        } &
        {\setlength{\tabcolsep}{0pt}%
         \renewcommand{\arraystretch}{0}%
         \begin{tabular}{@{}c@{}c@{}c@{}}
            \multicolumn{3}{@{}c@{}}{\includegraphics[width=0.141\linewidth]{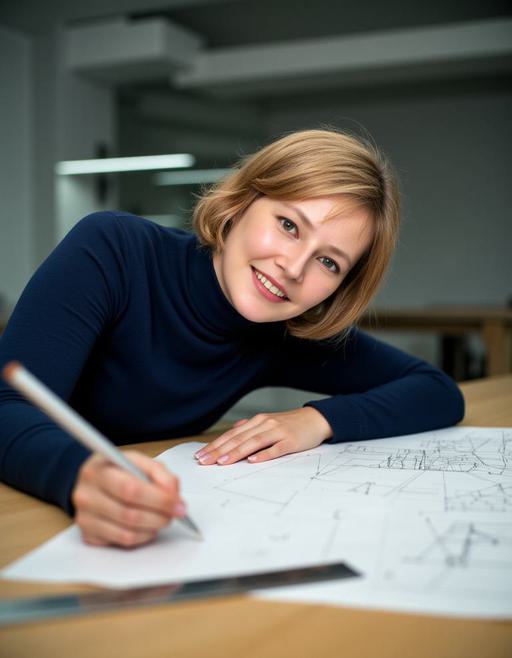}} \\
            \includegraphics[width=0.047\linewidth]{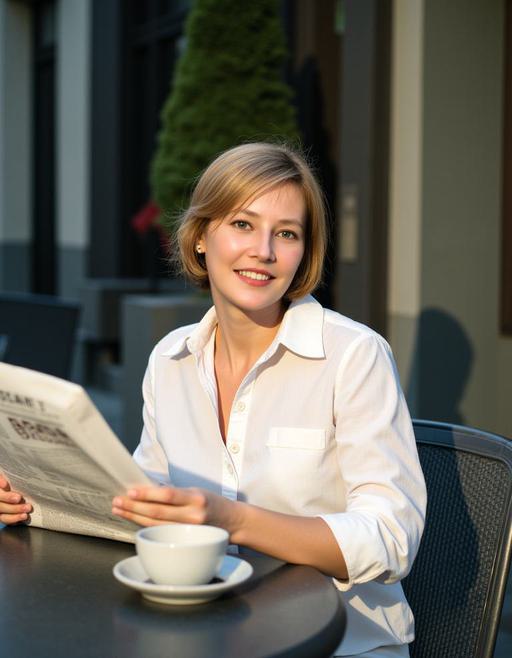} &
            \includegraphics[width=0.047\linewidth]{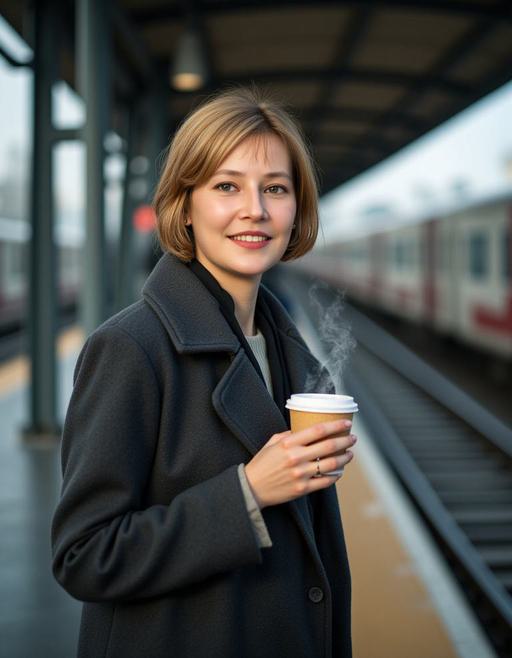} &
            \includegraphics[width=0.047\linewidth]{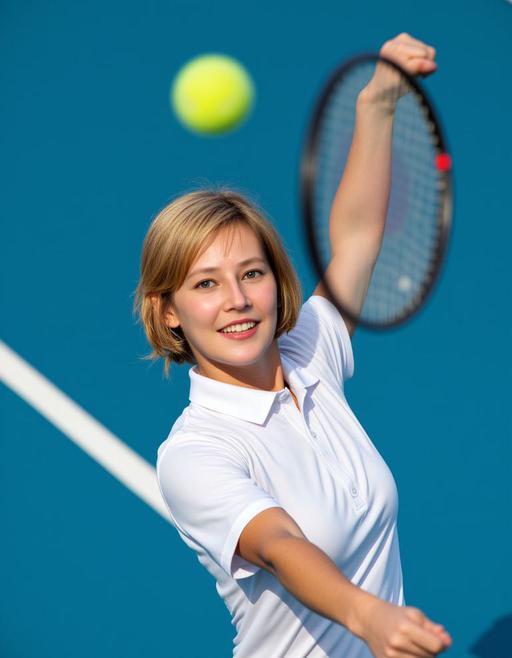}
         \end{tabular}%
        } &
        {\setlength{\tabcolsep}{0pt}%
         \renewcommand{\arraystretch}{0}%
         \begin{tabular}{@{}c@{}c@{}c@{}}
            \multicolumn{3}{@{}c@{}}{\includegraphics[width=0.141\linewidth]{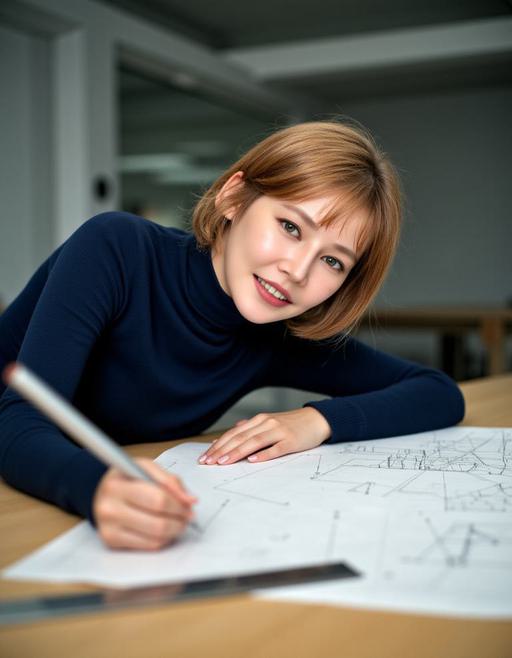}} \\
            \includegraphics[width=0.047\linewidth]{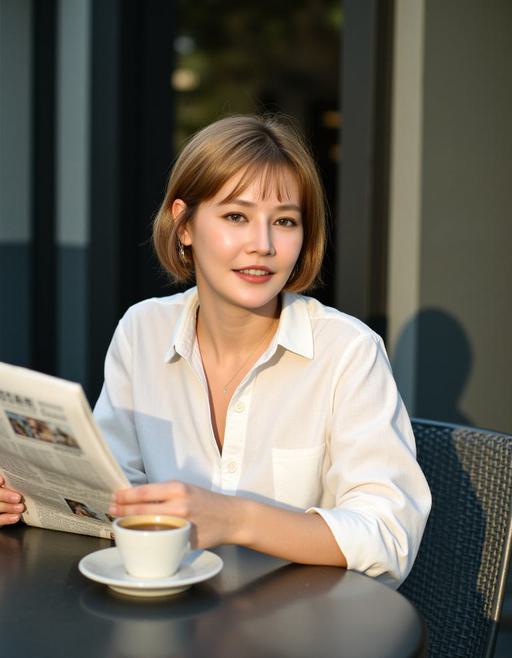} &
            \includegraphics[width=0.047\linewidth]{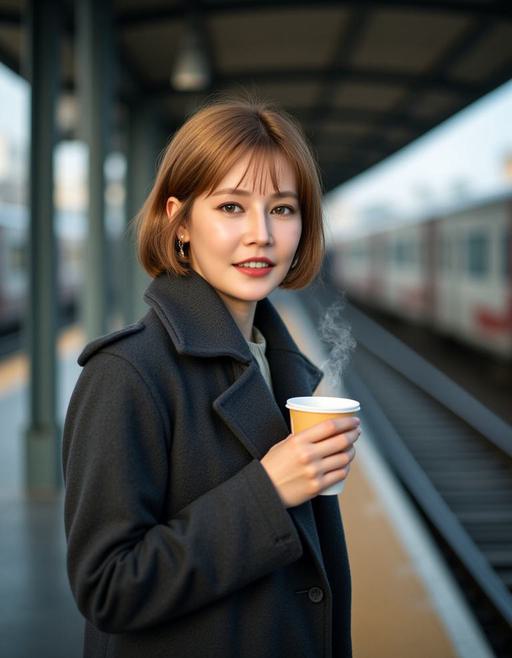} &
            \includegraphics[width=0.047\linewidth]{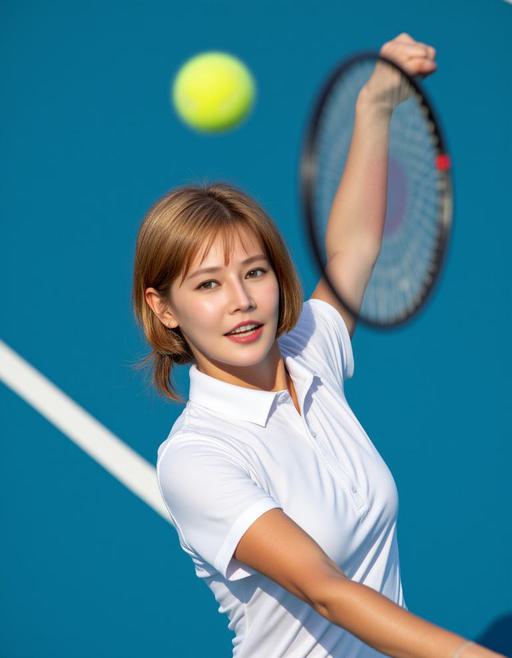}
         \end{tabular}%
        } &
        {\setlength{\tabcolsep}{0pt}%
         \renewcommand{\arraystretch}{0}%
         \begin{tabular}{@{}c@{}c@{}c@{}}
            \multicolumn{3}{@{}c@{}}{\includegraphics[width=0.141\linewidth]{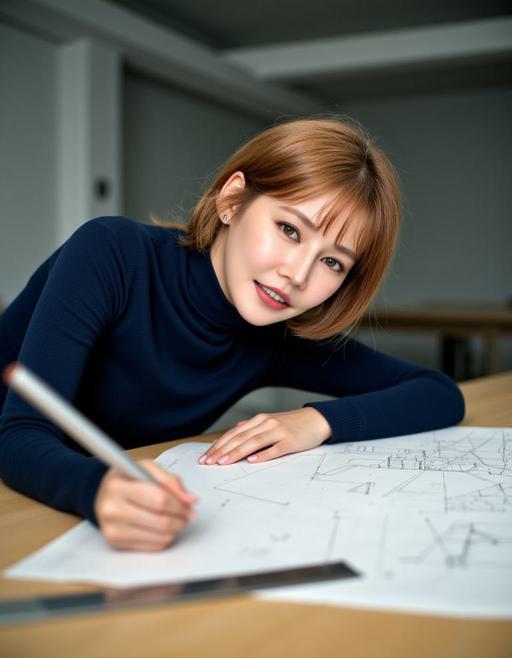}} \\
            \includegraphics[width=0.047\linewidth]{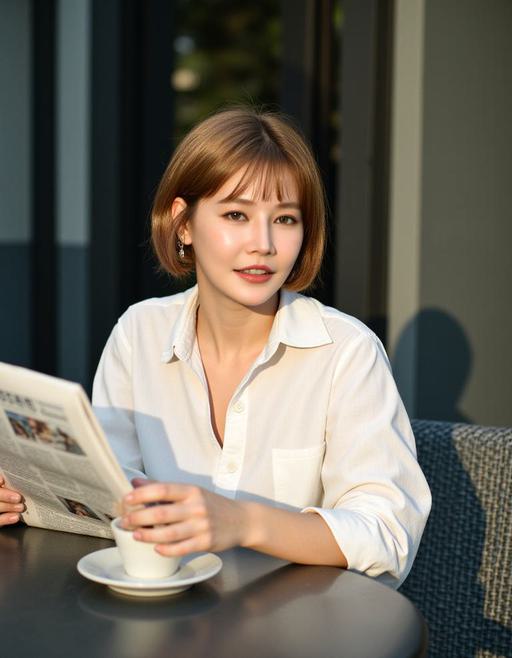} &
            \includegraphics[width=0.047\linewidth]{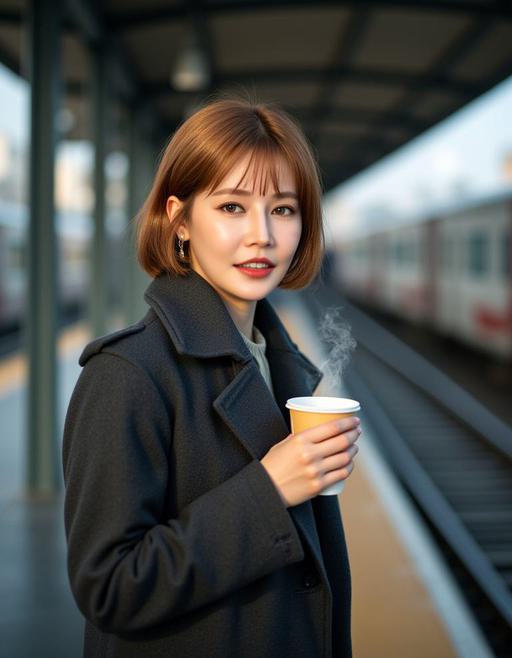} &
            \includegraphics[width=0.047\linewidth]{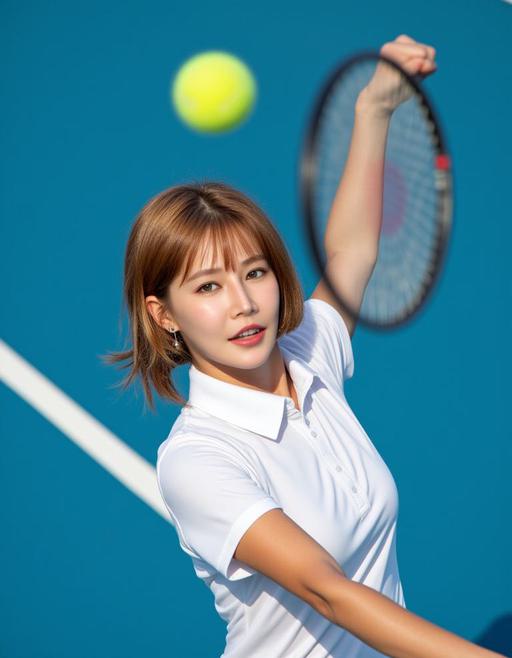}
         \end{tabular}%
        } &
        {\setlength{\tabcolsep}{0pt}%
         \renewcommand{\arraystretch}{0}%
         \begin{tabular}{@{}c@{}c@{}c@{}}
            \multicolumn{3}{@{}c@{}}{\includegraphics[width=0.141\linewidth]{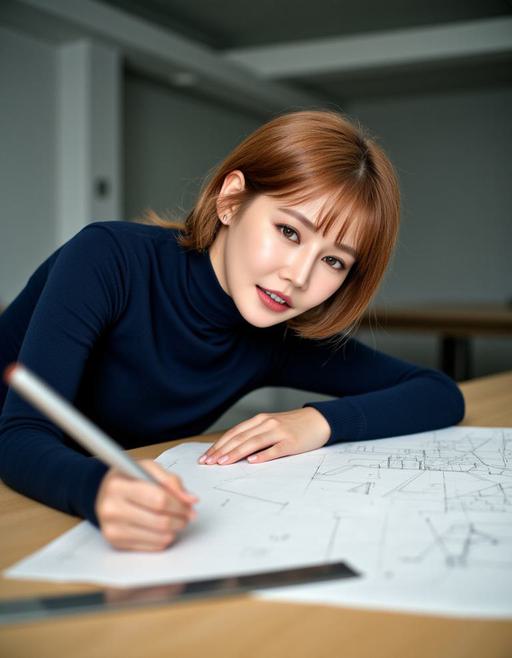}} \\
            \includegraphics[width=0.047\linewidth]{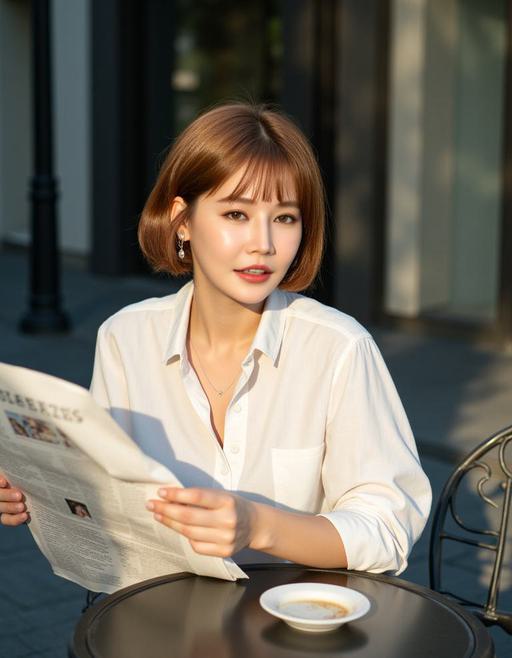} &
            \includegraphics[width=0.047\linewidth]{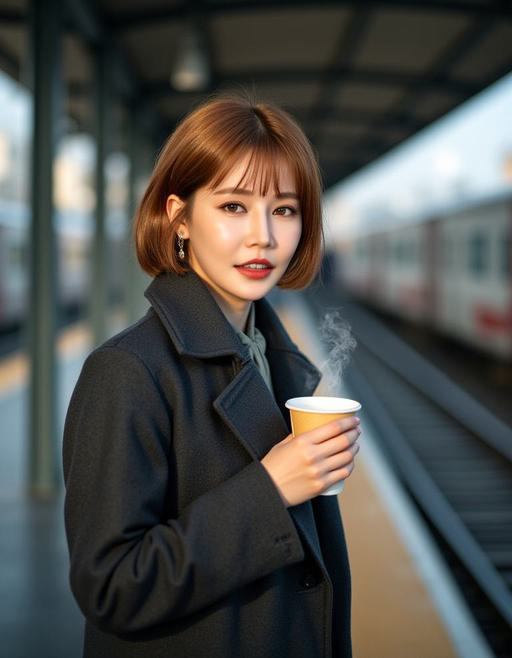} &
            \includegraphics[width=0.047\linewidth]{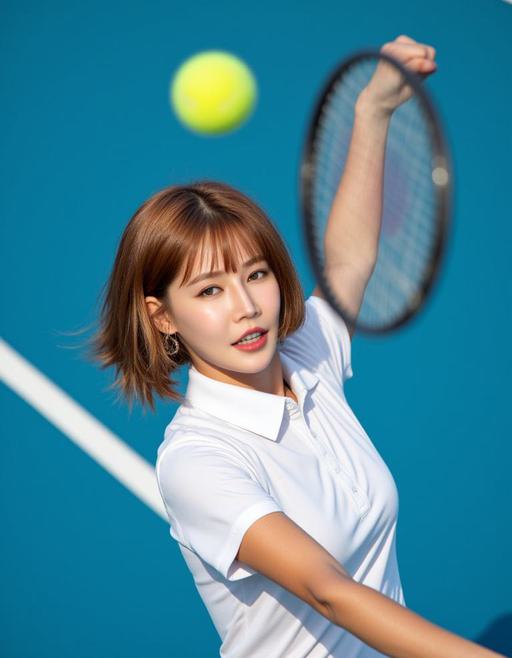}
         \end{tabular}%
        } &
        {\setlength{\tabcolsep}{0pt}%
         \renewcommand{\arraystretch}{0}%
         \begin{tabular}{@{}c@{}c@{}c@{}}
            \multicolumn{3}{@{}c@{}}{\includegraphics[width=0.141\linewidth]{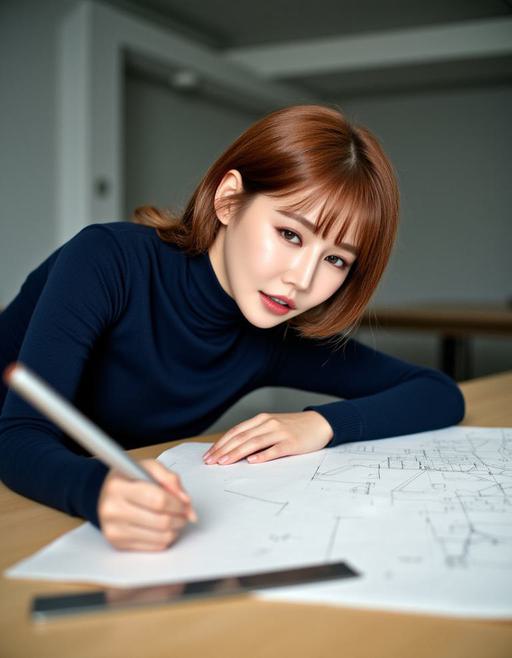}} \\
            \includegraphics[width=0.047\linewidth]{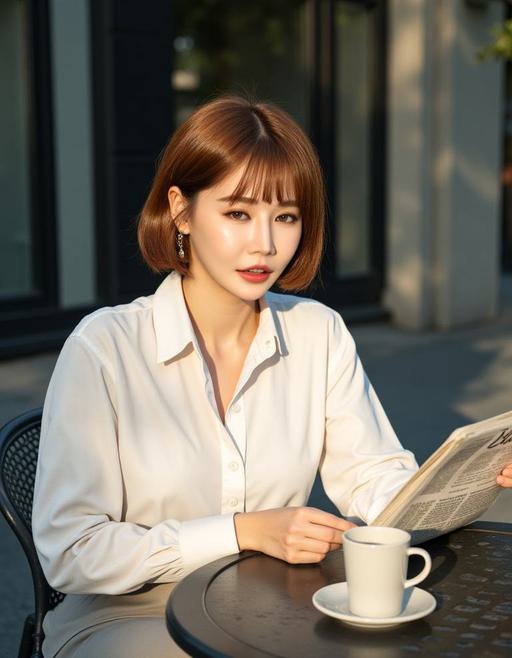} &
            \includegraphics[width=0.047\linewidth]{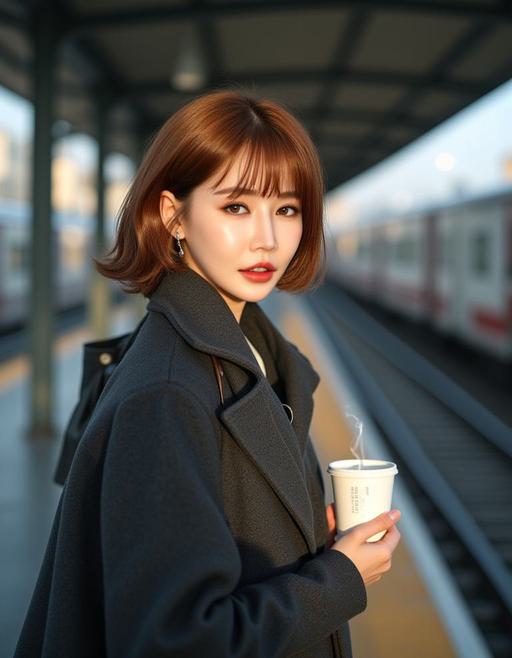} &
            \includegraphics[width=0.047\linewidth]{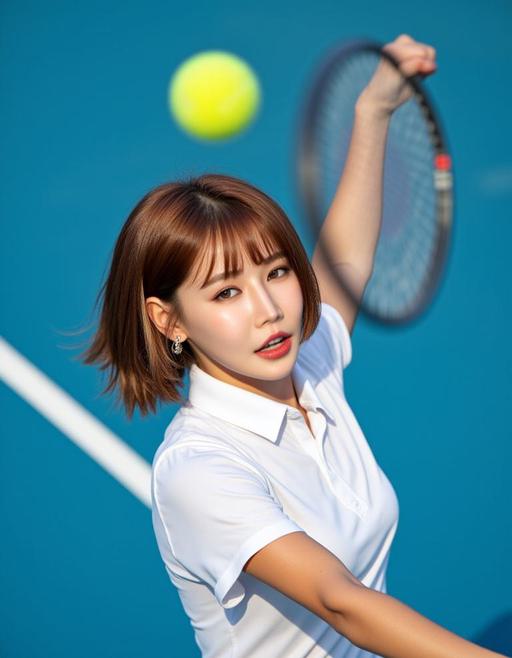}
         \end{tabular}%
        } \\ \\

        {\setlength{\tabcolsep}{0pt}%
         \renewcommand{\arraystretch}{0}%
         \begin{tabular}{@{}c@{}c@{}c@{}}
            \multicolumn{3}{@{}c@{}}{\includegraphics[width=0.141\linewidth]{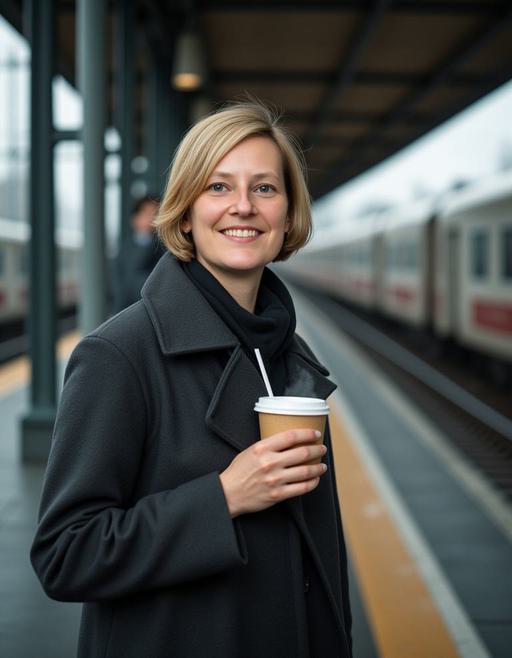}} \\
            \includegraphics[width=0.047\linewidth]{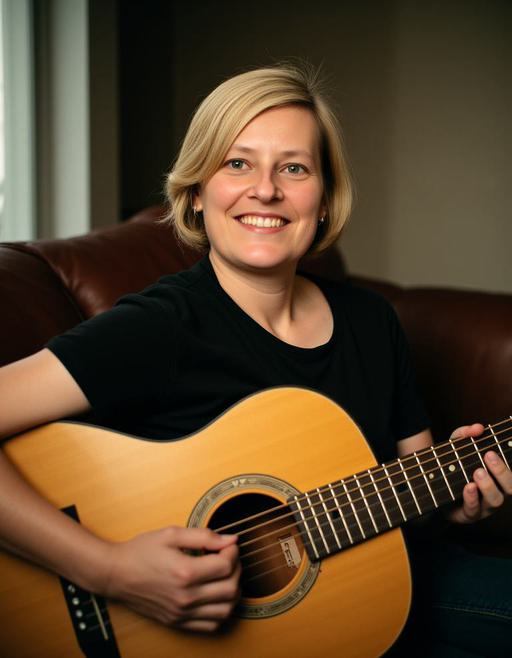} &
            \includegraphics[width=0.047\linewidth]{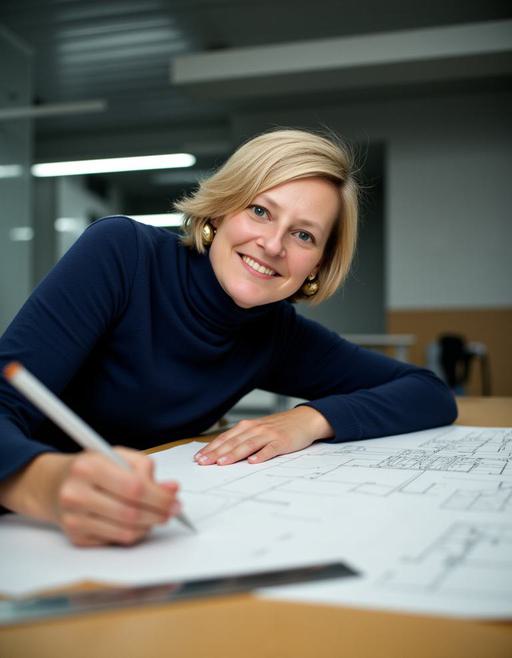} &
            \includegraphics[width=0.047\linewidth]{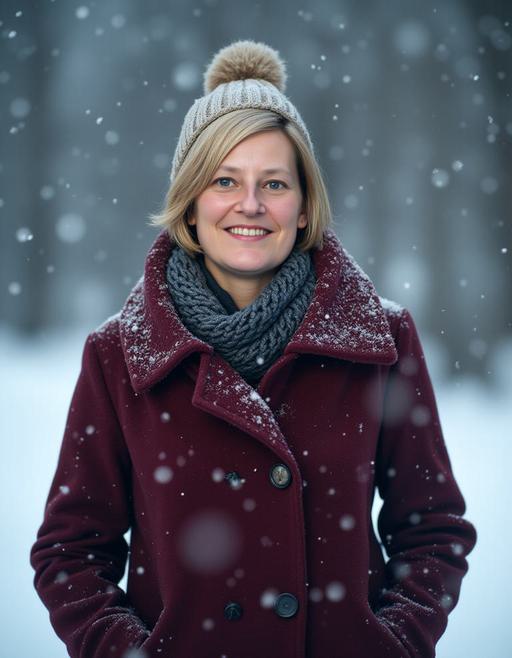}
         \end{tabular}%
        } &
        {\setlength{\tabcolsep}{0pt}%
         \renewcommand{\arraystretch}{0}%
         \begin{tabular}{@{}c@{}c@{}c@{}}
            \multicolumn{3}{@{}c@{}}{\includegraphics[width=0.141\linewidth]{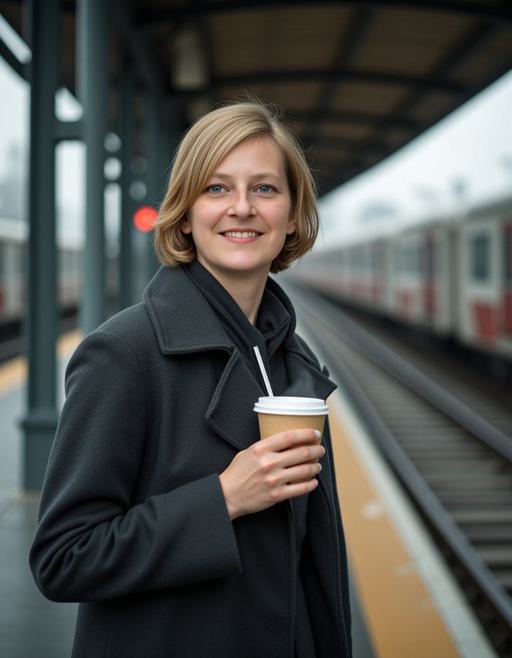}} \\
            \includegraphics[width=0.047\linewidth]{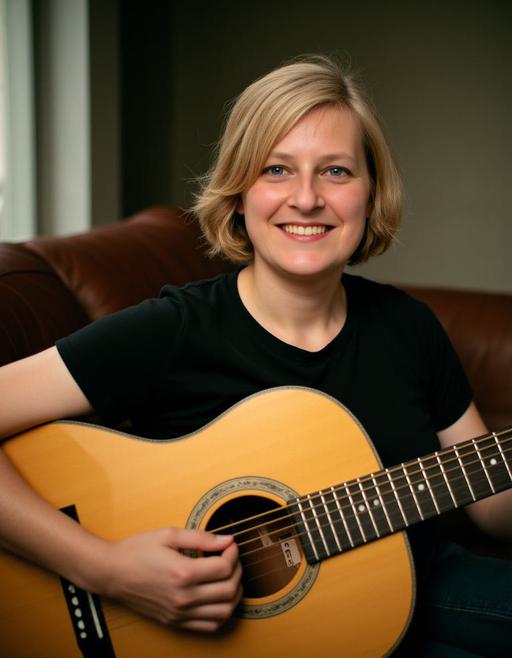} &
            \includegraphics[width=0.047\linewidth]{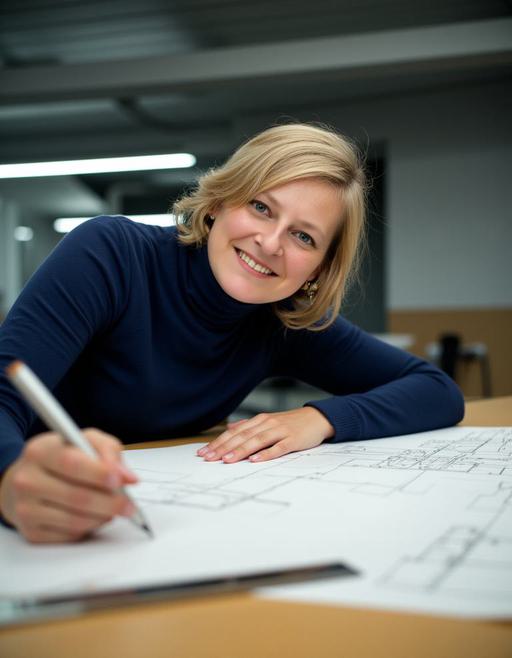} &
            \includegraphics[width=0.047\linewidth]{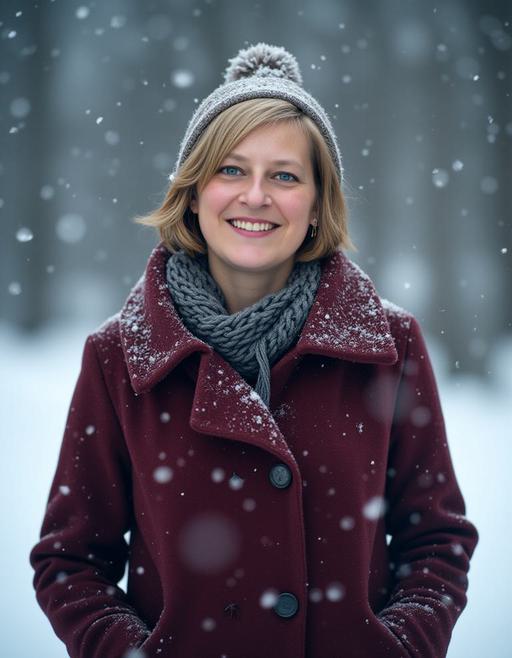}
         \end{tabular}%
        } &
        {\setlength{\tabcolsep}{0pt}%
         \renewcommand{\arraystretch}{0}%
         \begin{tabular}{@{}c@{}c@{}c@{}}
            \multicolumn{3}{@{}c@{}}{\includegraphics[width=0.141\linewidth]{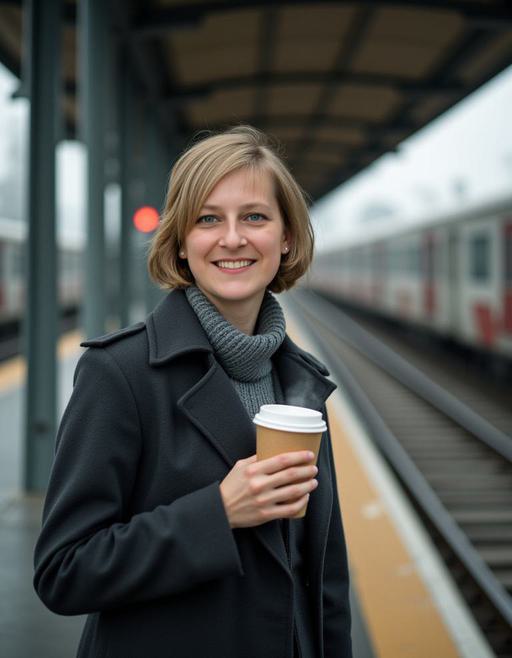}} \\
            \includegraphics[width=0.047\linewidth]{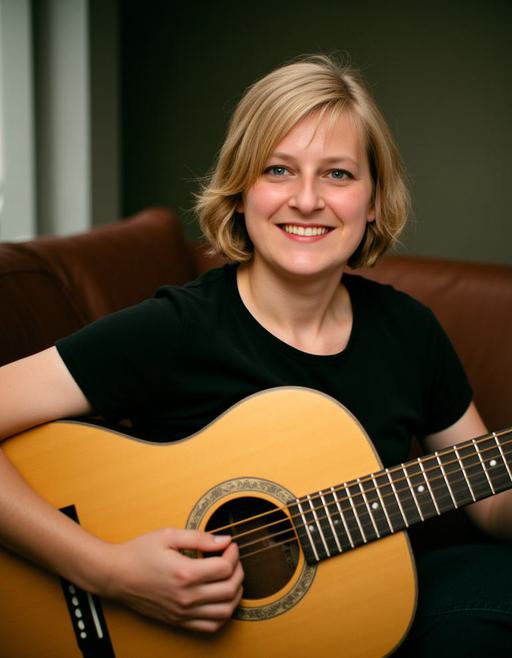} &
            \includegraphics[width=0.047\linewidth]{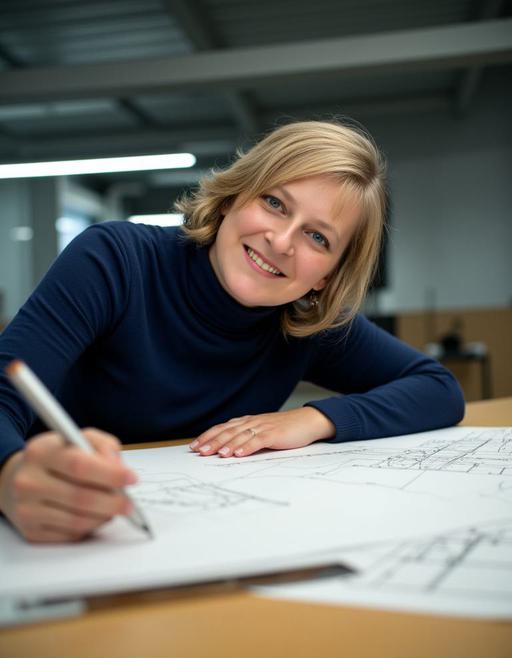} &
            \includegraphics[width=0.047\linewidth]{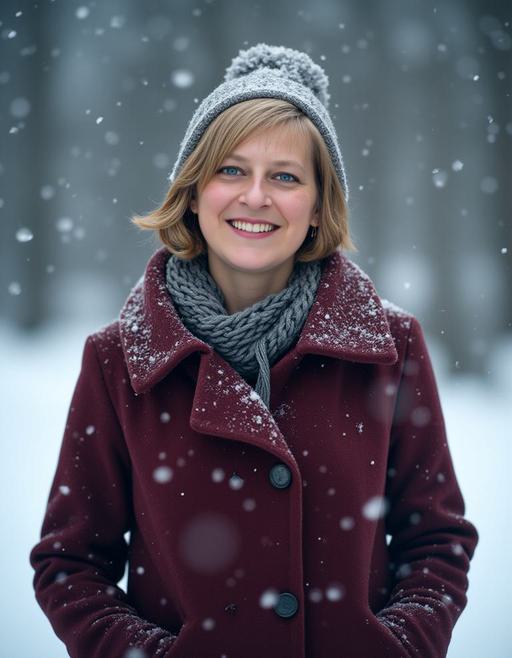}
         \end{tabular}%
        } &
        {\setlength{\tabcolsep}{0pt}%
         \renewcommand{\arraystretch}{0}%
         \begin{tabular}{@{}c@{}c@{}c@{}}
            \multicolumn{3}{@{}c@{}}{\includegraphics[width=0.141\linewidth]{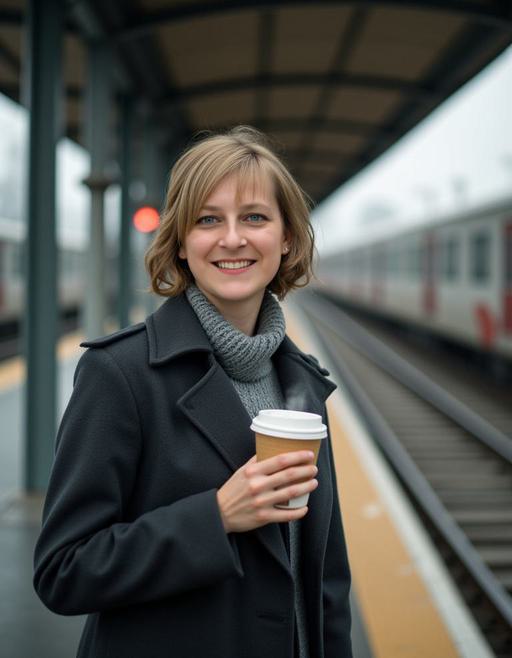}} \\
            \includegraphics[width=0.047\linewidth]{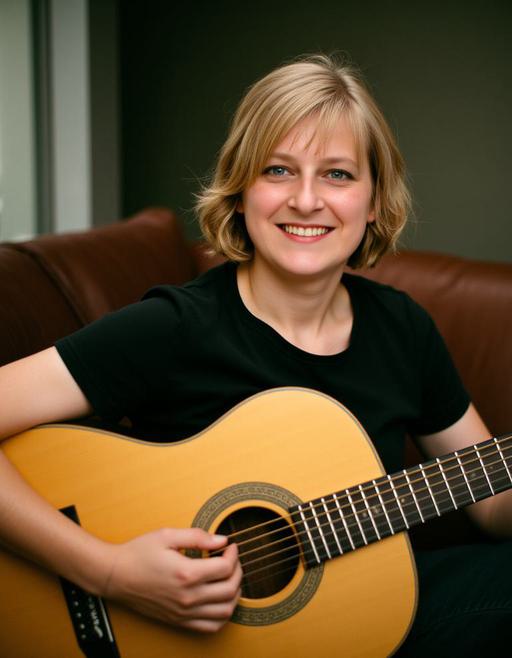} &
            \includegraphics[width=0.047\linewidth]{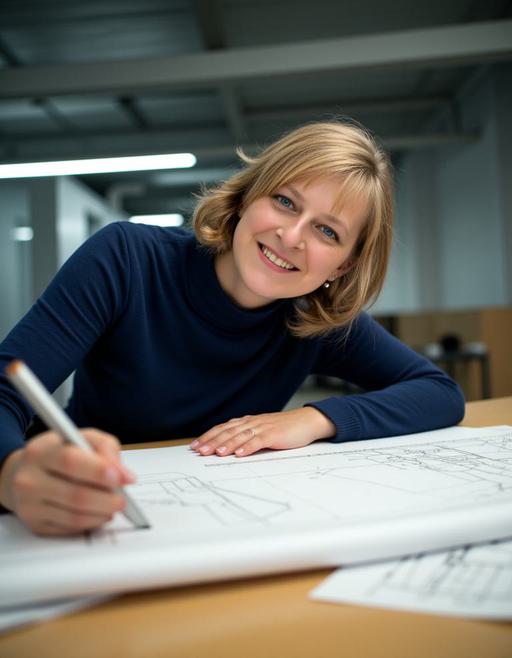} &
            \includegraphics[width=0.047\linewidth]{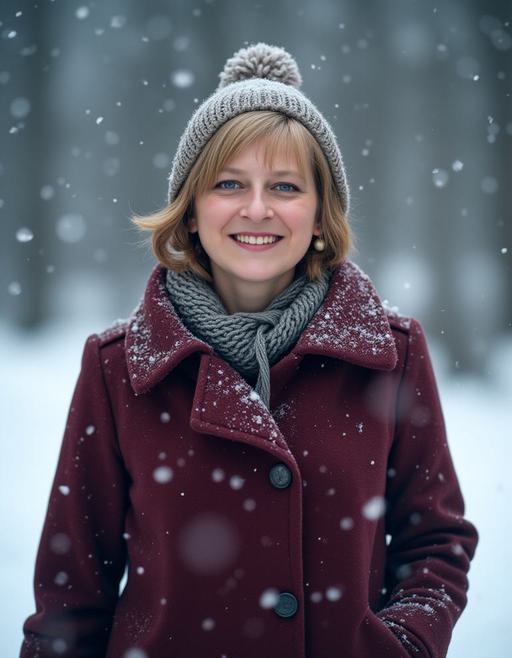}
         \end{tabular}%
        } &
        {\setlength{\tabcolsep}{0pt}%
         \renewcommand{\arraystretch}{0}%
         \begin{tabular}{@{}c@{}c@{}c@{}}
            \multicolumn{3}{@{}c@{}}{\includegraphics[width=0.141\linewidth]{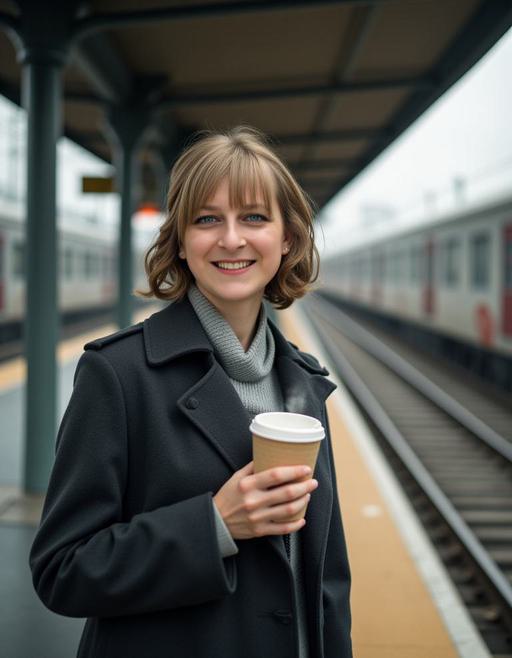}} \\
            \includegraphics[width=0.047\linewidth]{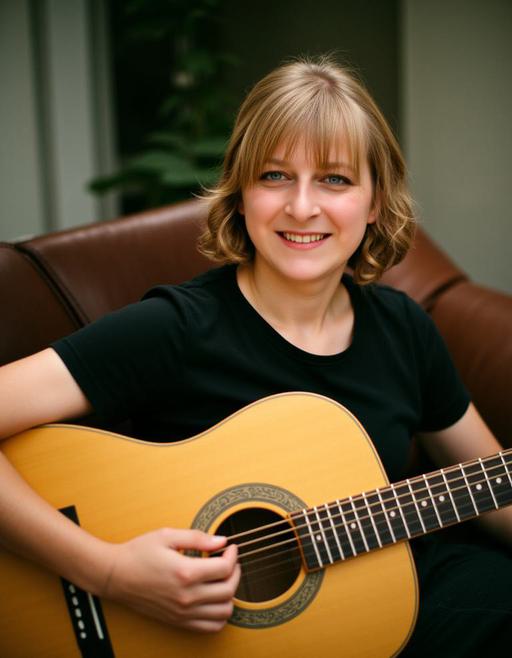} &
            \includegraphics[width=0.047\linewidth]{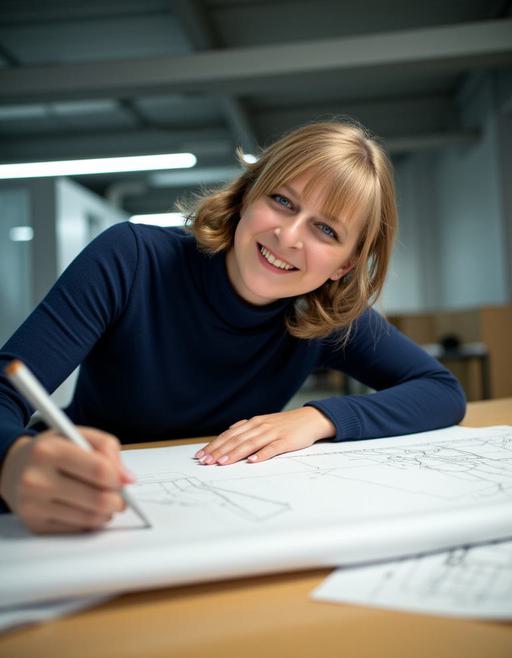} &
            \includegraphics[width=0.047\linewidth]{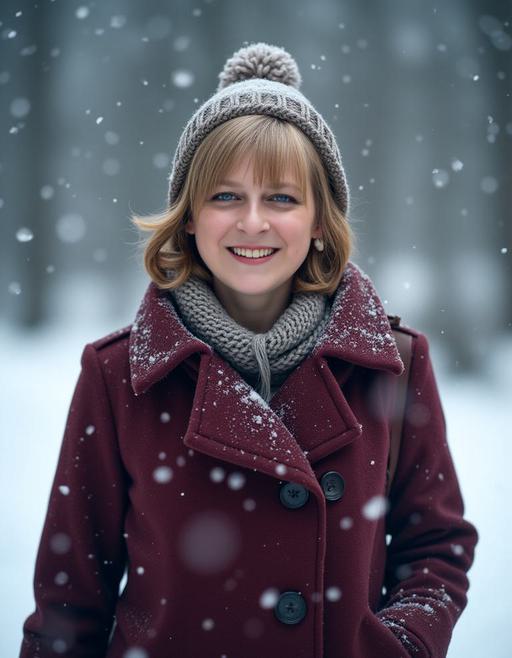}
         \end{tabular}%
        } &
        {\setlength{\tabcolsep}{0pt}%
         \renewcommand{\arraystretch}{0}%
         \begin{tabular}{@{}c@{}c@{}c@{}}
            \multicolumn{3}{@{}c@{}}{\includegraphics[width=0.141\linewidth]{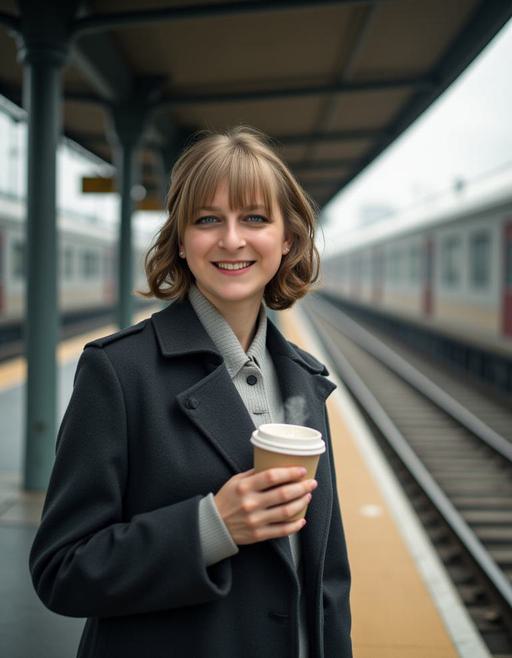}} \\
            \includegraphics[width=0.047\linewidth]{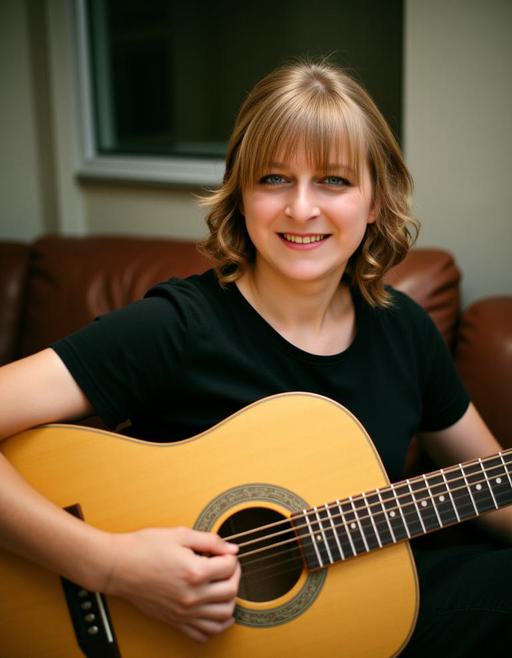} &
            \includegraphics[width=0.047\linewidth]{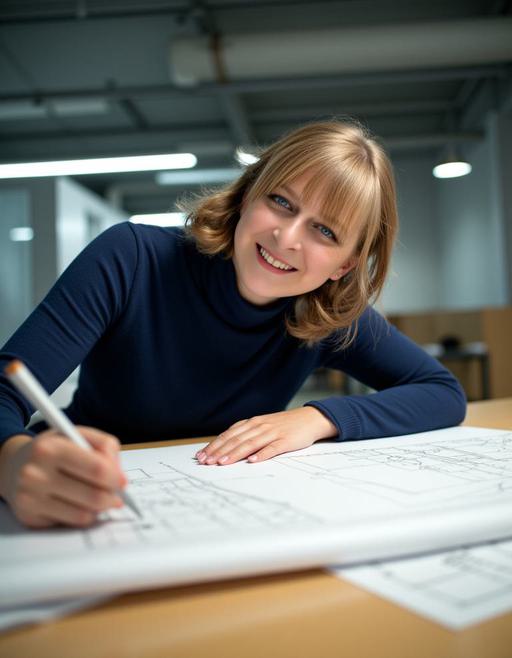} &
            \includegraphics[width=0.047\linewidth]{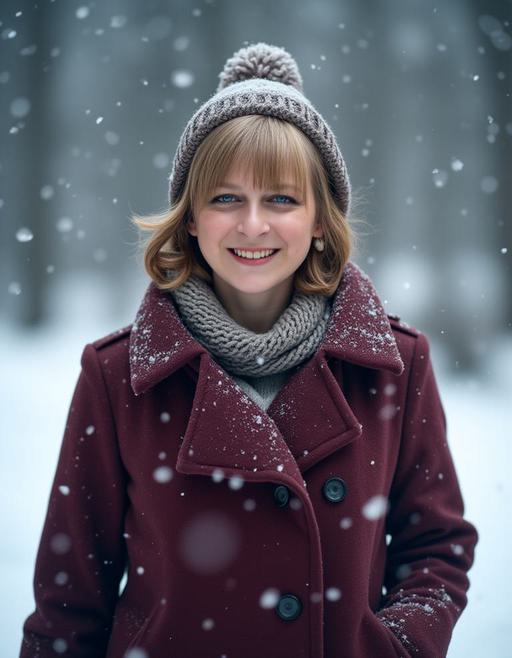}
         \end{tabular}%
        } &
        {\setlength{\tabcolsep}{0pt}%
         \renewcommand{\arraystretch}{0}%
         \begin{tabular}{@{}c@{}c@{}c@{}}
            \multicolumn{3}{@{}c@{}}{\includegraphics[width=0.141\linewidth]{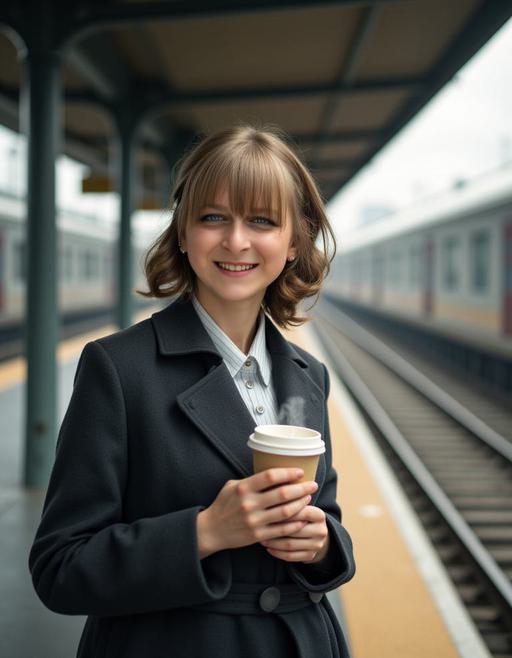}} \\
            \includegraphics[width=0.047\linewidth]{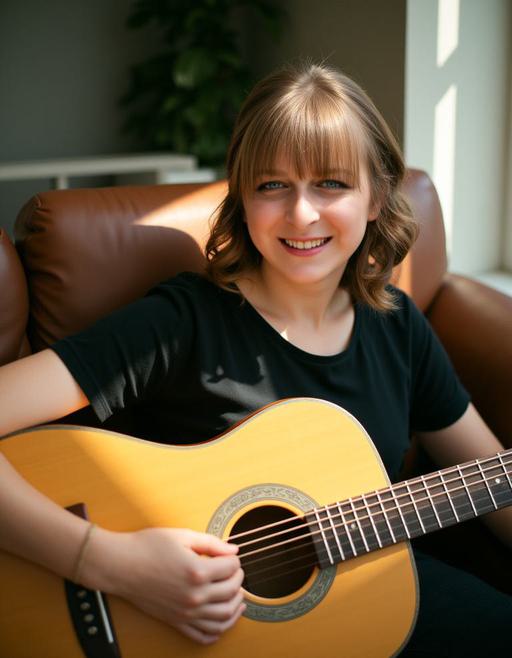} &
            \includegraphics[width=0.047\linewidth]{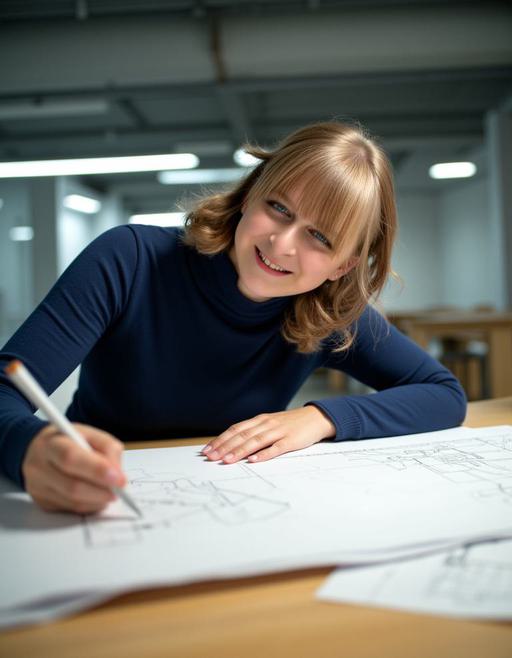} &
            \includegraphics[width=0.047\linewidth]{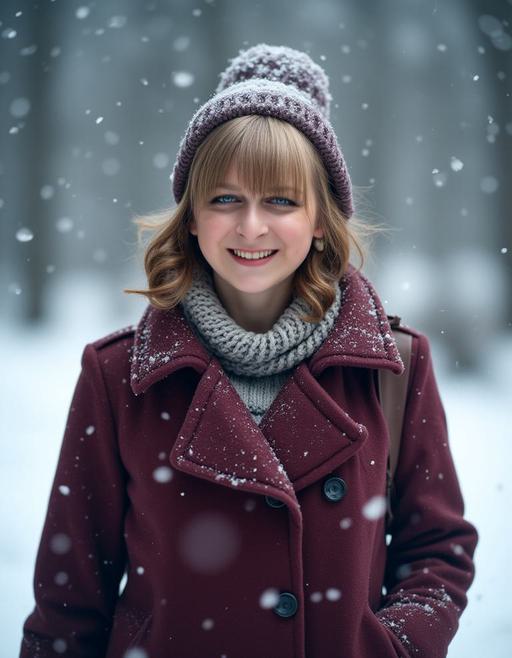}
         \end{tabular}%
        } \\ \\

        {\setlength{\tabcolsep}{0pt}%
         \renewcommand{\arraystretch}{0}%
         \begin{tabular}{@{}c@{}c@{}c@{}}
            \multicolumn{3}{@{}c@{}}{\includegraphics[width=0.141\linewidth]{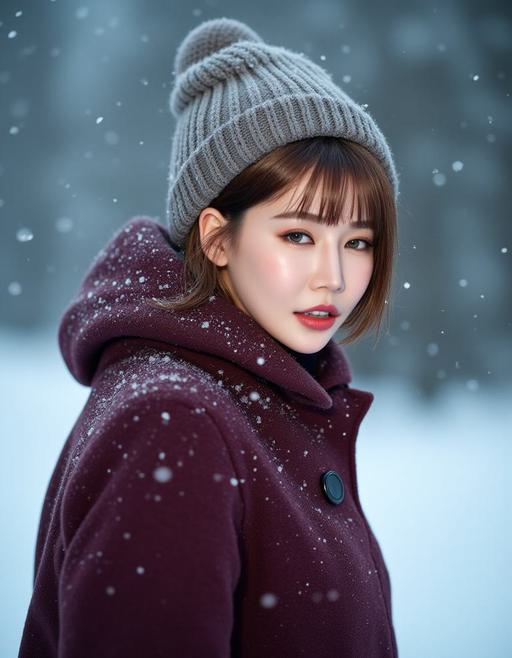}} \\
            \includegraphics[width=0.047\linewidth]{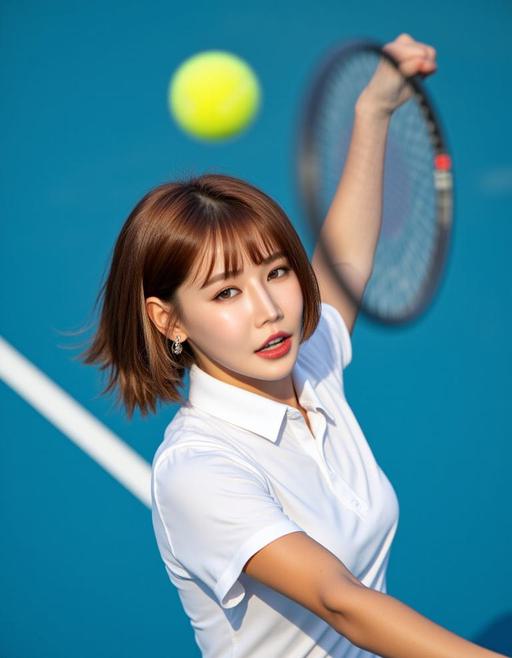} &
            \includegraphics[width=0.047\linewidth]{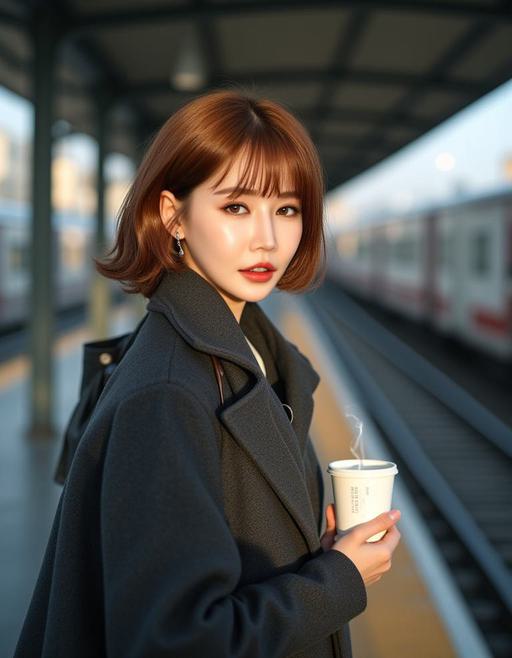} &
            \includegraphics[width=0.047\linewidth]{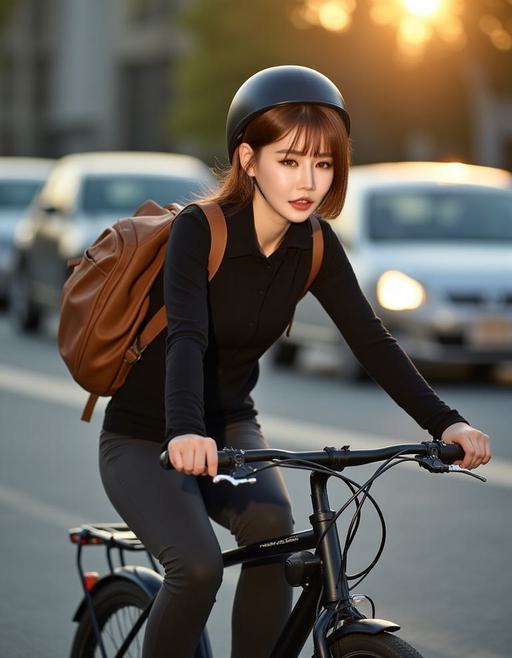}
         \end{tabular}%
        } &
        {\setlength{\tabcolsep}{0pt}%
         \renewcommand{\arraystretch}{0}%
         \begin{tabular}{@{}c@{}c@{}c@{}}
            \multicolumn{3}{@{}c@{}}{\includegraphics[width=0.141\linewidth]{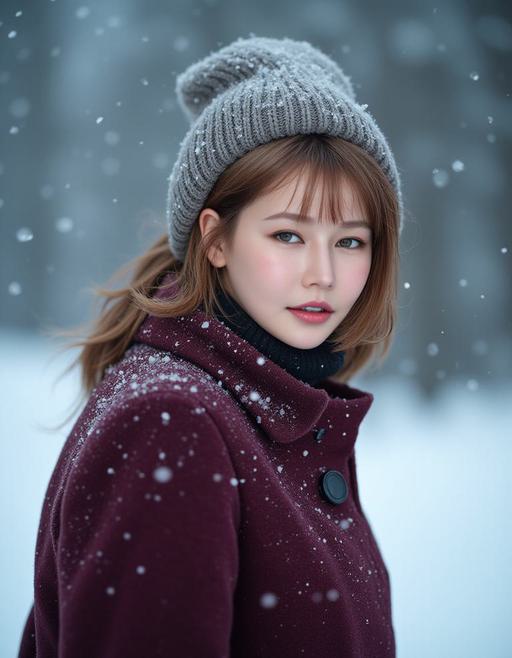}} \\
            \includegraphics[width=0.047\linewidth]{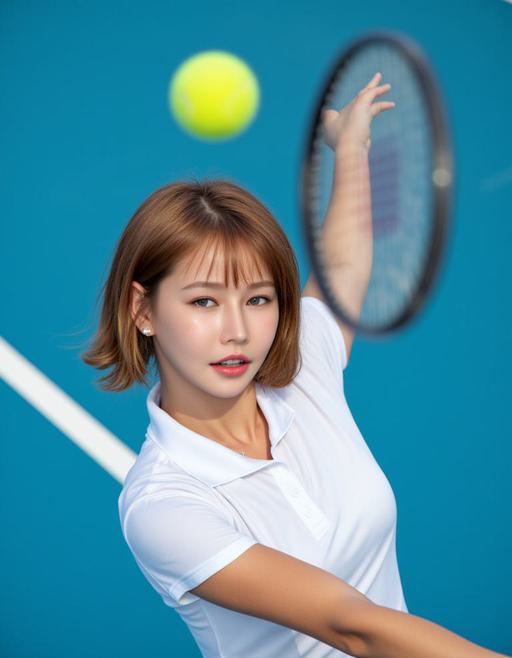} &
            \includegraphics[width=0.047\linewidth]{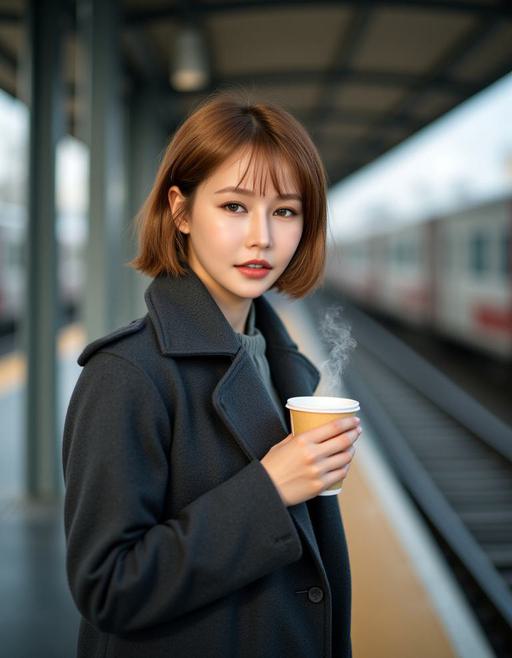} &
            \includegraphics[width=0.047\linewidth]{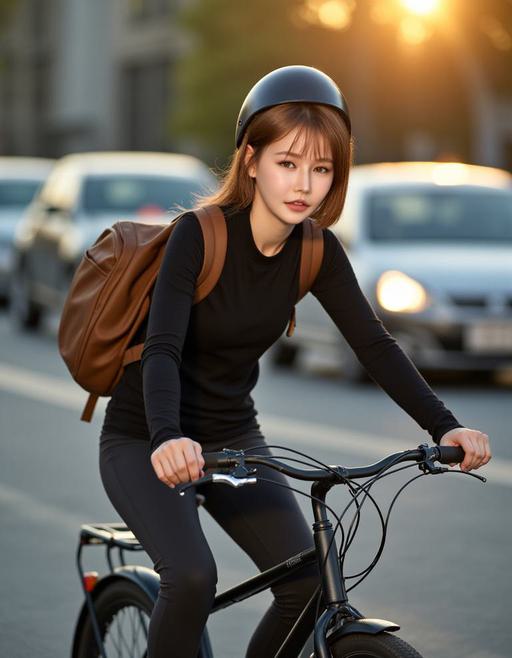}
         \end{tabular}%
        } &
        {\setlength{\tabcolsep}{0pt}%
         \renewcommand{\arraystretch}{0}%
         \begin{tabular}{@{}c@{}c@{}c@{}}
            \multicolumn{3}{@{}c@{}}{\includegraphics[width=0.141\linewidth]{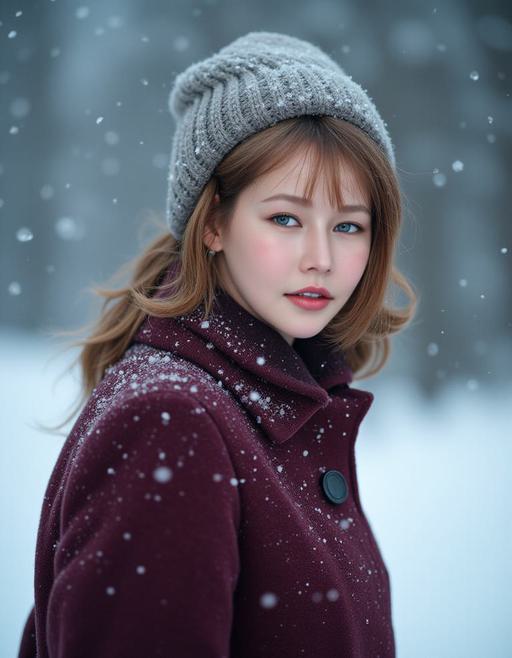}} \\
            \includegraphics[width=0.047\linewidth]{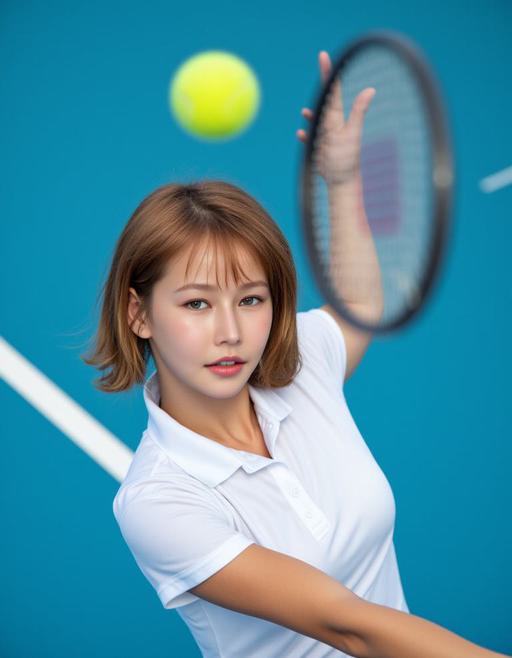} &
            \includegraphics[width=0.047\linewidth]{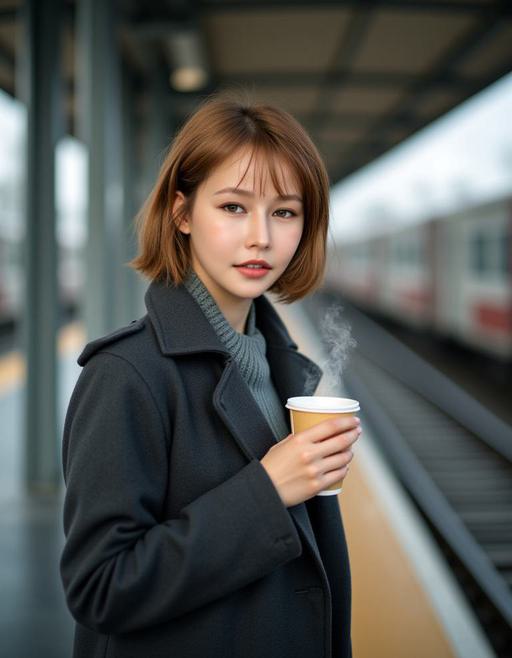} &
            \includegraphics[width=0.047\linewidth]{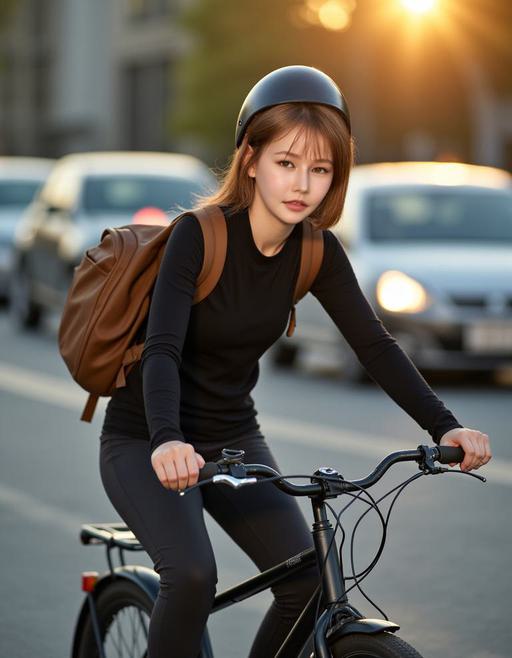}
         \end{tabular}%
        } &
        {\setlength{\tabcolsep}{0pt}%
         \renewcommand{\arraystretch}{0}%
         \begin{tabular}{@{}c@{}c@{}c@{}}
            \multicolumn{3}{@{}c@{}}{\includegraphics[width=0.141\linewidth]{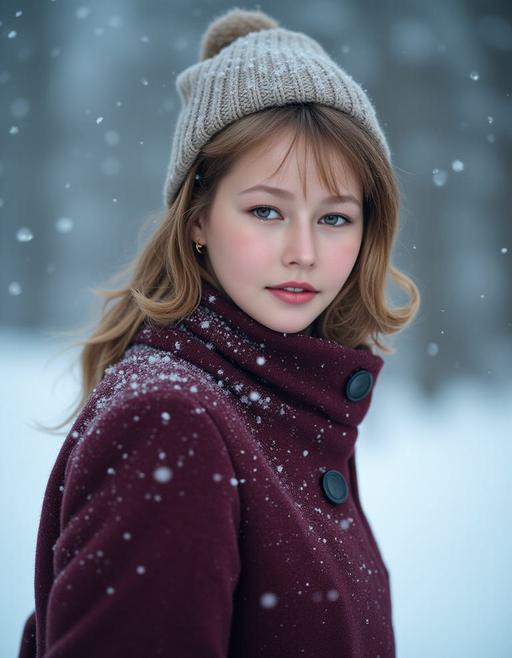}} \\
            \includegraphics[width=0.047\linewidth]{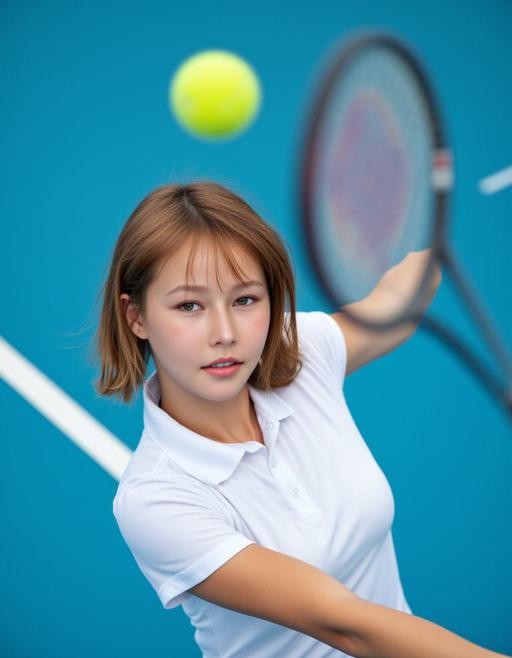} &
            \includegraphics[width=0.047\linewidth]{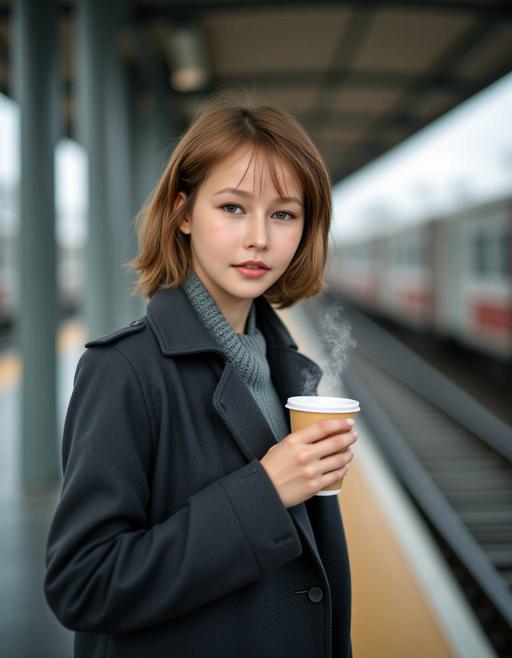} &
            \includegraphics[width=0.047\linewidth]{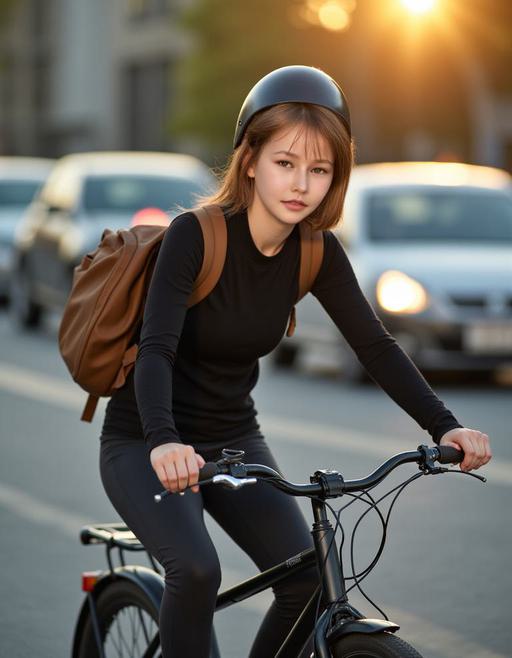}
         \end{tabular}%
        } &
        {\setlength{\tabcolsep}{0pt}%
         \renewcommand{\arraystretch}{0}%
         \begin{tabular}{@{}c@{}c@{}c@{}}
            \multicolumn{3}{@{}c@{}}{\includegraphics[width=0.141\linewidth]{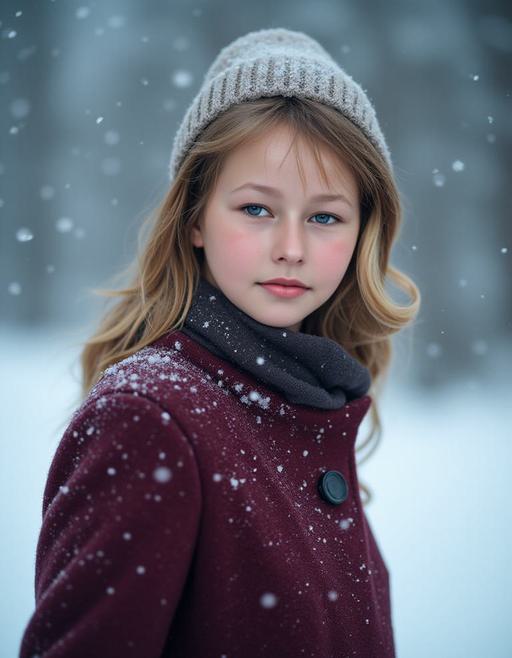}} \\
            \includegraphics[width=0.047\linewidth]{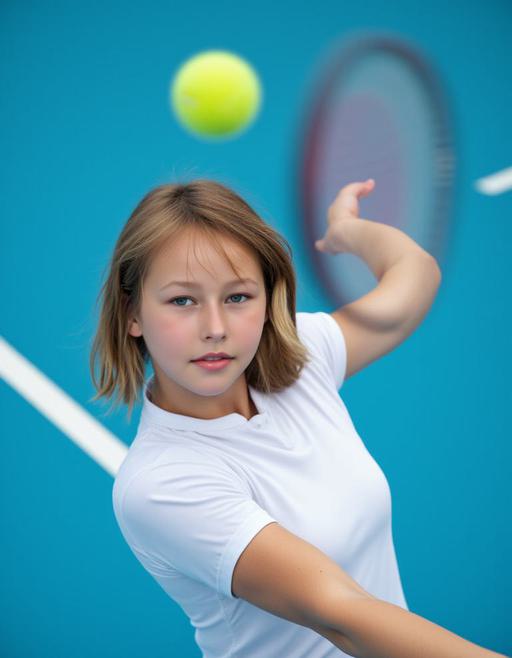} &
            \includegraphics[width=0.047\linewidth]{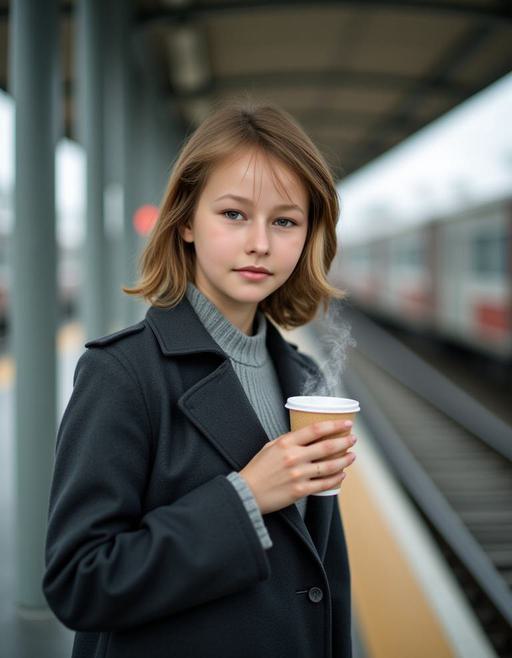} &
            \includegraphics[width=0.047\linewidth]{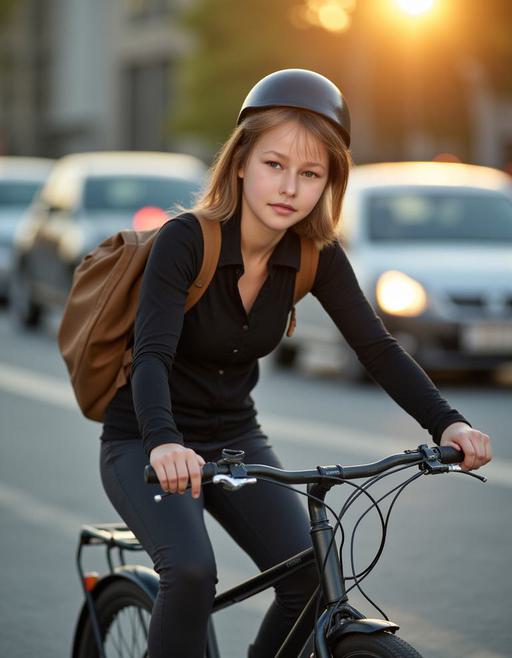}
         \end{tabular}%
        } &
        {\setlength{\tabcolsep}{0pt}%
         \renewcommand{\arraystretch}{0}%
         \begin{tabular}{@{}c@{}c@{}c@{}}
            \multicolumn{3}{@{}c@{}}{\includegraphics[width=0.141\linewidth]{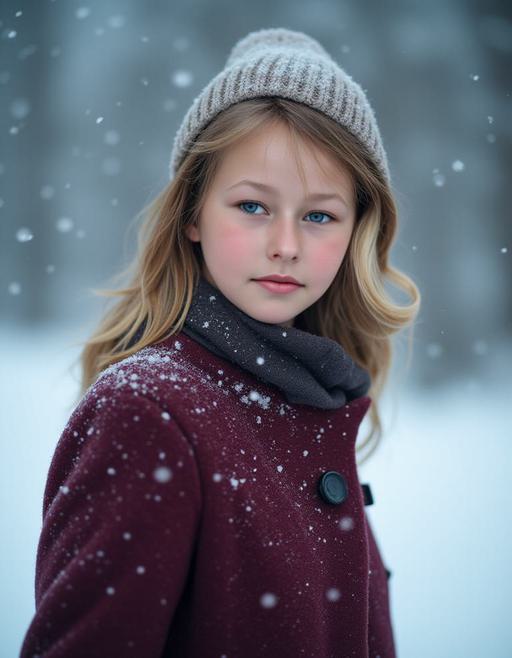}} \\
            \includegraphics[width=0.047\linewidth]{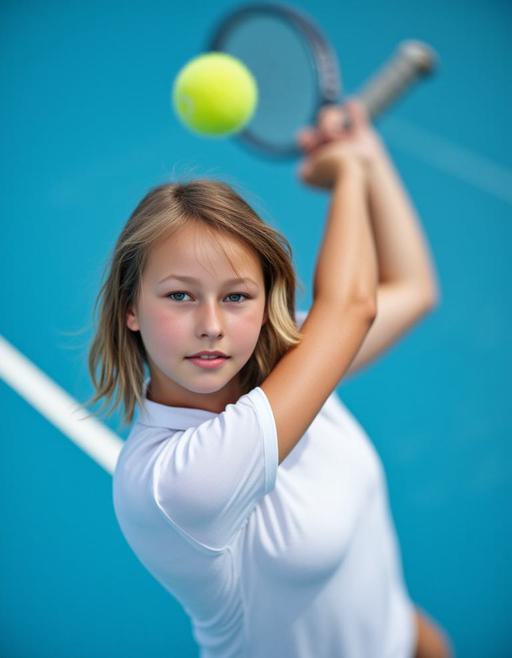} &
            \includegraphics[width=0.047\linewidth]{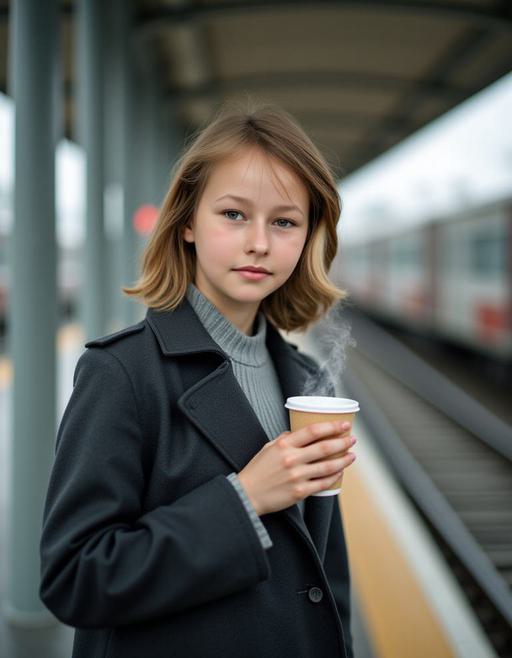} &
            \includegraphics[width=0.047\linewidth]{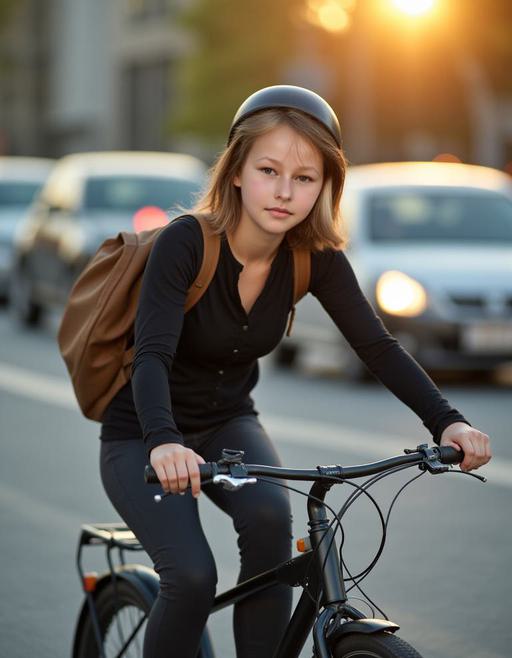}
         \end{tabular}%
        } &
        {\setlength{\tabcolsep}{0pt}%
         \renewcommand{\arraystretch}{0}%
         \begin{tabular}{@{}c@{}c@{}c@{}}
            \multicolumn{3}{@{}c@{}}{\includegraphics[width=0.141\linewidth]{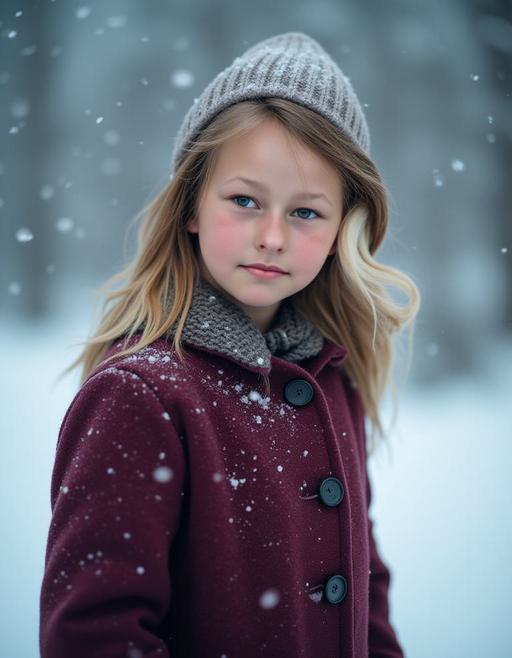}} \\
            \includegraphics[width=0.047\linewidth]{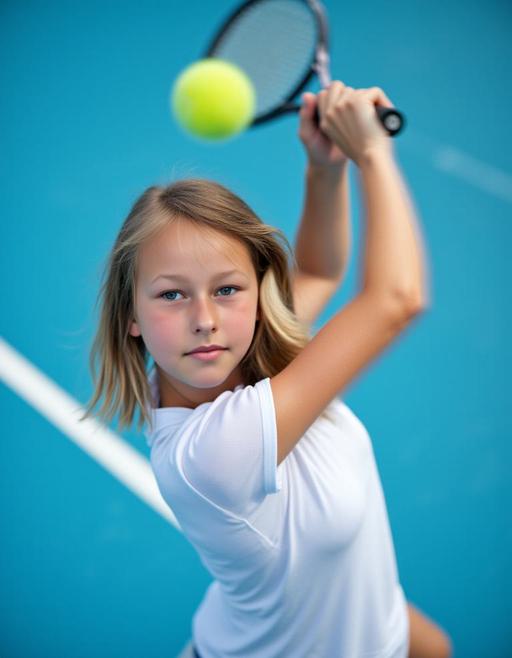} &
            \includegraphics[width=0.047\linewidth]{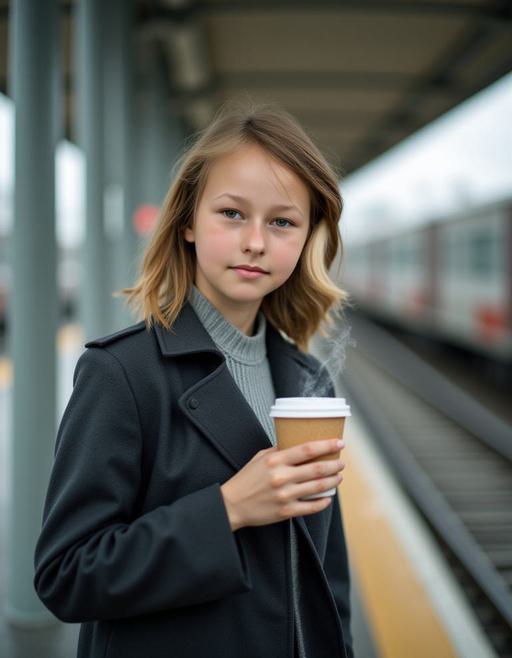} &
            \includegraphics[width=0.047\linewidth]{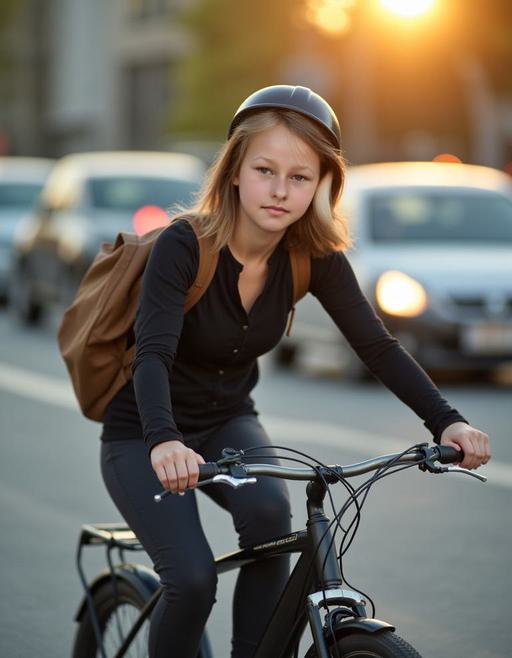}
         \end{tabular}%
        } \\
        
        Identity A &&&&&& Identity B \\

    \end{tabular}
    \caption{Continuous identity space via interpolation. The leftmost and rightmost columns use the encoding of two different identities (A and B). Intermediate columns are generated by linearly interpolating between their identity embeddings and applying the mixed embedding to all tokens. The identity changes gradually without abrupt jumps, illustrating the continuity of the learned identity representation across varied scenes.}
    \label{fig:pulid_blend}
\end{figure}

\begin{figure}
    \setlength{\tabcolsep}{0pt}
    \scriptsize{
        \begin{tabular}{cccc}
            \includegraphics[width=0.26\linewidth]{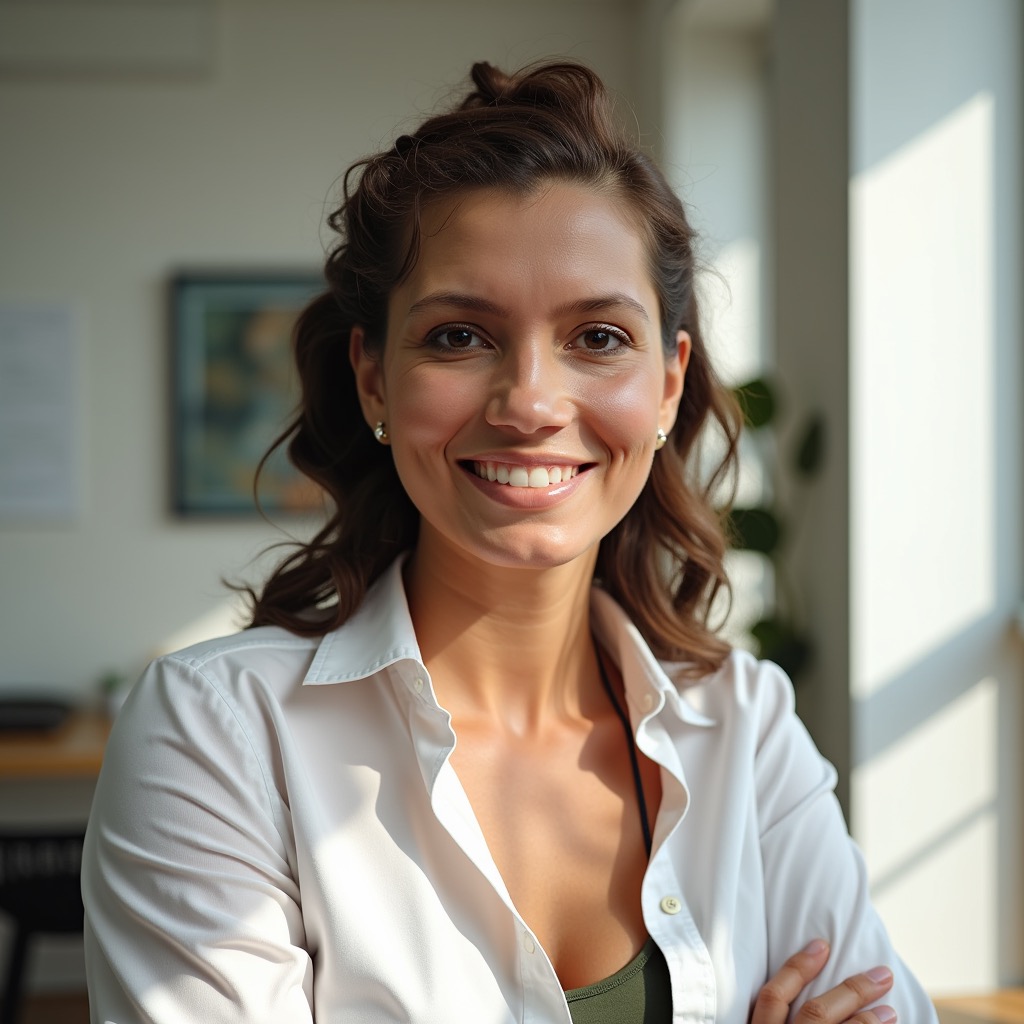} &
            \includegraphics[width=0.26\linewidth]{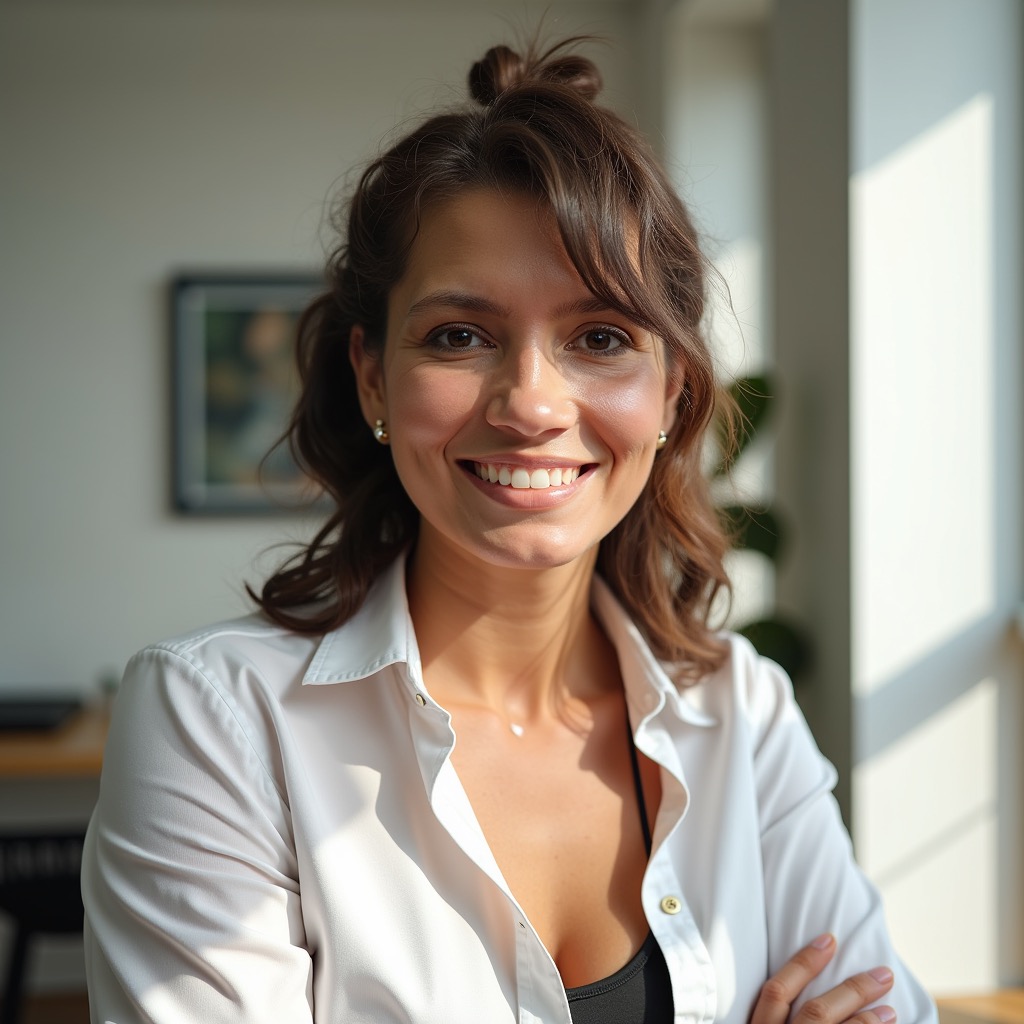} &
            \includegraphics[width=0.26\linewidth]{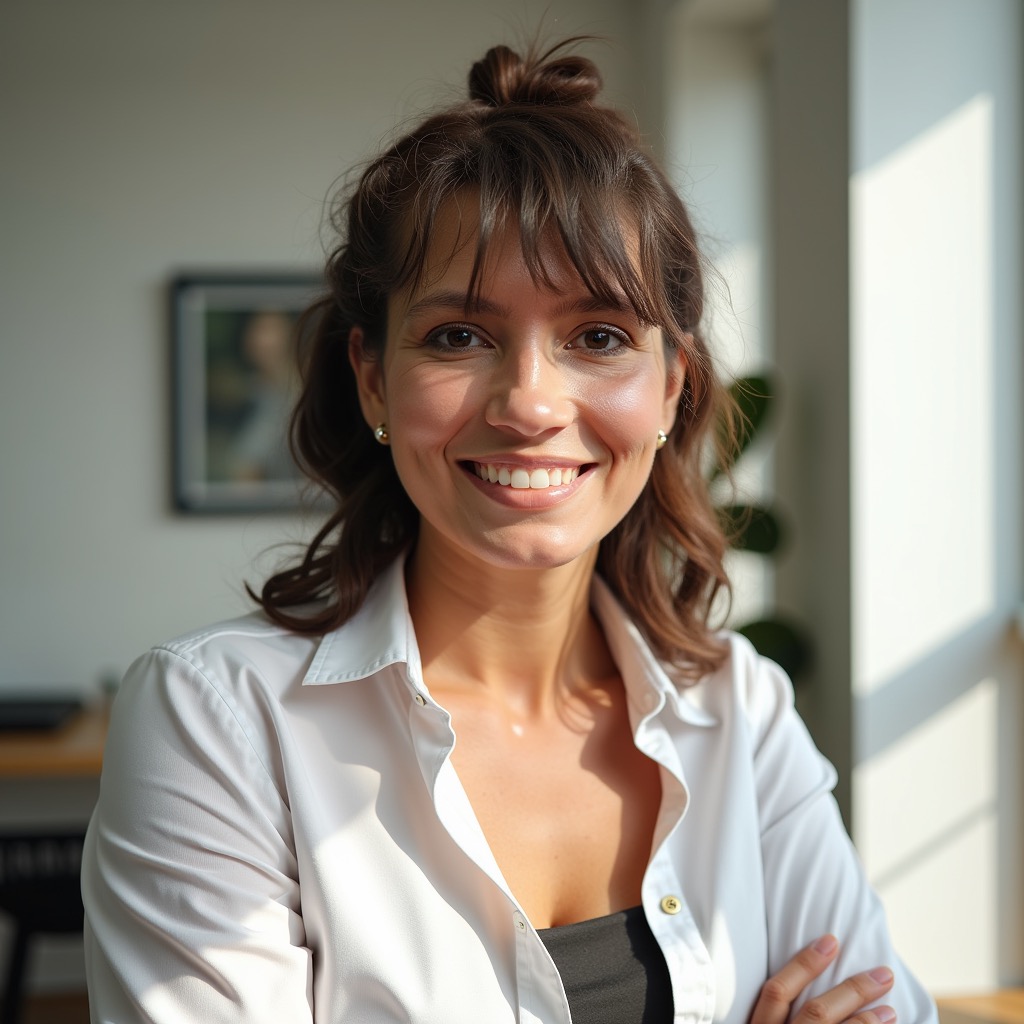} &
            \includegraphics[width=0.26\linewidth]{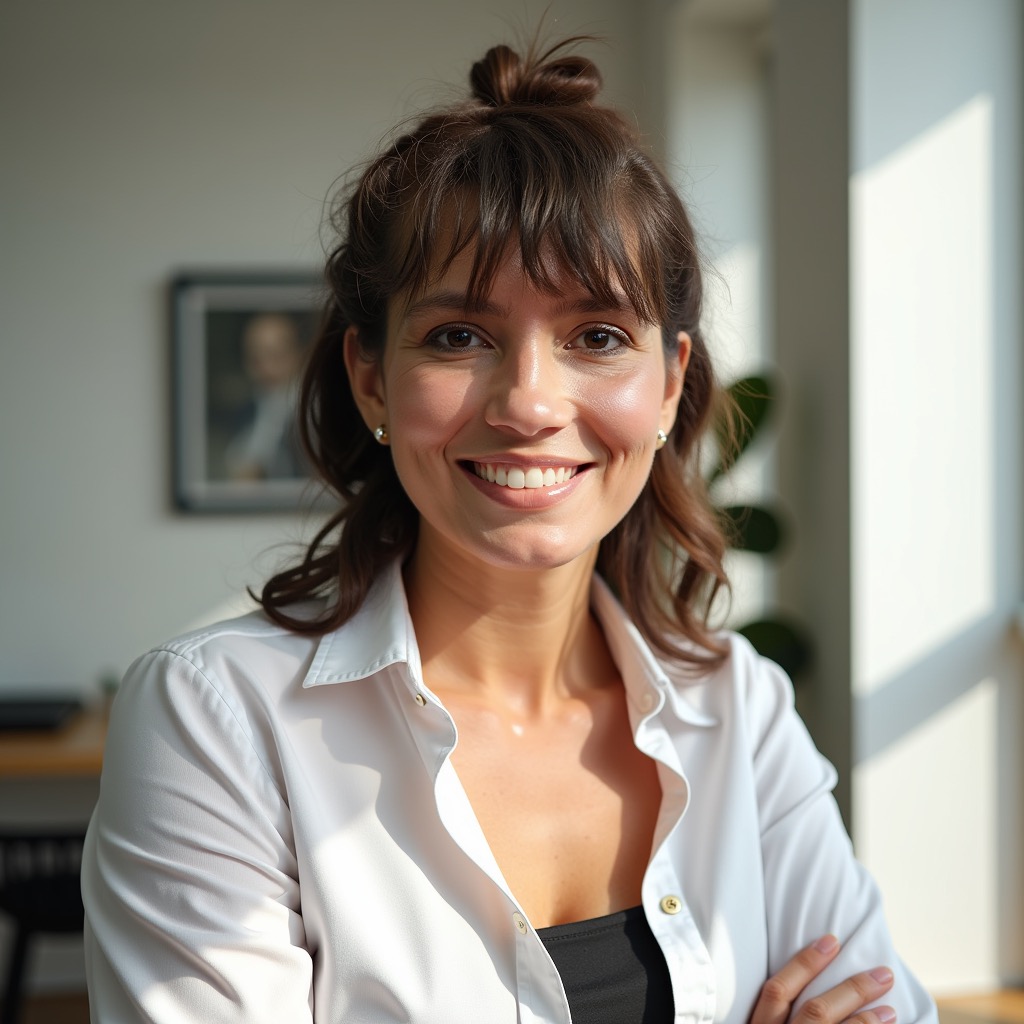} \\
            \includegraphics[clip, viewport=384bp 756bp 640bp 820bp, width=0.26\linewidth]{images/slider/woman_bangs/woman_bangs_0.jpg} &
            \includegraphics[clip, viewport=384bp 756bp 640bp 820bp, width=0.26\linewidth]{images/slider/woman_bangs/woman_bangs_75.jpg} &
            \includegraphics[clip,  viewport=384bp 756bp 640bp 820bp, width=0.26\linewidth]{images/slider/woman_bangs/woman_bangs_100.jpg} &
            \includegraphics[clip,  viewport=384bp 756bp 640bp 820bp, width=0.26\linewidth]{images/slider/woman_bangs/woman_bangs_150.jpg} \\
            Original Identity &
            \multicolumn{3}{c}{bangs + $\xrightarrow{\hspace{15em}}$} \\
            
            \includegraphics[clip, viewport=256bp 425bp 768bp 937bp,width=0.26\linewidth]{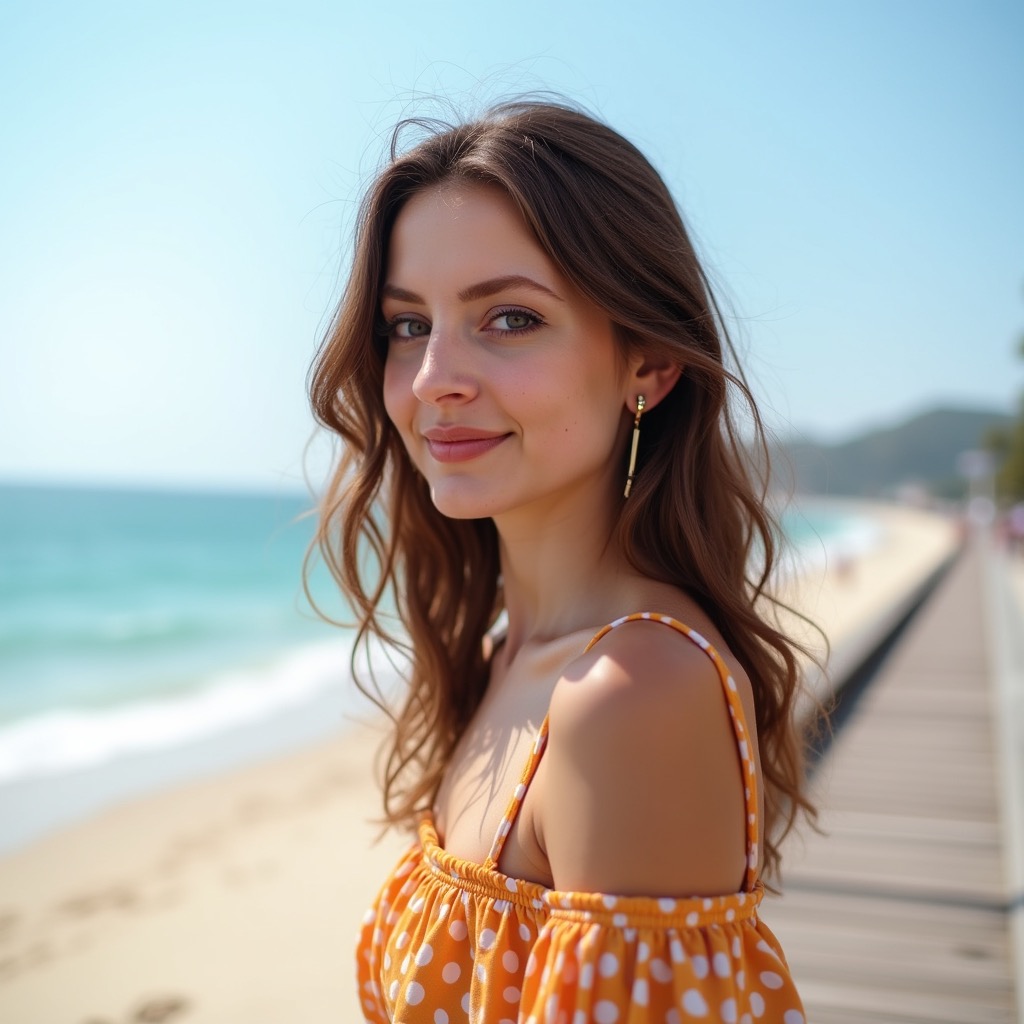} &
            \includegraphics[clip, viewport=256bp 425bp 768bp 937bp, width=0.26\linewidth]{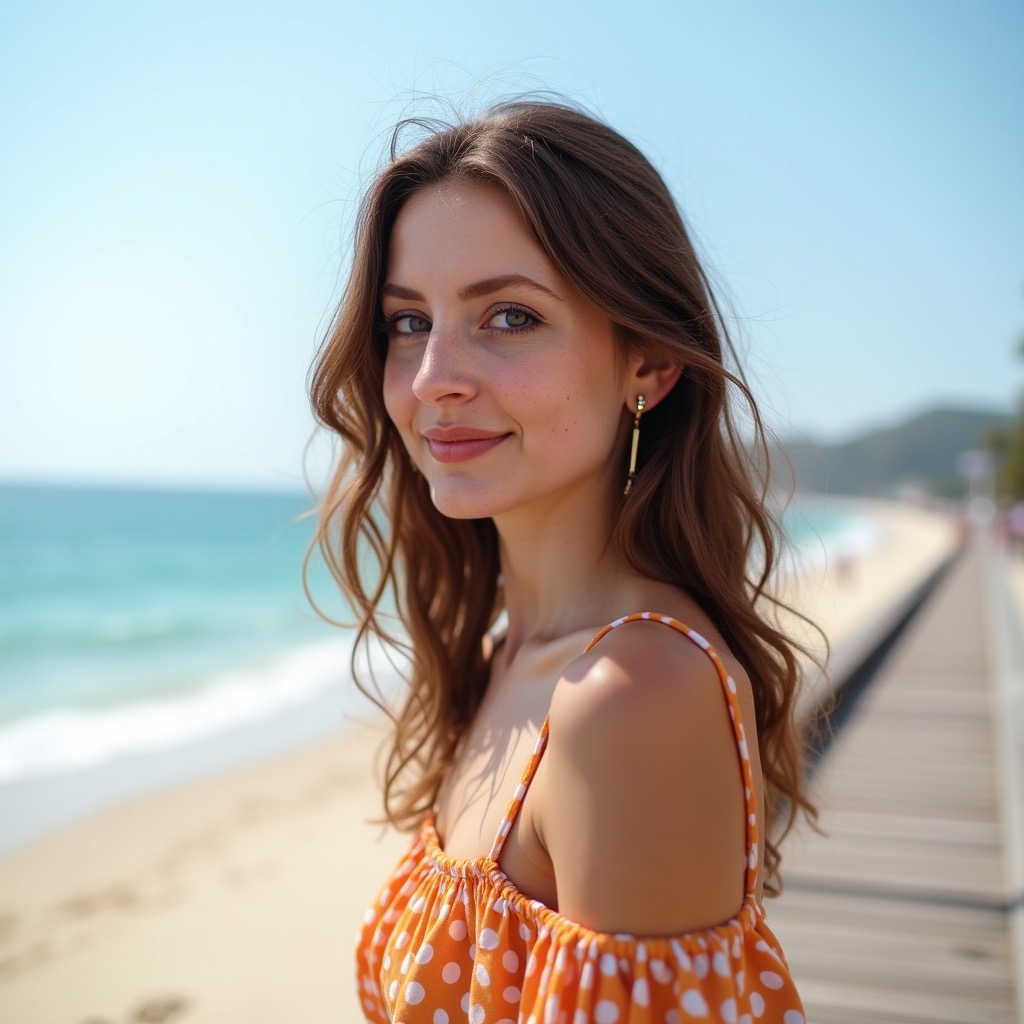} &
            \includegraphics[clip, viewport=256bp 425bp 768bp 937bp, width=0.26\linewidth]{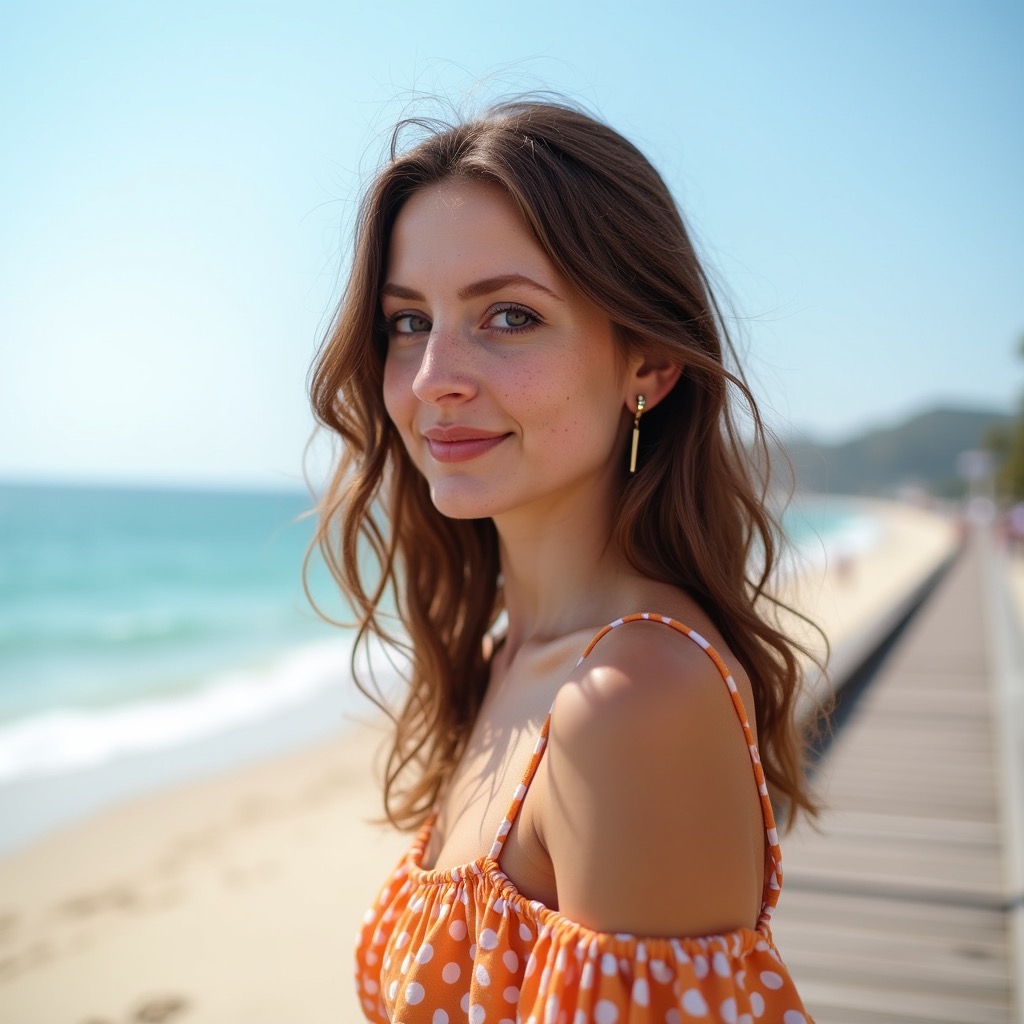} &
            \includegraphics[clip, viewport=256bp 425bp 768bp 937bp, width=0.26\linewidth]{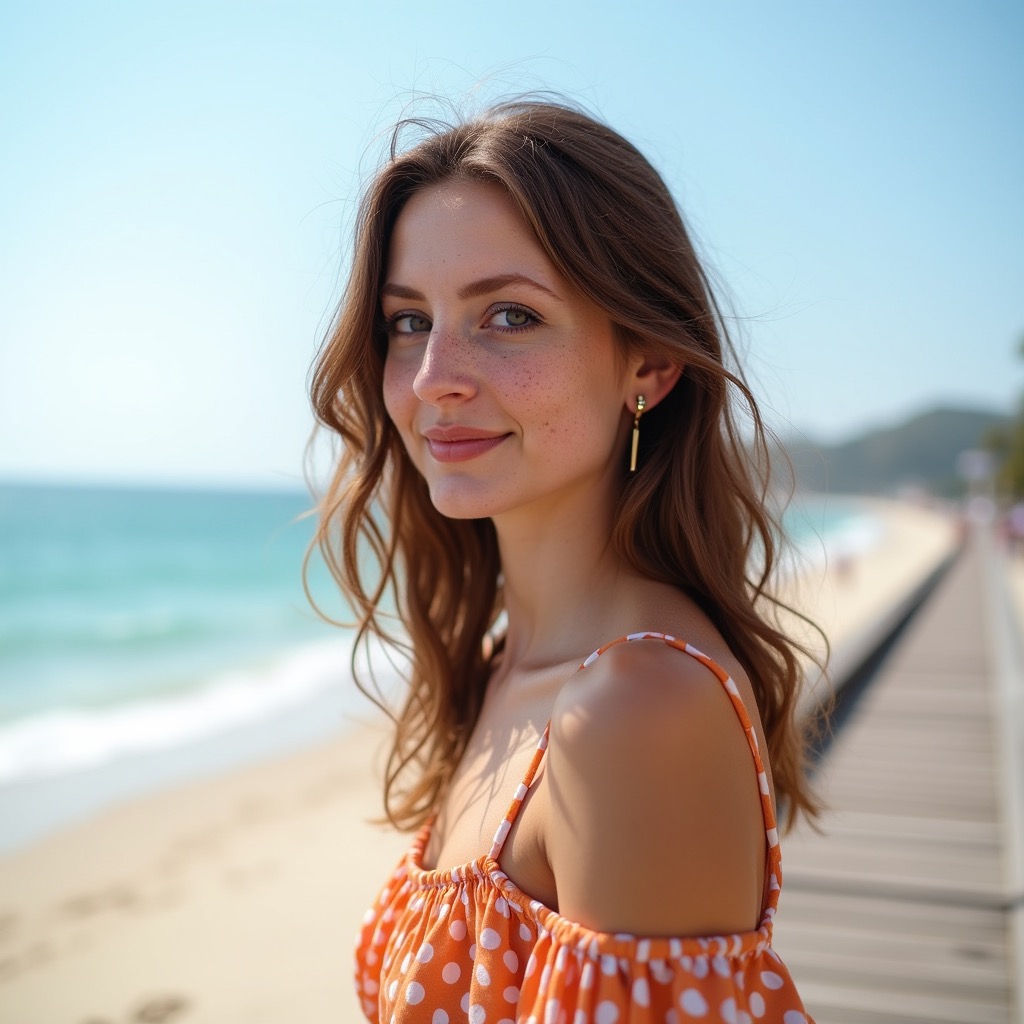} \\
            \includegraphics[clip, viewport=384bp 608bp 640bp 672bp, width=0.26\linewidth]{images/slider/freckles/freckles_0.jpg} &
            \includegraphics[clip, viewport=384bp 608bp 640bp 672bp, width=0.26\linewidth]{images/slider/freckles/freckles_50.jpg} &
            \includegraphics[clip, viewport=384bp 608bp 640bp 672bp, width=0.26\linewidth]{images/slider/freckles/freckles_75.jpg} &
            \includegraphics[clip, viewport=384bp 608bp 640bp 672bp, width=0.26\linewidth]{images/slider/freckles/freckles_150.jpg} \\
            Original Identity &
            \multicolumn{3}{c}{freckles + $\xrightarrow{\hspace{15em}}$} \\
            
            \includegraphics[clip, viewport=256bp 425bp 768bp 937bp, width=0.26\linewidth]{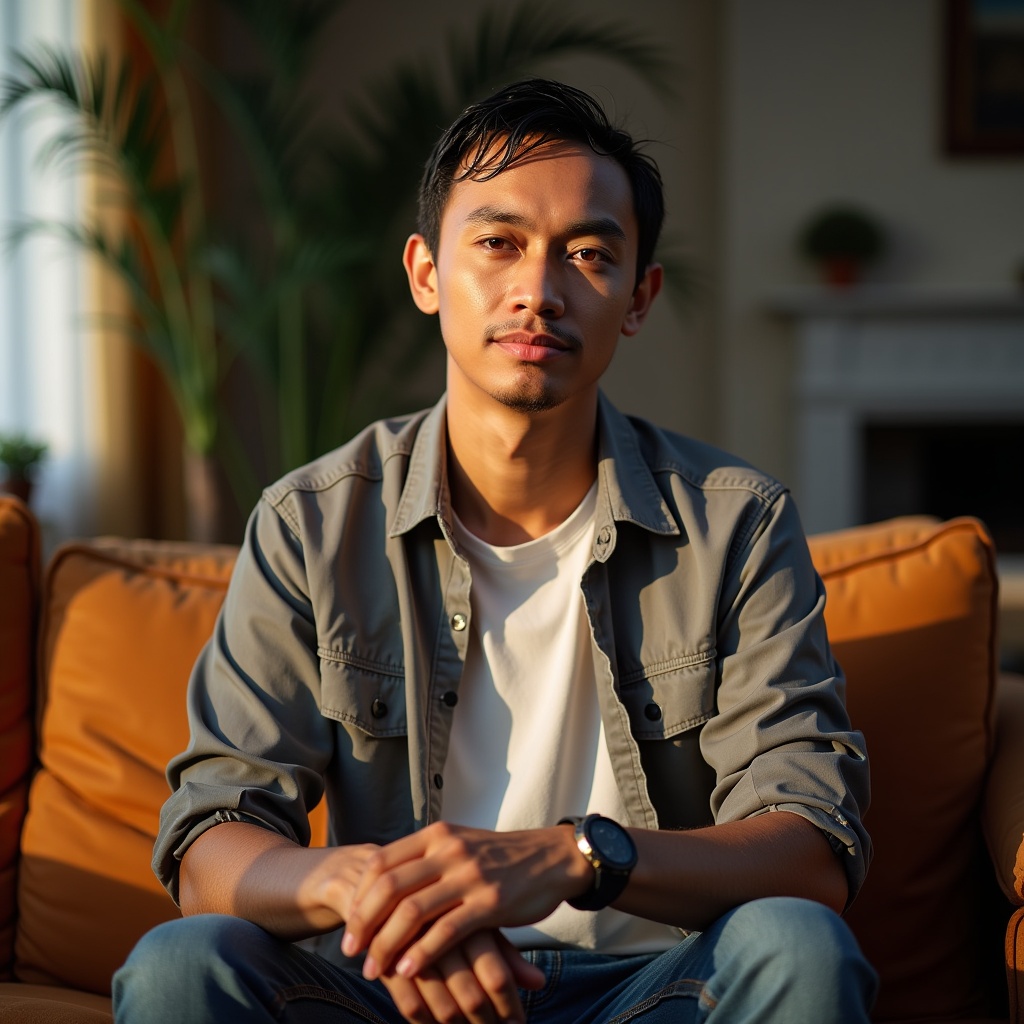} &
            \includegraphics[clip, viewport=256bp 425bp 768bp 937bp, width=0.26\linewidth]{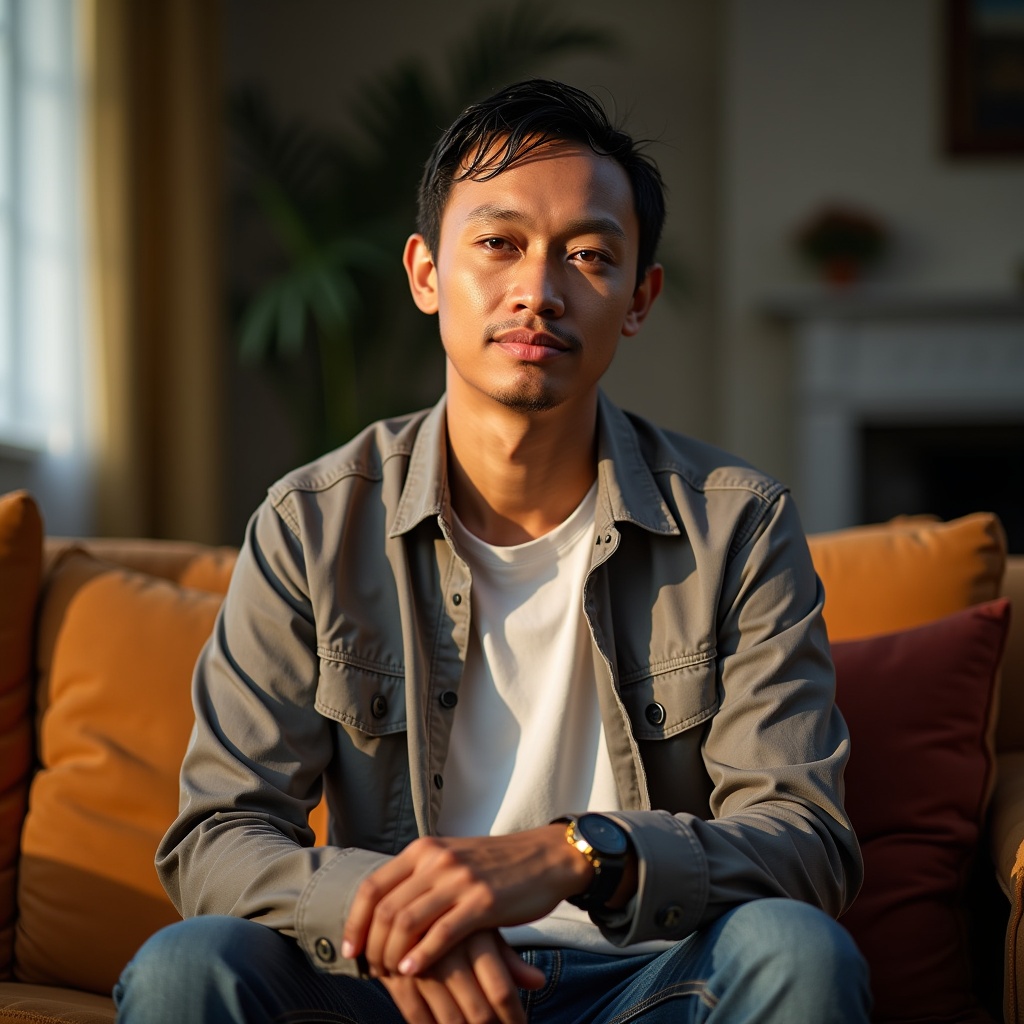} &
            \includegraphics[clip, viewport=256bp 425bp 768bp 937bp, width=0.26\linewidth]{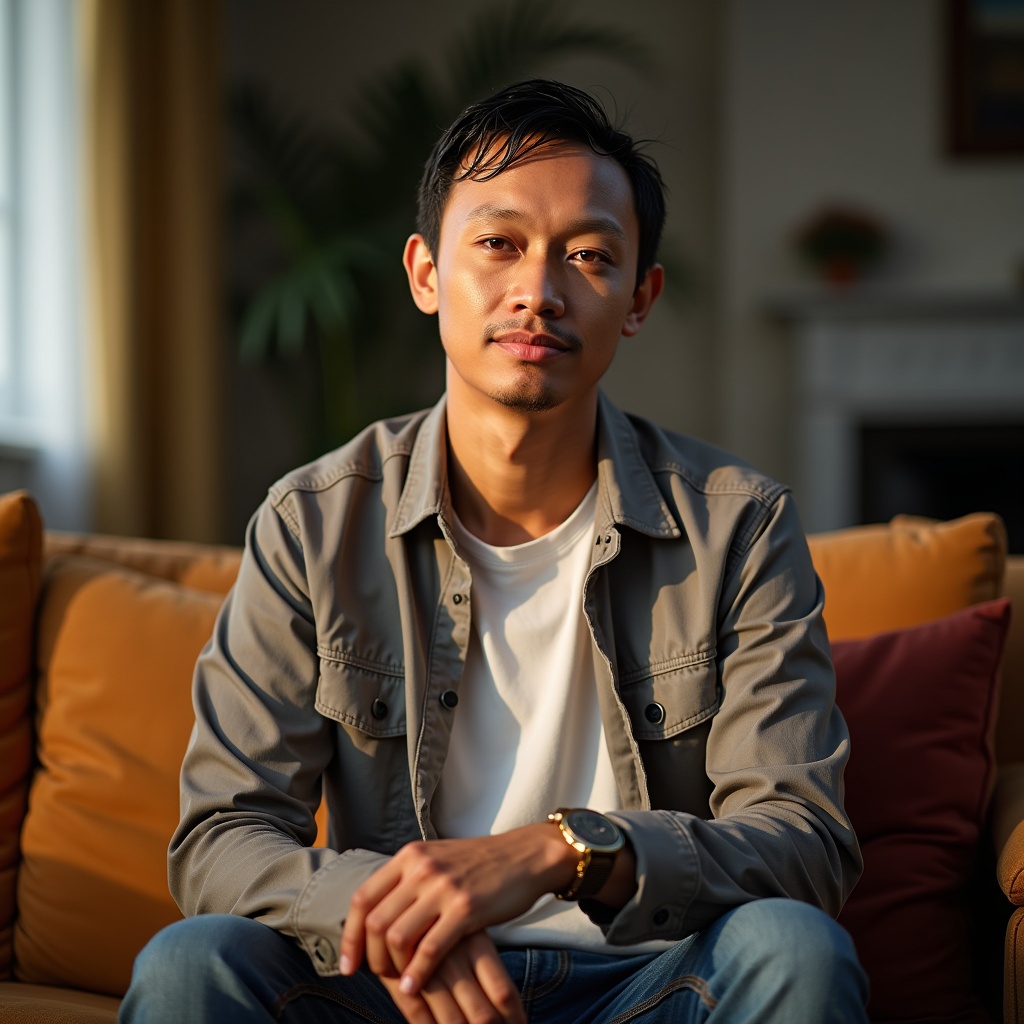} &
            \includegraphics[clip, viewport=256bp 425bp 768bp 937bp, width=0.26\linewidth]{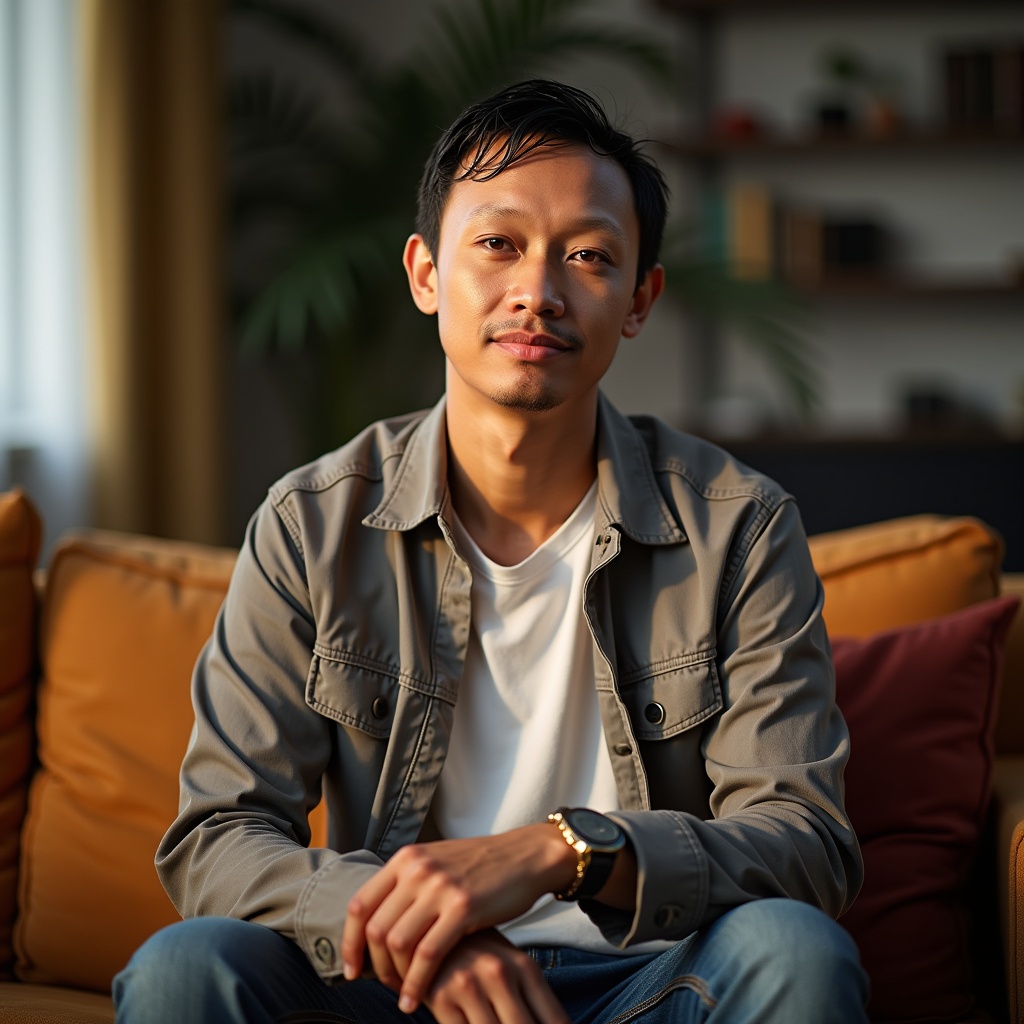} \\
            \includegraphics[clip, viewport=384bp 756bp 640bp 820bp, width=0.26\linewidth]{images/slider/eyebrows2/original.jpg} &
            \includegraphics[clip, viewport=384bp 756bp 640bp 820bp, width=0.26\linewidth]{images/slider/eyebrows2/pc5_scaleneg50.0.jpg} &
            \includegraphics[clip, viewport=384bp 756bp 640bp 820bp, width=0.26\linewidth]{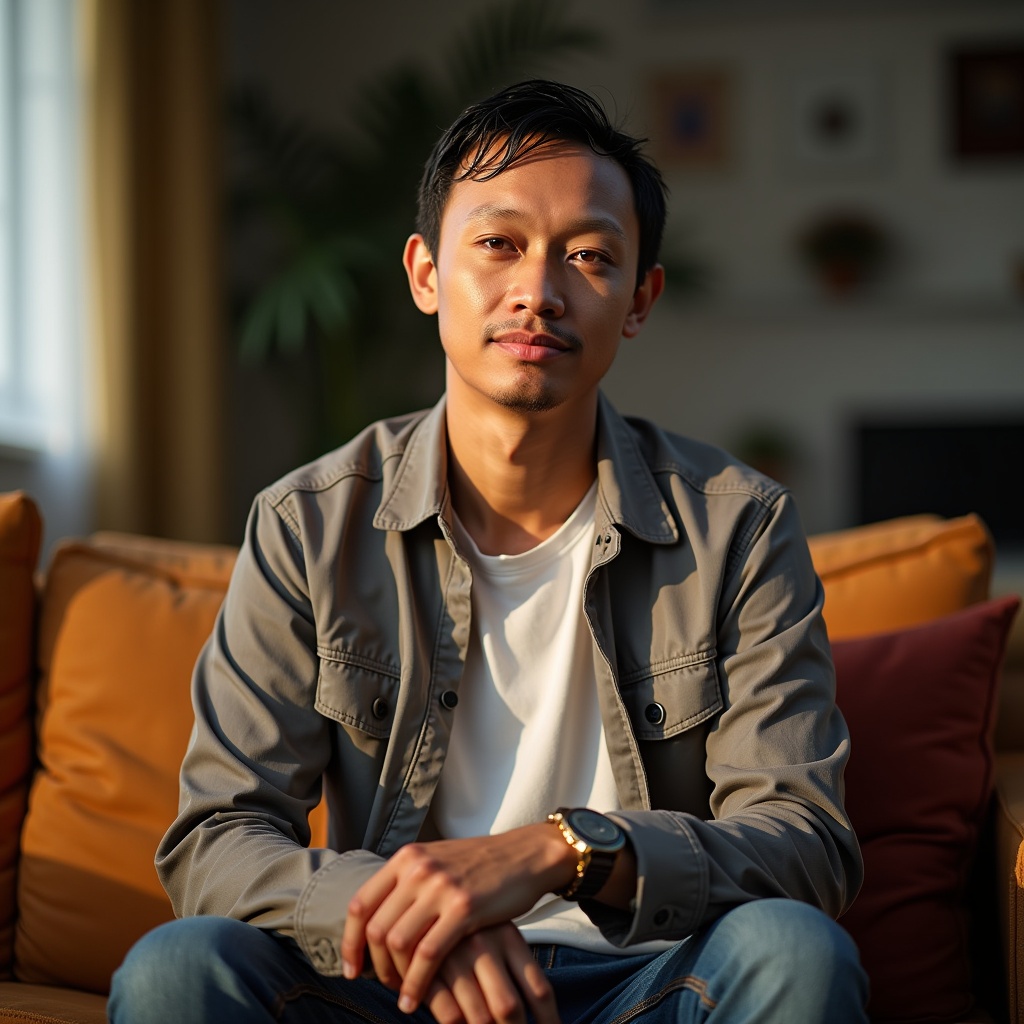} &
            \includegraphics[clip, viewport=384bp 756bp 640bp 820bp, width=0.26\linewidth]{images/slider/eyebrows2/pc5_scaleneg200.0.jpg} \\
            Original Identity &
            \multicolumn{3}{c}{Eyebrows - $\xrightarrow{\hspace{15em}}$} \\
            
        \end{tabular}
    }
    \vspace{-5pt}
    \caption{
        Slider-based results. Each row shows the gradual nature of our edits.
    }
    \vspace{-10pt}
    \label{fig:slider-results}
\end{figure}

To tune an identity, we perform vector addition in the identity latent space, allowing us to control the intensity of the edit using a scale parameter.
\cref{fig:sup_global_blend}
(Omni-ID) and
\cref {fig:pulid_blend}
(PuLID) illustrate a smooth transformation obtained by interpolating between two identities.
\cref{fig:slider-results}
shows three scale levels for each edit direction, derived either in an unsupervised manner via PCA (changing eyebrows) or in a supervised manner via attribute-aligned mean-difference (adding bangs) or SVM-based directions (adding freckles). These directions enable fine-grained, monotonic control over the strength of each edit.

\subsection{Q-Former Attention Maps}
\cref{fig:supp_attn_tokens}
presents the attention maps for the PuLID~\cite{guo2024pulid} encoder, which exhibit patterns similar to those of Omni-ID~\cite{qian2025omniidholisticidentityrepresentation} shown in
Figure 2 in the main paper.
In both cases, each learned token consistently focuses on a specific semantic region of the face, extracting localized information from areas such as the cheeks, forehead, eyebrows, nose, and eyes.

\begin{figure}
    \centering
    \setlength{\tabcolsep}{0pt}
    \scriptsize{
        \begin{tabular}{ccccc}
             \includegraphics[width=0.247\linewidth]{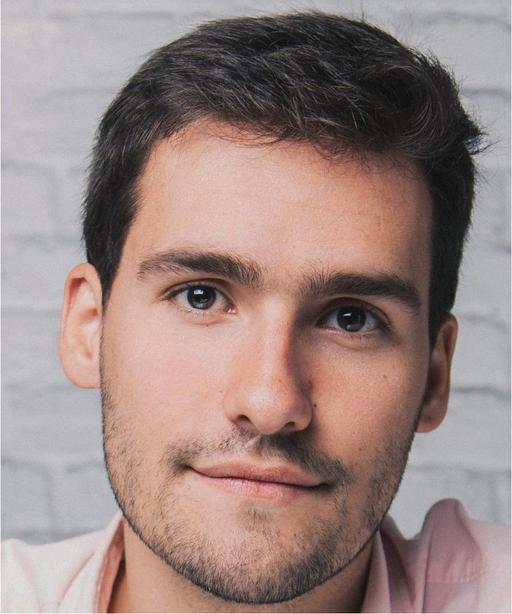} & { } &
             \includegraphics[width=0.247\linewidth]{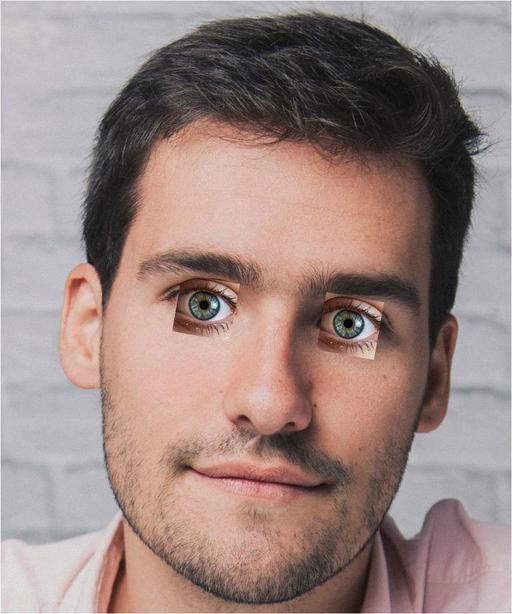} &
             \includegraphics[width=0.247\linewidth]{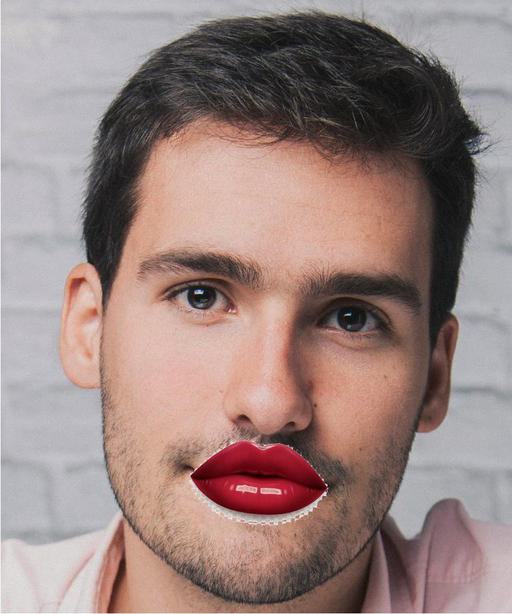} &
             \includegraphics[width=0.247\linewidth]{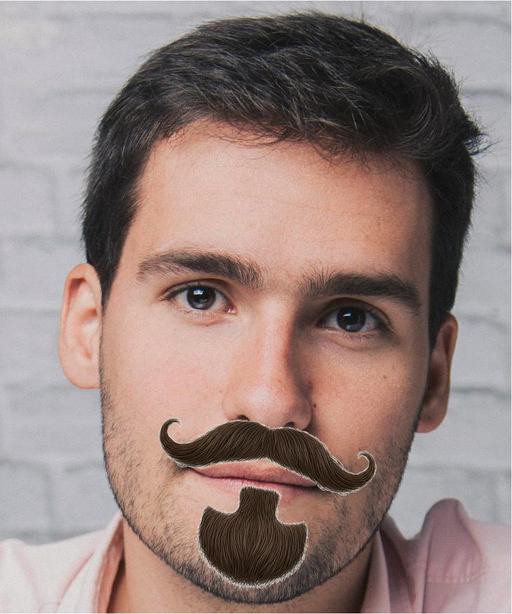} 
        \end{tabular}
        \begin{tabular}{cc cc cc c}
             \raisebox{15pt}{\rotatebox{90}{``... hiking''}} & { } &
             \includegraphics[width=0.31\linewidth]{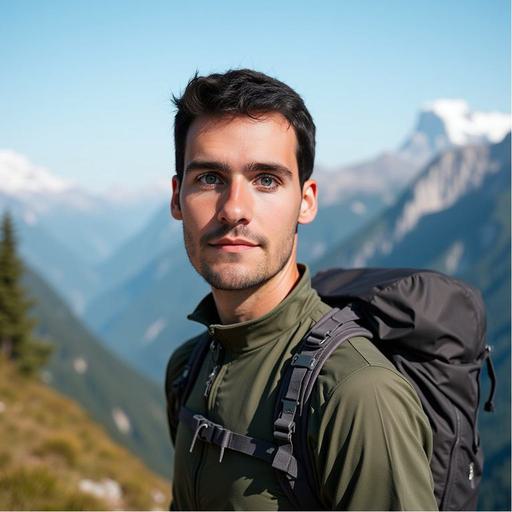} & { } &
             \includegraphics[width=0.31\linewidth]{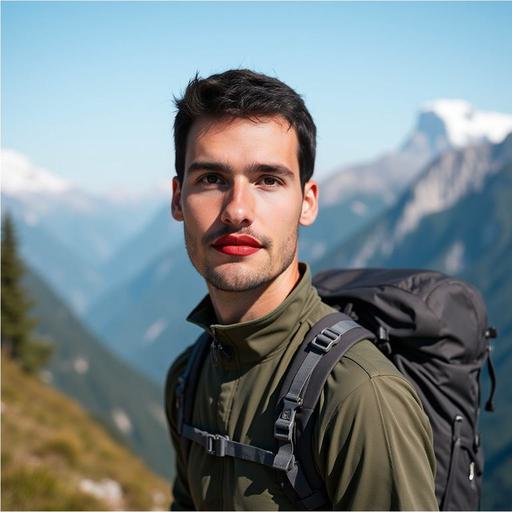} & { } &
             \includegraphics[width=0.31\linewidth]{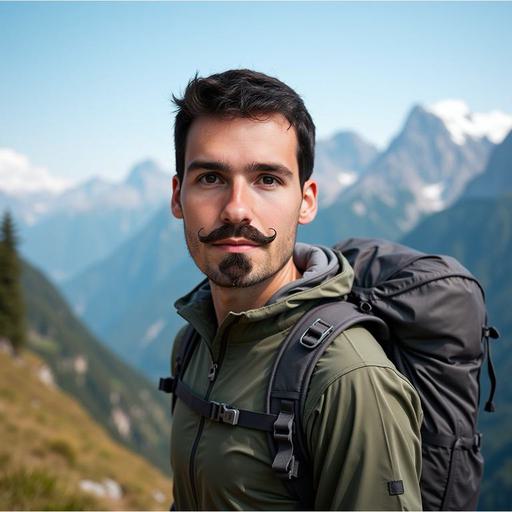} \\
             \raisebox{12pt}{\rotatebox{90}{``... in the city''}} & { } &
             \includegraphics[width=0.31\linewidth]{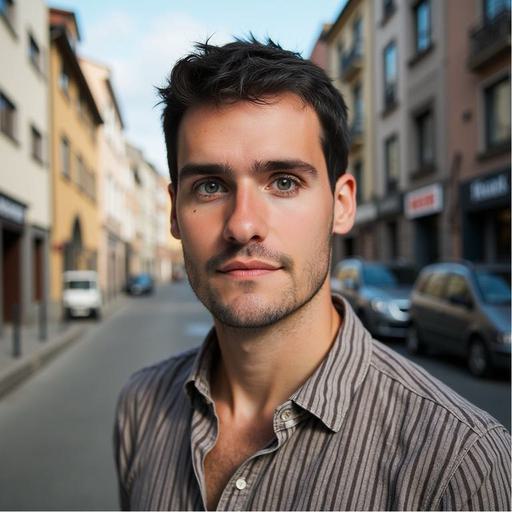} & { } &
             \includegraphics[width=0.31\linewidth]{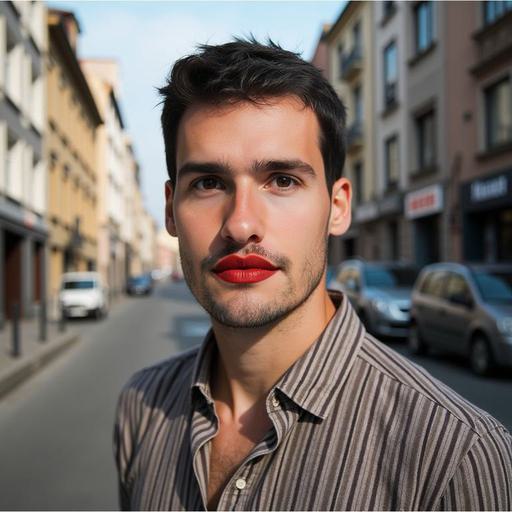} & { } &
             \includegraphics[width=0.31\linewidth]{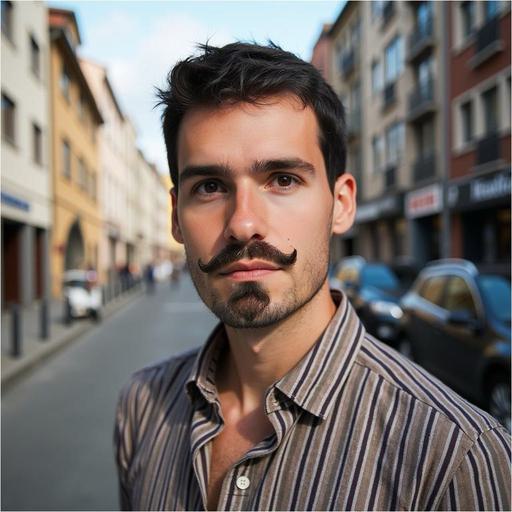} \\
             && Eyes && Lips && Van Dyke Beard
        \end{tabular}
    }
    \caption{Patch-based identity editing across multiple attributes and prompts. Top: we modify the source image by pasting different facial patches (eyes, lips, and a Van Dyke beard). Bottom: each modified image is encoded into the identity space to produce an edited identity representation, which is then used to generate diverse images under different prompts.}
    \vspace{-20pt}
    \label{fig:patch_editing_application}
\end{figure}

\subsection{Patch Editing}
\label{sec:patch_editing_app_supp}
We explore identity tuning through a patch-based operation. We perform a pixel-level manipulation of an input image, as shown in
\cref{fig:patch_editing_application},
and use the manipulated image’s representation as a new identity representation. We find that the encoder integrates the pasted patch smoothly into the representation, producing coherent identities that differ from the original only in the modified region. 
This method enables local identity edits and offers fine-grained control over the appearance of the edited part. It serves as a useful tool for targeted identity refinements, allowing edits that are difficult to specify through text, such as selecting a specific pair of eyeglasses or a particular mustache style.

\cref{fig:patch_editing_application}
presents three example edit types. In each, our method performs a localized modification of the identity, enabling consistent generation of the tuned identity across diverse contexts. 
The appearance of the pasted attributes is accurately preserved, for example in the style of the beard, while the remaining facial characteristics stay unchanged.

\begin{figure*}
    \centering
    \scriptsize
    \setlength{\tabcolsep}{0.5pt}
    \renewcommand{\arraystretch}{0.3}
    \addtolength{\belowcaptionskip}{-5pt}
    {

    \hspace{-0.2cm}
    \begin{minipage}{0.5\textwidth}
        \centering
        \begin{tabular}{c c c c c}
        &
        \multicolumn{2}{c}{
            \textbf{Original Identity}:\,
            \raisebox{-0.45\height}{%
                \includegraphics[width=0.15\textwidth]{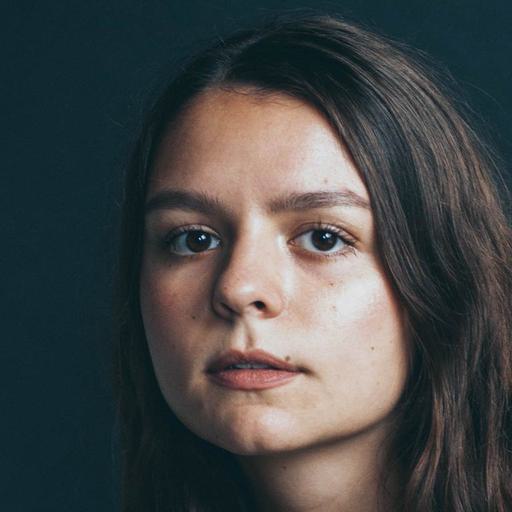}%
            }
        } &
        \multicolumn{2}{c}{\textbf{Edit}: Add Bangs} \\ \\

        \raisebox{0pt}{
            \rotatebox{90}{
            \begin{tabular}{c} PreciseControl \end{tabular}
            }
        } &
        \includegraphics[width=0.23\textwidth]{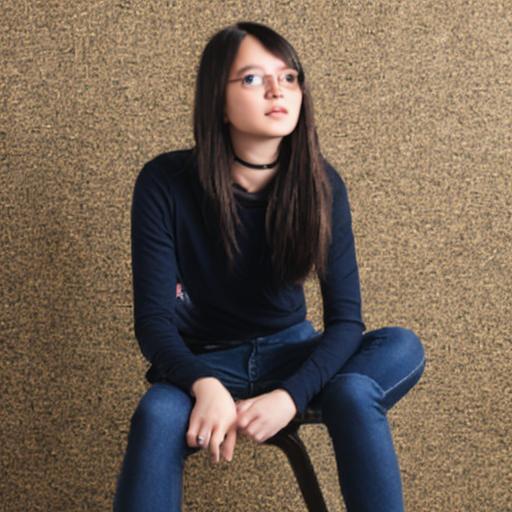} &
        \includegraphics[width=0.23\textwidth]{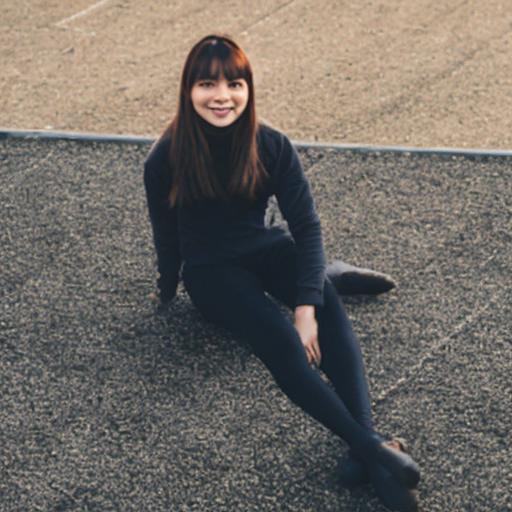} &
        \includegraphics[width=0.23\textwidth]{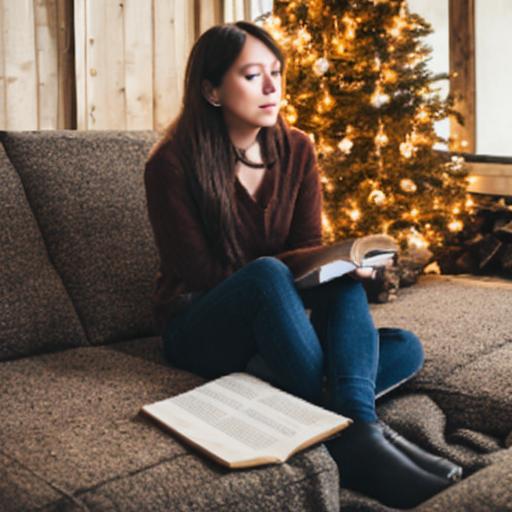} &
        \includegraphics[width=0.23\textwidth]{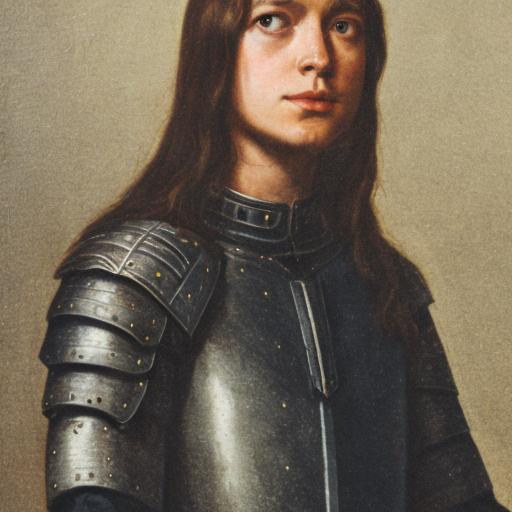} \\

        \raisebox{12pt}{
            \rotatebox{90}{
            \begin{tabular}{c} W2W  \end{tabular}
            }
        } &
        \includegraphics[width=0.23\textwidth]{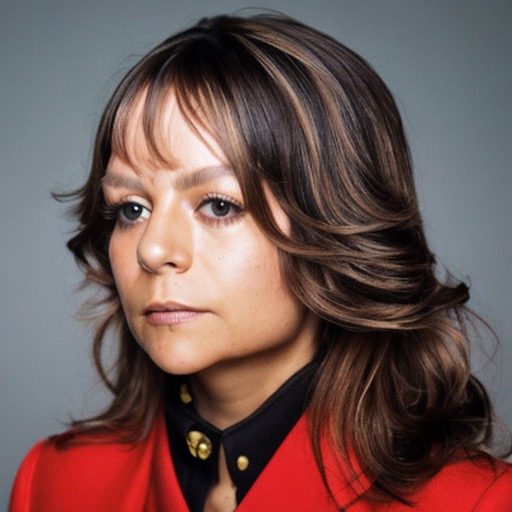} &
        \includegraphics[width=0.23\textwidth]{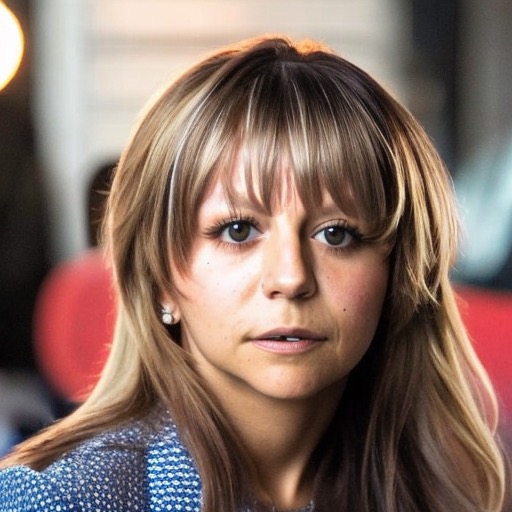} &
        \includegraphics[width=0.23\textwidth]{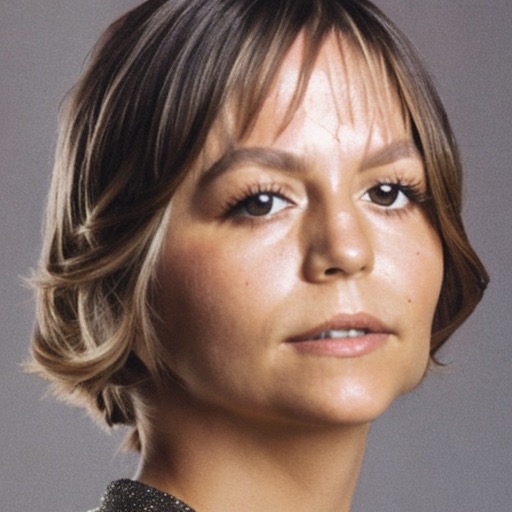} &
        \includegraphics[width=0.23\textwidth]{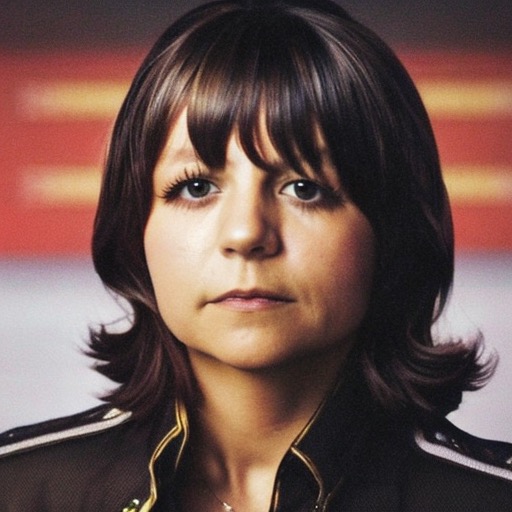} \\

        \raisebox{3pt}{
            \rotatebox{90}{
            \begin{tabular}{c} Kont. Direct \end{tabular}
            }
        } &
        \includegraphics[width=0.23\textwidth]{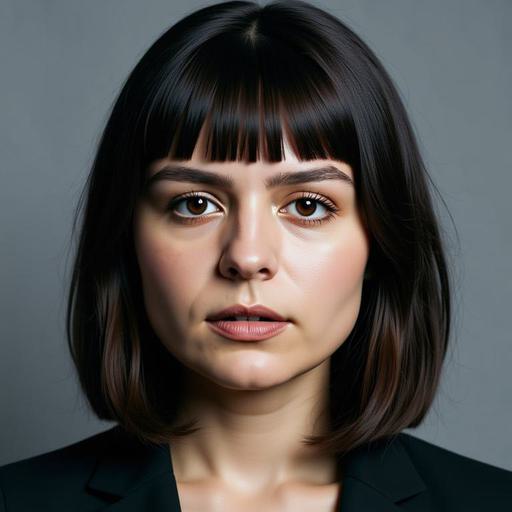} &
        \includegraphics[width=0.23\textwidth]{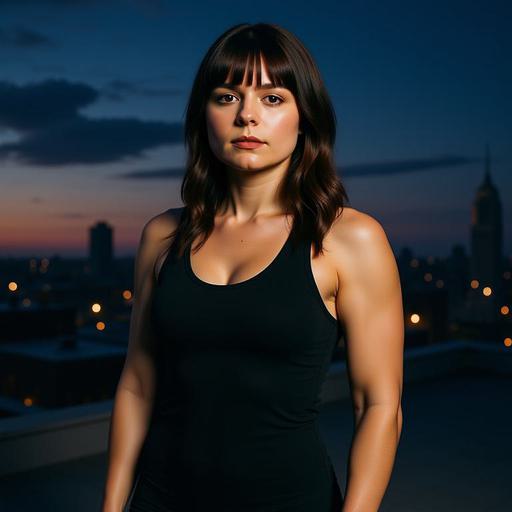} &
        \includegraphics[width=0.23\textwidth]{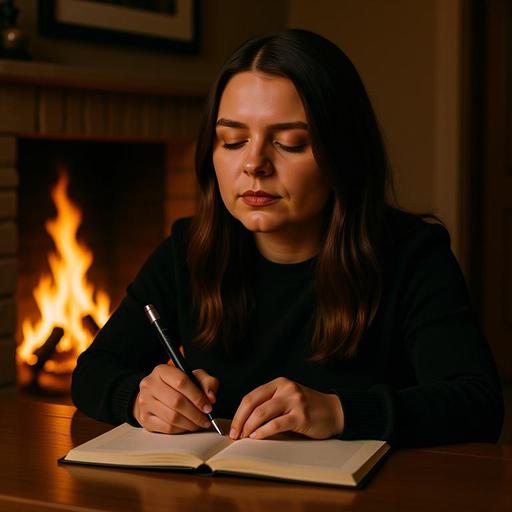} &
        \includegraphics[width=0.23\textwidth]{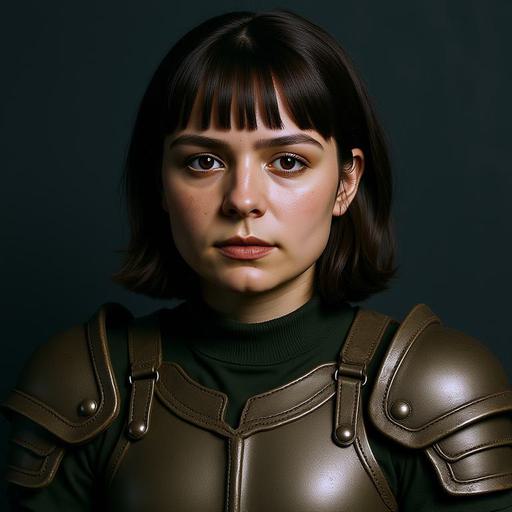} \\

        \raisebox{6pt}{
            \rotatebox{90}{
            \begin{tabular}{c} Kont. Seq. \end{tabular}
            }
        } &
        \includegraphics[width=0.23\textwidth]{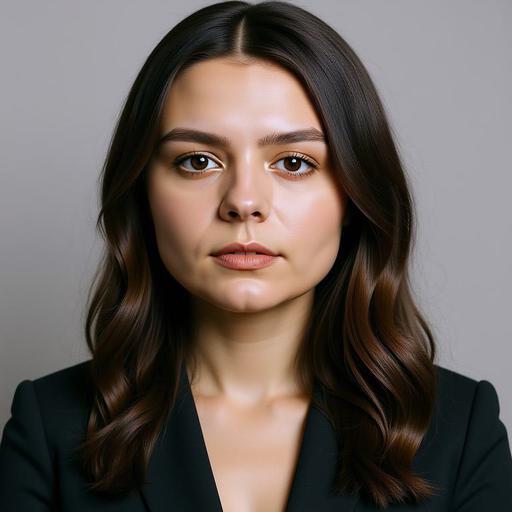} &
        \includegraphics[width=0.23\textwidth]{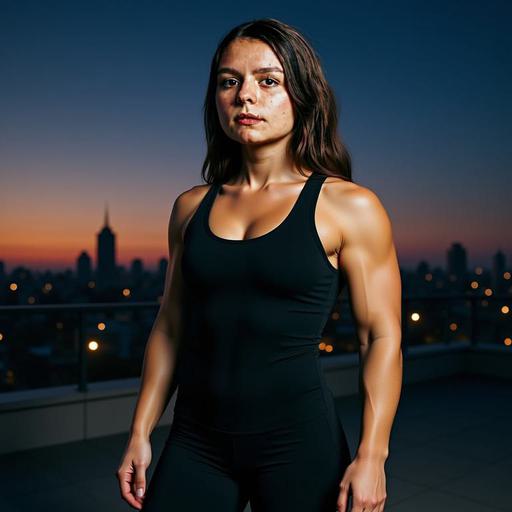} &
        \includegraphics[width=0.23\textwidth]{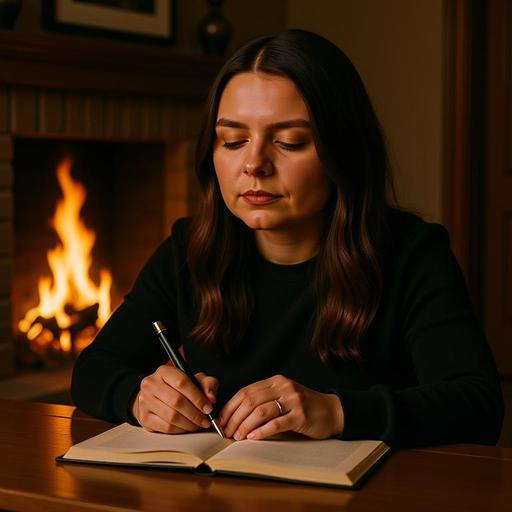} &
        \includegraphics[width=0.23\textwidth]{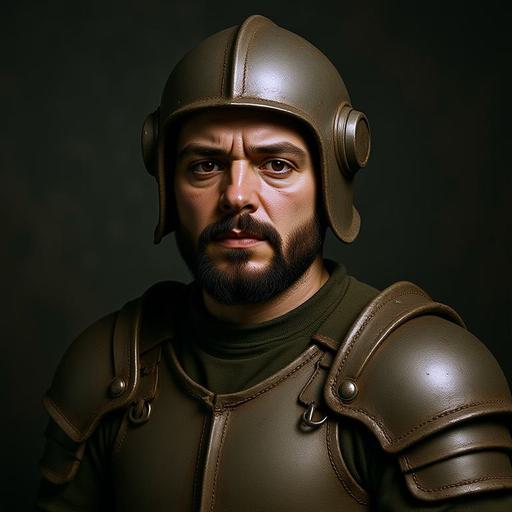} \\

        \raisebox{12pt}{
            \rotatebox{90}{
            \begin{tabular}{c} Ours  \end{tabular}
            }
        } &
        \includegraphics[width=0.23\textwidth]{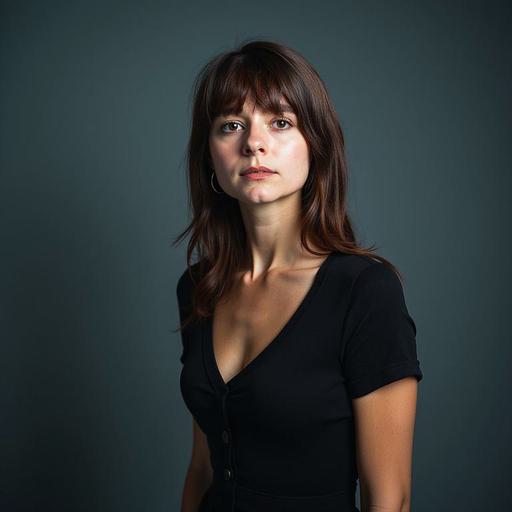} &
        \includegraphics[width=0.23\textwidth]{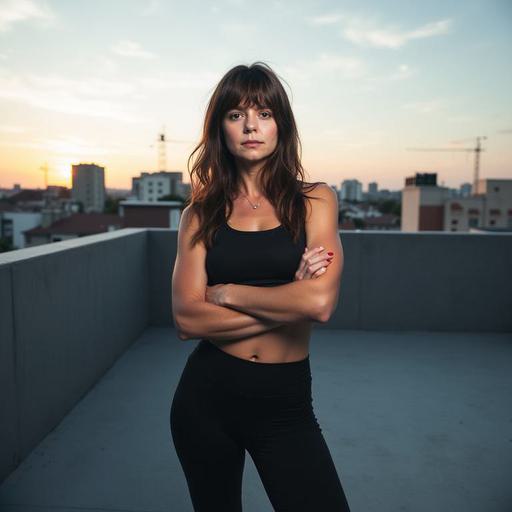} &
        \includegraphics[width=0.23\textwidth]{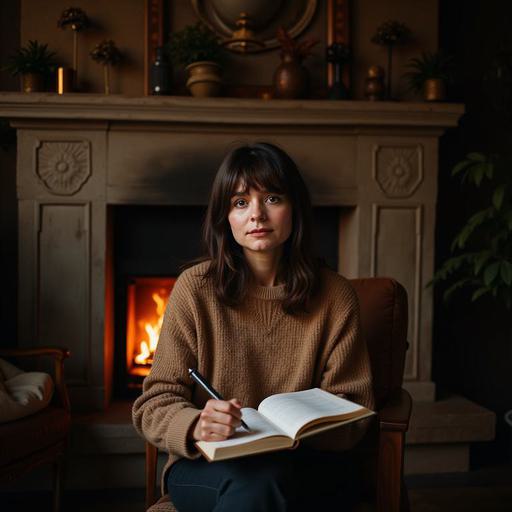} &
        \includegraphics[width=0.23\textwidth]{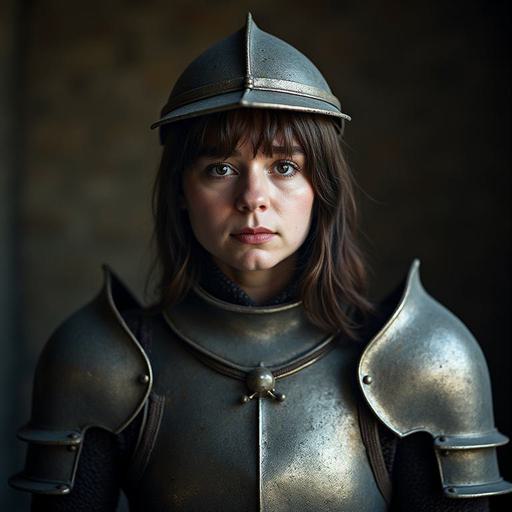} \\

        \end{tabular}
    \end{minipage}%
    \begin{minipage}{0.5\textwidth}
        \centering
        \begin{tabular}{c c c c c}
        &
        \multicolumn{2}{c}{
            \textbf{Original Identity}:\,
            \raisebox{-0.45\height}{%
                \includegraphics[width=0.15\textwidth]{images/qual_comp/bangs/37986.jpg}%
            }
        } &
        \multicolumn{2}{c}{\textbf{Edit}: Add Beard} \\ \\

        \raisebox{0pt}{
            \rotatebox{90}{
            \begin{tabular}{c} PreciseControl  \end{tabular}
            }
        } &
        \includegraphics[width=0.23\textwidth]{images/qual_comp/beard/PreciseControl/0_seed42.jpg} &
        \includegraphics[width=0.23\textwidth]{images/qual_comp/beard/PreciseControl/2_seed1337.jpg} &
        \includegraphics[width=0.23\textwidth]{images/qual_comp/beard/PreciseControl/7_seed3141.jpg} &
        \includegraphics[width=0.23\textwidth]{images/qual_comp/beard/PreciseControl/12_seed2718.jpg} \\

        \raisebox{12pt}{
            \rotatebox{90}{
            \begin{tabular}{c} W2W  \end{tabular}
            }
        } &
        \includegraphics[width=0.23\textwidth]{images/qual_comp/beard/W2W/0_seed42.jpg} &
        \includegraphics[width=0.23\textwidth]{images/qual_comp/beard/W2W/2_seed1337.jpg} &
        \includegraphics[width=0.23\textwidth]{images/qual_comp/beard/W2W/7_seed3141.jpg} &
        \includegraphics[width=0.23\textwidth]{images/qual_comp/beard/W2W/12_seed2718.jpg} \\

        \raisebox{3pt}{
            \rotatebox{90}{
            \begin{tabular}{c} Kont. Direct \end{tabular}
            }
        } &
        \includegraphics[width=0.23\textwidth]{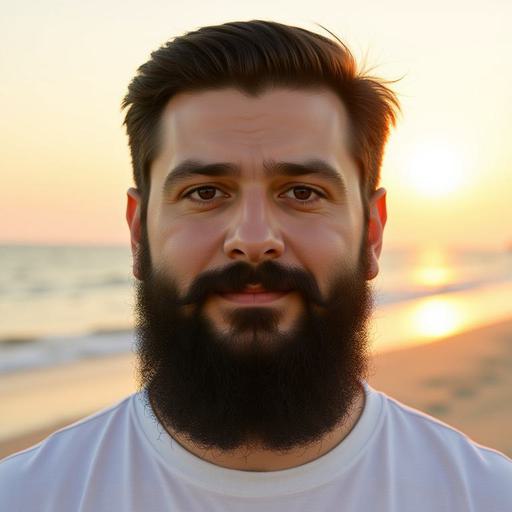} &
        \includegraphics[width=0.23\textwidth]{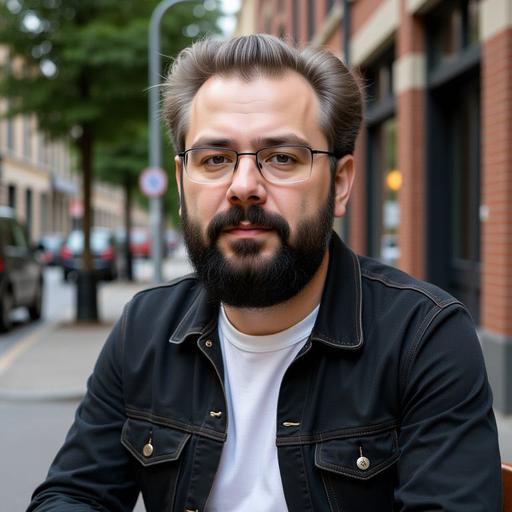} &
        \includegraphics[width=0.23\textwidth]{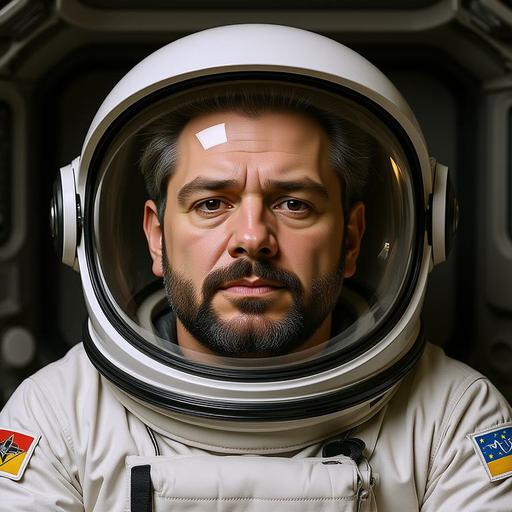} &
        \includegraphics[width=0.23\textwidth]{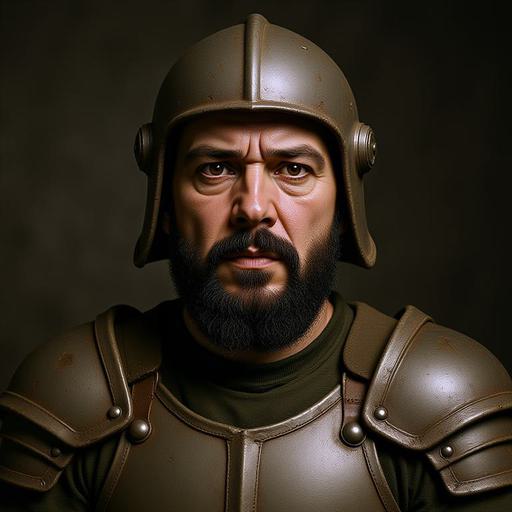} \\

        \raisebox{6pt}{
            \rotatebox{90}{
            \begin{tabular}{c} Kont. Seq. \end{tabular}
            }
        } &
        \includegraphics[width=0.23\textwidth]{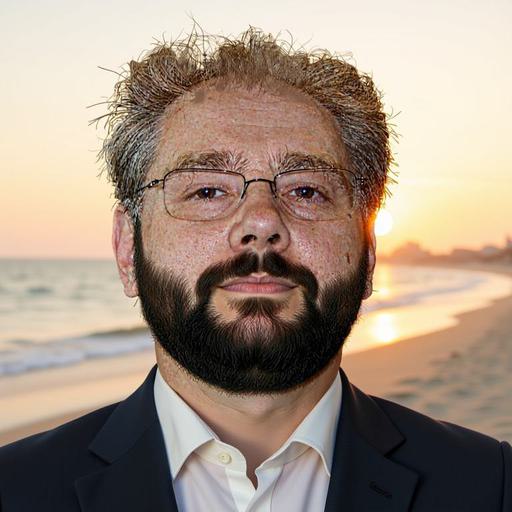} &
        \includegraphics[width=0.23\textwidth]{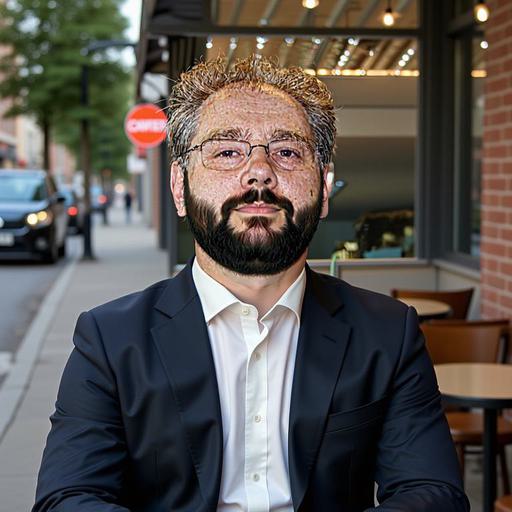} &
        \includegraphics[width=0.23\textwidth]{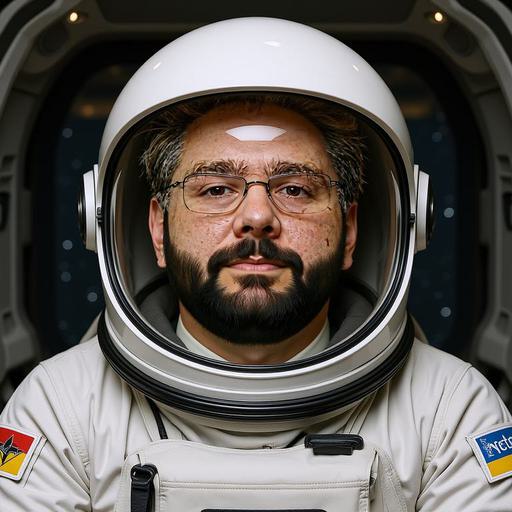} &
        \includegraphics[width=0.23\textwidth]{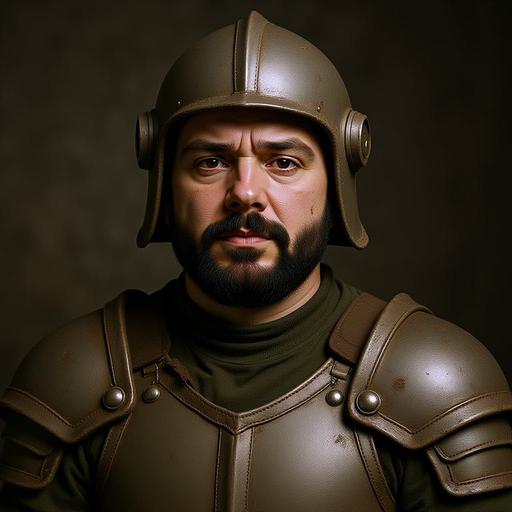} \\

        \raisebox{12pt}{
            \rotatebox{90}{
            \begin{tabular}{c} Ours  \end{tabular}
            }
        } &
        \includegraphics[width=0.23\textwidth]{images/qual_comp/beard/ours/0_seed42.jpg} &
        \includegraphics[width=0.23\textwidth]{images/qual_comp/beard/ours/2_seed1337.jpg} &
        \includegraphics[width=0.23\textwidth]{images/qual_comp/beard/ours/7_seed3141.jpg} &
        \includegraphics[width=0.23\textwidth]{images/qual_comp/beard/ours/12_seed2718.jpg} \\

        \end{tabular}
        
    \end{minipage}
    }
    \caption{
    Qualitative comparison. Each grid presents the results of an identity tuning of a single attribute, the left grid adds bangs and the right grid adds a beard. Each grid shows results for four prompts per method as columns. Our method consistently applies the target attribute across diverse prompts while keeping the edited identity coherent.
    }
    \label{fig:sup_qual_compare_figure}
\end{figure*}

\subsection{Comparisons}
\label{sec:comparisons_sup}

\paragraph{Qualitative Comparisons}
\cref{fig:sup_qual_compare_figure}
presents a qualitative comparison across methods. For each method, we show results for two identities, applying a different edit to each. We generate four images per edited identity under diverse prompts to assess identity consistency across prompts.

Competing methods often apply the requested edit yet fail to preserve identity across prompts.
In Flux Kontext Direct~\cite{labs2025flux1kontextflowmatching} this is expected: without an explicit identity embedding, attributes are not applied consistently from one generation to the next.
The identity often drifts even between the input and its edited output, especially for large edits such as adding a beard.
Flux Kontext Sequential improves identity coherence relative to Direct, but in cases when an edit fails (for example, bangs) it fail across all samples.
We also observe spillover into unrelated regions, such as unintended skin changes. In contrast, our approach preserves a coherent identity while reliably expressing the intended edit across prompts.

\paragraph{Benchmark}
The benchmark contains 20 identity images sampled from FFHQ, five basic identity edits (the common edits supported by Precise Control and Weights2Weights), and seven advanced edits that require finer control. We use 13 prompts generated by an LLM~\citep{openai2025chatgpt}, each describing a scene that includes a person. Applying all edits to each identity yields 100 basic identity edits and 140 advanced edits. For each method, we render 13 images per edit, resulting in 1{,}300 basic and 1{,}820 advanced images per method.
\cref{tb:benchmark_table}
presents the prompts which used to generate the results images, the basic and the advanced edit instructions.

As summarized in
Table 2 in the main paper,
our method was preferred over all baselines across all categories.
These results indicate that users judged our approach to better preserve the original identity, maintain consistency across prompts, and adhere to both the specified edit and the target prompt.
That indicates the superiority of our approach for the task of identity tuning.

\subsection{Flux Kontext implementation details}
\label{sec:supp:kontext}

We evaluate two Flux Kontext baselines~\citep{labs2025flux1kontextflowmatching} that we defined for qualitative and quantitative analyses:

\textbf{Kontext Direct (single step).}
Given a single reference image and a target scene prompt, Kontext applies the edit and renders the scene in one pass. Template:
\texttt{\{EDIT\} this person and generate an image of \{BASE\_PROMPT\}}.
Example (freckles):
\texttt{Add freckles to this person and generate an image of \{BASE\_PROMPT\}}.

\textbf{Kontext Sequential (two steps).}
We separate the local edit from the scene rendering. Step 1 edits the portrait in isolation; Step 2 conditions on the edited image to render the scene.
Step 1: \texttt{\{EDIT\} this person}
$\rightarrow$
Step 2: \texttt{Generate an image of this person \{BASE\_PROMPT\}}.
Example (freckles): \texttt{Add freckles to this person} $\rightarrow$ \texttt{Generate an image of this person \{BASE\_PROMPT\}}.

\textbf{Settings.}
Official checkpoint with default inference parameters, the reference portrait is the only image input.

\begin{table}
    \centering
    \scriptsize
    \vspace{-5pt}
    \caption{
        The complete set of source prompts and edit instruction used for our quantitative evaluation and user study.
    }
    \begin{tabular}{c}

        \toprule
        \textbf{Prompts} \\
        \midrule
        Professional headshot of one person on a beach at golden hour, \\ centered composition, eyes looking at camera, crisp focus \\\\
        Studio fashion portrait of one person on a seamless backdrop, \\ centered composition, three-point lighting, tack-sharp details \\\\
        Street portrait of one person at a sidewalk café, \\ centered composition, natural light, subtle bokeh \\\\
        Athletic portrait of one person on a rooftop at dusk, \\ centered composition, dynamic stance, clean background separation \\\\
        Renaissance-inspired oil painting of one person, half-length, \\ centered composition, soft chiaroscuro, ornate textures \\\\
        Cyberpunk portrait of one person under neon signs, \\ centered composition, teal–magenta glow, shallow DOF \\\\
        Outdoor lifestyle portrait of one person in a meadow, \\ centered composition, airy pastel tones, gentle bokeh \\\\
        Astronaut portrait of one person inside a spacecraft, \\ centered composition, visor lifted, panel reflections, high realism \\\\
        Cozy indoor portrait of one person journaling by a fireplace, \\ centered composition, warm tungsten glow, soft textures \\\\
        Surreal portrait of one person hovering above desert dunes, \\ centered composition, pastel sky, tilt-shift effect \\\\
        Hip-hop album cover portrait of one person by a classic car at night, \\ centered composition, colored gels, subtle film grain \\\\
        Minimalist editorial portrait of one person in a red coat \\ against a white wall, centered composition, strong negative space \\\\
        Medieval knight portrait of one person in weathered armor, \\ centered composition, raised visor, cinematic realism \\
               
        \bottomrule
        \textbf{Basic Edits} \\
        \midrule
        Add/Remove beard \\
        Make the person bald \\
        Change the person gender \\
        Add/Remove eyeglasses \\
        Add/Remove bangs \\
        \bottomrule
        \textbf{Advanced Edits} \\
        \midrule
        Add/Remove freckles \\
        Make the eyebrows arched \\
        Make the eyebrows bushy \\
        Add/Remove big lips \\
        Add/Remove mustache \\
        Make the skin pale \\
        Add rosy cheeks \\
        \bottomrule
    \end{tabular}
    \label{tb:benchmark_table}
\end{table}

\section{Limitations}
\label{sec:limitations}
A limitation of our approach relates to the token capacity of the Q-Former. Our localized editing performs optimally when a sufficient number of queries are used, enabling the natural disentanglement of facial regions. Conversely, a low token count forces the model to encode multiple facial aspects into a single token. This compression prevents isolated editing through single or group token manipulation. While our discovered semantic directions (both supervised and unsupervised) can still navigate this compressed space to edit specific attributes without affecting others, direct token swapping is heavily penalized. Consequently, operations like local blending between identities produce entangled results if the underlying tokens contain information from multiple distinct facial parts. Ultimately, these findings suggest that future personalization models should train the Q-Former with an adequate token count to preserve natural disentanglement, thereby unlocking greater controllability for downstream editing tasks.

\section{Broader Impacts}
\label{sec:broader_impacts}
Our work advances the fundamental understanding of identity representation in generative models, enabling more precise and controllable editing. While this offers significant benefits for creative applications and personalized media, it also carries potential societal risks. Similar to other identity manipulation techniques, fine-grained control over facial attributes could be misused to generate misleading content or alter a person's likeness without consent. To mitigate these risks, we encourage the continued development of robust detection algorithms and responsible deployment practices, such as requiring watermarks for generated media and adhering to strict consent protocols when utilizing personal reference images.

\begin{figure*}
    \centering
    \small
    
    \setlength{\tabcolsep}{0pt}
    \begin{tabular}{@{}c@{\hspace{6pt}}c@{}c@{}c@{}c@{}c@{}}
    
        \includegraphics[width=0.16\textwidth]{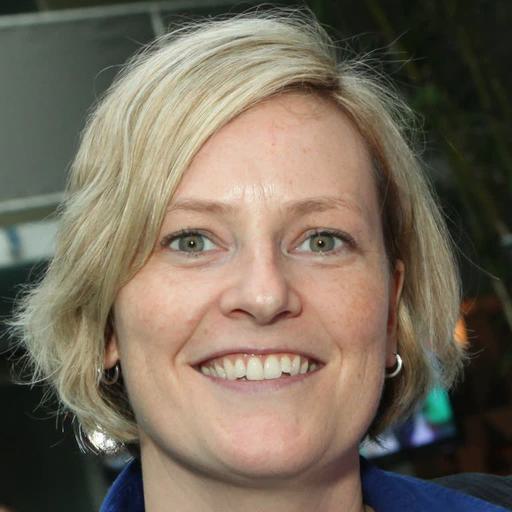} &
        \includegraphics[width=0.16\textwidth]{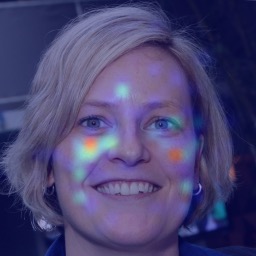} &
        \includegraphics[width=0.16\textwidth]{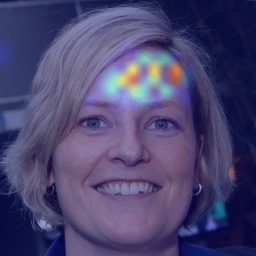} &
        \includegraphics[width=0.16\textwidth]{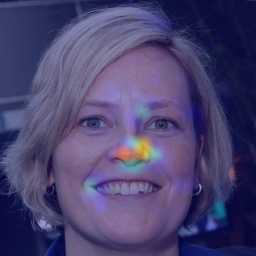} &
        \includegraphics[width=0.16\textwidth]{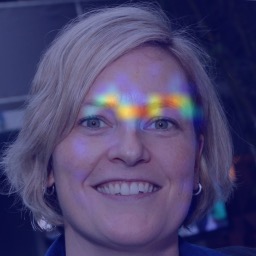} &
        \includegraphics[width=0.16\textwidth]{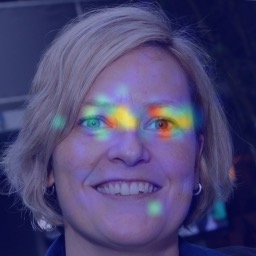} \\
    
        \includegraphics[width=0.16\textwidth]{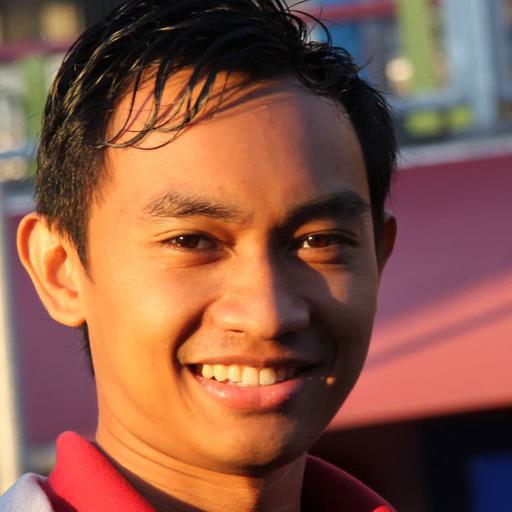} &
        \includegraphics[width=0.16\textwidth]{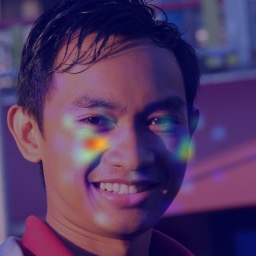} &
        \includegraphics[width=0.16\textwidth]{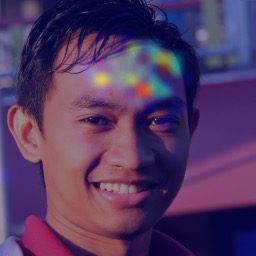} &
        \includegraphics[width=0.16\textwidth]{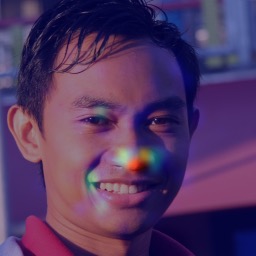} &
        \includegraphics[width=0.16\textwidth]{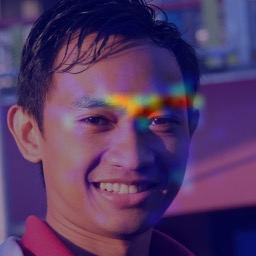} &
        \includegraphics[width=0.16\textwidth]{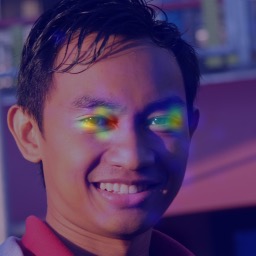} \\
    
        \includegraphics[width=0.16\textwidth]{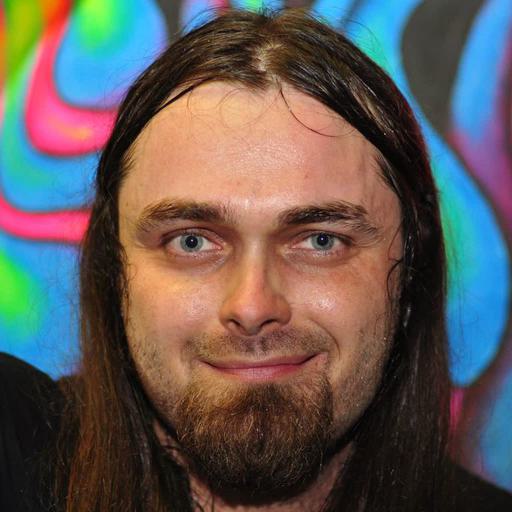} &
        \includegraphics[width=0.16\textwidth]{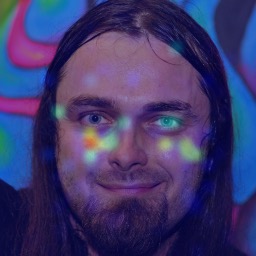} &
        \includegraphics[width=0.16\textwidth]{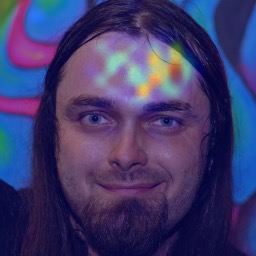} &
        \includegraphics[width=0.16\textwidth]{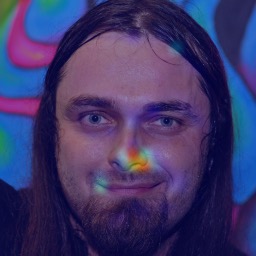} &
        \includegraphics[width=0.16\textwidth]{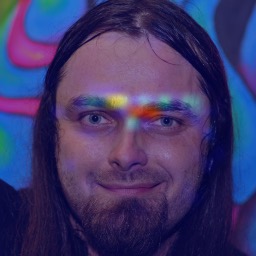} &
        \includegraphics[width=0.16\textwidth]{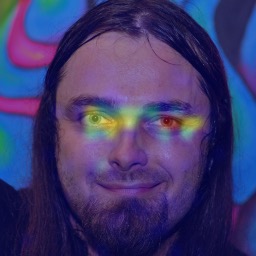} \\
    
        \includegraphics[width=0.16\textwidth]{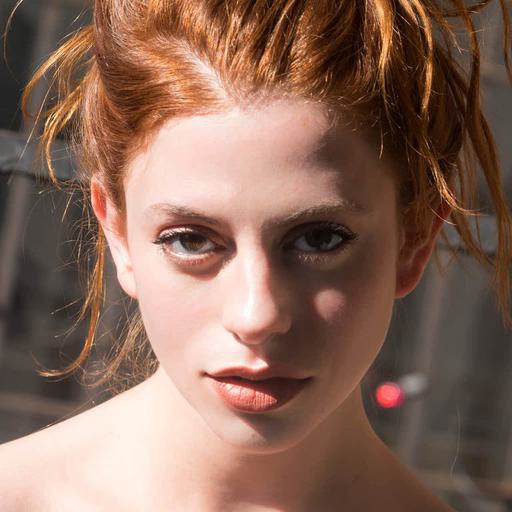} &
        \includegraphics[width=0.16\textwidth]{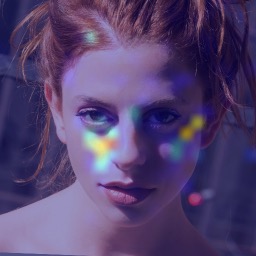} &
        \includegraphics[width=0.16\textwidth]{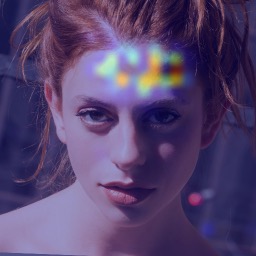} &
        \includegraphics[width=0.16\textwidth]{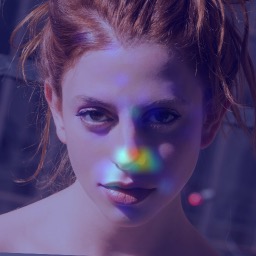} &
        \includegraphics[width=0.16\textwidth]{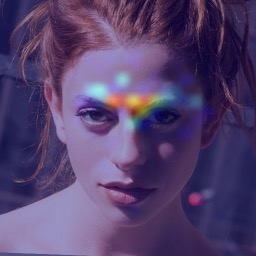} &
        \includegraphics[width=0.16\textwidth]{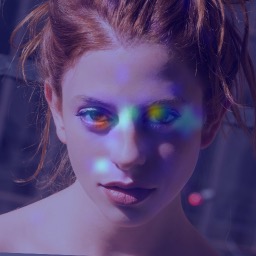} \\
    
        \includegraphics[width=0.16\textwidth]{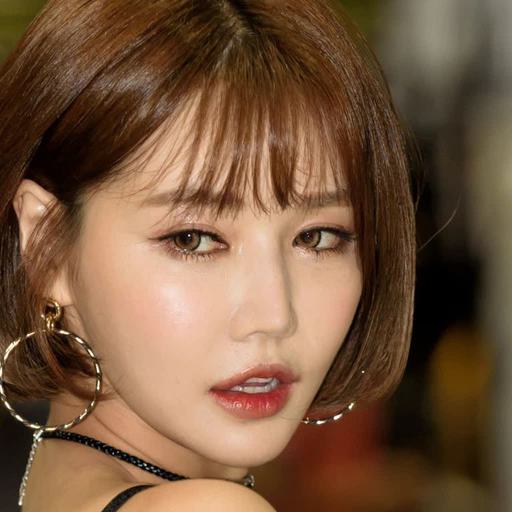} &
        \includegraphics[width=0.16\textwidth]{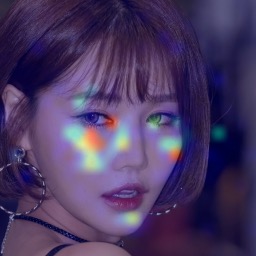} &
        \includegraphics[width=0.16\textwidth]{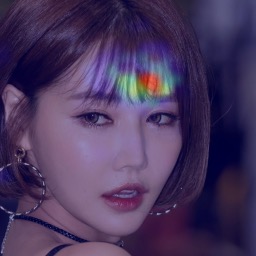} &
        \includegraphics[width=0.16\textwidth]{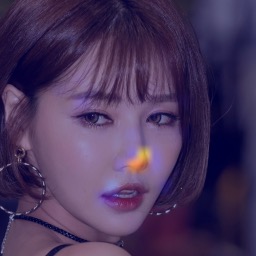} &
        \includegraphics[width=0.16\textwidth]{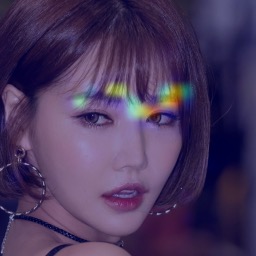} &
        \includegraphics[width=0.16\textwidth]{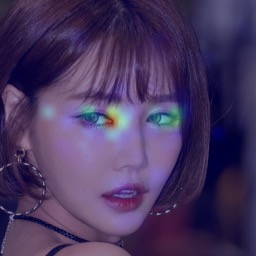} \\
    
        \includegraphics[width=0.16\textwidth]{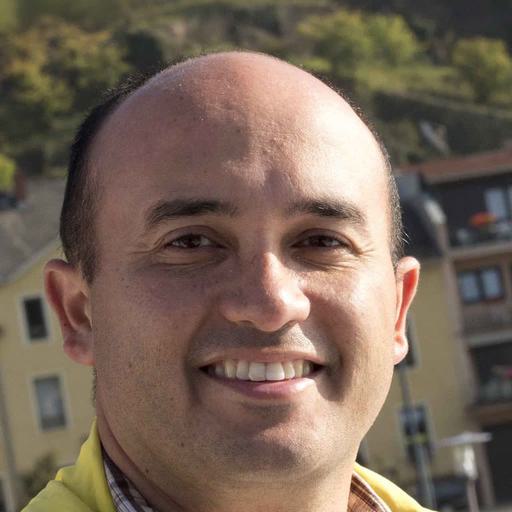} &
        \includegraphics[width=0.16\textwidth]{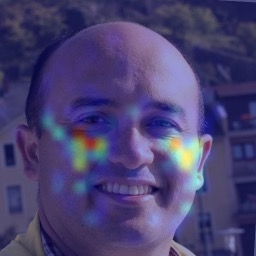} &
        \includegraphics[width=0.16\textwidth]{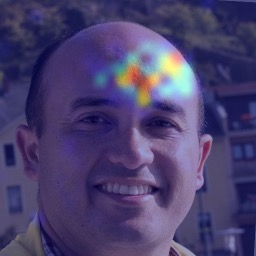} &
        \includegraphics[width=0.16\textwidth]{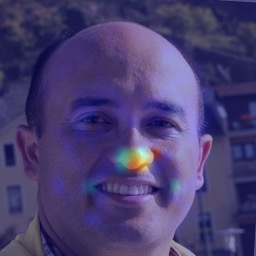} &
        \includegraphics[width=0.16\textwidth]{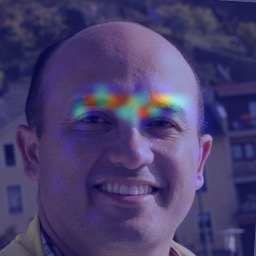} &
        \includegraphics[width=0.16\textwidth]{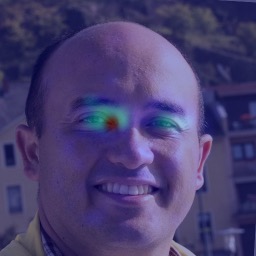} \\
    
        \includegraphics[width=0.16\textwidth]{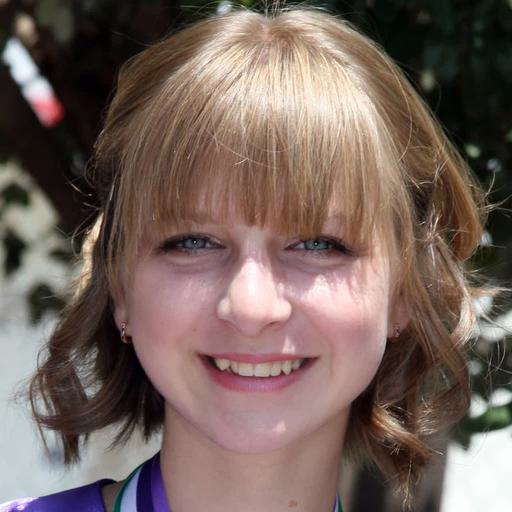} &
        \includegraphics[width=0.16\textwidth]{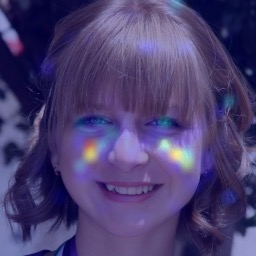} &
        \includegraphics[width=0.16\textwidth]{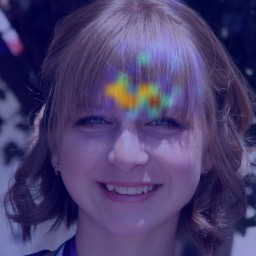} &
        \includegraphics[width=0.16\textwidth]{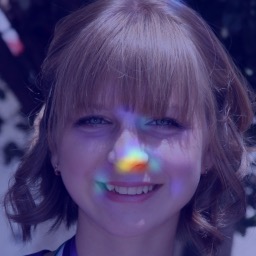} &
        \includegraphics[width=0.16\textwidth]{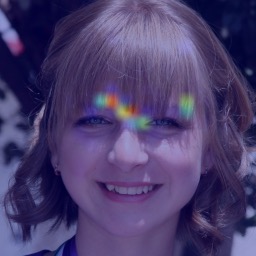} &
        \includegraphics[width=0.16\textwidth]{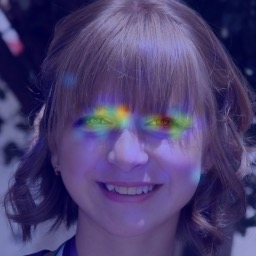} \\

        Input Image &
        Cheeks Token & 
        Forehead Token & 
        Nose Token & 
        Eyebrows Token & 
        Eyes Token \\

    \end{tabular}
    \caption{
        Token behavior in the face encoder. Q-Former attention maps for five learned query tokens reveal localized semantics. The same token keeps the same role across both identities, showing consistent token-to-region mapping.}
    \label{fig:supp_attn_tokens}
\end{figure*}

\begin{figure*}[t]
    \centering
    \includegraphics[width=0.6\linewidth]{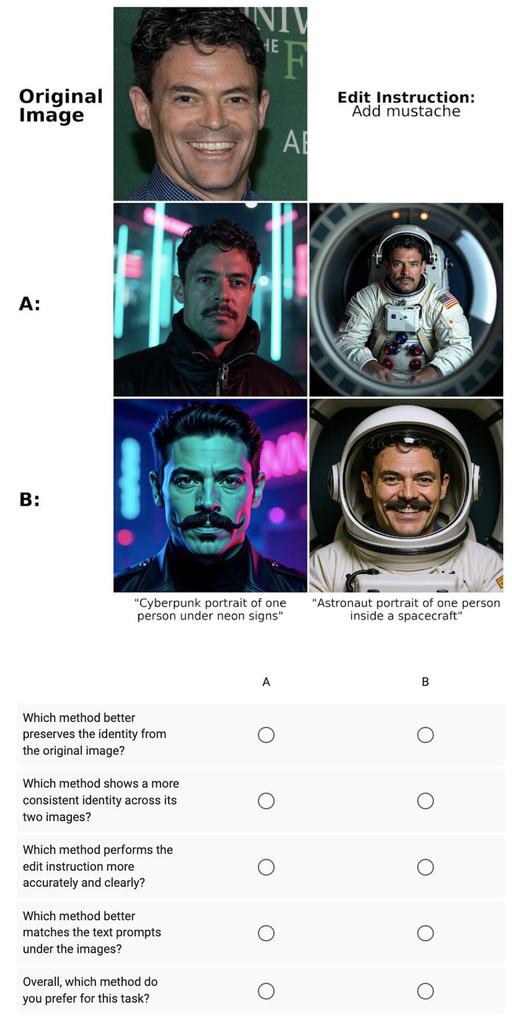}
    \caption{An example of a question in the user study
    }
    \label{fig:user_study_example}
\end{figure*}

\end{appendices}

\end{document}